\documentclass[runningheads]{llncs}

\usepackage[mobile]{eccv}

\definecolor{cvprblue}{rgb}{0.21,0.49,0.74}
\usepackage[pagebackref,breaklinks,colorlinks,allcolors=cvprblue]{hyperref}

\usepackage{url}
\usepackage{bm} %
\usepackage{wrapfig}

\usepackage[ruled]{algorithm2e} %

\SetAlFnt{\small}
\SetAlCapFnt{\small}
\SetAlCapNameFnt{\small}
\SetAlCapHSkip{0pt}

\newcommand{\name}{\textsc{SuperFlex}}

\PassOptionsToPackage{table,dvipsnames}{xcolor}

\usepackage{graphicx,calc} %
\usepackage{color}    %
\usepackage{ifthen}
\usepackage{paralist}
\usepackage{longtable}
\usepackage{colortbl}
\usepackage{nomencl}
\usepackage{mathtools}
\usepackage{booktabs}
\usepackage{overpic}
\usepackage{pgfplots}
\usepackage{pgfplotstable}
\usepackage{amsfonts}
\usepackage{subcaption}
\usepackage{mathrsfs}
\usepackage[scr=boondox]{mathalpha}

\newcommand{\eg}{\textit{e.g.}}

\usepackage{pifont}%

\newcommand{\cmark}{\ding{51}}%
\newcommand{\xmark}{\ding{55}}%

\renewcommand*{\sin}[1]{\text{sin}\left( #1 \right)}
\renewcommand*{\cos}[1]{\text{cos}\left( #1 \right)}

\newcommand*{\loss}{\mathcal{L}}

\newcommand{\R}{\mathbb{R}}

\definecolor{customgreen}{HTML}{91C799} %
\definecolor{customred}{HTML}{EBB1AB} %
\definecolor{customgray}{HTML}{CCCCCC} %

\renewcommand{\vec}[1]{\mathbf{#1}}

\NewDocumentCommand{\fnPr}{}{\mathbb{P}}
\RenewDocumentCommand{\Pr}{om}{\fnPr\IfValueT{#1}{_{#1}}\parentheses*{#2}}

\NewDocumentCommand{\E}{somo}{\ensuremath{\mathbb{E}\IfValueT{#2}{_{#2}}{} \IfBooleanTF{#1}{#3}{\IfValueTF{#4}{\left[#3\ \middle|\ #4\right]}{\brackets*{#3}}}}}
\NewDocumentCommand{\Cov}{som}{\mathrm{Cov}\IfValueT{#2}{_{#2}}{} \IfBooleanTF{#1}{#3}{\brackets*{#3}}}

\NewDocumentCommand{\N}{omm}{\mathcal{N}(\IfValueT{#1}{#1;}{} #2, #3)}

\NewDocumentCommand{\uSN}{o}{\mathcal{N}(\IfValueT{#1}{#1;}{} 0, 1)}
\NewDocumentCommand{\KF}{ommmmm}{\mathcal{KF}(\IfValueT{#1}{#1;}{} #2, #3, #4, #5, #6)}
\NewDocumentCommand{\GP}{omm}{\mathcal{GP}(\IfValueT{#1}{#1;}{} #2, #3)}
\NewDocumentCommand{\Unif}{om}{\mathrm{Unif}(\IfValueT{#1}{#1;}{} #2)}
\NewDocumentCommand{\Bern}{om}{\mathrm{Bern}(\IfValueT{#1}{#1;}{} #2)}
\NewDocumentCommand{\Bin}{omm}{\mathrm{Bin}(\IfValueT{#1}{#1;}{} #2, #3)}
\NewDocumentCommand{\Beta}{omm}{\mathrm{Beta}(\IfValueT{#1}{#1;}{} #2, #3)}
\NewDocumentCommand{\Laplace}{omm}{\mathrm{Laplace}(\IfValueT{#1}{#1;}{} #2, #3)}

\usepackage{url}

\usepackage{pgfplots}
\pgfplotsset{compat=1.18}

\pgfplotsset{every axis/.append style={
title={\small \textbf{Overall Preference}},
    title style={at={(0.5,1)}, anchor=south, yshift=0pt},
    ymin=0, ymax=100,
    ybar stacked, bar width=15pt,
    xtick=data,
    ytick={},
    yticklabels={},
    grid=major,
    major grid style={line width=0.3pt, gray!30},
    axis x line=bottom,
    axis y line=left,
    axis line style = {draw=none},
    xticklabels={\tiny Spice-E, \tiny Spice-E-T}, %
    x tick label style={rotate=0, anchor=north},
    enlarge x limits=0.35,
    nodes near coords={\pgfmathprintnumber{\pgfplotspointmeta}\%},
    every node near coord/.append style={font=\tiny, text=black},
    width=3cm, height=4cm,
    tick label style={font=\small},
}}

\definecolor{my_blue}{RGB}{95, 126, 172}
\definecolor{my_violet}{RGB}{108, 60, 132}

\usepackage{tikz}
\usetikzlibrary{arrows.meta}

\usetikzlibrary{shapes.geometric, decorations.markings, calc}

\usepackage{eccvabbrv}

\usepackage{graphicx}
\usepackage{booktabs}

\usepackage[accsupp]{axessibility}  %

\usepackage{hyperref}
\usepackage{orcidlink}

\newcommand{\method}{\textsf{SuperFlex}}

\begin{document}

\title{\method{}: Deformable Superquadrics for\\ Point Cloud Decomposition} 

\titlerunning{\method}

\author{
\centering
{Gabriel Tavernini}$^{1*}$ \hspace{7px} 
{Elisabetta Fedele}$^{1*}$ \hspace{7px}
{Tiago Novello}$^{2,3}$\\
{Leonidas Guibas}${^2}$  \hspace{7px}%
{Marc Pollefeys}${^{1}}$ \hspace{7px}
{Francis Engelmann}${^{4}}$
}
{
\small
\institute{$^1$ETH Zurich \hspace{10px}
$^2$Stanford University \hspace{10px}
$^3$IMPA \hspace{10px}
$^4$USI Lugano}
}

\authorrunning{Tavernini et al.}

\maketitle

\begin{abstract}
Superquadrics have proven to provide a compact, geometrically meaningful representation for 3D objects.
However, existing methods suffer from limited reconstruction accuracy, are restricted to rigid primitives, and lack robustness to partial point clouds.
In this work, we present \method{}, an enhanced framework that expands the expressive power and applicability of superquadric decompositions. First, we introduce a novel loss formulation which significantly improves reconstruction accuracy. Second, we include bending and tapering deformations, enabling high-fidelity representation of curved and asymmetric geometries. Finally, we leverage these high-quality decompositions as supervision to train a model that is robust to partial real-world point clouds. Experiments demonstrate substantial improvements in reconstruction accuracy over both optimization- and learning-based baselines while maintaining a highly compact primitive representation. Project page: \href{https://superflex3d.github.io}{https://superflex3d.github.io}.

\end{abstract}

\vspace{-25px}

\begin{figure}[b!]
\centering
\setlength{\tabcolsep}{3pt}
\renewcommand{\arraystretch}{1.0}

\begin{footnotesize}
\begin{tabular}{
>{\centering\arraybackslash}m{0.1\linewidth}
>{\centering\arraybackslash}m{0.2\linewidth}
>{\centering\arraybackslash}m{0.3\linewidth}
>{\centering\arraybackslash}m{0.3\linewidth}
}

\rotatebox{90}{\shortstack{Input\\Point Cloud}}
&
\includegraphics[width=0.8\linewidth,height=1.8cm,keepaspectratio]{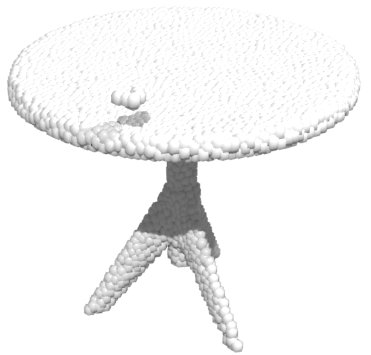}
&
\includegraphics[width=\linewidth,height=1.8cm,keepaspectratio]{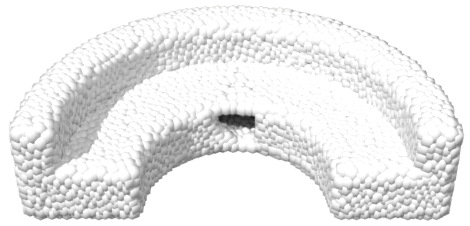}
&
\includegraphics[width=\linewidth,height=1.8cm,keepaspectratio]{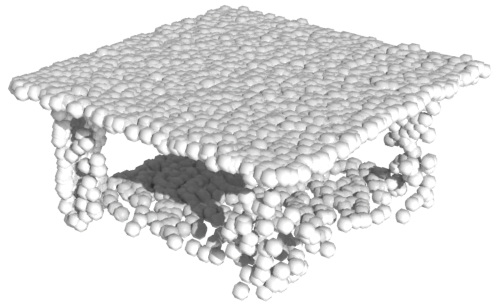}
\\
\rotatebox{90}{\shortstack{SuperDec\\~\cite{fedele2025superdec}}}
&
\includegraphics[width=0.8\linewidth,height=1.8cm,keepaspectratio]{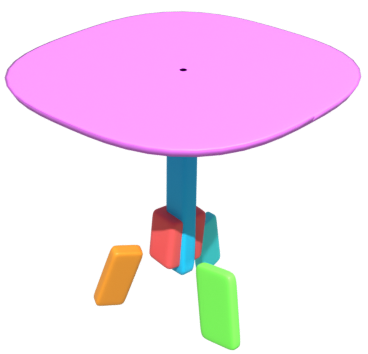}
&
\includegraphics[width=\linewidth,height=1.8cm,keepaspectratio]{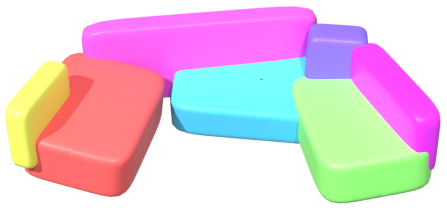}
&
\includegraphics[width=\linewidth,height=1.8cm,keepaspectratio]{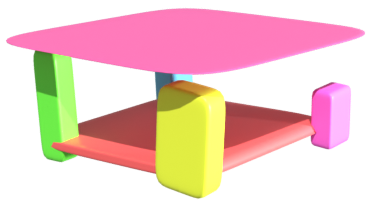}
\\
\rotatebox{90}{\shortstack{\textbf{\method{}}\\(Ours)}}
&
\includegraphics[width=0.8\linewidth,height=1.8cm,keepaspectratio]{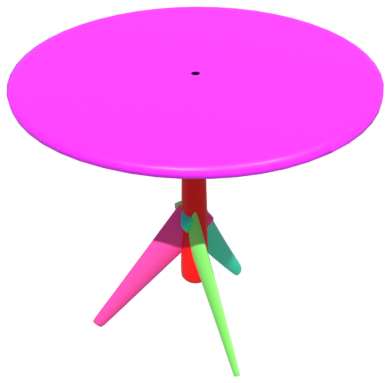}
&
\includegraphics[width=\linewidth,height=1.8cm,keepaspectratio]{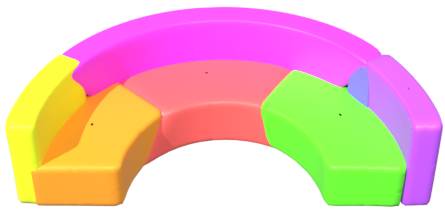}
&
\includegraphics[width=\linewidth,height=1.8cm,keepaspectratio]{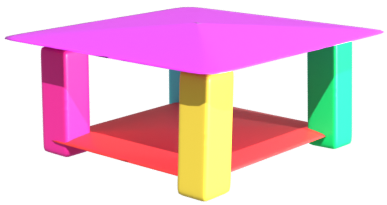}\\[-2pt]
&
\textit{Reconstruction}
&
\textit{Expressiveness}
&
\textit{Robustness}
\end{tabular}
\end{footnotesize}

\vspace{-2mm}
\caption{\textbf{3D Object Decomposition with Deformable Superquadrics.} Given a 3D point cloud of an object, our \method{} decomposes it into geometrically accurate \emph{superquadric} primitives.
Our loss significantly improves reconstruction quality \emph{(left)}, while \emph{bending} and \emph{tapering} further increase the expressiveness of superquadrics \emph{(center)}. We also show robustness to real-world partial point clouds \emph{(right)}.}
\label{fig:teaser}
\end{figure}

\section{Introduction}
\label{sec:intro}

Obtaining compact, interpretable, and geometrically meaningful representation of 3D objects remains a fundamental challenge in computer vision. While polygon meshes, point clouds, and implicit representations offer high geometric fidelity, they often lack the structural abstraction required for high-level reasoning in robotics and scene understanding. Superquadrics~\cite{Barr1981SuperquadricsAA} offer a compact alternative: they can represent a wide range of shapes, from ellipsoids to cylinders and boxes, using only a small set of explicit parameters.

\noindent Existing superquadric decomposition methods have traditionally been divided into learning- and optimization-based. Learning-based approaches are often limited to specific object categories~\cite{paschalidou2019superquadrics, Yang2021UnsupervisedLF}, while optimization-based methods~\cite{liu2023marching, liu2022robust} struggle to jointly address part segmentation and primitive fitting, making them sensitive to local minima and hyperparameter tuning. Recently, SuperDec~\cite{fedele2025superdec} demonstrated the potential of combining these paradigms by using a transformer-based model to segment inputs into parts and predict their corresponding superquadric parameters.

\noindent Despite these advances, fundamental limitations remain. First, SuperDec primarily relies on Chamfer-distance supervision, which can lead to gaps between primitives and suboptimal shape fitting. Second, it predicts only \textit{rigid} superquadrics, which are inherently convex and symmetric, limiting their ability to faithfully represent bent and tapered object parts. Finally, it is only trained to unsupervisely decompose complete point clouds, lacking robustness to occlusions and partial data common in real-world scenes.
To address these challenges, we introduce \method{}, a framework for deformable superquadrics decomposition and robust primitive prediction. \method{} as shown in Fig.~\ref{fig:teaser} improves both reconstruction fidelity and robustness to partial observations.

\noindent In summary, our contributions are the following:

\begin{itemize}
    \item \textbf{Improved feed-forward reconstruction.} We introduce a novel training loss that significantly improves the reconstruction accuracy of feed-forward superquadric decomposition networks.
    \item \textbf{Deformable superquadrics.} We integrate bending and tapering deformations into a learning-based superquadric decomposition, substantially increasing the expressive power of individual primitives while maintaining a compact parametric representation.
    \item \textbf{Robust feed-forward prediction.} We use our high-quality superquadric decompositions as supervision to train a robust predictor that learns to output complete primitive structure even from partial observations. 
\end{itemize}

\noindent We compare \method{} against existing methods on objects from ShapeNet, a standard object-level benchmark, and on the Aria Synthetic Environment for scene-level evaluation with partial point clouds. Our results demonstrate that deformable superquadrics provide substantially higher reconstruction accuracy while maintaining a similarly compact representation, and enable robust primitive prediction from partial observations.

\section{Related Work}
The task of decomposing complex 3D shapes into elementary primitives is a long-standing problem. It requires jointly solving two coupled tasks: (1) segmenting the input into parts that can each be represented by a single primitive, and (2) fitting a primitive to each part. Traditionally, this problem was formulated as a per-instance optimization problem and solved independently for every input shape. More recently, self-supervised learning-based approaches have shown that neural networks can jointly solve these tasks while learning reusable geometric representations across objects, leading to improved decomposition quality and significantly faster inference.

\subsubsection{Optimization-based methods}
originate from the superquadric fitting literature~\cite{jaklic2000segmentation}. Recent works have addressed several limitations of classical fitting algorithms. EMS~\cite{liu2022robust} introduced a probabilistic formulation that improves robustness to noise and outliers. However, to obtain multiple primitives, it recursively fits a dominant superquadric and subsequently processes the residual clusters, limiting its ability to represent objects with non-hierarchical structures. Marching Primitives~\cite{liu2023marching} and Light-SQ~\cite{wang2025light} instead leverage signed distance fields (SDFs) to improve fitting accuracy. While effective, their reliance on accurate SDFs limits their applicability to real-world settings, where only partial point clouds are typically available. More recently, SuperFrusta~\cite{ganeshan2026residual} extended this optimization paradigm to a more expressive primitive representation, enabling accurate reconstruction of a broader range of geometries through differentiable optimization. Despite their strong reconstruction accuracy, optimization-based methods cannot learn reusable shape priors, are susceptible to local minima, and require expensive optimization at inference~time.

\vspace{-10px}
\subsubsection{Learning-based methods} predict primitive parameters directly from point clouds~\cite{paschalidou2021neural}, images~\cite{paschalidou2020learning}, or rendered depth maps~\cite{genova2019learning}. They are typically trained in a self-supervised manner using reconstruction objectives. Early learning-based methods showed that, with appropriate reconstruction losses, neural networks can decompose shapes into a small set of primitives for specific object categories. Tulsiani \etal~\cite{tulsiani2017learning} introduced a CNN-based approach for cuboid decomposition, later extended to more expressive primitives such as superquadrics~\cite{paschalidou2019superquadrics}. CSA~\cite{Yang2021UnsupervisedLF} further improved interpretability by using stronger point encoders and jointly predicting primitive parameters and part segmentations. However, these approaches often rely on category-specific training and global shape features, which limits generalization to out-of-category objects and complex real-world scenes. SuperDec~\cite{fedele2025superdec} alleviates these issues with a category-agnostic set predictor and sparsity regularization. \method{} builds on this line by introducing deformable superquadrics and a novel loss function to significantly improve the reconstruction fidelity.

\newcommand{\existparm}{\textbf{e}}
\newcommand{\lsdf}{\loss_{\text{SDF}}}
\newcommand{\liou}{\loss_{\text{IoU}}}
\newcommand{\lbbox}{\loss_{\text{bbox}}}
\newcommand{\loverlap}{\loss_{\text{overlap}}}
\newcommand{\lreg}{\loss_{\text{reg}}}
\newcommand{\lsparsity}{\loss_{\text{sparsity}}}
\newcommand{\lexist}{\loss_{\text{exist}}}
\newcommand{\npoints}{N}
\newcommand{\nprimitives}{P}

\section{Preliminaries}

\noindent
\textbf{Superquadrics} are a family of shapes introduced by Barr \etal~\cite{Barr1981SuperquadricsAA} and widely adopted in computer vision, graphics, and robotics 
\cite{jaklic2000segmentation, paschalidou2019superquadrics, paschalidou2020learning, vezzani2017grasping, fedele2025spacecontrol, sella2026prox, monnier2023differentiable, zhao2026sq}. In canonical pose, a superquadric is fully described by five parameters: three scale values $\mathbf{s} = (s_x, s_y, s_z)$ and two shape parameters $\boldsymbol{\epsilon} \!=\!(\epsilon_1, \epsilon_2)$.
The implicit equation of a superquadric, defining its surface, is given by:
\begin{equation}\label{e-can-sq}
q_{\text{can}}(\mathbf{x}; \mathbf{s}, \boldsymbol{\epsilon}) :=
\left(
\left(\frac{x}{s_x}\right)^{\frac{2}{\epsilon_2}}
+
\left(\frac{y}{s_y}\right)^{\frac{2}{\epsilon_2}}
\right)^{\frac{\epsilon_2}{\epsilon_1}}
+
\left(\frac{z}{s_z}\right)^{\frac{2}{\epsilon_1}}
= 1 \, .
\end{equation}

\noindent Extending this representation to a global coordinate system requires an additional rigid transformation consisting of translation $\mathbf{t} \in \mathbb{R}^3$ and rotation $R \in SO(3)$, resulting in a total of 11 parameters per superquadric. The implicit representation of the superquadrics then becomes
$
q_{\text{can}}\big(R^{-1}(\mathbf{x}-\mathbf{t}); \mathbf{s}, \boldsymbol{\epsilon}\big)=1$.
While this formulation provides a compact representation, real-world objects frequently exhibit \textit{tapered} and \textit{bent} geometries. Next, we describe how to handle these cases by extending the canonical model $q_{\text{can}}$ with such deformations.

\begin{figure}[!b]
\centering
    \begin{overpic}[height=0.11\linewidth]{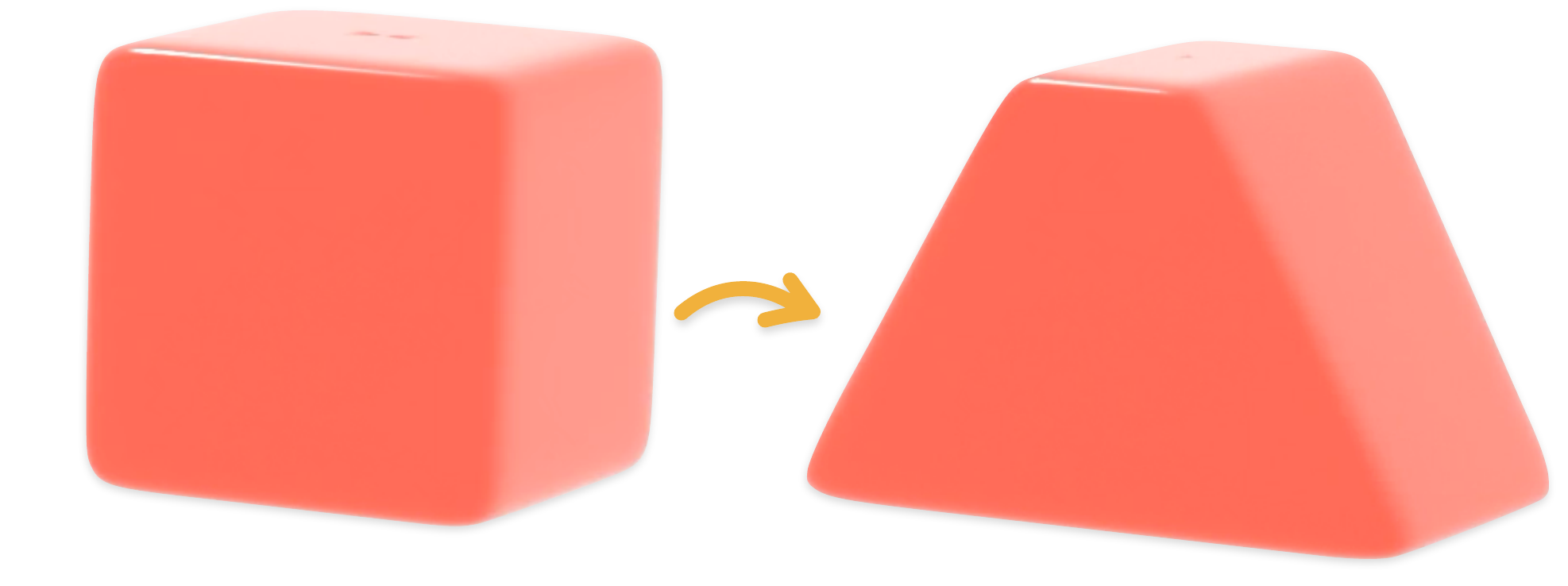}
        \put(45, 23){\footnotesize $\footnotesize \bm \tau$}
    \end{overpic}
\hspace{5px}
\raisebox{-0.015\linewidth}{\resizebox{0.145\linewidth}{!}{%
\begin{tikzpicture}[scale=0.5]

\definecolor{shapeFill}{RGB}{250,165,150}
\definecolor{shapeEdge}{RGB}{240,60,0}

\def\W{2.5}
\def\T{0.6}
\def\H{2.8}

\pgfmathsetmacro{\topW}{\W-\T*2}
\pgfmathsetmacro{\botW}{\W+\T*2}

\coordinate (TL) at (-\topW/2,\H);
\coordinate (TR) at ( \topW/2,\H);
\coordinate (BL) at (-\botW/2,0);
\coordinate (BR) at ( \botW/2,0);

\filldraw[
    fill=shapeFill,
    fill opacity=0.1,
    draw=shapeEdge,
    draw opacity=0.3,
    line width=1.3pt,
    rounded corners=3pt
]
(-\W/2, \H) --
(\W/2, \H) --
(\W/2, 0.0) --
(-\W/2, 0.0) -- cycle;

\filldraw[
    fill=shapeFill,
    fill opacity=0.7,
    draw=shapeEdge,
    line width=1.3pt,
    rounded corners=3pt
]
(TL) --
(TR) --
(BR) --
(BL) -- cycle;

\draw[black!80, line width=1.5pt, {Stealth[length=2.5mm]}-]
    ($(BR)+(.0,-0.0)$)
    --
    ($(BR)-(\T,-0.0)$)
    node[midway,below=0pt] {$\tau_x$};

\draw[black!80, line width=1.5pt, {Stealth[length=2.5mm]}-]
    ($(TR)+(.0,-0.0-0.0)$) --
    ($(TR)+(\T,-0.0-0.0)$)
    node[midway,above=0pt] {$\tau_x$};

\draw[densely dashed] ($(BR)-(\T,0)$) -- ++(0,\H);

\coordinate (C) at (0,\H/2);

\draw[->,black,thick] (C) -- ++(0, 0.75) node[left=3pt] {$z$};
\draw[->,black,thick] (C) -- ++(0.75, 0) node[below=3pt] {$x$};
\draw[very thin,fill=white] (C) circle (0.22) node[left=3pt] at (C) {$y$};
\node at (C) {$\otimes$};

\end{tikzpicture}
}}
\hspace{25px}
    \begin{overpic}[width=0.15\linewidth]{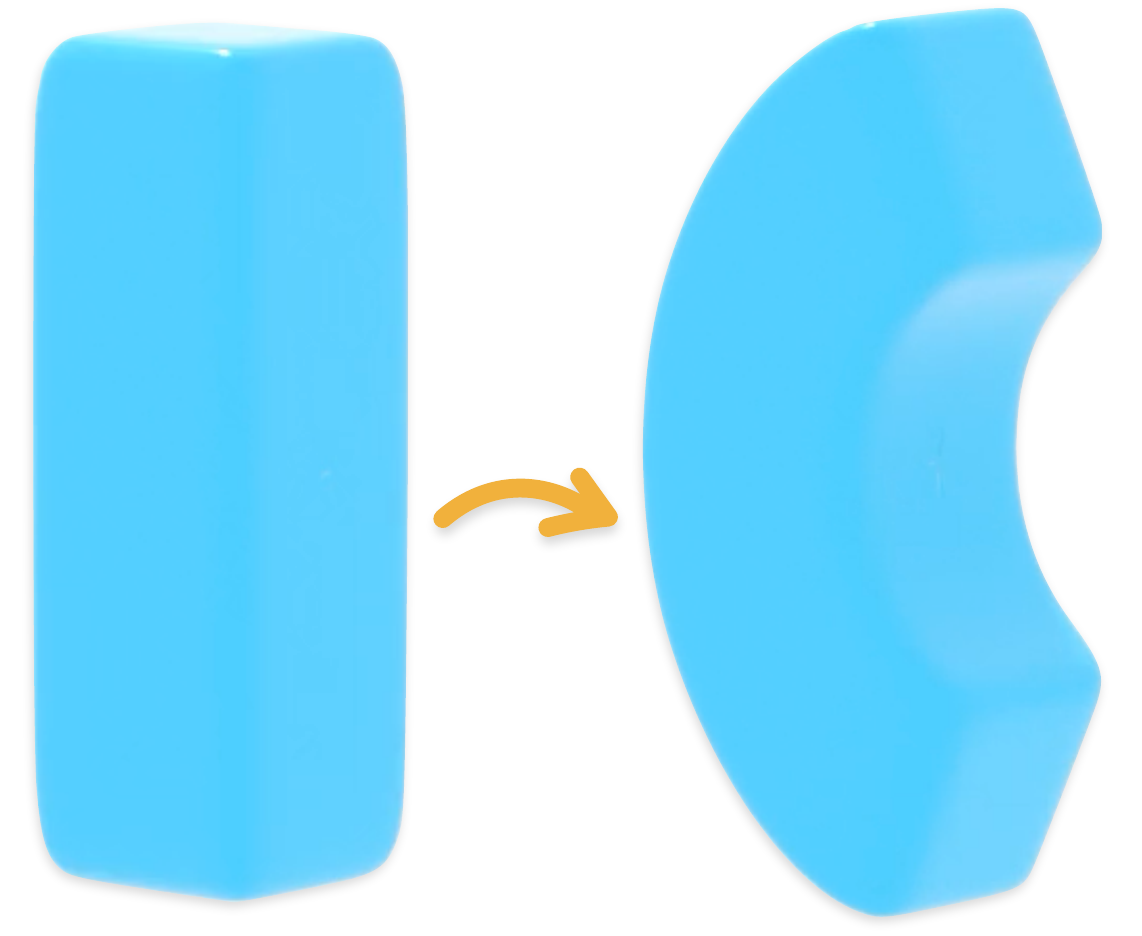}
        \put(38, 47){\footnotesize $\small \bm \beta_z$}
    \end{overpic}
\resizebox{0.27\linewidth}{!}{%
\begin{tikzpicture}[line cap=round, line join=round, scale=0.6]
  \definecolor{bandfill}{RGB}{206,228,244}
  \definecolor{bandline}{RGB}{83,160,210}
  \definecolor{centerdot}{RGB}{120,180,230}
  \pgfmathsetmacro{\Rc}{2.5}
  \pgfmathsetmacro{\halfw}{0.5}
  \pgfmathsetmacro{\Rout}{\Rc+\halfw}
  \pgfmathsetmacro{\Rin}{\Rc-\halfw}
  \pgfmathsetmacro{\archt}{3}
  \pgfmathsetmacro{\halftheta}{(\archt/\Rc)*180/pi/2}
  \pgfmathsetmacro{\phitop}{180-\halftheta}
  \pgfmathsetmacro{\phibot}{180+\halftheta}
  \pgfmathsetmacro{\extdeg}{8}
  \pgfmathsetmacro{\phitopext}{\phitop-\extdeg}
  \pgfmathsetmacro{\phibotext}{\phibot+\extdeg}
  \coordinate (C) at (9,3);
  \coordinate (O) at ($(C)+(180:\Rc)$);

  \coordinate (OXY) at (3.5,3);
  \pgfmathsetmacro{\Rxy}{1.3}
  \pgfmathsetmacro{\alphaz}{45}
 
  \draw[black!25, densely dashed, line width=1pt] (OXY) circle (\Rxy);

   \draw[bandline, fill=bandfill, fill opacity=0.2, draw opacity=0.8, line width=1.2pt, rounded corners=3pt]
    ($(OXY)+(-\halfw,-\halfw)$) rectangle ($(OXY)+(\halfw,\halfw)$);

  \node[circle, fill=black!70, inner sep=2pt] at (OXY) {};
 
  \draw[black, thick, ->] (OXY) -- ++(\Rxy+0.35,0) node[right] {$x$};
  \draw[black, thick, ->] (OXY) -- ++(0,\Rxy+0.35) node[above] {$y$};

  \draw[black!70, line width=1pt] ($(OXY)+(0:0.55)$) arc (0:\alphaz:0.55);
  \node[anchor=south west, text=black] at ($(OXY)+(0.5*\alphaz:0.85)+(-0.3,-0.3)$) {$\alpha_z$};
 
  \draw[black!70, line width=1.2pt, -{Stealth[length=3mm]}] (OXY) -- ++(\alphaz:\Rxy+0.1) node[above right] {\!\!\!\!$\alpha$\! axis};

 \draw[very thin,fill=white] (OXY) circle (0.12) node[below left=0pt] at (OXY) {$z$};
 \node at (OXY) {$\odot$};

  \draw[bandline, fill=bandfill, fill opacity=0.2, draw opacity=0.5, line width=1.2pt, rounded corners=4pt]
    ($(O)+(-\halfw,-\archt/2)$) rectangle ($(O)+(\halfw,\archt/2)$);

  \fill[bandfill, rounded corners=4pt, fill opacity=0.7]
    ($(C)+(\phibot:\Rout)$) arc (\phibot:\phitop:\Rout)
    -- ($(C)+(\phitop:\Rin)$) arc (\phitop:\phibot:\Rin)
    -- cycle;

  \draw[bandline!70, line width=1.2pt, rounded corners=4pt]
    ($(C)+(\phibot:\Rout)$) arc (\phibot:\phitop:\Rout)
    -- ($(C)+(\phitop:\Rin)$) arc (\phitop:\phibot:\Rin)
    -- cycle;

  \draw[centerdot, dotted, line width=1.2pt] ($(C)+(\phibotext:\Rc)$) arc (\phibotext:\phitopext:\Rc);

  \node[circle, fill=black!70, inner sep=2pt] (origin) at (C) {};
\draw[black!70, line width=1.2pt, -{Stealth[length=3mm]}]
  (C) -- ($(C)+(\phitop:\Rc)$);
\node[anchor=south west] at ($(C)+0.45*(\phitop:\Rc)+(0.15,0.05)$) {$\dfrac{1}{k_z}$};

  \draw[->,black,thick] (O) -- ++(0, 0.75) node[left=0pt] {$z$};
  \draw[->,black,thick] (O) -- ++(0.75, 0) node[below=0pt] {$\alpha$};

\end{tikzpicture}
}

\caption{Illustration of superquadric deformations: \emph{tapering (red)} and \emph{bending (blue)}. Both deformations are shown along a single axis (the $z$-axis).
For tapering along $z$, the parameters are $\tau_x$ (shown) and $\tau_y$ (not shown).
For bending, we select a bending angle $\alpha_z$ around the $z$ axis, and bend to the curvature defined by $k_z$.}
\label{f-tapering-bending}
\end{figure}
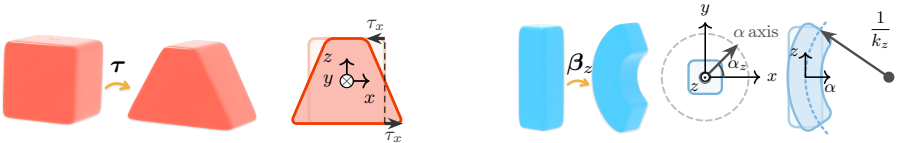

\subsubsection{Parametric Deformations.}
The expressiveness of superquadrics can be further extended by introducing global shape deformations \cite{barr1984global, solina1990recovery, jaklic2000segmentation}.
A deformation $\mathbf{D}$ explicitly maps surface points $\mathbf{x}$ of the undeformed shape to their deformed counterparts $\mathbf{X}$, i.e., $\mathbf{X} = \mathbf{D}(\mathbf{x})$.
In this work, we consider two such deformations, \emph{tapering} $\bm{\tau}$ and \emph{bending} $\bm{\beta}$ (see Fig.~\ref{f-tapering-bending}).
Tapering progressively narrows or widens a shape along a specified axis, while bending warps a straight line into a circular arc, whose curvature is defined by $k$.
The bending orientation is determined by a specified axis and an angle $\alpha$.
Further details are provided in Appendix~\ref{sec:sq-supp}.

\noindent In this work, we found single-axis tapering to be sufficient, using $\bm{\tau}=(\tau_x,\tau_y)$ to control tapering along the $x$- and $y$-directions with respect to the $z$-axis.
For bending, we use all three axes, leading to $\bm{\beta}=(\bm{\beta}_x,\bm{\beta}_y,\bm{\beta}_z)$, which comprises the parameter pairs $(k_x,\alpha_x)$, $(k_y,\alpha_y)$, and $(k_z,\alpha_z)$.
In total, the deformation model introduces eight additional parameters: two for tapering and six for bending.

\noindent Given a 3D point $\mathbf{x}$, the implicit function of a \textit{deformable superquadric}, parameterized by the 19 coefficients 
$\Theta=\{\mathbf{t}, R, \mathbf{s}, \boldsymbol{\epsilon}, \bm{\tau}, \bm{\beta}\}$, 
can be evaluated by first applying to $\mathbf{x}$ the inverse rigid transformation, followed by inverse bending with respect to the $z$-, $x$-, and $y$-axes and then inverse tapering. We denote this composition of inverse deformation operators by $\mathbf{D}^{-1}_{\bm{\tau},\bm{\beta}}$. Finally for all the surface points $\mathbf{x}$ of a deformed superquadric, the following equations holds true:
\begin{align}\label{e-deformable-superquadrics}
q(\mathbf{x}; \Theta)
=
q_{\text{can}}\Big( \mathbf{D}^{-1}_{\bm{\tau},\bm{\beta}}\big(R^{-1}(\mathbf{x}-\mathbf{t})\big); \mathbf{s}, \boldsymbol{\epsilon}\Big)\, = 1. 
\end{align}

\subsubsection{Radial Distance.} To fit superquadrics to observed surface points, we need a distance from a point $\mathbf{x}$ to the superquadric surface. The true Euclidean signed distance function (SDF) has no closed-form solution for superquadrics and is expensive to compute during training.
Instead, we use the \textit{radial distance}: the distance from $\mathbf{x}$ to where the ray from the superquadric's center through $\mathbf{x}$ hits the surface, which can be computed explicitly as follows~\cite{van1998fitting}:

\begin{align}\label{e-radial-distance}
d(\mathbf{x};\Theta) = \|R^{-1}(\mathbf{x}-\mathbf{t}) \| \left(1 -
q(\mathbf{x}; \Theta)^{-\frac{\epsilon_1}{2}}
\right)\, .
\end{align}

\section{Method}
We aim to decompose object point clouds into a set of deformable superquadric primitives. To achieve this, we first present a novel reconstruction loss, which we use both to supervise a feed-forward model (Sec.~\ref{sec:train}) and to refine its predictions through test-time optimization (Sec.~\ref{sec:optim}). However, real-world point clouds are rarely complete: sensor noise and occlusions typically result in partial observations. To address this, we show how the high-quality decompositions obtained on full point clouds can serve as supervision to train a model that is robust to partial, real-world point clouds (Sec.~\ref{sec:occlusion-robustness}).

\subsection{Feed-forward Neural Network}\label{sec:train}

\subsubsection{Model architecture.}
\begin{figure}[t]
    \centering
    \vspace{15px}
    \includegraphics[width=1\linewidth]{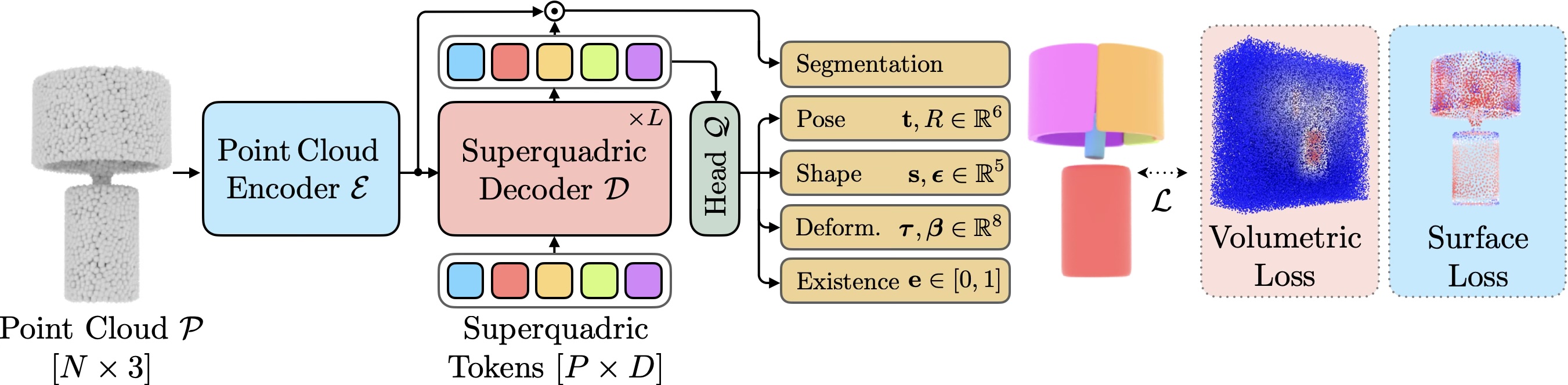}
    \caption{\textbf{Illustration of the \method{} model.}
Given an input point cloud, \name{} decomposes the object into a set of superquadric primitives, each defined by its pose $(\mathbf{t}, R)$, shape $(\mathbf{s}, \bm{\epsilon})$, and deformations $(\bm{\tau}, \bm{\beta})$.
The model is trained via self-supervised joint volumetric and surface losses. 
A subsequent (optional) object-specific optimization can further improve the superquadric decomposition quality using the same losses. 
    }
    \label{fig:overview}
\end{figure}

Fig.~\ref{fig:overview} illustrates the \method{} model, a Transformer-based architecture closely following the design of SuperDec~\cite{fedele2025superdec}.
Given an input point cloud $\mathcal{P}\,\in \mathbb{R}^{\npoints \times 3}$, we first extract point features $\mathcal{F}_{\text{PC}} \in \mathbb{R}^{\npoints \times D}$ using a point cloud encoder $\mathcal{E}$ (PVCNN~\cite{liu2019point}). In parallel, we initialize $\nprimitives$ superquadric tokens $\mathcal{F}_{\text{SQ}} \in \mathbb{R}^{\nprimitives \times D}$ using sinusoidal positional encodings. A transformer decoder $\mathcal{D}$ iteratively refines these queries through self-attention among the superquadric tokens $\mathcal{F}_{\text{SQ}}$, cross-attention with the point features $\mathcal{F}_{\text{PC}}$, and feed-forward layers for $L$ times. 
A \textit{regression head} $\mathscr{Q}$ maps the refined tokens to the parameters $\Theta$ of $\nprimitives$ deformable superquadrics and their existence probabilities $\existparm$:
\begin{align}\label{e-model}
f(\mathcal{P}) =
\mathscr{Q}\big(\mathcal{D}(\mathcal{F}_{\text{SQ}}, \mathcal{F}_{\text{PC}})\big)
=
\{(\Theta_p,\existparm_p)\}_{p=1}^{\nprimitives},
\end{align}
where $\Theta_p\in\mathbb{R}^{19}$ denotes the geometric parameters of the $p$-th deformable superquadric and $\existparm_p\in[0,1]$ denotes its existence probability.

\noindent Finally, a \textit{segmentation head} predicts a soft assignment matrix $W \in \mathbb{R}^{N \times P}$ as
$W = \sigma\big(\mathcal{F}_{\text{PC}}\cdot \phi(\mathcal{F}_{\text{SQ}})^\top\big)$,
where $\phi(\mathcal{F}_{\text{SQ}}) \in \mathbb{R}^{P\times D}$ is a learned projection of the refined superquadric queries and $\sigma$ is the softmax which for each point defines a probability distribution over the $P$ superquadrics.

\subsubsection{Loss function.}\label{sec:loss}
To train the model $f$, recent methods such as SuperDec~\cite{fedele2025superdec} rely primarily on the Chamfer distance between the input point cloud and points sampled from the superquadric surfaces. Although effective for coarse alignment, this loss biases the optimization towards overly rounded primitives, limiting the model's ability to capture sharp geometric details.

To address this limitation, we introduce an alternative loss that combines a differentiable Intersection-over-Union loss $\liou$ with a signed distance function (SDF) loss $\lsdf$. The former encourages the union of superquadrics to better align with the overall shape, resulting in fewer holes in the reconstruction. The latter provides geometry-aware gradients within a narrow band around the surface, enabling the recovery of fine details.
We also add the regularization term $\lreg$, described below, and define the overall training objective as:
\begin{align}
\loss = \lambda_{\text{IoU}}\liou + \lambda_{\text{SDF}}\lsdf + \lreg \,.
\end{align}

\paragraph{The Volumetric Loss $\liou$}
 encourages the predicted set of superquadrics to accurately reconstruct the target geometry. For a point $\mathbf{x}\in\R^3$, we define the occupancy of each predicted superquadric $\Theta_p$ using a soft indicator function~\cite{deng2020cvxnet}:
\begin{equation}\label{e-sdf-sq}
\mathcal{I}_p(\mathbf{x}) =
\sigma\left(-\frac{d(\mathbf{x};\Theta_p)}{\tau_{\text{IoU}}}\right),
\end{equation}
where $d(\mathbf{x};\Theta_p)$ is the radial distance to the $p$-th superquadric surface (Eq.~\ref{e-radial-distance}), $\sigma$ is the sigmoid function, and $\tau_{\text{IoU}}$ is a temperature parameter.
Thus, the global occupancy field associated with the predicted $\nprimitives$ primitives $f(\mathcal{P})$ is defined by:
\begin{align}\label{e-occupancy}
\hat{\mathcal{O}}(\mathbf{x}) = 1 - \prod_{p=1}^\nprimitives \bigl(1 - \existparm_p\cdot\mathcal{I}_p(\mathbf{x})\bigr) \,.
\end{align}

The final $\liou$ loss penalizes the discrepancy between the predicted occupancy $\hat{\mathcal{O}}(\mathbf{x})$ and the ground-truth occupancy $\mathcal{O}(\mathbf{x})$:
\begin{equation}\label{e-iou-loss}
\liou = -\log \left(
\frac{\sum_{\mathbf{x}\in\mathcal{V}} \hat{\mathcal{O}}(\mathbf{x}) \;\mathcal{O}(\mathbf{x})}
{\sum_{\mathbf{x}\in\mathcal{V}} \left(\hat{\mathcal{O}}(\mathbf{x}) + \mathcal{O}(\mathbf{x}) - \hat{\mathcal{O}}(\mathbf{x})\;\mathcal{O}(\mathbf{x})\right)}
\right)\, .
\end{equation}
where $\mathcal{V}$ is a set of points sampled uniformly in a unit cube containing the object.

\paragraph{The Surface Loss $\lsdf$} additionally encourages local surface alignment by minimizing the distance between the input point cloud $\mathcal{P}=\{\mathbf{x}_i\}_{i=1}^{\npoints}$ and the surfaces of the $\nprimitives$ predicted superquadrics:
\begin{equation}\label{e-loss-sdf}
\mathcal{L}_{\text{SDF}}
=
\frac{1}{\npoints} \sum_{i=1}^{\npoints}
\sum_{p=1}^{\nprimitives}
w_{ip} \,
\psi\Big(d(\mathbf{x}_i; \Theta_p)\Big)\, .
\end{equation}
Here, $w_{ip}$ is the predicted segmentation assignment weight that controls the influence of primitive $p$ on point $\mathbf{x}_i$. To improve stability and robustness to outliers, we pass the distances $d(\mathbf{x}_i; \Theta_p)$ through an activation function $\psi$, described in more detail in Appendix~\ref{sdf-activation}.

\paragraph{Regularization Losses.}
We further regularize the primitive prediction with
\begin{align}
\lreg = \lambda_o\loverlap + \lambda_s\lsparsity + \lambda_e\lexist \, .
\end{align}
\noindent The overlap loss $\loverlap$ reduces spatial redundancy by penalizing regions where multiple primitives explain the same volume. Following~\cite{deng2020cvxnet}, we aggregate the occupancy indicators $\mathcal{I}_p(\mathbf{x})$ and penalize values above one:
\begin{equation}
\loverlap
=
\mathbb{E}_{\mathbf{x}}
\left[
\mathrm{ReLU}\!\left(
\sum_{p=1}^{\nprimitives} \mathcal{I}_p(\mathbf{x}) - 1
\right)
\right]\, .
\end{equation}

\noindent For compactness and primitive selection, we use the same sparsity and existence regularizers as SuperDec~\cite{fedele2025superdec}. Specifically, $\lsparsity$ penalizes the $0.5$-norm of the primitive usage
$m_p \!=\! \frac{1}{\npoints}\sum_{i=1}^{\npoints} w_{ip}$,
encouraging parsimonious decompositions. The existence loss $\lexist$ supervises the predicted existence probability $\existparm_p$ using binary cross-entropy, with a self-generated target $\hat{\existparm}_p = \mathbb{I}(m_p > \epsilon_{\existparm})$ obtained by thresholding the usage metric.
This keeps existence predictions consistent with each primitive's actual geometric contribution.

\subsection{Test-time Optimization}\label{sec:optim}
To further refine individual object reconstructions, we perform additional test-time optimization of the primitive parameters $\Theta$. This refinement uses the same reconstruction loss as at training time, but differs in two aspects: (1)~we only optimize primitives predicted as existent, and (2)~we no longer rely on the assignment matrix for SDF weighting. We detail these two modifications below.

\subsubsection{Active Primitive Selection.} 
To simplify the optimization landscape, we treat the existence probabilities as a binary mask. We select a set of active primitives $\mathcal{M}_{\text{active}}$ by thresholding the network's predictions at $\existparm_p > 0.5$. During this refinement phase  we only optimize the geometric parameters $\Theta_p$ of the active set.

\subsubsection{Global Distance and Union Representation.} Unlike the training phase, where the model relies on a predicted assignment matrix $w_k(\mathbf{x})$ for $\mathcal{L}_{\text{SDF}}$, the test-time optimization treats the primitives as a single global shape. To improve the gradient flow during optimization, we replace the hard minimum with a LogSumExp approximation~\cite{deng2020cvxnet}. To evaluate the volumetric consistency $\liou$ during optimization, we bypass the probabilistic aggregation of individual primitives. Instead, we compute a single global occupancy probability $\hat{\mathcal{O}}(\mathbf{x})$ by directly applying the soft indicator function to the unified distance field. This formulation ensures that the IoU penalty is driven purely by the boundary of the active set $\mathcal{M}_{\text{active}}$, maintaining differentiability while approximating the geometric union of the primitives. The effect of test-time optimization is shown in Tab.~\ref{tab:opti}.

\subsection{Robustness to Occlusions} \label{sec:occlusion-robustness}
Up to this point, we assumed complete object point clouds. In practice, however, real-world scans are typically noisy and partial due to occlusions. To enable structural completion from partial inputs, we use the decompositions predicted by our test-time-optimized model on complete point clouds (Sec.~\ref{sec:optim}) as pseudo-ground truth, and finetune \method{} to predict these same decompositions from partially observed versions of the same point clouds.

\paragraph{Training Objective.}
In a deterministic setting, direct regression of superquadric parameters is ill-posed due to the inherent symmetries of the representation (\eg{}, axis permutations and sign flips), which produce conflicting gradients during training. To address this, we frame the task as a set-prediction problem. We employ \textit{Hungarian matching} to find the optimal one-to-one assignment between the predicted set of superquadrics $\{\hat{\Theta}_p\}$ and the ground truth set $\{\Theta_p\}$. After matching, we minimize a geometric loss based on the Chamfer Distance (CD) \cite{brazil2023omni3dlargebenchmarkmodel} only for the matched pairs where the ground-truth primitive exists:
\begin{equation}
\mathcal{L}_{\text{geom}} = \sum_{p=1}^M \text{CD}\left( \mathcal{S}(\hat{\Theta}_p), \mathcal{S}(\Theta_p) \right)\, ,
\label{eq:geom-cd}
\end{equation}
where $\mathcal{S}(\Theta)$ denotes a set of points sampled from the surface of superquadric~$\Theta$. 
This formulation makes the loss invariant to parametric permutations and directly emphasizes surface alignment. To regularize training, we also include the volumetric loss $\liou$ (Eq.~\ref{e-iou-loss}), computed from the full-object occupancy. Furthermore, we introduce an auxiliary geometric loss $\mathcal{L}_{\text{rigid}}$ which is computed as in Eq.~\ref{eq:geom-cd}, but without considering tapering and bending deformations. This acts as a shape prior, encouraging the model to first capture the coarse rigid structure before refining the complex deformations. The final loss is therefore defined as $\mathcal{L}_{\text{sup}} = \mathcal{L}_{\text{geom}} +\lambda_{\text{IoU}} \liou + \mathcal{L}_{\text{rigid}}$.

\section{Evaluation}
We first compare \method{} with previous learning- and optimization-based approaches for single-object decomposition (Sec.~\ref{sec:sota-comparison}). Then, starting from the predictions of \method{}, we evaluate the effect of additional test-time optimization (Sec.~\ref{sec:optimization}). Finally, we demonstrate the robustness of our finetuned variant of \method{} on real-world point clouds (Sec.~\ref{sec:occlusion-evaluation}).

\subsection{Comparison with State-of-the-art Methods}\label{sec:sota-comparison}
\subsubsection{Dataset.}
We evaluate on the ShapeNet dataset~\cite{chang2015shapenet} which is widely used in prior primitive-decomposition work, enabling direct comparison with existing baselines. We use the 13 classes and train/val/test splits defined by Choy \etal{}~\cite{choy20163d}, and for each object, we sample 4096 points using Farthest Point Sampling (FPS)~\cite{qi2017pointnet++}. All objects are pre-aligned to a canonical orientation.

\subsubsection{Methods in Comparison.}
We compare \method{} against several state-of-the-art primitive-based decomposition methods. Among the \textit{learning-based} approaches, we consider SQ~\cite{paschalidou2019superquadrics}, the seminal method for neural superquadric prediction; CSA~\cite{Yang2021UnsupervisedLF}, which represents objects using cuboids with a consistency-aware assignment; and SuperDec~\cite{fedele2025superdec}, a recent feed-forward approach for superquadric decomposition. These methods achieve fast inference but often struggle to accurately capture complex local geometry. We also compare against \textit{aoptimization-based} methods, including EMS~\cite{liu2022robust}, which focuses on robust primitive estimation, and Marching Primitives~\cite{liu2023marching}, which performs more extensive optimization to achieve high IoU at the expense of substantially higher runtime and a larger number of primitives. We evaluate \method{} in two variants: a rigid version using standard superquadrics and the full version, which additionally incorporates tapering and bending deformations.

\subsubsection{Training Details.}
Following SuperDec, all learning-based methods are trained jointly on the 13 ShapeNet categories. We train our network for $1000$ epochs starting from SuperDec's ShapeNet checkpoint, with hyperparameters $P = 16$, $N = 4096$, $D=128$, $L=3$, 
$\epsilon_{\existparm} = 24$,
$\tau_{\text{IoU}}=0.001$, $\lambda_{\text{IoU}}=0.2$, $\lambda_{\text{SDF}}=3.2$, $\lambda_{o}=5$, $\lambda_{s} = 1.26$, $\lambda_{e} = 0.01$. We use a learning rate of $3 \times 10^{-4}$ for standard model training, and reduce it to $1 \times 10^{-4}$ for the robust fine-tuning. 
The model is trained on 4 GPUs on an NVIDIA GH200 GPU Node with a batch size of 32.

\subsubsection{Metrics.}
We compute different metrics to compare \textit{reconstruction accuracy}, \textit{compactness}, and \textit{speed} of different methods. We evaluate reconstruction accuracy using L1 and L2 Chamfer distances, Intersection over Union (IoU), and F-score~\cite{TankAndTamples}. We evaluate compactness in terms of average number of active primitives (\# Prim.).

\begin{table}[t]
    \centering
    \resizebox{\linewidth}{!}{%
    \begin{tabular}{llc| cccccc}
        \toprule
        \textbf{Method}  & \textbf{Prim. Type} & \textbf{T\&B} &\textbf{IoU}$\uparrow$ & \textbf{F}~\cite{TankAndTamples}$\uparrow$ & \textbf{L1}$\downarrow$ & \textbf{L2}$\downarrow$ & \textbf{\#\,Prim}.$\downarrow$ & \textbf{Runtime}$\downarrow$ \\
        \midrule
        SQ~\cite{paschalidou2019superquadrics}  & Superquadrics&\xmark &$0.27$& $0.13$ & $3.67$ & $0.29$ & $10$& $0.0065$ s \\
        CSA~\cite{Yang2021UnsupervisedLF}  & Cuboids&\xmark &$0.42$& $0.14$& $3.52$& $0.30$& $7.85$& $\mathbf{0.004}$ s\\
        EMS~\cite{liu2022robust} & Superquadrics &\xmark & $0.40$ & $0.29$ & $4.40$ & $1.06$& $5.67$ & $7.82$ s\\
        March.Prim.~\cite{liu2023marching} & Superquadrics &\xmark& $0.56$ & $0.18$ & $2.08$ & $0.10$& $24.10$ & $163.41$ s\\
        SuperDec \cite{fedele2025superdec}  & Superquadrics&\xmark & $0.59$ & $0.28$ & $1.77$ & $0.050$ & $5.79$ & $0.0075$ s\\ 
        \method{} (ours) & Superquadrics&\xmark                    
        & $0.70$ & $0.35$ & $1.75$ & $0.049$ & $6.04$ & $0.0075$ s \\
        \method{} (ours) & Superquadrics & \cmark 
        & $\mathbf{0.72}$ & $\mathbf{0.37}$ & $\mathbf{1.54}$ & $\mathbf{0.043}$ & $\textbf{5.64}$ & $0.0082$ s\\
        \bottomrule
    \end{tabular}
    }
    \vspace{5px}
    \caption{\textbf{Quantitative results on ShapeNet.} We compare our model to other baselines approaches on ShapeNet. T\&B indicates whether tapering and bending deformations are enabled. L1 and L2 scores are multiplied by $10^2$.}
    \label{tab:main}
\end{table}

\subsubsection{Quantitative Evaluation.}
We report quantitative results in Tab.~\ref{tab:main}. \method{} significantly outperforms both learning-based and optimization-based methods, particularly in terms of IoU. The only method that achieves comparable IoU to our base model is Marching Primitives~\cite{liu2023marching}. However, it requires over $4\times$ more primitives and is approximately $10{,}000\times$ slower. We consider IoU to be the most representative metric for evaluating primitive-based decompositions, as it directly measures how well the union of the predicted primitives matches the occupancy of the ground truth shape. %

\begin{figure}[t]
\centering
\begin{small}
\rotatebox{90}{\footnotesize \shortstack{Input \\ Point Cloud}}
\includegraphics[height=1.7cm,trim={200px 100px 190px 130px},clip]{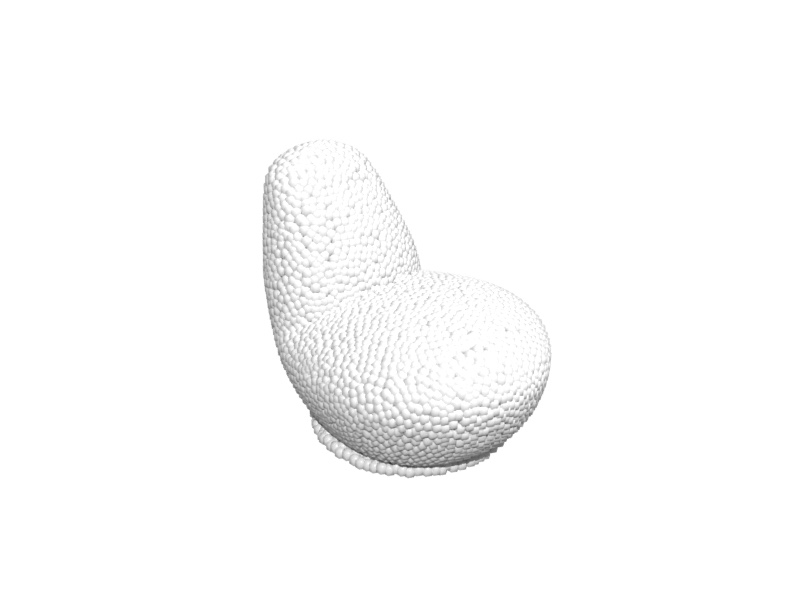}\hspace{-9px}
\includegraphics[height=1.7cm,trim={120px 70px 220px 80px},clip]{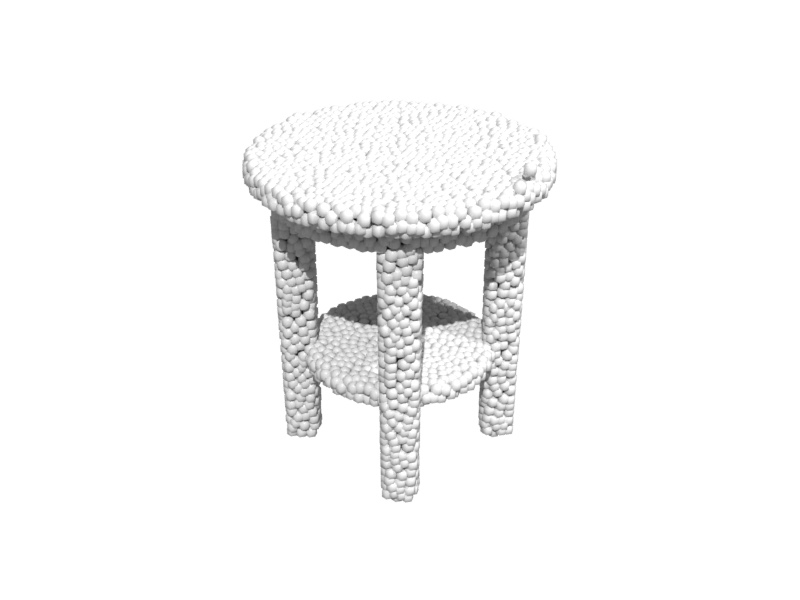}
\includegraphics[height=1.7cm,trim={220px 120px 220px 120px},clip]{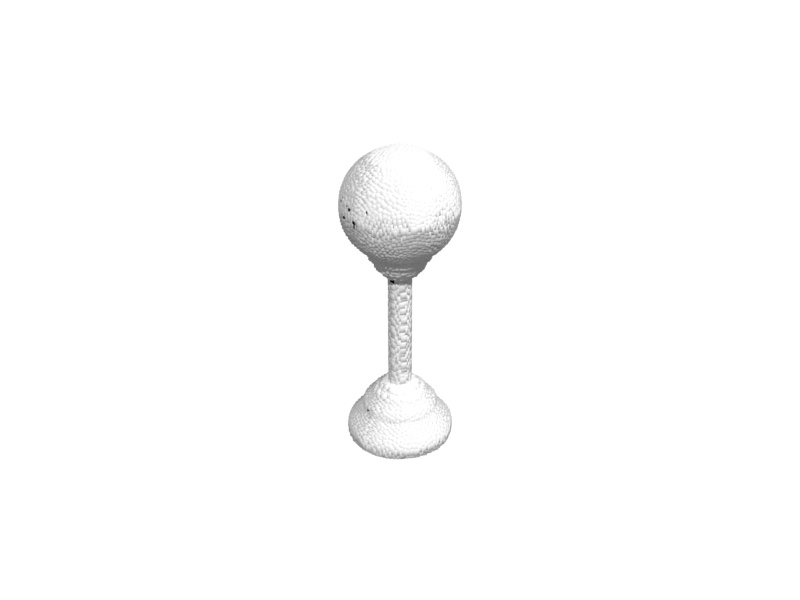}
\includegraphics[height=1.7cm,trim={220px 120px 220px 90px},clip]{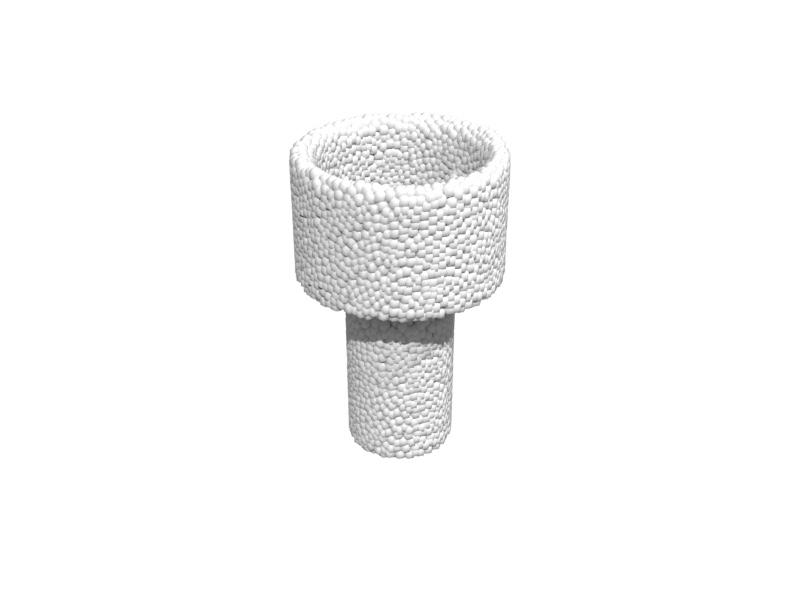}
\includegraphics[height=1.7cm,trim={260px 120px 230px 210px},clip]{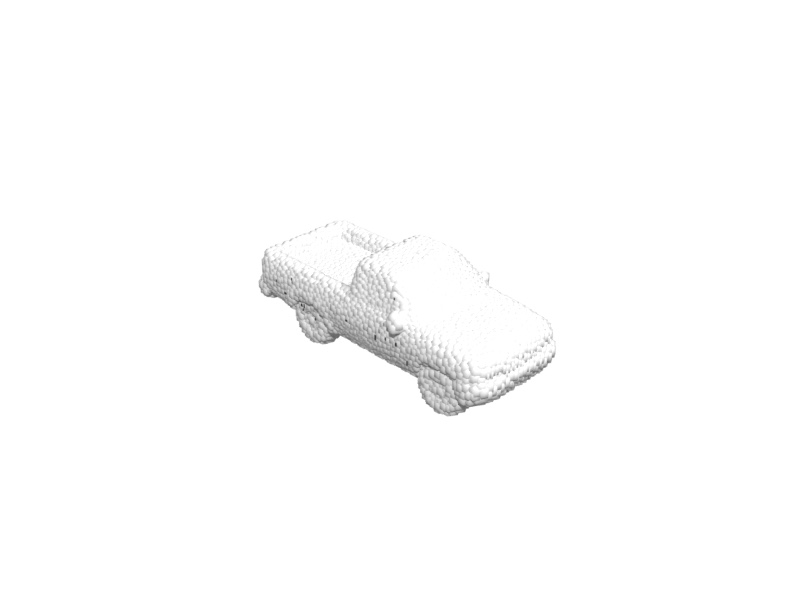}
\includegraphics[height=1.7cm,trim={240px 150px 260px 220px},clip]{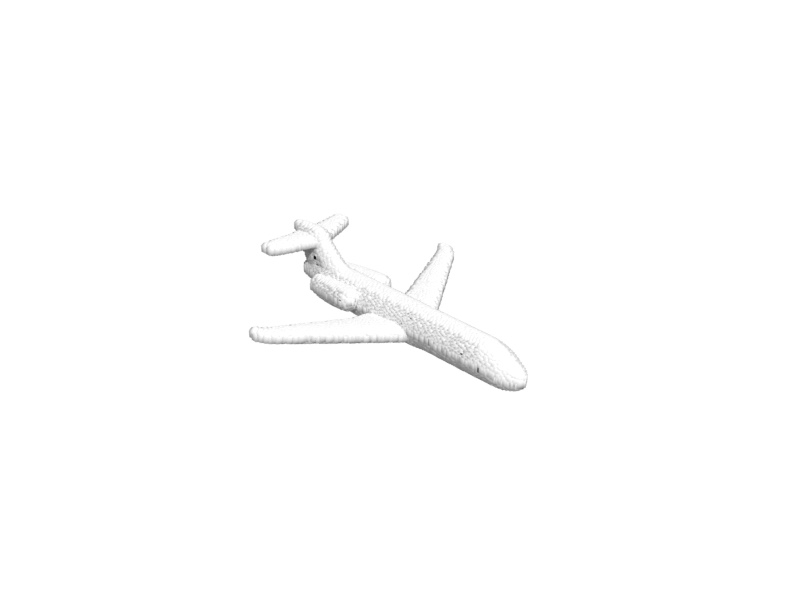}\\

\rotatebox{90}{\footnotesize \shortstack{Marching \\ Prim.\cite{liu2023marching}}}
\includegraphics[height=1.7cm,trim={200px 100px 190px 130px},clip]{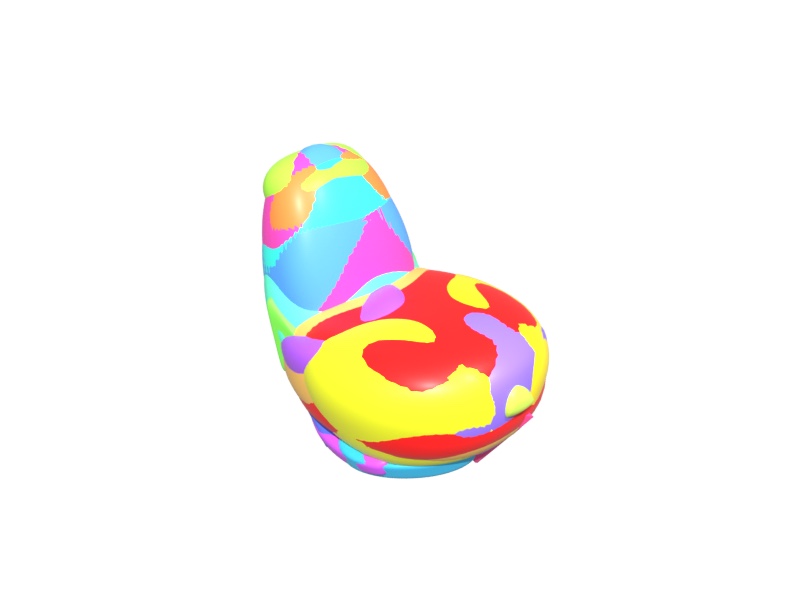}\hspace{-9px}
\includegraphics[height=1.7cm,trim={120px 70px 220px 80px},clip]{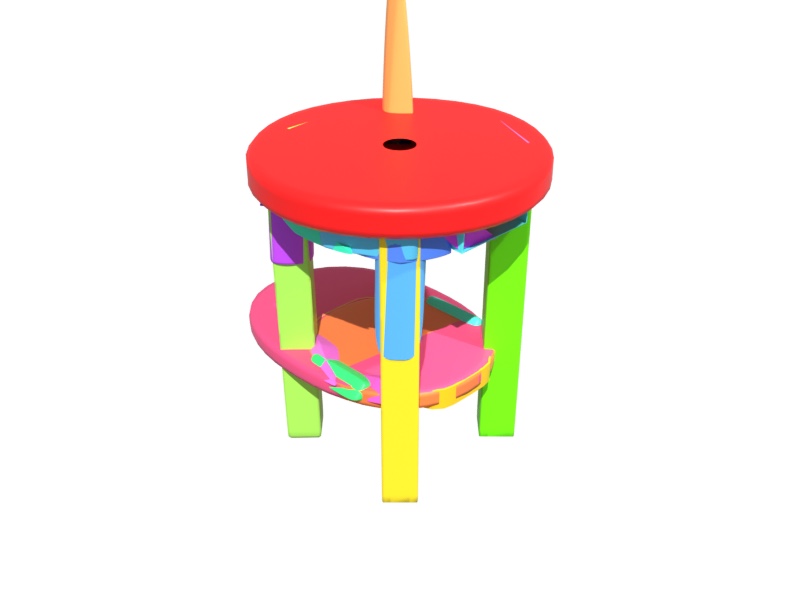}
\includegraphics[height=1.7cm,trim={220px 120px 220px 120px},clip]{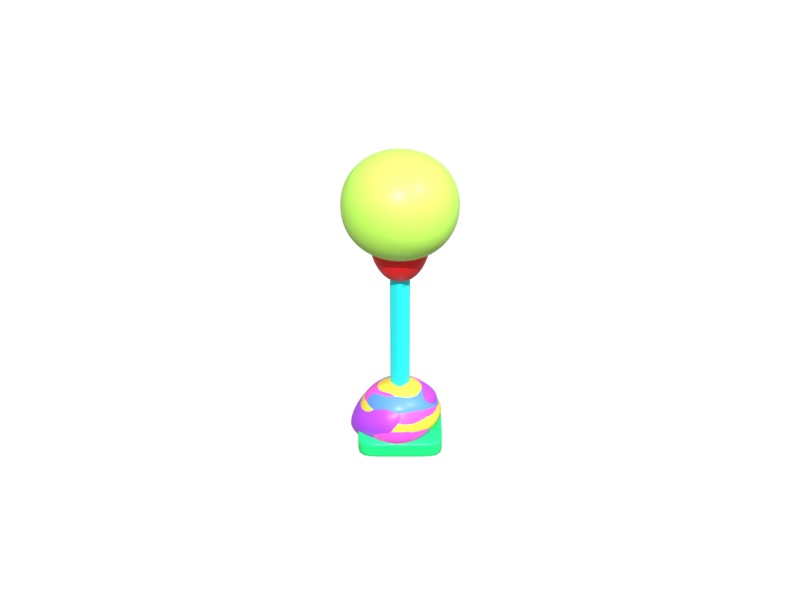}
\includegraphics[height=1.7cm,trim={220px 120px 220px 90px},clip]{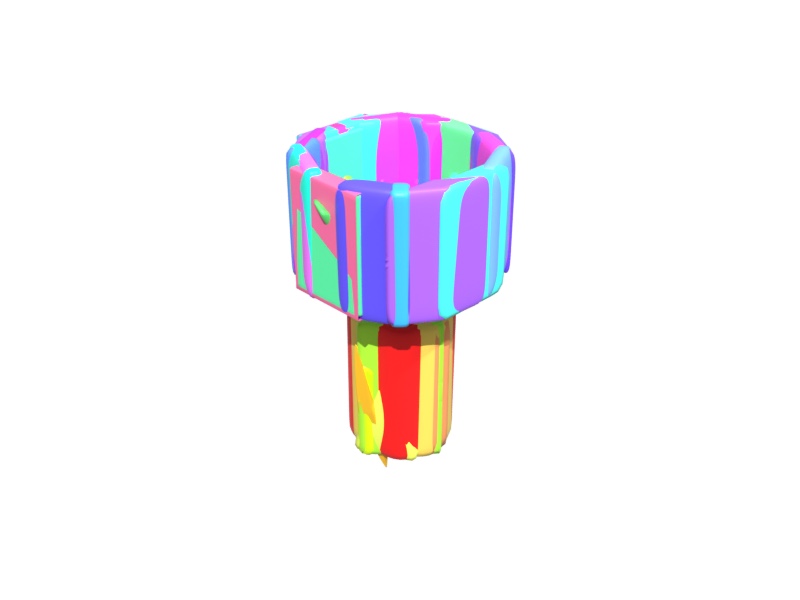}
\includegraphics[height=1.7cm,trim={260px 120px 230px 210px},clip]{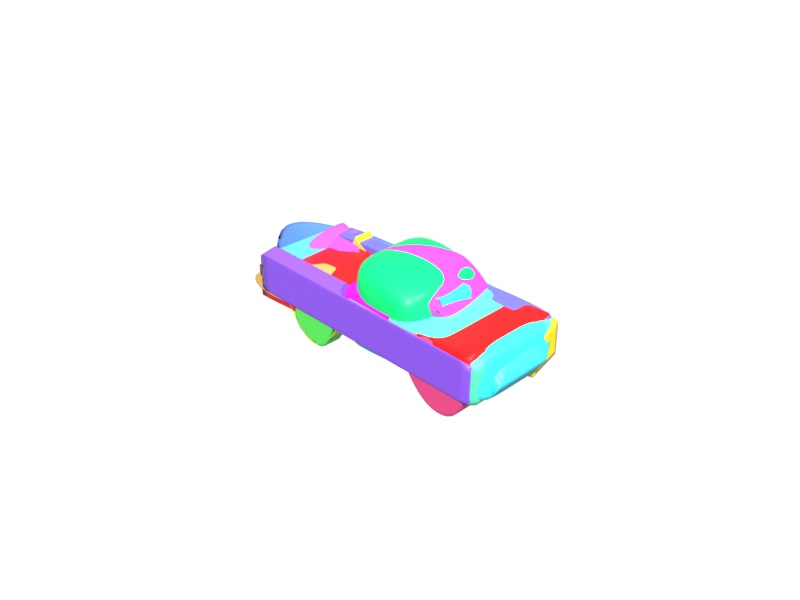}
\includegraphics[height=1.7cm,trim={240px 150px 260px 220px},clip]{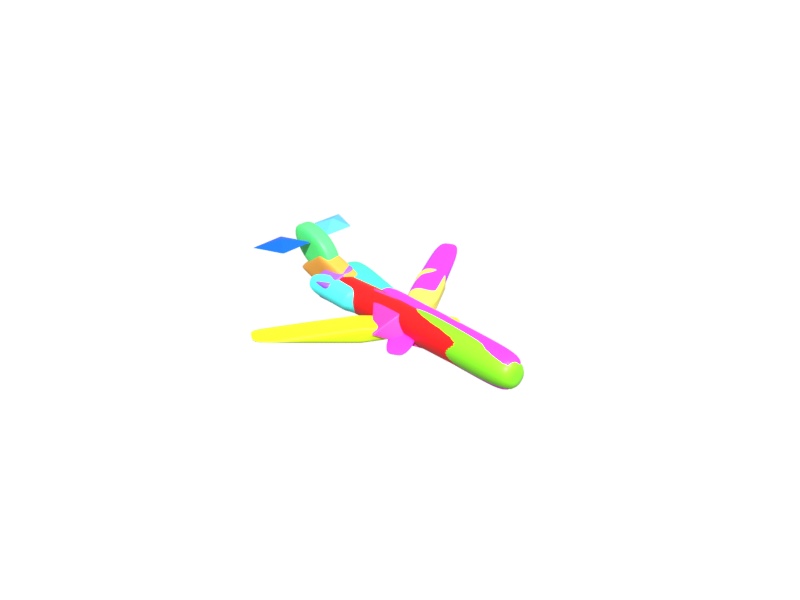}\\
\rotatebox{90}{\footnotesize \shortstack{ \hspace{2px}SuperDec\\ \cite{fedele2025superdec}}}\hspace{-4px}
\includegraphics[height=1.7cm,trim={200px 100px 190px 130px},clip]{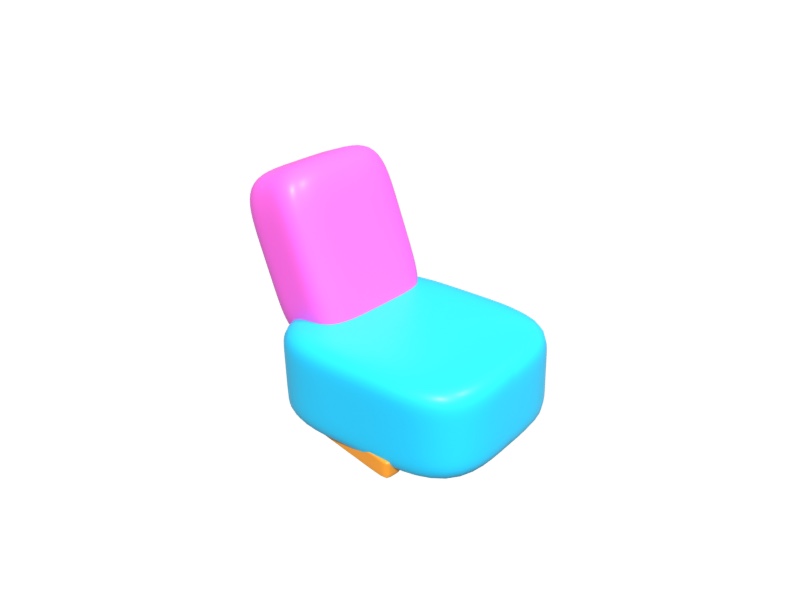}\hspace{-9px}
\includegraphics[height=1.7cm,trim={120px 70px 220px 80px},clip]{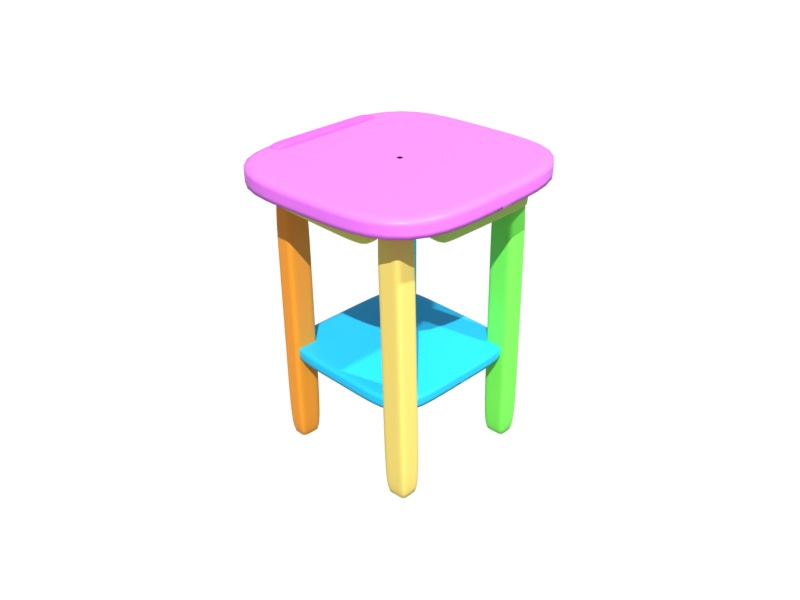}
\includegraphics[height=1.7cm,trim={220px 120px 220px 120px},clip]{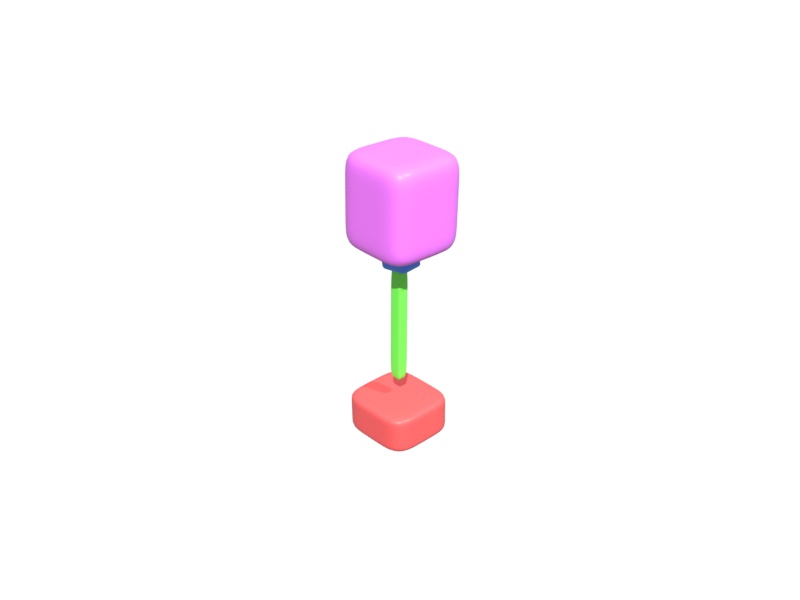}
\includegraphics[height=1.7cm,trim={220px 120px 220px 90px},clip]{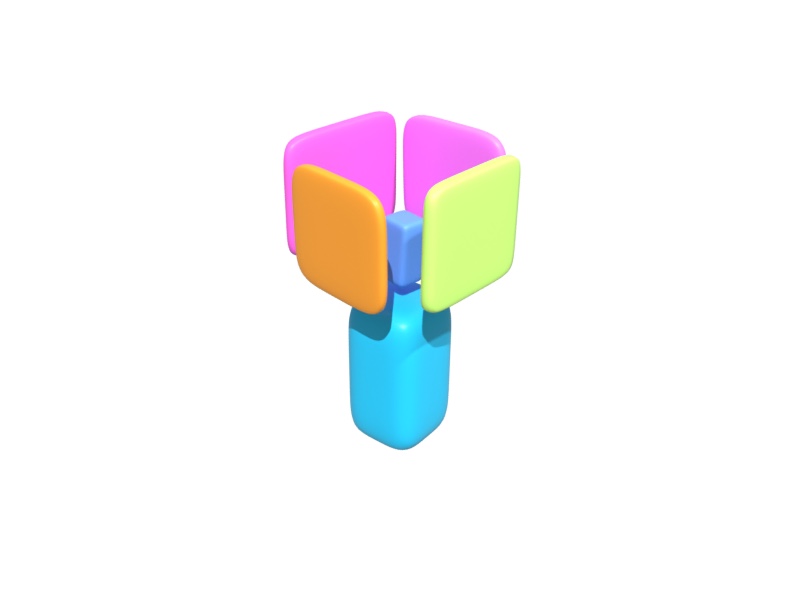}
\includegraphics[height=1.7cm,trim={260px 120px 230px 210px},clip]{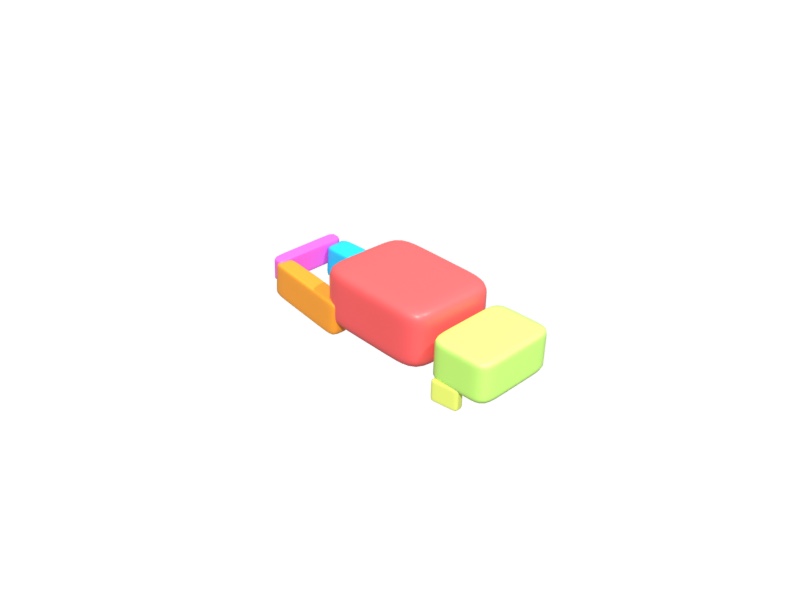}
\includegraphics[height=1.7cm,trim={240px 150px 260px 220px},clip]{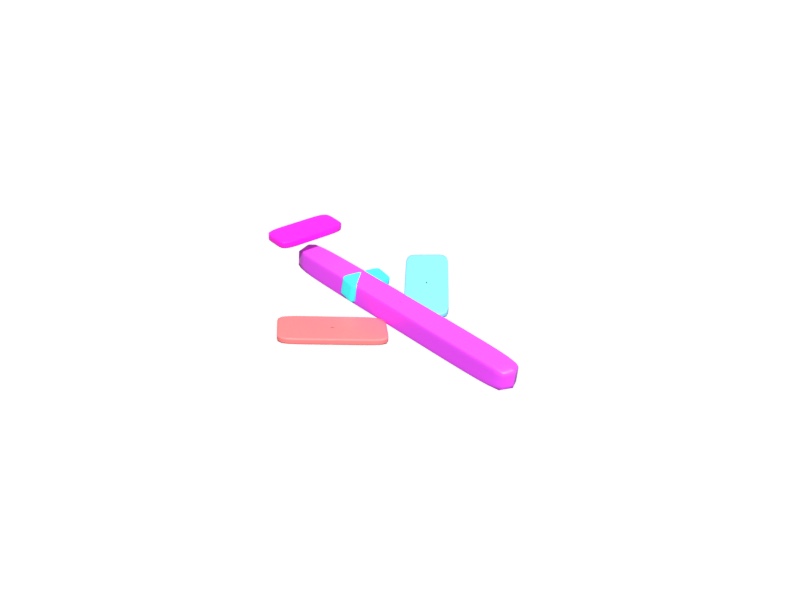}\\
\rotatebox{90}{\footnotesize \shortstack{ \hspace{2px}\method{}\\\scriptsize{(No T\&B)}}}\hspace{-4px}
\includegraphics[height=1.7cm,trim={200px 100px 190px 130px},clip]{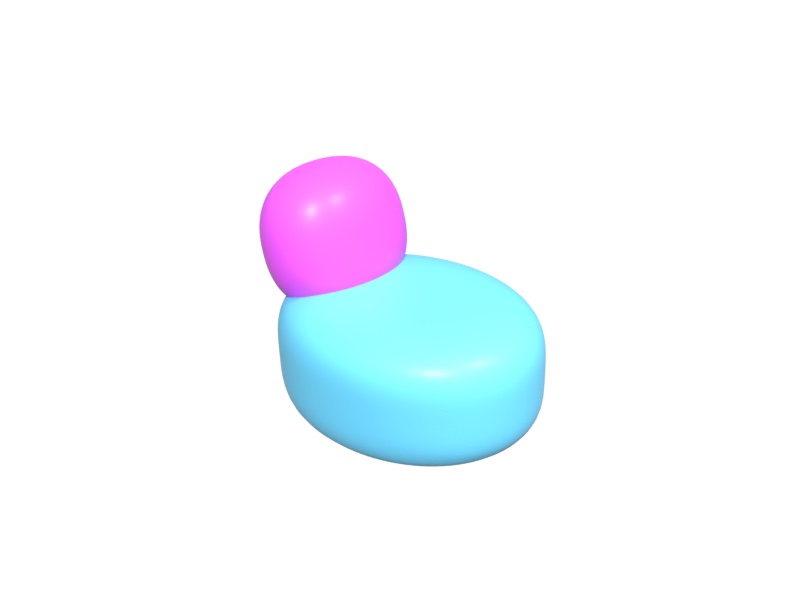}\hspace{-9px}
\includegraphics[height=1.7cm,trim={120px 70px 220px 80px},clip]{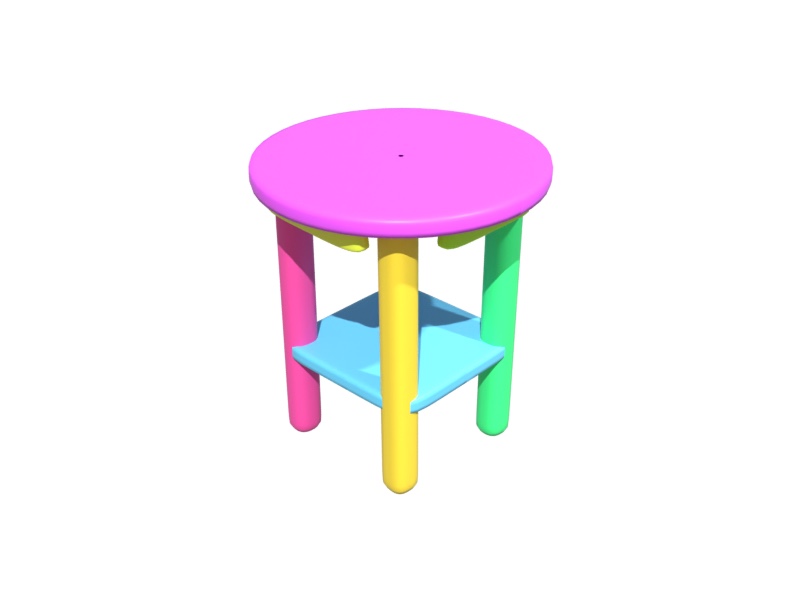}
\includegraphics[height=1.7cm,trim={220px 120px 220px 120px},clip]{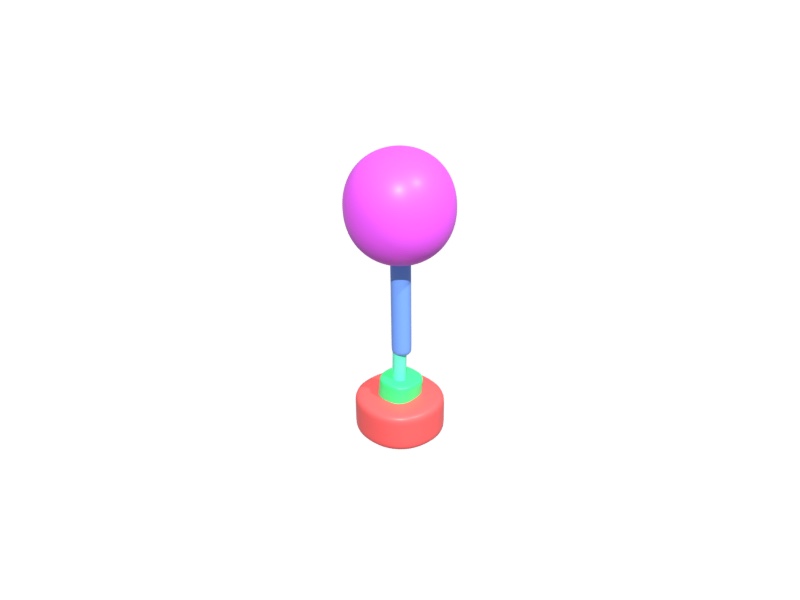}
\includegraphics[height=1.7cm,trim={220px 120px 220px 90px},clip]{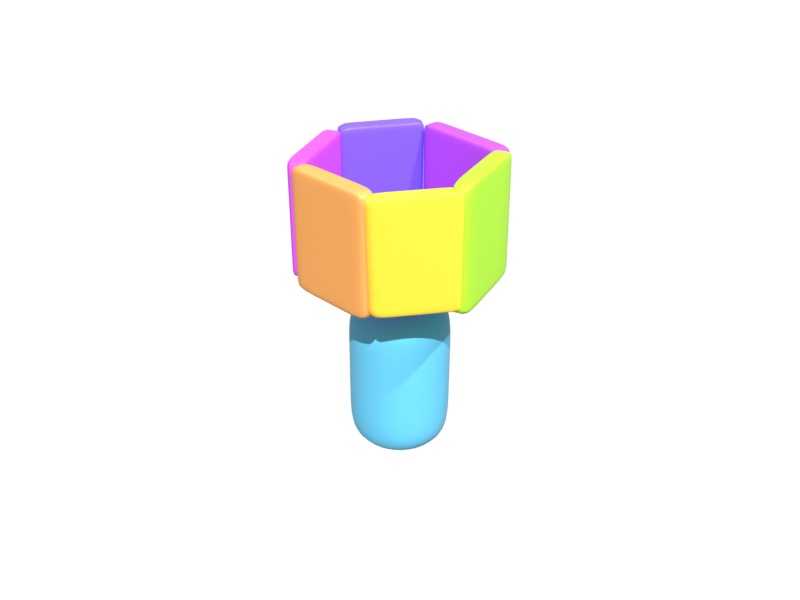}
\includegraphics[height=1.7cm,trim={260px 120px 230px 210px},clip]{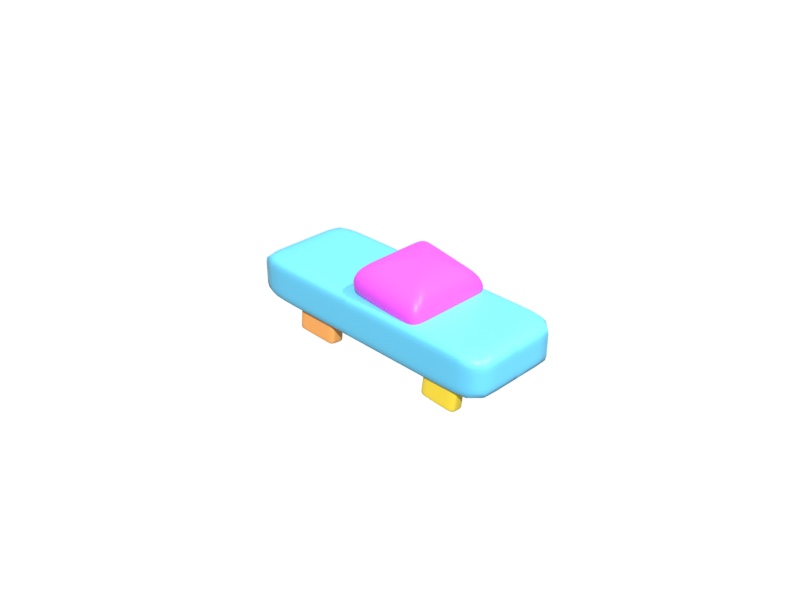}
\includegraphics[height=1.7cm,trim={240px 150px 260px 220px},clip]{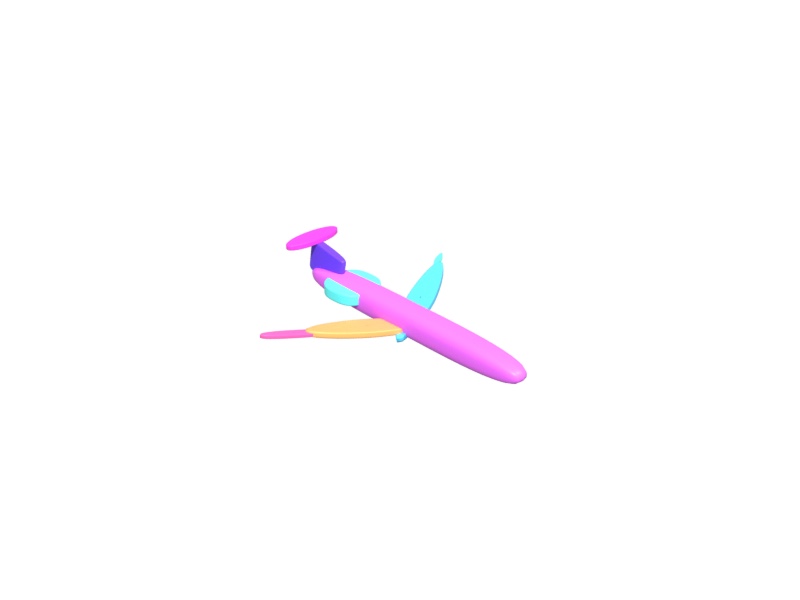}\\

\rotatebox{90}{\footnotesize \shortstack{\\ \hspace{5px}\method{}}}\hspace{-4px}
\includegraphics[height=1.7cm,trim={200px 100px 190px 130px},clip]{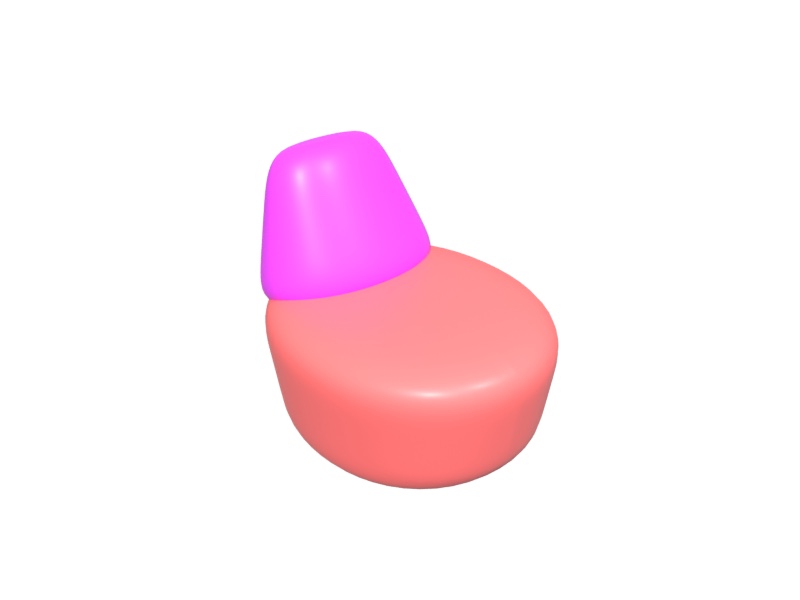}\hspace{-9px}
\includegraphics[height=1.7cm,trim={120px 70px 220px 80px},clip]{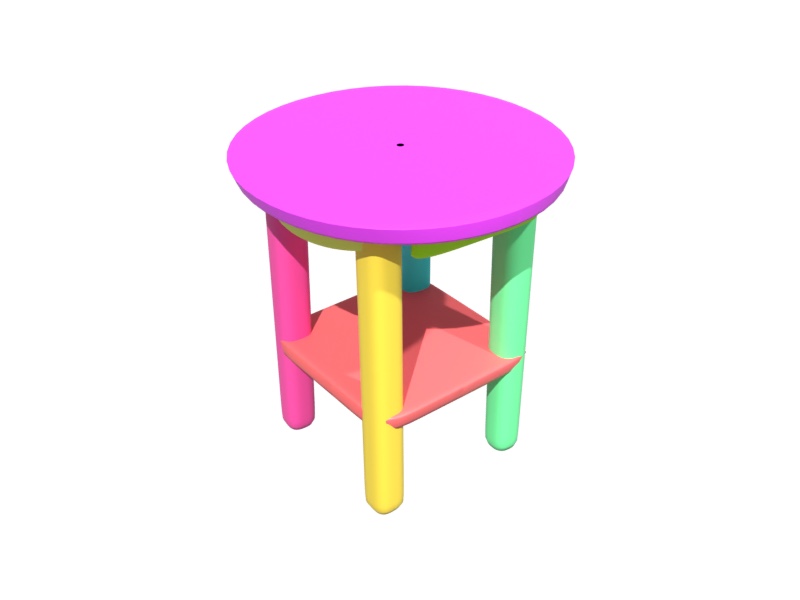}
\includegraphics[height=1.7cm,trim={220px 120px 220px 120px},clip]{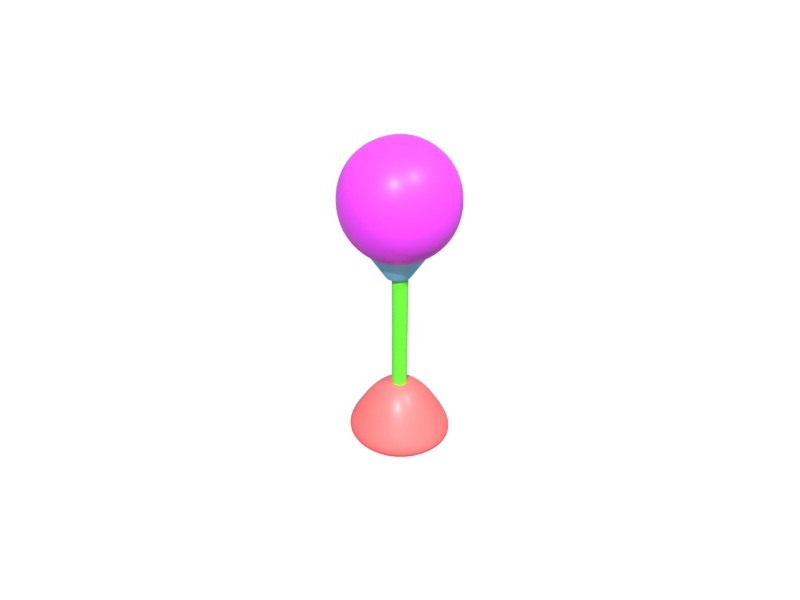}
\includegraphics[height=1.7cm,trim={220px 120px 220px 90px},clip]{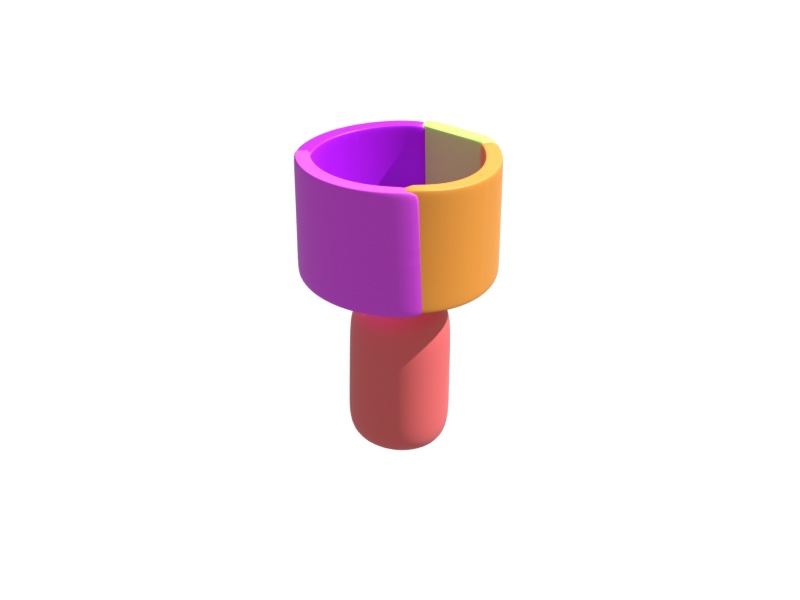}
\includegraphics[height=1.7cm,trim={260px 120px 230px 210px},clip]{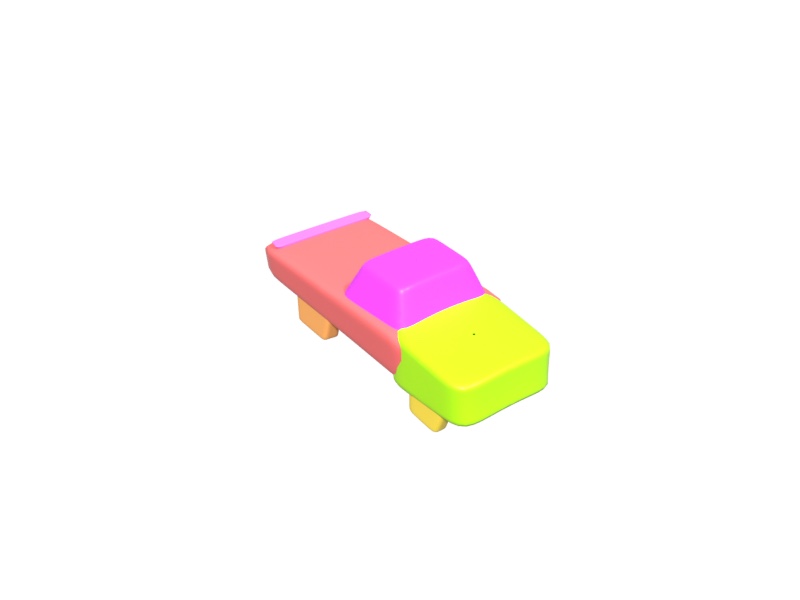}
\includegraphics[height=1.7cm,trim={240px 150px 260px 220px},clip]{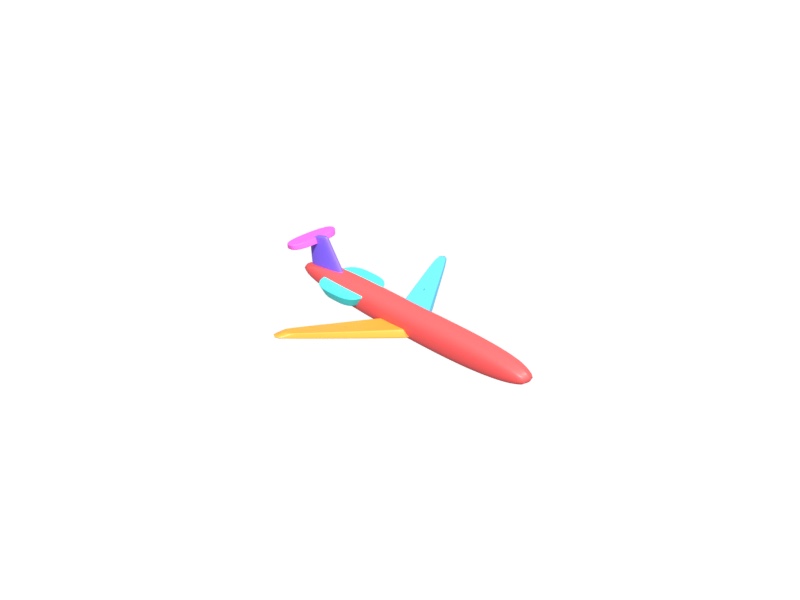}\\
\end{small}
\caption{\textbf{Qualitative Results on ShapeNet.} \emph{Top row:} input point clouds. \emph{Below:} the outputs of baselines and our methods. Different colors indicate different primitives. For \name{}, we compare variants with and without tapering and bending (T\&B) deformation parameters.}
\label{fig:quali-shapenet}
\end{figure}

\subsubsection{Qualitative Evaluation.}
We present qualitative results in Fig.~\ref{fig:quali-shapenet}. Compared to Marching Primitives~\cite{liu2023marching}, \method{} achieves comparable reconstruction quality while using substantially fewer primitives. Compared to SuperDec~\cite{fedele2025superdec}, it produces similarly compact decompositions while capturing the underlying geometry much more accurately. In particular, our method better represents rounded surfaces (first three columns) and avoids the large gaps between neighboring primitives often observed in prior work (last three columns).

\noindent The results of \method{} without tapering and bending demonstrate that our loss function alone substantially improves reconstruction quality while retaining the representational constraints of rigid superquadrics. Enabling tapering and bending further improves the reconstruction of asymmetric shapes and non-convex parts, respectively. Additional qualitative results and comparisons with more baselines are provided in Appendix~\ref{sec:quali-supp}.

\begin{table}[h!]
    \centering
    {\setlength{\tabcolsep}{5.5pt}%
    \begin{tabular}{c| cccccc}
        \toprule
        \textbf{Test-time Opt.} &\textbf{IoU}$\uparrow$ & \textbf{F}~\cite{TankAndTamples}$\uparrow$ & \textbf{L1}$\downarrow$ & \textbf{L2}$\downarrow$ & \textbf{\#\,Primitives}$\downarrow$ & \textbf{Runtime}$\downarrow$ \\
         \midrule
        \xmark & $0.72$ & $0.37$ & $1.54$ & $0.043$ & $\mathbf{5.64}$ & $\mathbf{0.0082}$ s\\
        \cmark & $\mathbf{0.87}$ & $\mathbf{0.49}$ & $\mathbf{1.32}$ & $\mathbf{0.036}$ & $\mathbf{5.64}$ & $5$ s \\ 
        \bottomrule
    \end{tabular}}
    \vspace{5px}
    \caption{\textbf{Test-time Optimization on ShapeNet.} We evaluate the impact of applying test-time optimization to our feed-forward predictions. The first row reports the original feed-forward predictions, while the second row reports the refined predictions after test-time optimization. Chamfer distances (L1 and L2) are multiplied by $10^2$.}
    \label{tab:opti}
\end{table}

\begin{figure}[h!]
    \centering
    \includegraphics[width=\linewidth]{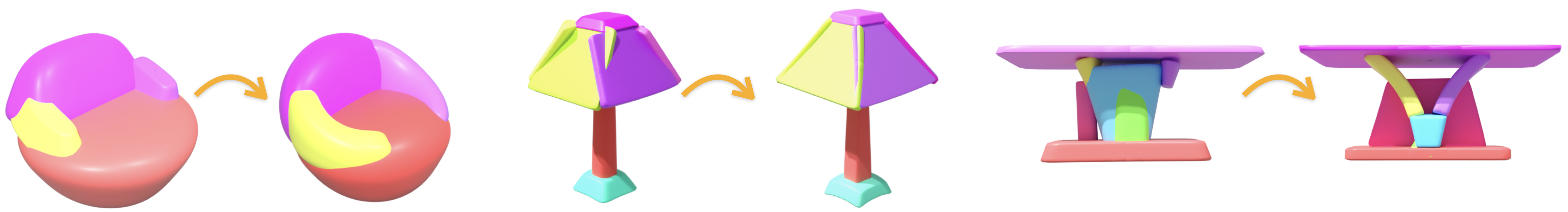}
    \vspace{-5mm}
    \caption{\textbf{Qualitative Test-time Optimization Results.} \method{} refinement fits curved parts with a compact set of flexible superquadric primitives.}
    \label{fig:optimization}
\end{figure}

\subsection{Test-time Optimization}\label{sec:optimization}
To evaluate our refinement method, we perform test-time optimization using Adam \cite{kingma2017adammethodstochasticoptimization} for 1000 iterations per object, using a temperature $\tau_{\text{opt}} = 0.01$.
Tab.~\ref{tab:opti} shows that our lightweight refinement stage, while substantially faster than prior optimization-based methods, further improves the predictions of the feed-forward model, yielding a relative improvement of $22\%$. Fig.~\ref{fig:optimization} qualitatively illustrates how the refinement produces more accurate primitive fits.

\begin{figure}[t]
\centering
\begin{small}
\rotatebox{90}{\footnotesize \shortstack{Input \\ Point Cloud \vspace{1.5px}}}
\includegraphics[height=1.7cm,trim={280px 320px 280px 130px},clip]{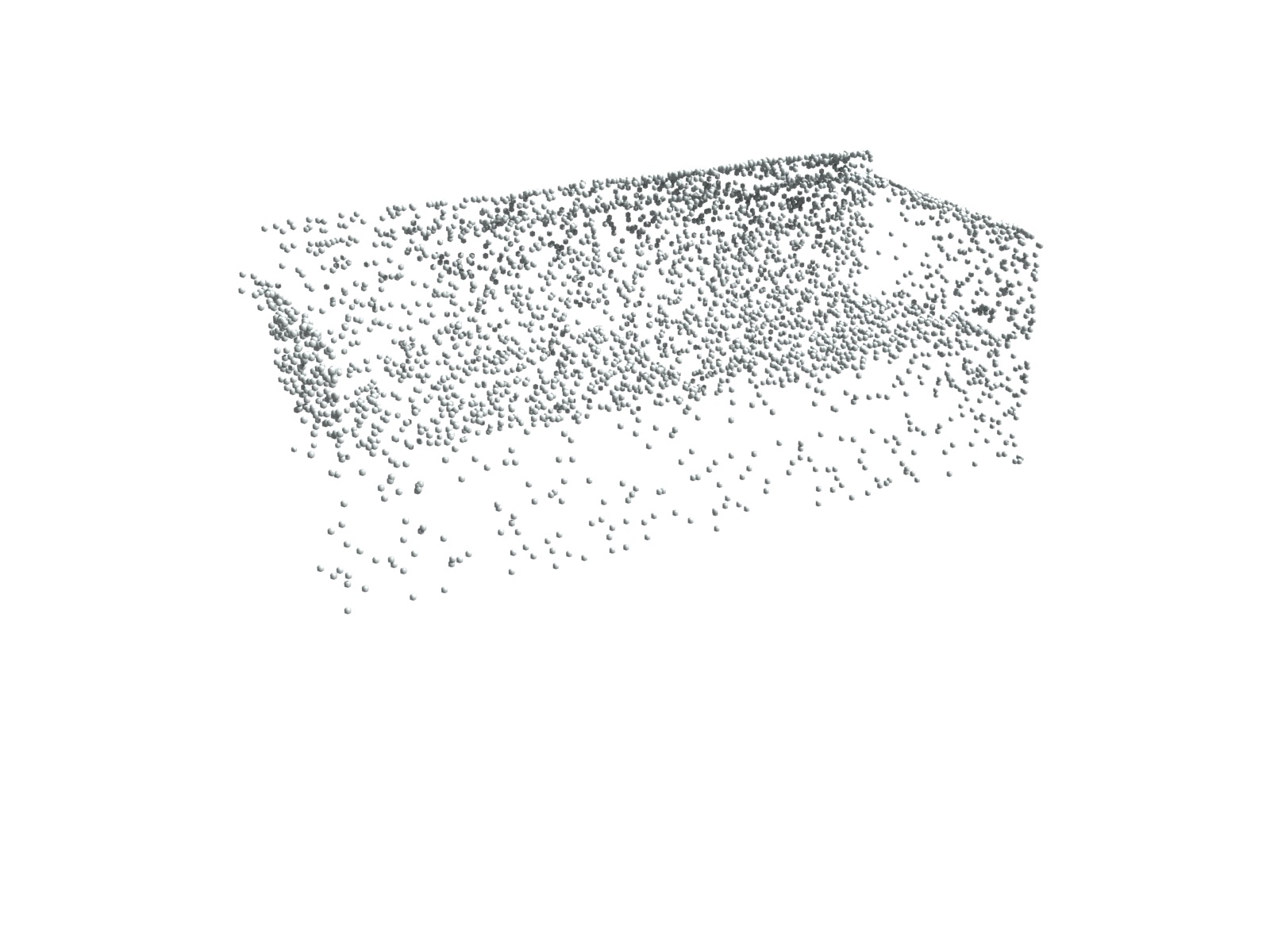}
\includegraphics[height=1.7cm,trim={700px 480px 700px 480px},clip]{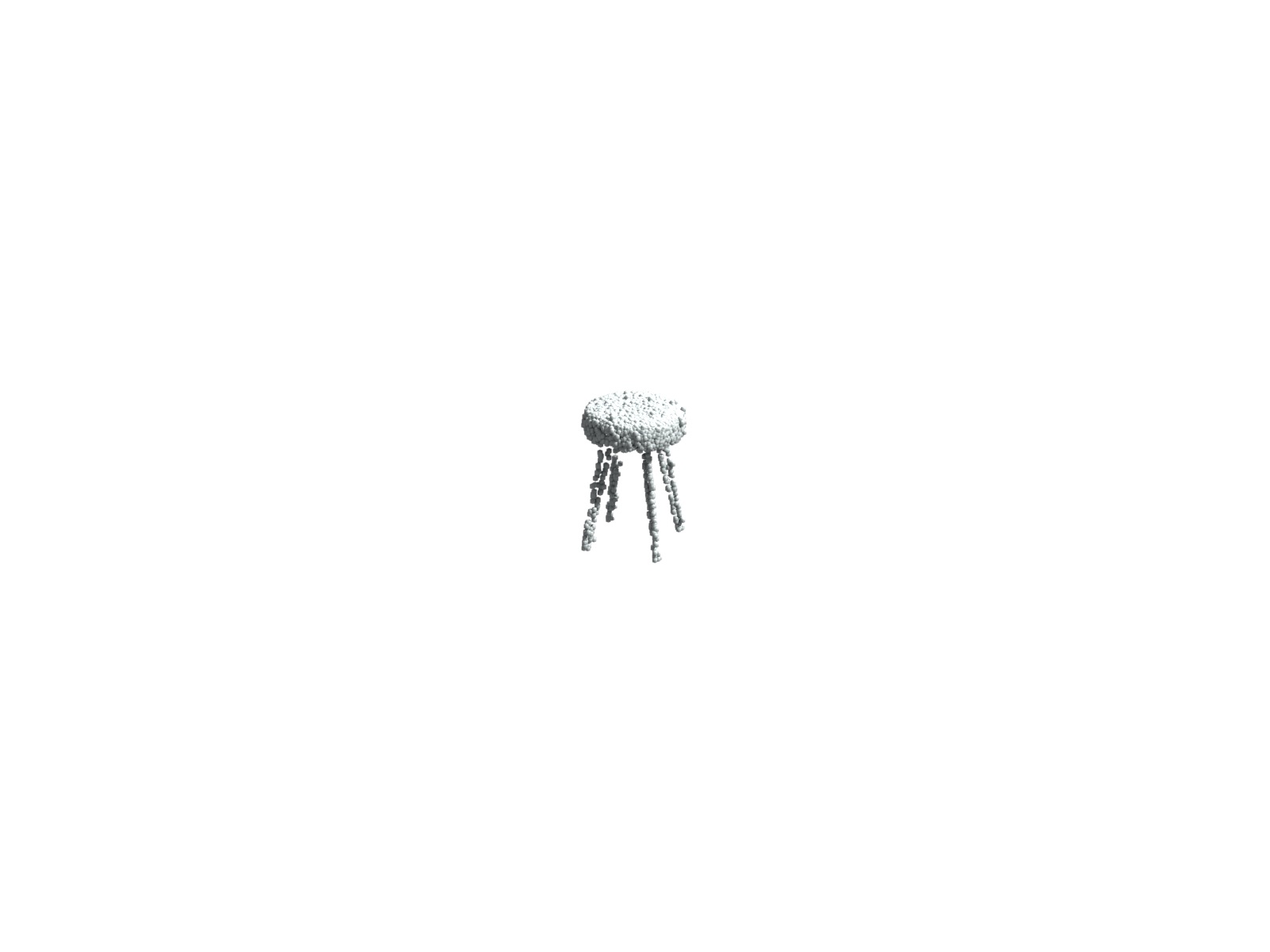}
\includegraphics[height=1.7cm,trim={580px 480px 580px 280px},clip]{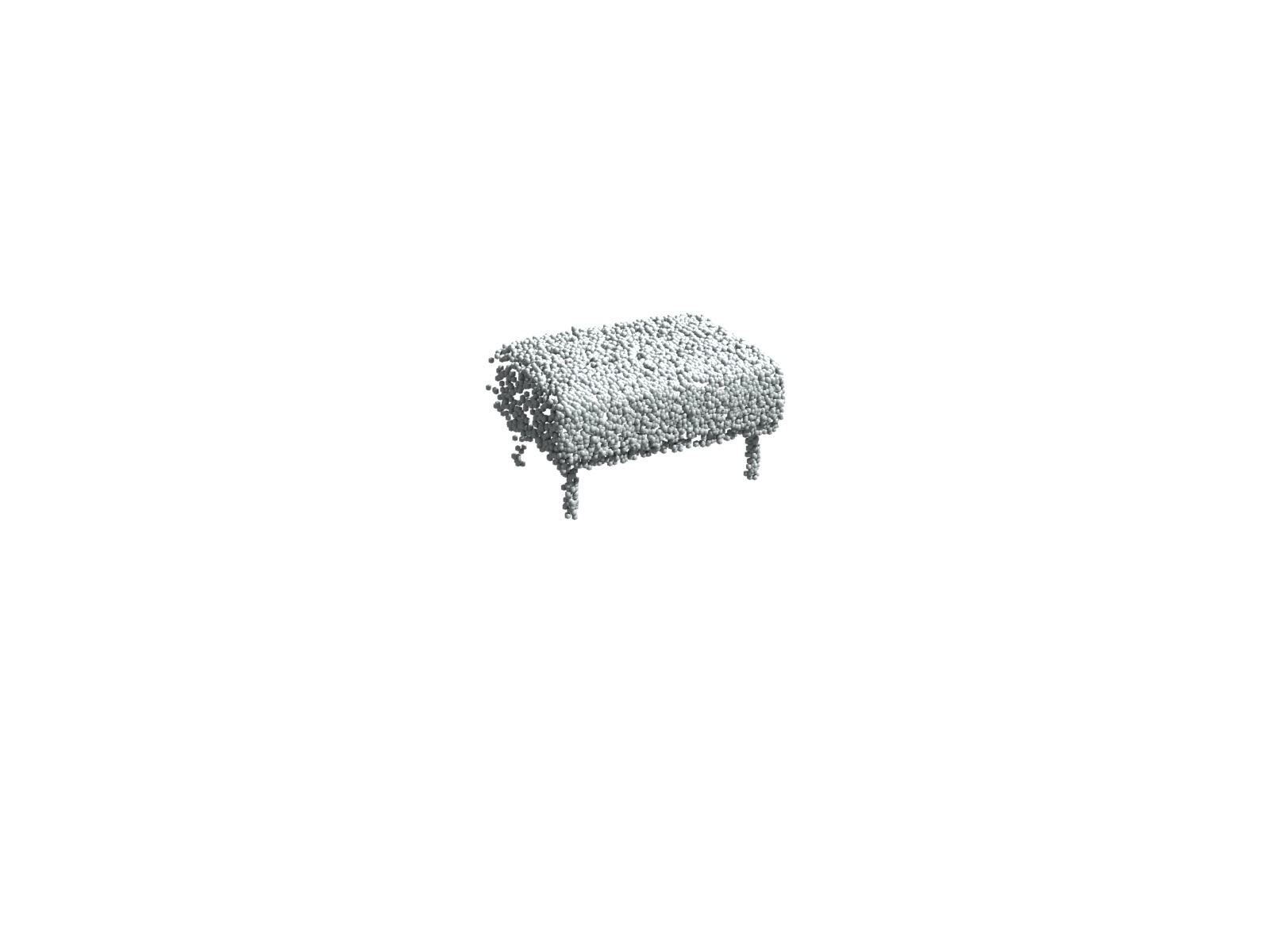}
\includegraphics[height=1.7cm,trim={480px 440px 480px 180px},clip]{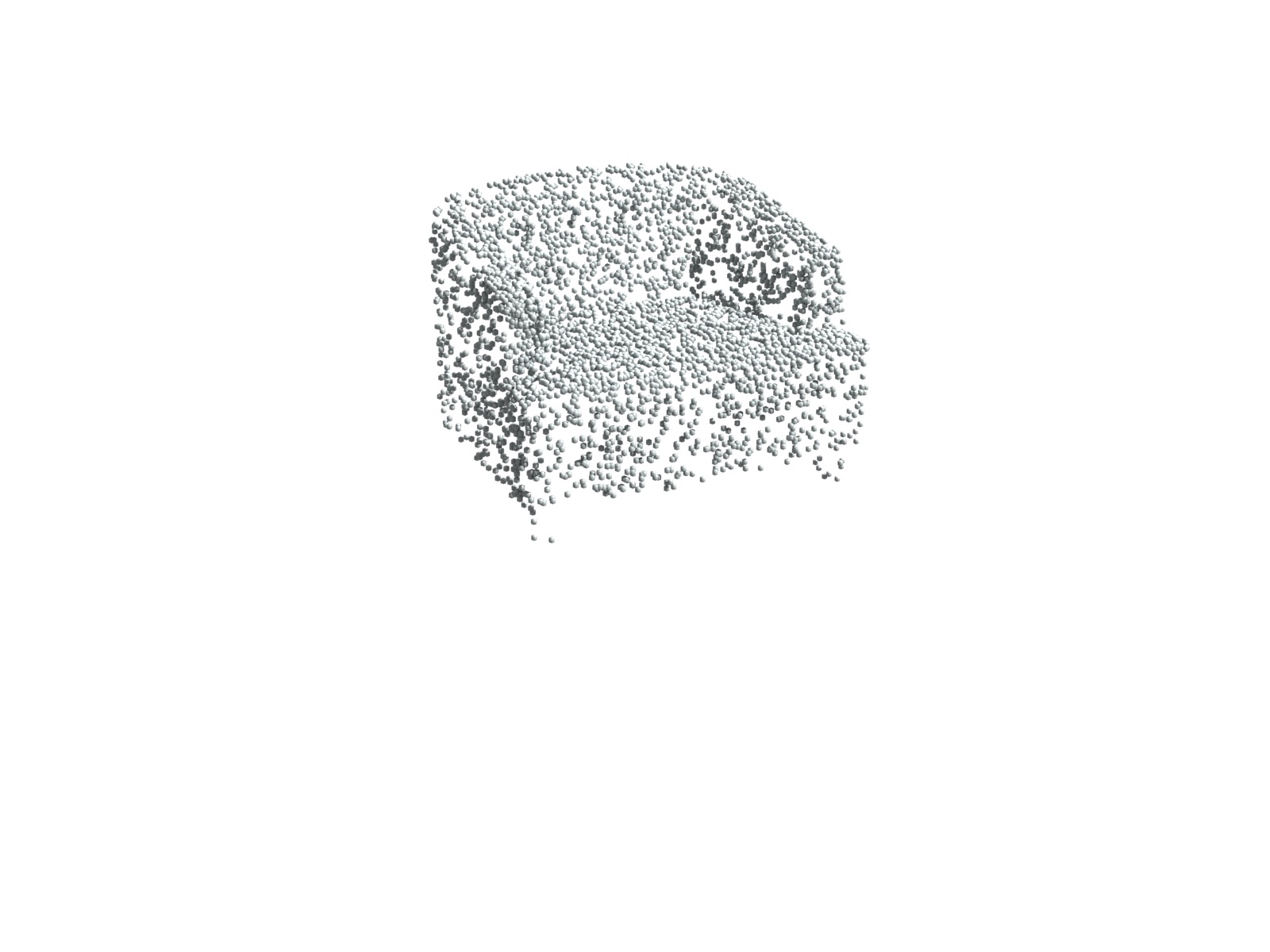}
\includegraphics[height=1.7cm,trim={580px 480px 580px 280px},clip]{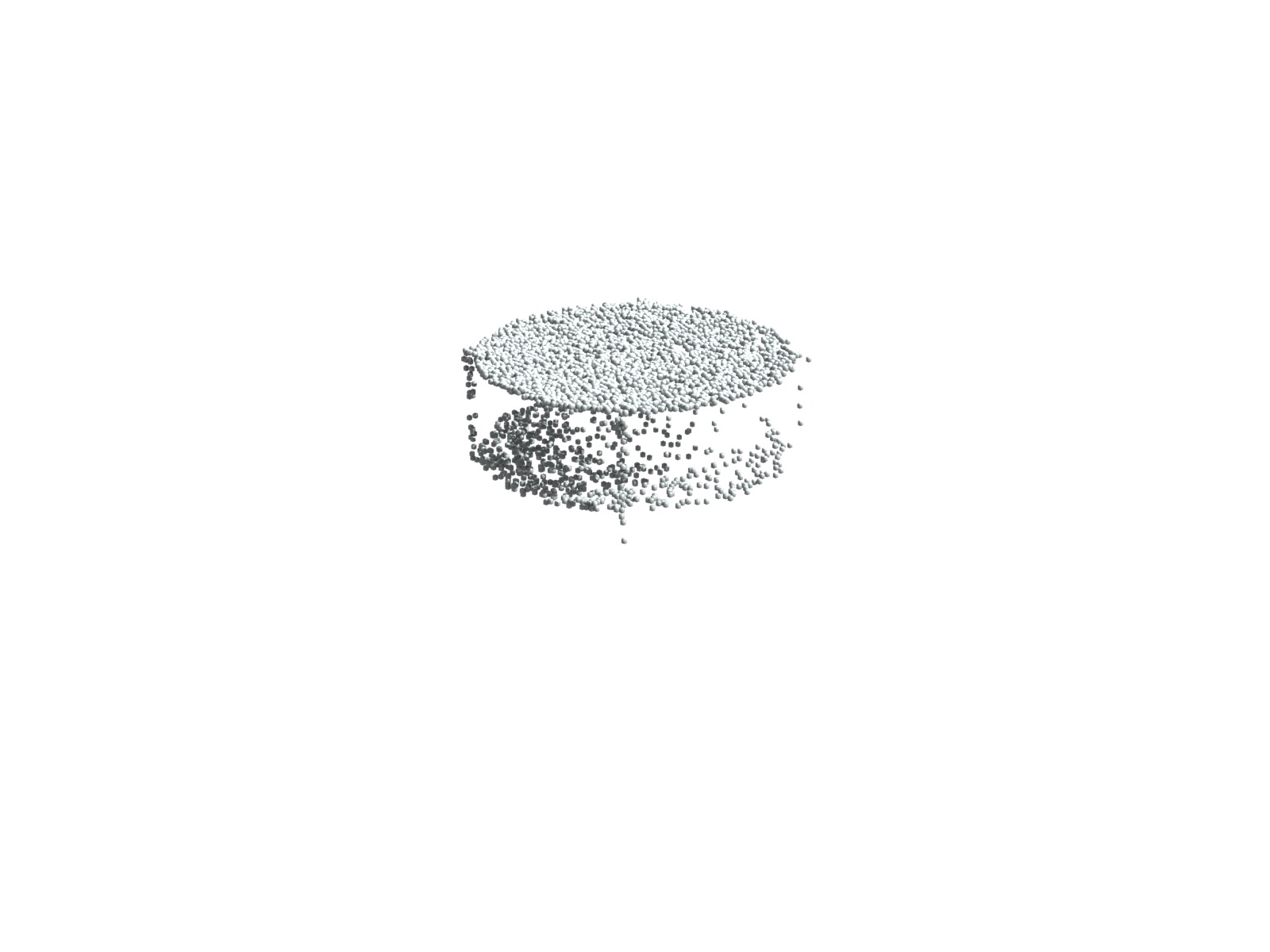}
\includegraphics[height=1.7cm,trim={680px 560px 680px 400px},clip]{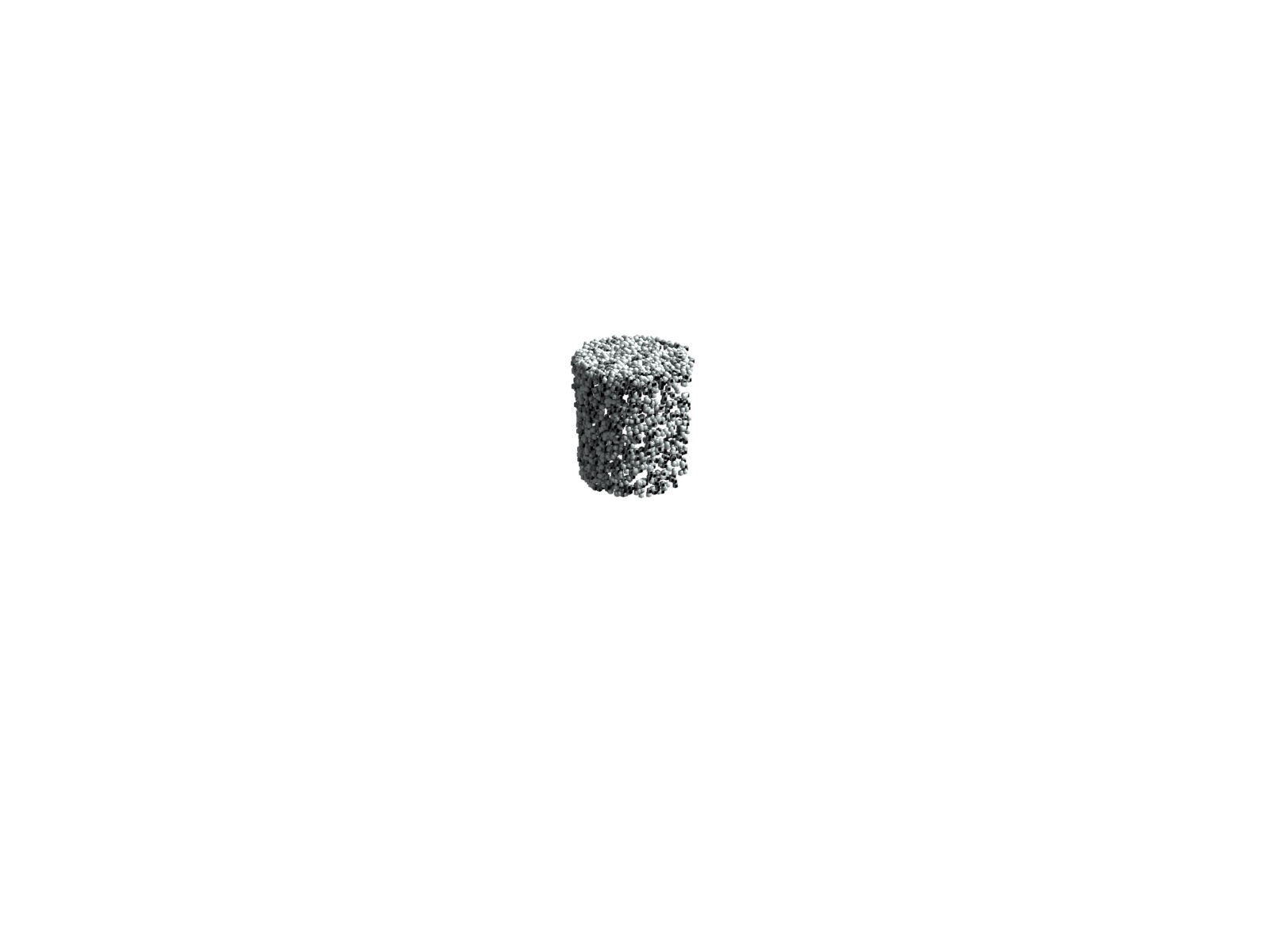}

\rotatebox{90}{\footnotesize \shortstack{\\ \vspace{6px} \method{}}}
\includegraphics[height=1.7cm,trim={280px 320px 280px 130px},clip]{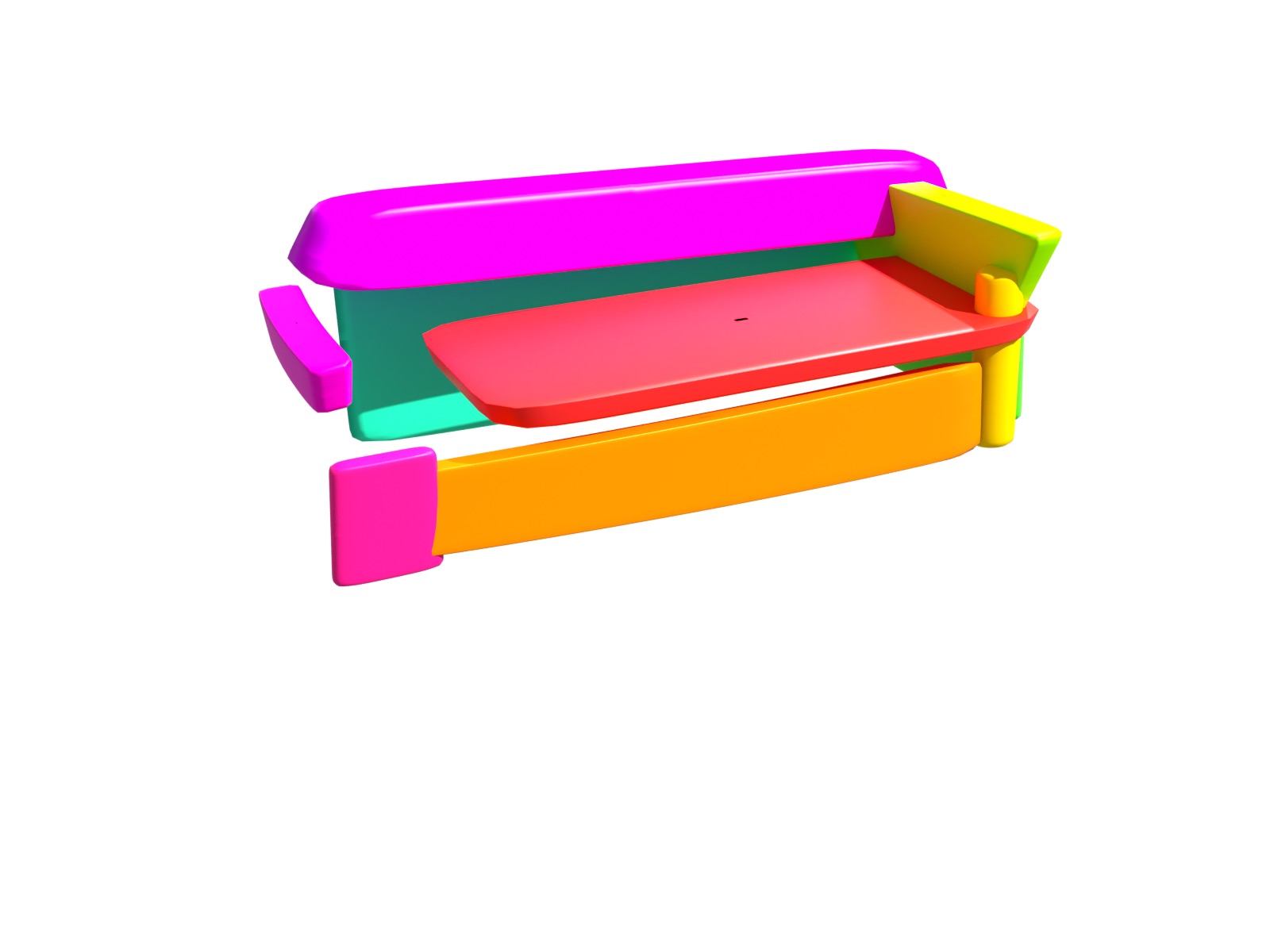}
\includegraphics[height=1.7cm,trim={700px 480px 700px 480px},clip]{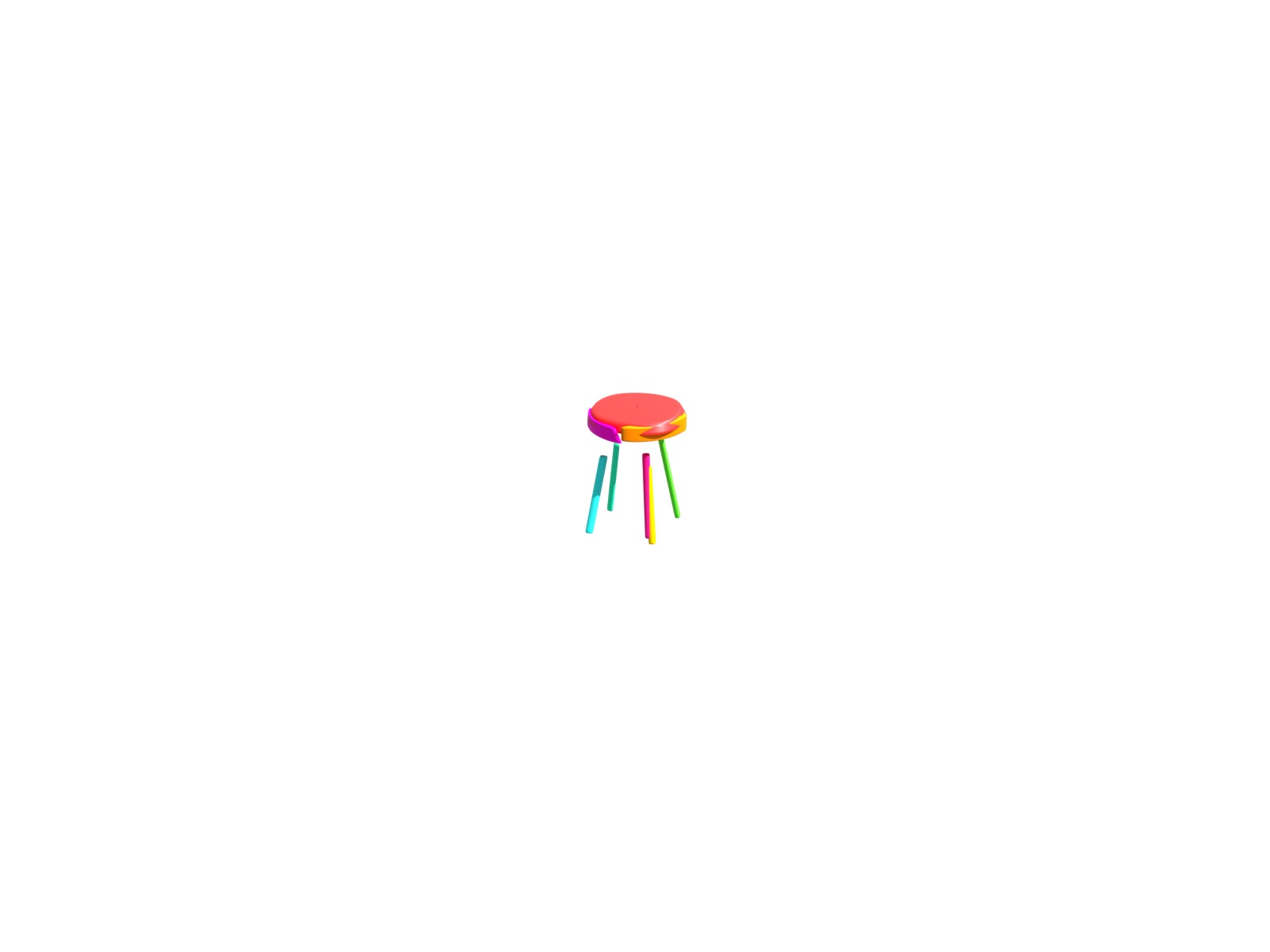}
\includegraphics[height=1.7cm,trim={580px 480px 580px 280px},clip]{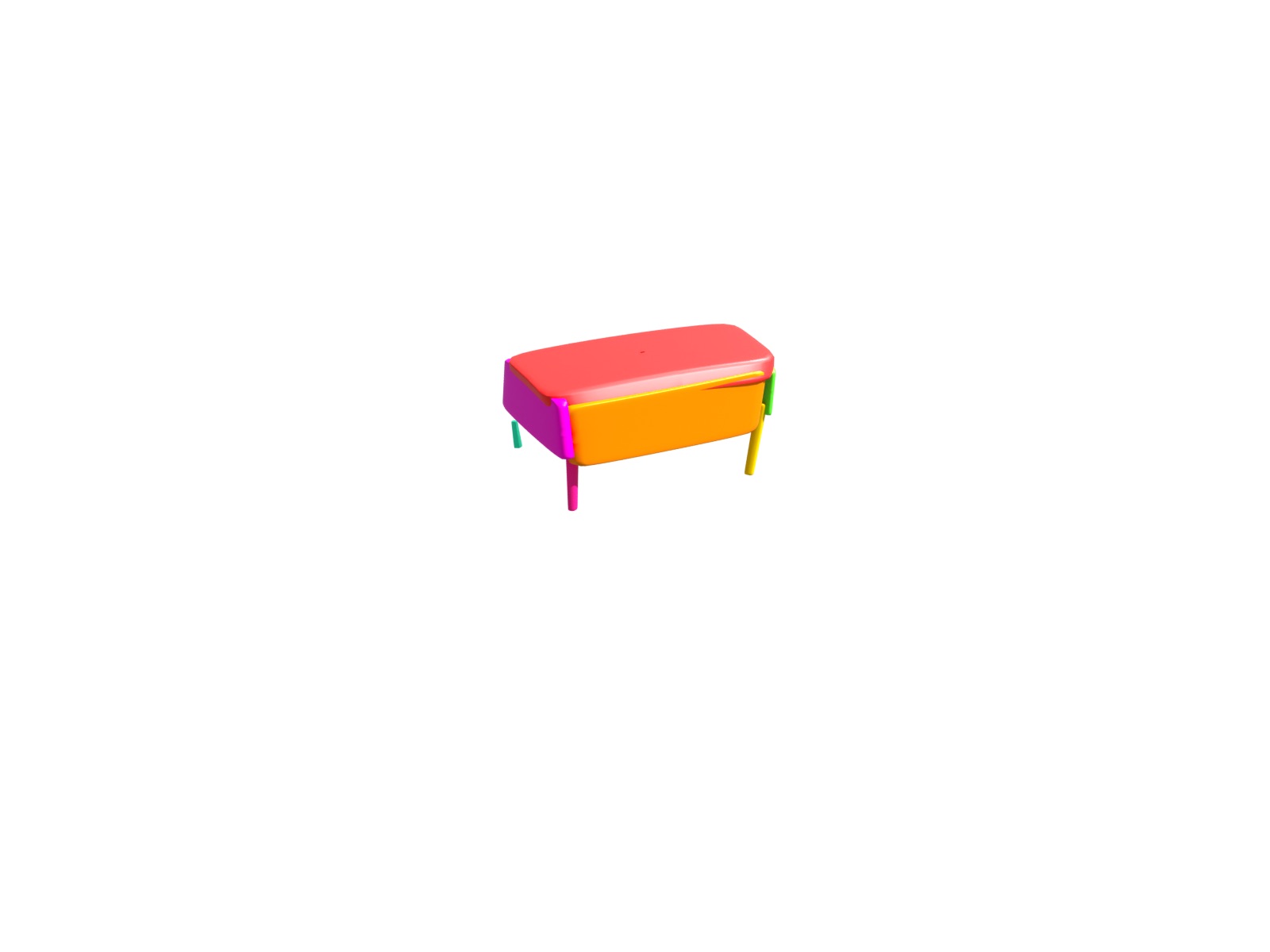}
\includegraphics[height=1.7cm,trim={480px 440px 480px 180px},clip]{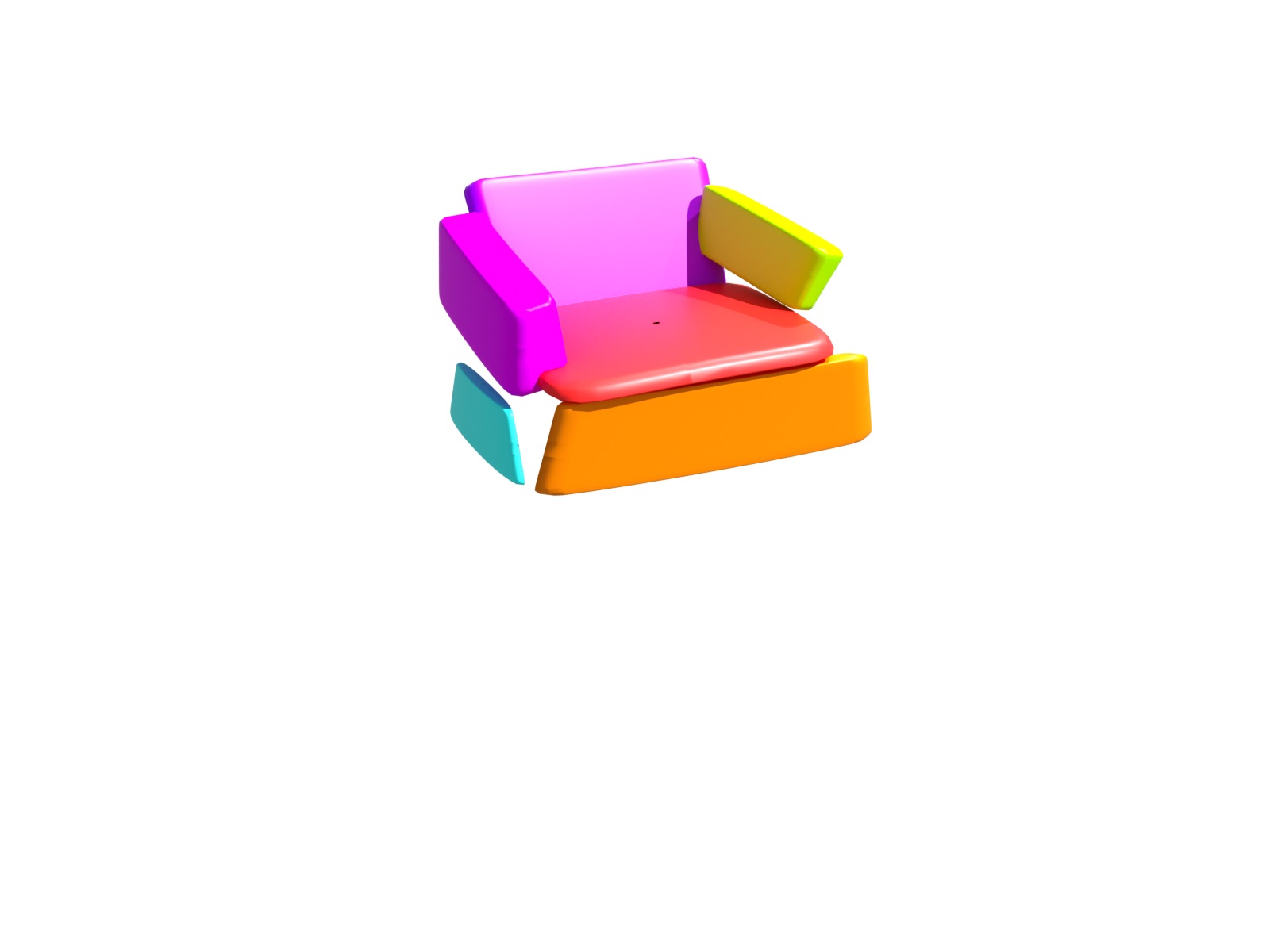}
\includegraphics[height=1.7cm,trim={580px 480px 580px 280px},clip]{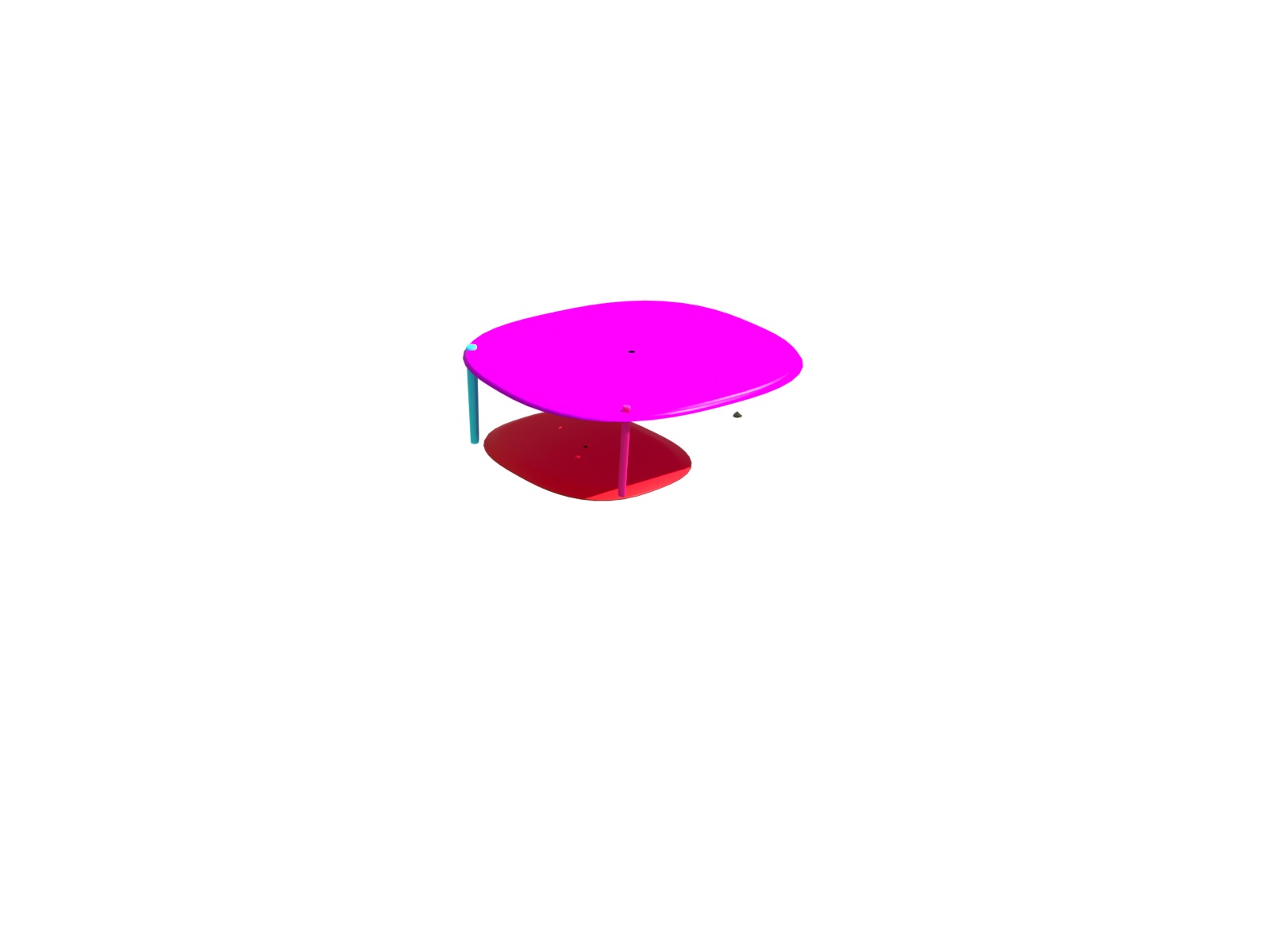}
\includegraphics[height=1.7cm,trim={680px 560px 680px 400px},clip]{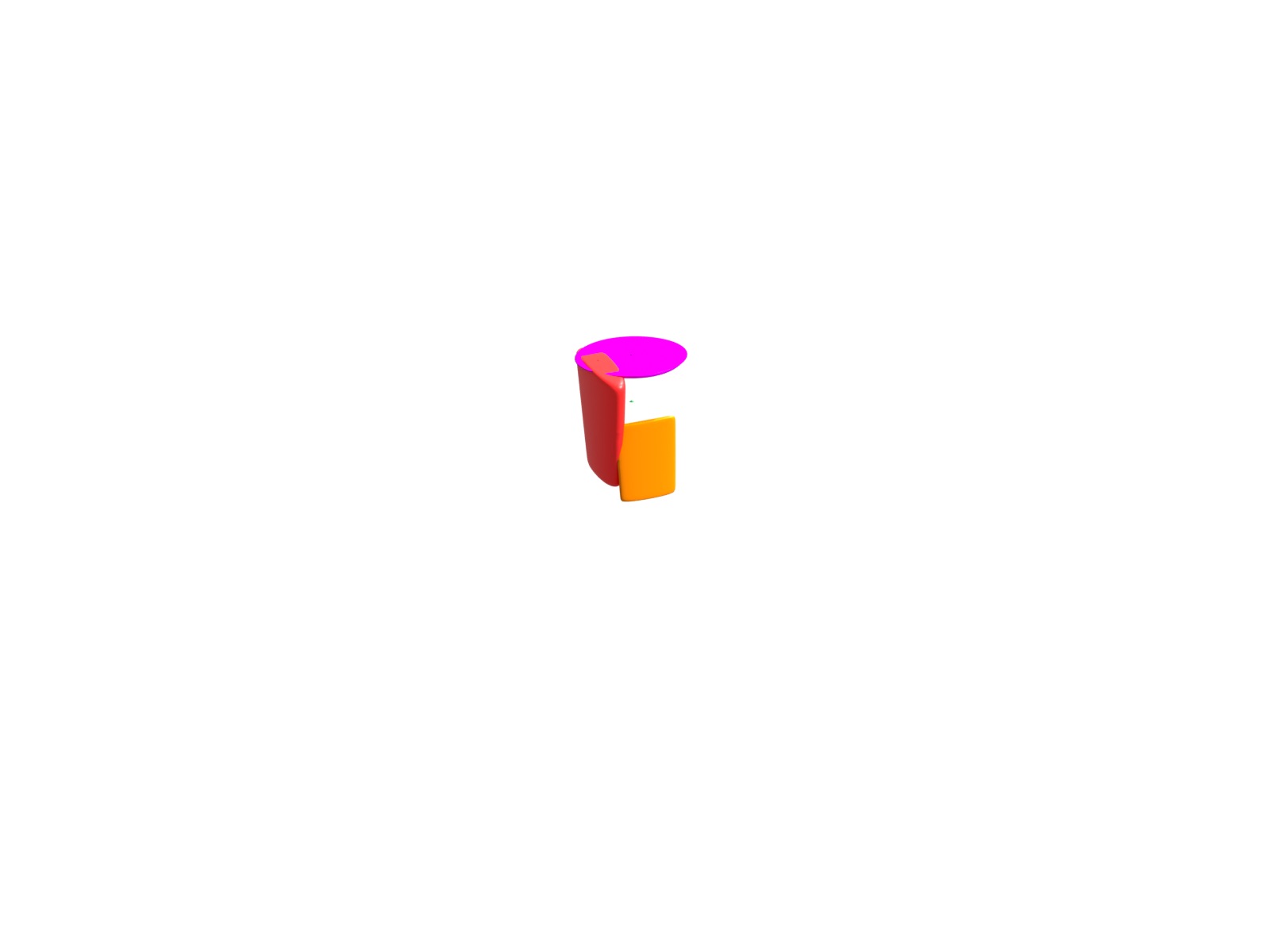}\\

\rotatebox{90}{\footnotesize \shortstack{Robust \\ \method{}}}
\includegraphics[height=1.7cm,trim={280px 320px 280px 130px},clip]{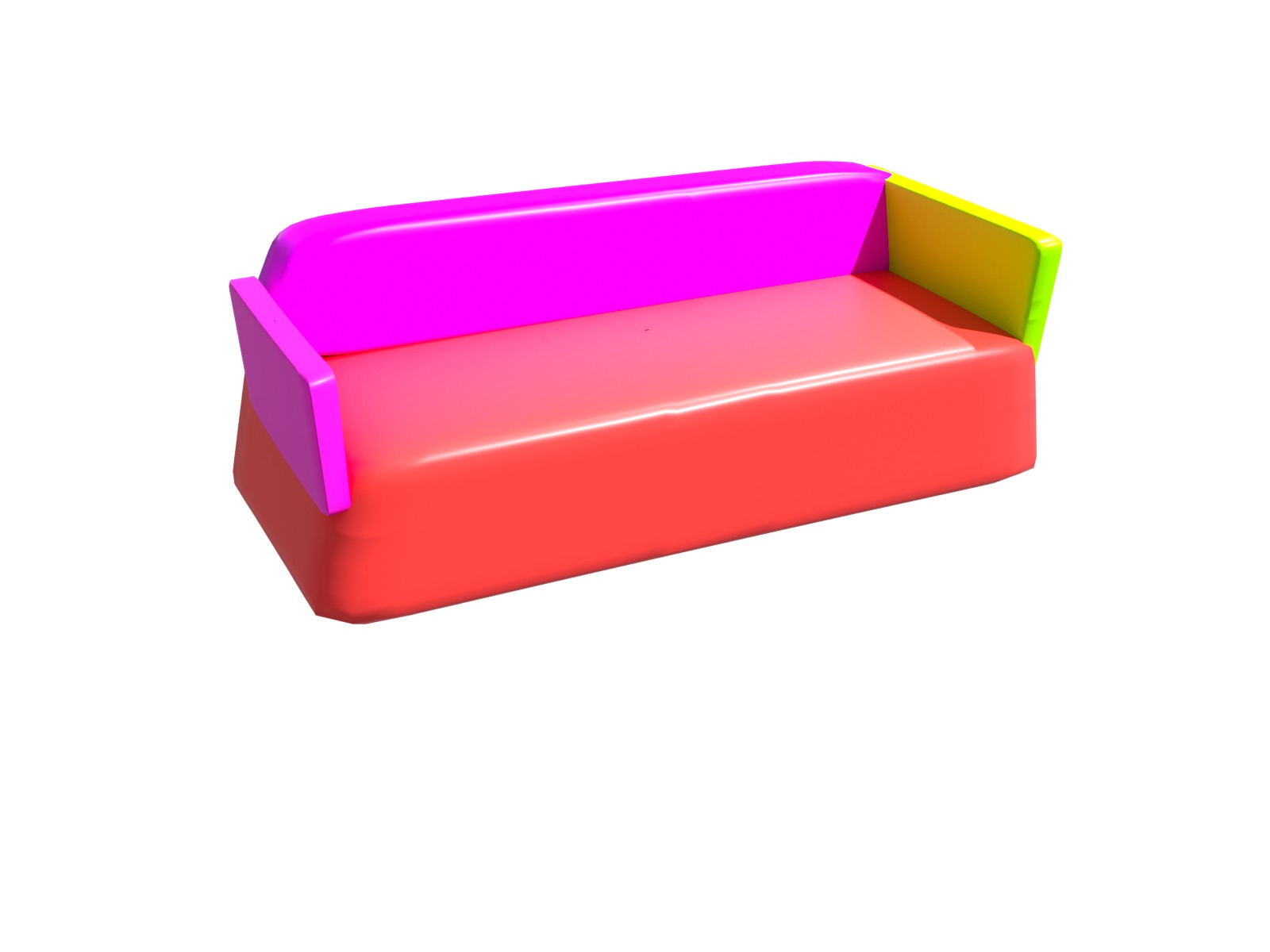}
\includegraphics[height=1.7cm,trim={700px 480px 700px 480px},clip]{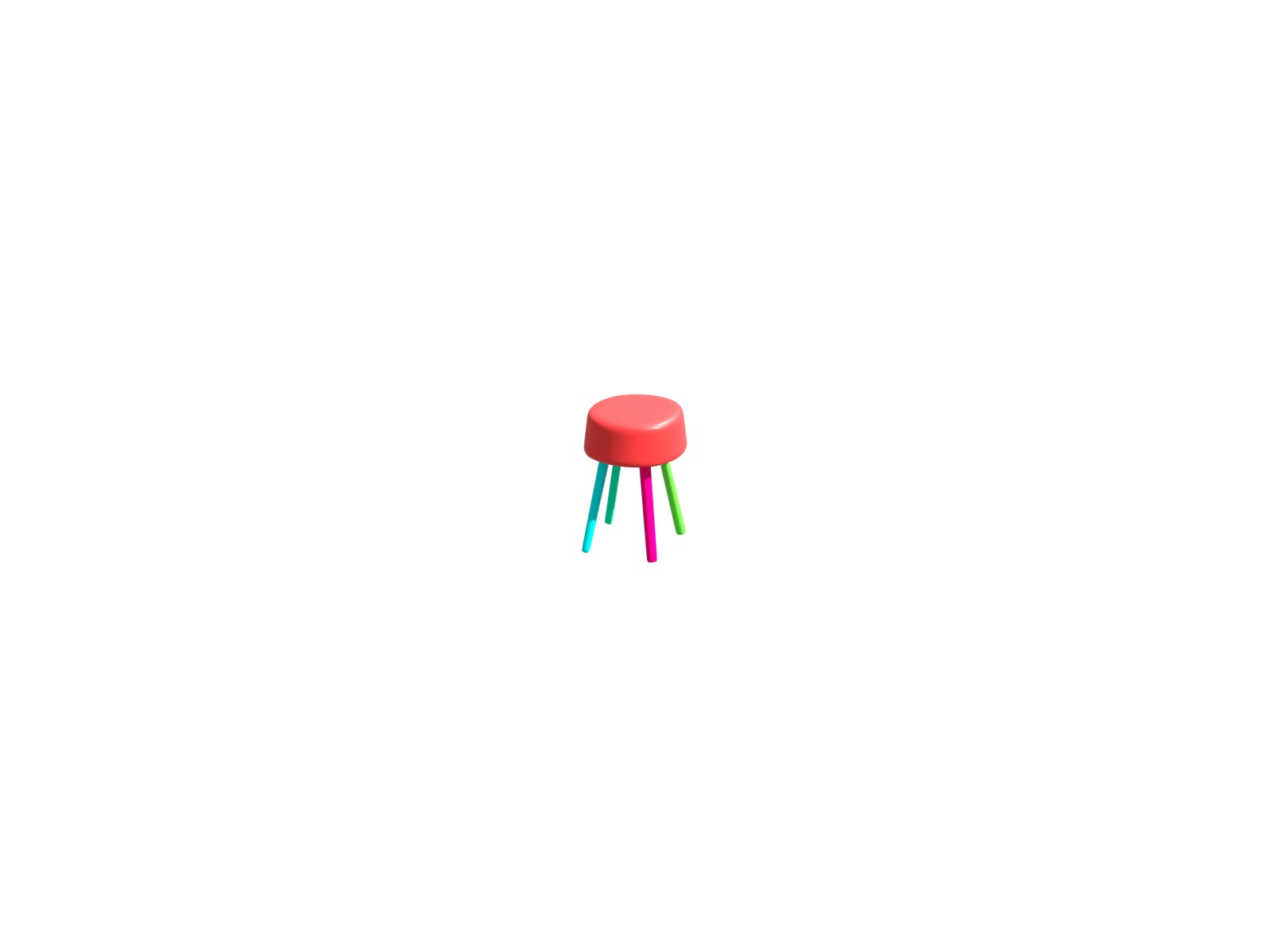}
\includegraphics[height=1.7cm,trim={580px 480px 580px 280px},clip]{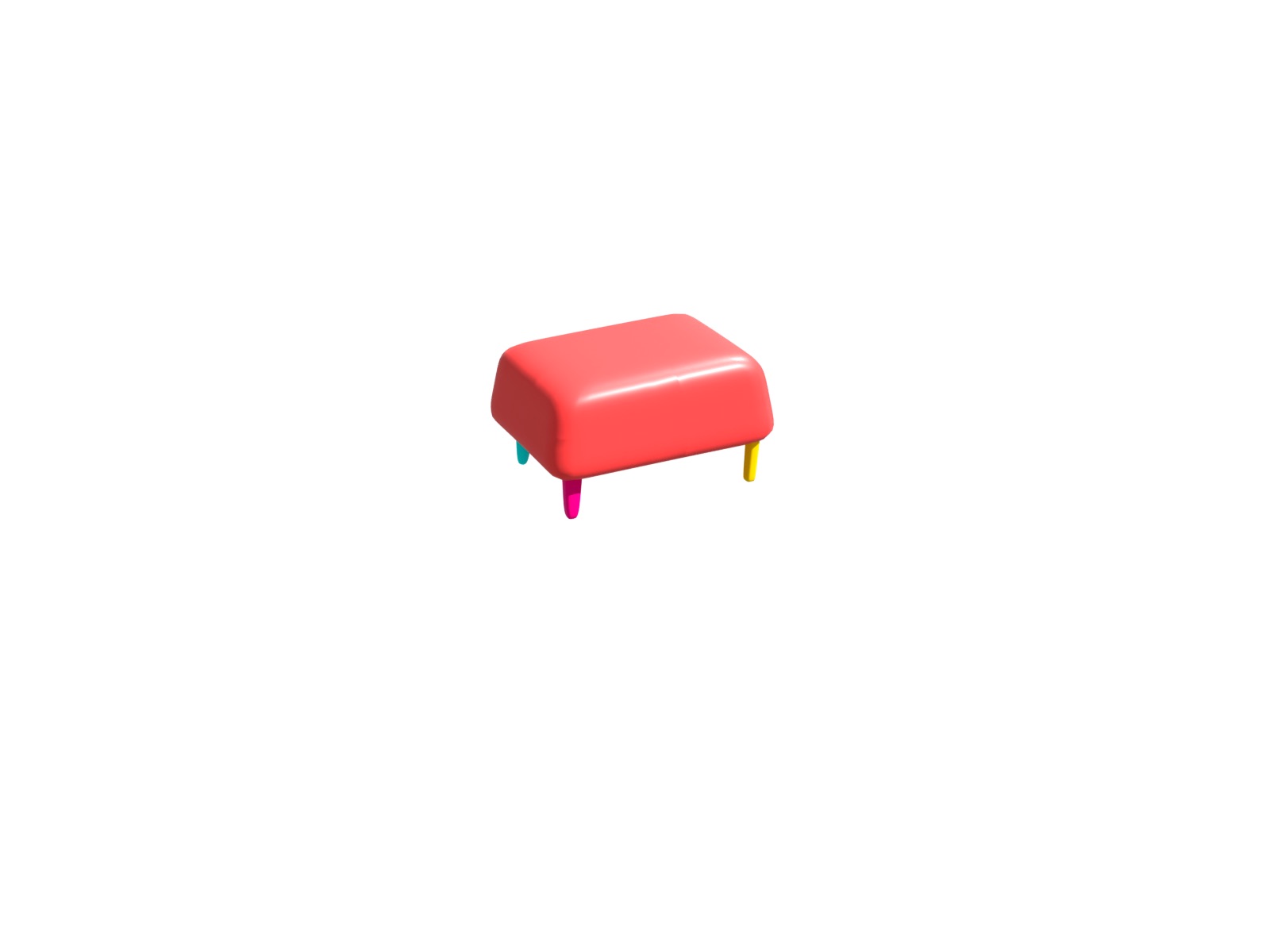}
\includegraphics[height=1.7cm,trim={480px 440px 480px 180px},clip]{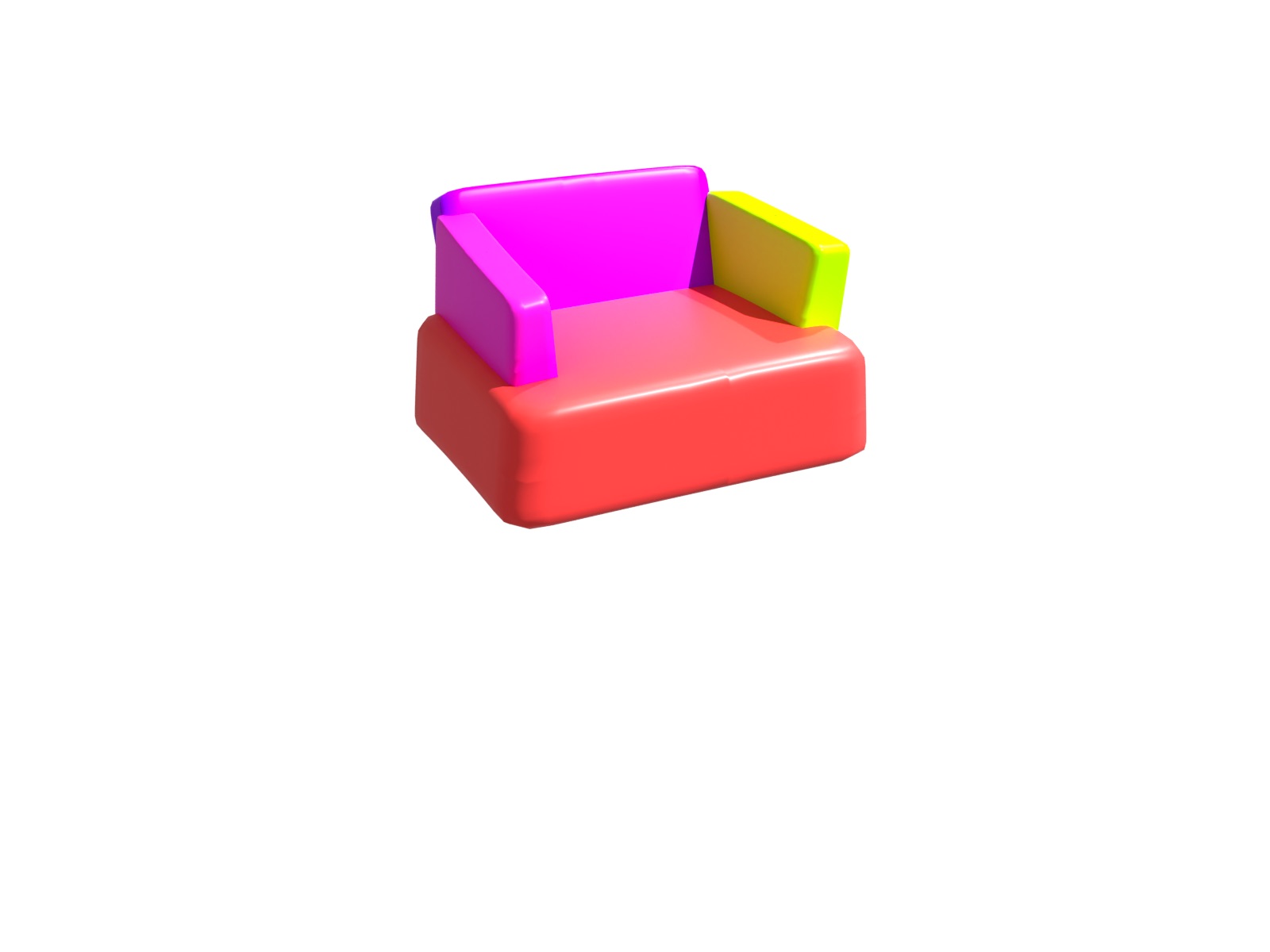}
\includegraphics[height=1.7cm,trim={580px 480px 580px 280px},clip]{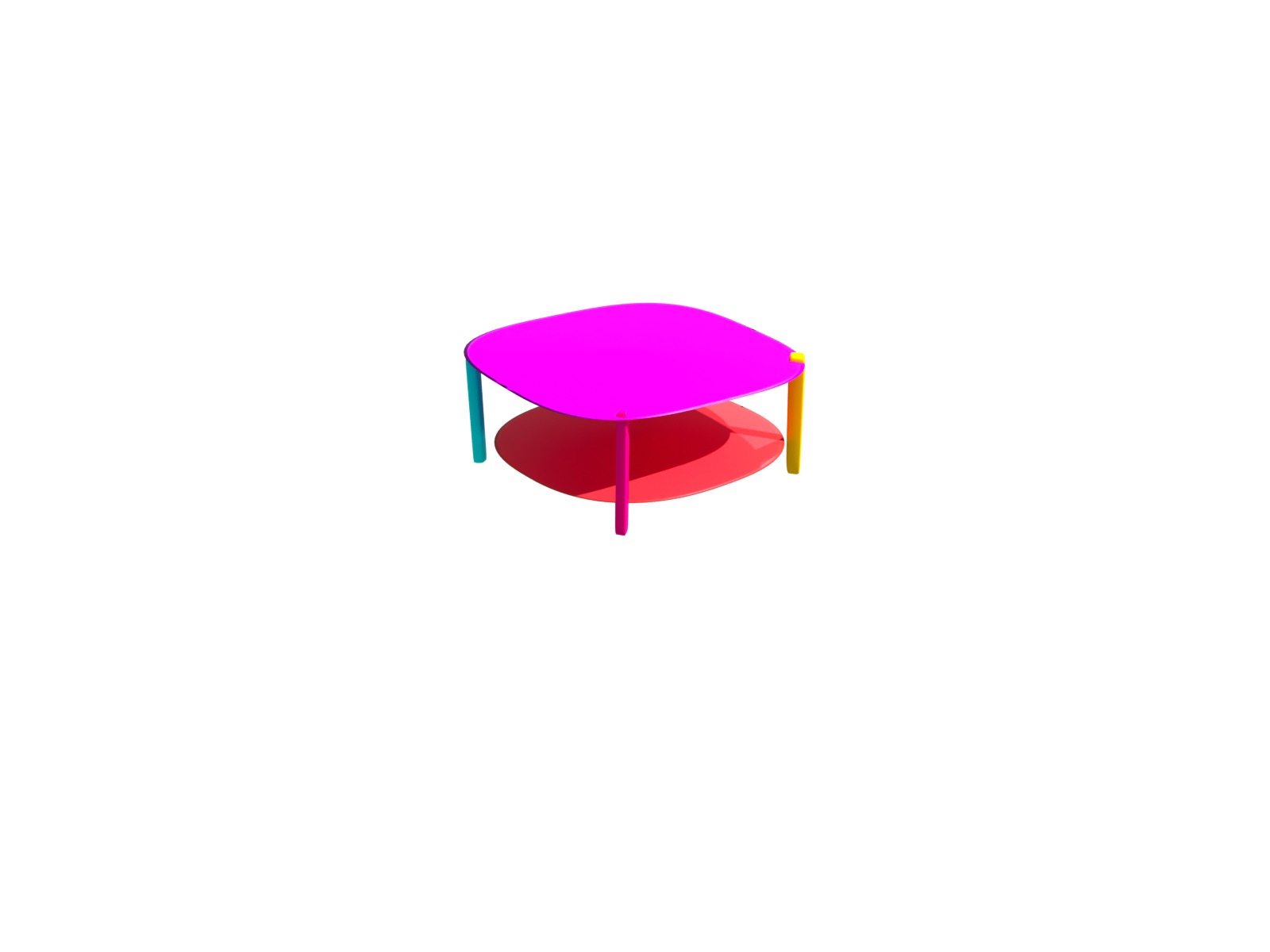}
\includegraphics[height=1.7cm,trim={680px 560px 680px 400px},clip]{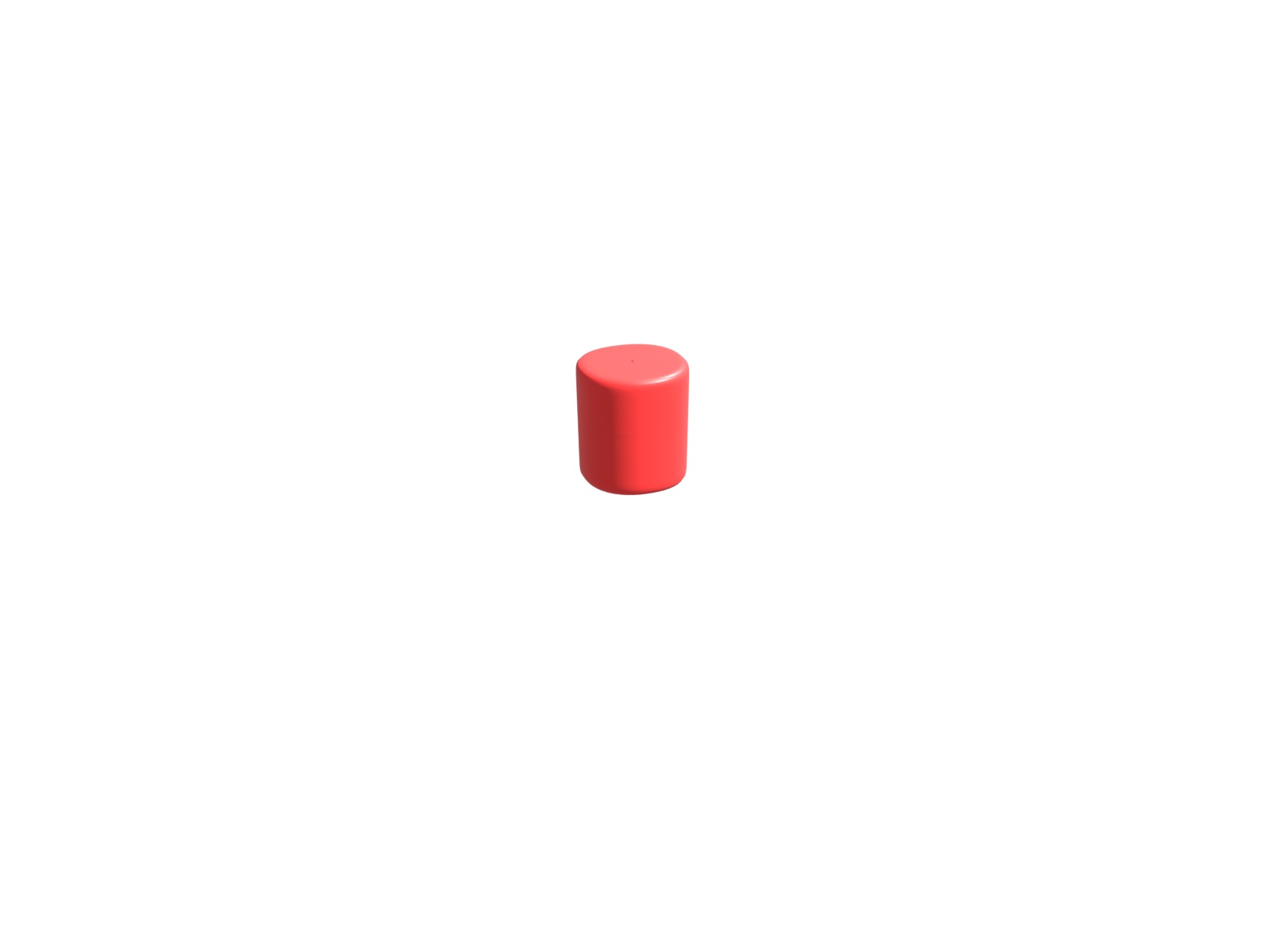}\\
\end{small}
\caption{\textbf{Qualitative Results on ASE.} \emph{Top row:} partial input  point clouds. \emph{Below:} the outputs of our standard \method{} and its robust fine-tuned variant.}
\label{fig:quali-robustness}
\end{figure}

\subsection{Partial Point Clouds}\label{sec:occlusion-evaluation}
\subsubsection{Datasets.}
We evaluate reconstruction under realistic sensing conditions using both quantitative and qualitative experiments. For quantitative evaluation, we use the Aria Synthetic Environments (ASE) dataset~\cite{pan2023aria}, while qualitative results on real-world scans are shown on ScanNet++~\cite{yeshwanth2023scannethighfidelitydataset3d}.
ASE provides complete ground truth geometry for every object, enabling accurate evaluation across all metrics. The dataset consists of complex multi-room indoor scenes populated with furniture models from the Amazon Berkeley Objects (ABO) dataset~\cite{collins2022abo}. We use the provided depth maps and instance segmentation masks to extract object point clouds exhibiting realistic viewpoint-dependent incompleteness, including self-occlusions, partial visibility, and occlusions caused by surrounding objects. These partial point clouds are provided as input to our model, and predictions are evaluated against the ground truth 3D meshes from ABO.

\subsubsection{Training Details.}
We train a robust variant of \method{} to predict deformable superquadrics from partially observed point clouds (see Sec.~\ref{sec:occlusion-robustness}). During training, we simulate realistic scanning artifacts by applying \textit{Hidden Point Removal}~\cite{katz2007direct} to model viewpoint-dependent visibility, and random spherical masking to mimic occlusions from foreground objects. To further reduce the domain gap between these synthetic augmentations and real-world occlusions, we perform a brief fine-tuning stage of 100 epochs on partial point clouds from the ASE dataset. This adaptation is especially helpful when large portions of an object are occluded.

\subsubsection{Quantitative Evaluation.}
Results are shown in Tab.~\ref{tab:occlusion}. Fine-tuning with occlusion augmentations more than doubles the IoU over the unsupervised \method{}, showing that pseudo-ground-truth superquadrics provide an effective supervision signal for partial observations. Fine-tuning on ASE further improves performance, highlighting the benefit of adapting to the target domain.

\begin{table}[t]
    \centering
        {\setlength{\tabcolsep}{6pt}%
    \begin{tabular}{lcc| cccccc}
        \toprule
        \textbf{Method} & \textbf{Occ.} & \textbf{ASE} &\textbf{IoU}$\uparrow$ & \textbf{F}~\cite{TankAndTamples}$\uparrow$ & \textbf{L1}$\downarrow$ & \textbf{L2}$\downarrow$ & \textbf{\#\,Prim}.$\downarrow$ \\
        \midrule
        SQ~\cite{paschalidou2019superquadrics} & \xmark & \xmark & $0.13$ & $0.10$& $8.62$ &$3.00$ &$10.0$\\
        CSA~\cite{Yang2021UnsupervisedLF} & \xmark & \xmark  & $0.13$ & $0.11$ & $6.81$ & $2.34$ &$8.66$\\
        EMS~\cite{liu2022robust} & \xmark & \xmark  & $0.20$& $\mathbf{0.19}$& $10.96$& $6.05$&$4.47$\\
        SuperDec \cite{fedele2025superdec} & \xmark & \xmark & $0.20$ & $0.14$ & $5.98$ & $2.45$ & $6.0$ \\ 
        \midrule
        \method{} & \xmark & \xmark & $0.20$ & $0.16$ & $5.85$ & $2.09$ & $6.3$ \\ 
        \method{} & \cmark & \xmark & $0.48$ & $0.16$ & $4.49$ & $1.62$ & $4.9$ \\ 
        \method{} & \cmark & \cmark & $\mathbf{0.54}$ & $0.18$ & $\mathbf{3.98}$ & $\mathbf{1.39}$ & $\mathbf{4.3}$ \\ 
        \bottomrule

    \end{tabular}}
    \vspace{5px}
    \caption{\textbf{Quantitative results on partial object point clouds from ASE~\cite{avetisyan2024scenescript}}. Occ. denotes fine-tuning with occlusion augmentations on ShapeNet, while ASE denotes fine-tuning on partial object point clouds from ASE.}
    \label{tab:occlusion}
\end{table}

\subsubsection{Qualitative Evaluation}
\begin{figure}[t!]
    \centering
    \includegraphics[width=0.49\linewidth]{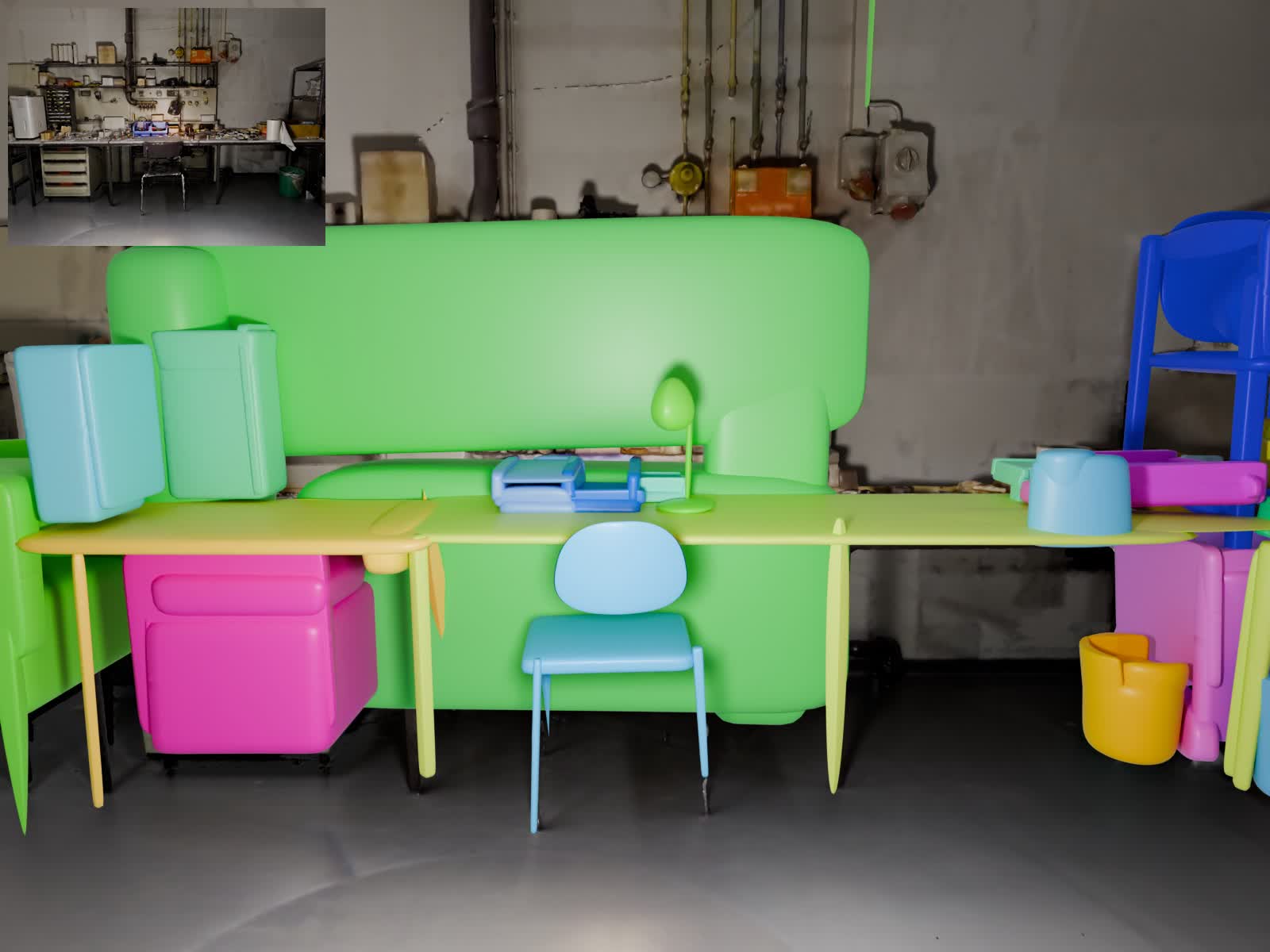}\hfill
    \includegraphics[width=0.49\linewidth]{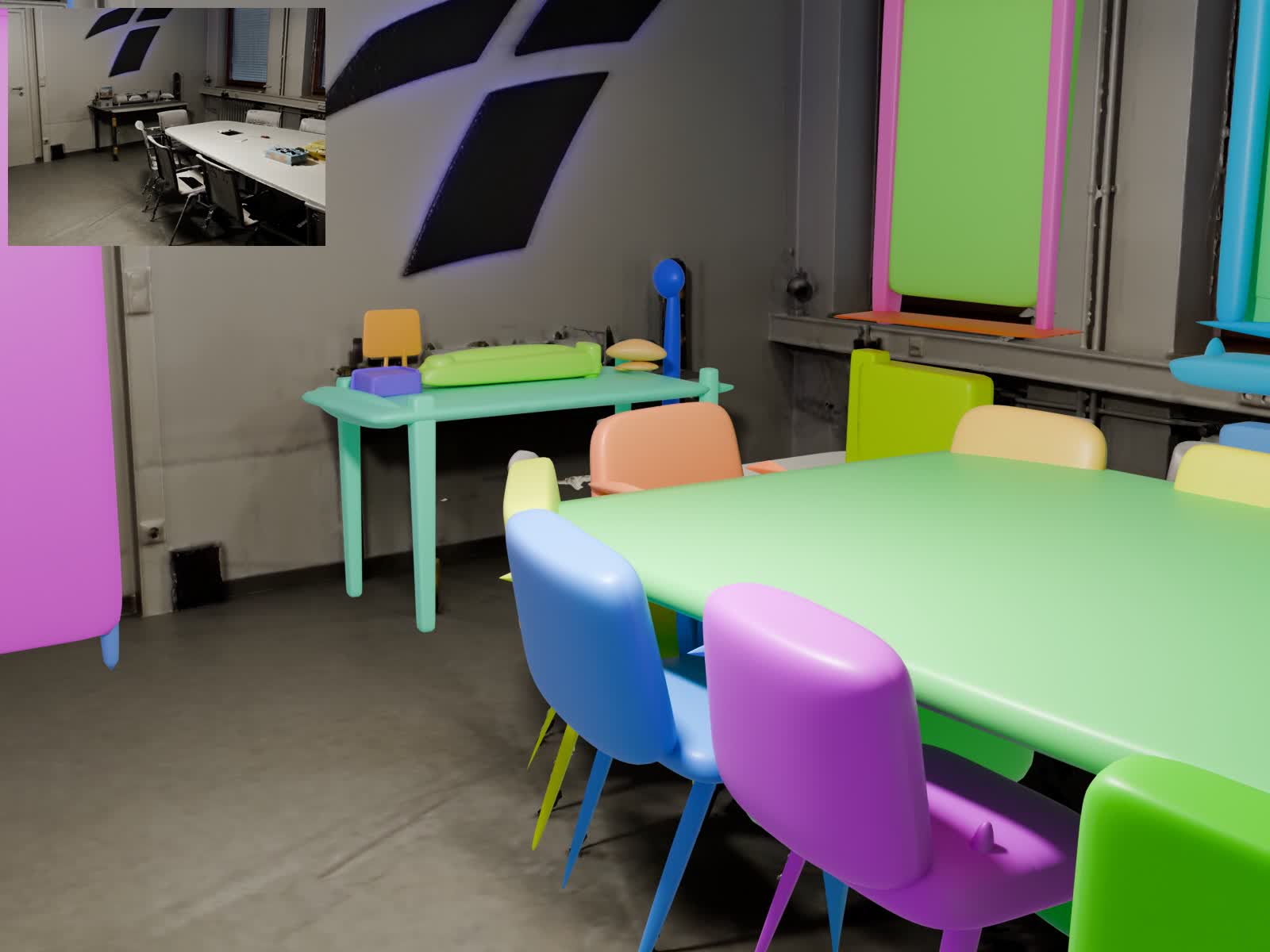}
    \vspace{-0.1cm}
\caption{\textbf{Robustness on real-world point clouds.} We show results from two ScanNet++ scenes where objects are decomposed into sets of superquadrics by our robust \method{} variant. Different colors indicate different object instances. A rendering of the original scene is shown in the top-left corner.}
    \label{fig:scenes}
\end{figure}

Figure~\ref{fig:quali-robustness} shows results on objects from the Aria Synthetic Environments (ASE) dataset~\cite{pan2023aria}, where partial point clouds are extracted from depth and instance masks with realistic occlusions and missing observations. Despite severe incompleteness, our occlusion-aware model reconstructs complete surfaces, demonstrating that the learned priors capture global object structure. Figure~\ref{fig:scenes} shows results on real-world ScanNet++ scenes, where objects are segmented using ground truth instance masks and decomposed into superquadrics with our method. These results demonstrate robustness in cluttered environments and generalization beyond synthetic data.

\subsection{Ablations}
\subsubsection{Losses.}
We conduct an ablation study to evaluate the contribution of each term in our loss function. The results, summarized in Table~\ref{loss-ablation}, demonstrate the impact of our volumetric and surface objectives on reconstruction quality and decomposition parsimony. We will explain the results line by line.
Firstly, replacing SuperDec’s~\cite{fedele2025superdec} Chamfer loss with our differentiable $\mathcal{L}_{\text{IoU}}$ improves reconstruction quality, increasing IoU (0.59 $\rightarrow$ 0.74) and F-score (0.29 $\rightarrow$ 0.40). However, this comes at the cost of increased geometric redundancy, reflected by higher overlap ($32\%$) and L2 error. This behavior arises because the volumetric objective primarily enforces global occupancy consistency, without explicitly constraining surface precision or discouraging multiple primitives from explaining the same region. As a result, thin or detailed structures are penalized less strongly than with point-based distance metrics.
Adding $\mathcal{L}_{\text{SDF}}$ introduces surface-level supervision that improves local geometric accuracy, substantially reducing L2 error (0.124 $\rightarrow$ 0.042). By aligning primitives with signed distance constraints, the model better captures fine-grained surface structure and exhibits reduced redundancy, lowering overlap from $32\%$ to $14\%$, while preserving strong volumetric performance.
Finally, $\mathcal{L}_{\text{overlap}}$ minimizes spatial redundancy, reducing overlap to $8\%$ and the average primitive count to $5.64$ without sacrificing reconstruction fidelity. These results confirm that combining volumetric, surface, and regularization terms is essential for achieving high-fidelity yet parsimonious 3D decompositions.

\begin{table}[t]
    \centering
    \setlength{\tabcolsep}{5pt}
    \begin{tabular}{l | cccccc}
        \toprule
        \textbf{Method} & \textbf{IoU}$\uparrow$ & \textbf{F}~\cite{TankAndTamples}$\uparrow$ & \textbf{L1}$\downarrow$ & \textbf{L2}$\downarrow$ & \textbf{\#\,Prim}.$\downarrow$ & \textbf{Overlap}$\downarrow$ \\
        \midrule
        $\mathcal{L}_{\text{SuperDec}}$ \cite{fedele2025superdec}  & $0.59$ & $0.29$ & $1.75$ & $0.050$ & $5.78$ & $4\%$ \\
        $\mathcal{L}_{\text{IoU}}$  & $\mathbf{0.74}$ & $\mathbf{0.40}$ & $1.76$ & $0.124$ & $5.94 $& $32\%$ \\ 
        $\mathcal{L}_{\text{IoU}} + \mathcal{L}_{\text{SDF}}$  & $0.72$ & $0.37$ & $1.54$ & $\mathbf{0.042}$ & $5.80$ & $14\%$ \\ 
        $\mathcal{L}_{\text{IoU}} + \mathcal{L}_{\text{SDF}} + \mathcal{L}_{\text{overlap}}$  & $0.72$ & $0.37$ & $\mathbf{1.54}$ & $0.043$ & $\mathbf{5.64}$ & $\mathbf{8\%}$ \\  
        \bottomrule
    \end{tabular}
    \vspace{5px}
    \caption{\textbf{Loss Ablation Study}. 
    \label{loss-ablation} We ablate the contribution of each term in our loss function. We report baseline performances training our model with the reconstruction loss from SuperDec (first row) and the improvements obtained by substituting it with our loss terms, which we incrementally introduce (following rows). We keep $\lsparsity$ and $\lexist$ fixed in all experiments, as they are essential to the model architecture.
    }
\end{table}

\subsubsection{Deformations}
\begin{table}[t]
    \centering
    \setlength{\tabcolsep}{12pt}
    \begin{tabular}{cc ccccc}
        \toprule
        \textbf{Tapering}  & \textbf{Bending} &\textbf{IoU}$\uparrow$ & \textbf{F}~\cite{TankAndTamples}$\uparrow$ & \textbf{L1}$\downarrow$ & \textbf{L2}$\downarrow$ \\
        \midrule
        \xmark & \xmark & $0.81$ & $0.44$ & $1.46$ & $0.041$ \\
        \cmark & \xmark & $0.83$ & $0.45$ & $1.40$ & $0.038$ \\
        \xmark & \cmark & $0.84$ & $0.46$ & $1.39$ & $0.037$ \\ 
        \cmark & \cmark & $\mathbf{0.85}$ & $\mathbf{0.47}$ & $\mathbf{1.35}$ & $\textbf{0.036}$ \\ 
        \bottomrule
    \end{tabular}
    \vspace{5px}
    \caption{\textbf{Ablation Study on Bending and Tapering}. 
    We evaluate the impact of tapering and bending deformations. We start from undeformed feed-forward predictions and we evaluate different configurations. Results are computed on ShapeNet, starting from undeformed predictions.
    }
    \label{tab:optimization-deformation}
\end{table}

To validate the impact of our refinement procedure and the additional expressivity introduced by the extended superquadric parameterization, we perform an ablation study on the ShapeNet test split. Starting from the undeformed feedforward initializations, we refine the decompositions using different deformation parameterizations. Specifically, we analyze the contribution of tapering and bending during the optimization stage (Table~\ref{tab:optimization-deformation}).
Introducing tapering alone leads to a modest improvement in reconstruction quality; similarly, enabling 3-axis bending results in a further gain across all metrics, indicating that these deformations capture geometric variations not well represented by the base superquadric formulation. When both deformations are enabled, the model achieves its best performance, increasing IoU from 0.81 to 0.85 while consistently reducing reconstruction errors. Although the quantitative gains, particularly for tapering, appear incremental, qualitative inspection (see Fig.~\ref{fig:quali-shapenet} and Fig.~\ref{fig:optimization}) reveals that these deformations are crucial for capturing primitives with gradually varying cross-sections or curved profiles. These local refinements are often highly effective visually, even if they are not the primary drivers of global metrics such as IoU.
Overall, the study confirms that deformation parameters substantially improve the expressivity of superquadric primitives during refinement.

\section{Conclusion}
\label{sec:conclusion}
We presented \name{}, a method for superquadric-based 3D reconstruction that introduces tapering and bending deformations to improve reconstruction fidelity. Our approach combines a novel joint volumetric and surface loss with an efficient refinement stage, enabling accurate primitive decompositions while remaining substantially faster than prior optimization-based methods. Leveraging these high-quality decompositions as pseudo-ground truth, we fine-tune a model variant that predicts complete object structures from noisy and partial real-world observations. Extensive experiments demonstrate that \name{} preserves the compactness of parametric representations while overcoming key limitations of prior superquadric-based approaches, making it an effective and scalable representation for 3D scene understanding.

\subsubsection{Acknowledgments:}
This work was supported by a SwissAI Grant for Small Projects and a GPU grant from NVIDIA. Elisabetta Fedele is supported by the ETH AI Center doctoral fellowship and by the Swiss National Science Foundation (SNSF) Advanced Grant 216260 (\textit{Beyond Frozen Worlds: Capturing Functional 3D Digital Twins from the Real World}).

\bibliographystyle{splncs04}
\bibliography{main}
\clearpage
\appendix

\begin{center}
    {\Large \textbf{\method{}: Deformable Superquadrics for Point Cloud Decomposition}}\\[1em]
    {\normalsize Supplementary Material}
\end{center}
\renewcommand\thesection{\Alph{section}}

\setcounter{page}{1}
\setcounter{section}{0}
\begin{abstract}
We provide further explanations on the superquadric parameterization in Sec.~\ref{sec:sq-supp}, additional method details in Sec.~\ref{sec:additionl-method-details}, additional qualitative results in Sec.~\ref{sec:quali-supp}, and an interesting downstream application of our method in Sec.~\ref{sec:sam3}.
\end{abstract}

\section{Superquadrics and Deformations}\label{sec:sq-supp}
\subsection{Tapering and Bending formulation}
\label{tb-formulation}
To evaluate the implicit form of a \textit{deformable superquadric}, parameterized by
$\Theta=\{\mathbf{t}, R, \mathbf{s}, \boldsymbol{\epsilon}, \boldsymbol{\tau}, \boldsymbol{\beta\}}$, 
we first apply the rigid transformation, followed by bending with respect to the $z$-, $x$-, and $y$-axes, then the tapering operator, and finally the implicit function for canonical superquadrics (Eq. \eqref{e-can-sq}):
\begin{align*}
q(\mathbf{x}; \Theta)
=
q_{\text{can}}\Big( \mathcal{T}^{-1}\circ \mathcal{B}_{y}^{-1}\circ\mathcal{B}_{x}^{-1}\circ \mathcal{B}_{z}^{-1}\big(R^\top(\mathbf{x}-\mathbf{t})\big); \mathbf{s}, \boldsymbol{\epsilon}\Big)\, .
\end{align*}
Here, the inverse tapering operator $\mathcal{T}^{-1}$ and the inverse bending operator $\mathcal{B}^{-1}_{z}$ are defined as:
\begin{align*}
\mathcal{T}^{-1}(x,y,z) \!=\!
\begin{pmatrix}
x/\left(\frac{\tau_x}{s_z}z+1 \right) \\
y/\left(\frac{\tau_y}{s_z}z+1 \right) \\
z
\end{pmatrix},\quad 
\mathcal{B}_{z}^{-1}(x,y,z) \!=\!
\begin{pmatrix}
x-(R-r)\cos{ \alpha_z} \\
y-(R-r)\sin {\alpha_z} \\
k_z^{-1}\gamma
\end{pmatrix}\, ,
\end{align*}
where $\boldsymbol{\tau}=(\tau_x,\tau_y)$ are the tapering parameters along the $x$- and $y$-axes, and $s_z$ is the superquadric scale along the $z$-axis.
For bending with curvature $k_z$ and bending plane orientation $\alpha_z$,
the auxiliary quantities are defined as follows:
\begin{align*}
\gamma\!=\! \arctan\frac{z}{k_z^{-1}\!\!-\!R},\,
r\!=\! k_z^{-1} \!\!-\! \|(z, k_z^{-1} \!\!-\! R)\|,\,\text{and }
R\!=\! \|(x,y)\| \cos{\alpha_z \!-\! \arctan{\frac{y}{x}}}\, .
\end{align*}
The remaining inverse bending operators $\mathcal{B}_{x}^{-1}$ and $\mathcal{B}_{y}^{-1}$ are defined analogously, yielding six bending parameters $ \boldsymbol{\beta}= (k_x, \alpha_x,k_y, \alpha_y, k_z, \alpha_z)$. 
We found the single-axis tapering to be expressive enough. Note that superquadrics also have an explicit function, where the same transformations must be applied in reverse order, namely tapering, then bending, followed by the inverse rigid transformation. We refer to \cite{jaklic2000segmentation} for additional details.

\subsection{Geometric interpretation of deformations}
This section provides a geometric interpretation of the tapering and bending deformation parameters. While in \ref{tb-formulation} we define the inverse operators $\mathcal{T}^{-1}$ and $\mathcal{B}^{-1}$ mathematically, the corresponding parameters can be understood more intuitively through their geometric effect on the canonical superquadric shape.

\subsubsection{Tapering Parameters.} 
The tapering operator linearly scales the cross-section of the superquadric along a specified primary axis, effectively breaking the constant-thickness symmetry of the canonical shape. The tapering coefficients \(\tau_x\) and \(\tau_y\) control the linear expansion or contraction of the primitive along the \(x\)- and \(y\)-axes, respectively, as a function of the position along the \(z\)-axis.

\subsubsection{Bending Parameters.} The bending operator maps the straight central axis of the superquadric onto a circular arc without altering the cross-sectional shape orthogonal to that axis. The curvature parameter $k_z$ quantifies the sharpness of the shape's bend; its inverse, $\frac{1}{k_z}$, represents the radius of the circular arc formed by the deformed $z$-axis. The angle $\alpha_z$ dictates the azimuthal direction of the bend. It defines the orientation of the plane (relative to the $x$ and $y$ axes ) into which the superquadric is deformed.

\subsection{Is the bending direction needed?}

To investigate the necessity of the bending plane orientation $\alpha$, we evaluate its impact on reconstruction fidelity. The bending operator $\mathcal{B}^{-1}$ relies on $\alpha$ to define the orientation of the bending plane. As shown in Table~\ref{tab:optimization-overparameterization}, we tested a configuration where bending is enabled, but the displacement parameter $\alpha$ is omitted. This setup performs worse than the full parametrization, highlighting that simply defining a curvature $k$ is insufficient for geometric expressiveness. The specific mathematical formulation of the bend, enabled by $\alpha$, is necessary to shift the center of deformation and capture the curvature of complex parts. By including $\alpha$, the model effectively decouples a primitive's rigid rotation from its bending deformation, simplifying both the learning and optimization process.

\begin{table}[t]
    \centering
    \setlength{\tabcolsep}{4pt}
    \begin{tabular}{lc ccccc}
        \toprule
        Method &\textbf{IoU}$\uparrow$ & \textbf{F}~\cite{TankAndTamples}$\uparrow$ & \textbf{L1}$\downarrow$ & \textbf{L2}$\downarrow$ \\
        \midrule
        Without $\alpha$ & 0.83 & 0.47 & 1.36 & 0.036 \\
        With $\alpha$ & \textbf{0.86} & \textbf{0.49} & \textbf{1.31} & \textbf{0.034} \\
        \bottomrule
    \end{tabular}
    \vspace{5px}
    \caption{\textbf{Bending Direction Ablation Study}. 
    We compare model performance with and without the bending direction parameters $\alpha$. Results are reported on a subset of the ShapeNet test set. L1 and L2 are multiplied by $10^2$.
    }
    \label{tab:optimization-overparameterization}
\end{table}

\subsection{Necessity of Multi-Axis Bending}

Our multi-axis bending formulation addresses the inherent sensitivity of single-axis bending to a primitive's initial rigid orientation. By decoupling the bending deformation from the superquadric's local coordinate system, both the feed-forward model and the optimization can apply curvature along any spatial direction, regardless of the primitive's current alignment, significantly simplifying the optimization landscape.

\begin{table}[t]
    \centering
    \setlength{\tabcolsep}{6pt}
    \begin{tabular}{c cccc}
        \toprule
        Axis Pruned & \textbf{IoU}$\uparrow$ & \textbf{F}~\cite{TankAndTamples}$\uparrow$ & \textbf{L1}$\downarrow$ & \textbf{L2}$\downarrow$ \\
         \midrule
        0 & $\mathbf{0.87}$ & $\mathbf{0.49}$ & $\mathbf{1.32}$ & $\mathbf{0.036}$ \\
        1 & $0.86$ & $\mathbf{0.49}$ & $\mathbf{1.32}$ & $\mathbf{0.036}$ \\
        2 & $0.84$ & $0.48$ & $1.35$ & $0.037$ \\
        \bottomrule
    \end{tabular}
    \vspace{5px}
    \caption{\textbf{Bending Axis Ablation Study}. 
    We compare model performance before and after pruning the bending axis with the smallest magnitude. Results are reported on the ShapeNet test set. L1 and L2 are multiplied by $10^2$.
    }
    \label{tab:bending-pruning}
\end{table}

As shown in Table \ref{tab:bending-pruning}, we evaluate the impact of this flexibility by pruning the bending axes with the smallest magnitudes after the optimization stage. While retaining all three axes yields the highest performance, we note that pruning the weakest axis (leaving 2 active axes) incurs a negligible penalty. This suggests that while all three axes are utilized for maximal fidelity, many primitives can be represented more compactly without a significant loss in reconstruction accuracy. However, a sharper drop occurs when restricted to a single axis (2 axes pruned), indicating that 2 bent axis are not uncommon.

\section{Additional Method Details}
\label{sec:additionl-method-details}
\subsection{Unsupervised Occlusion Robust Training}
\label{sec:unsup-robust}
While the main text focuses on supervised structural completion, we also explore an unsupervised robust variant of \method{} aimed at evaluating local fidelity under noise and partial visibility. Rather than predicting missing geometry through learned structural priors, the unsupervised model applies the standard loss $\mathcal{L}$ (Sec.~\ref{sec:loss}) directly to the augmented, occluded point clouds. Because this regime lacks ground-truth targets for occluded regions, the model is not expected to hallucinate missing structures,  it instead acts as a robust surface fitter. This approach yields a clean, primitive-based decomposition optimized to tightly capture only the visible fragments while remaining resilient to scanning artifacts, making it particularly effective for high-fidelity local alignment or real-world complex scenes.

\subsubsection{Evaluation}
We compare our unsupervised robust model against the supervised network on partially occluded point clouds. Quantitatively (Table~\ref{tab:occlusion-unsup}), the supervised approach achieves a $21.5\%$ lower $L_2$ error ($1.39$ vs $1.77$), an $11.1\%$ lower $L_1$ error ($3.98$ vs $4.48$), and an improved F-score ($0.18$ vs $0.16$). 

This performance gap stems directly from the training regimes. Supervising the network with complete pseudo-ground truths enables it to learn structural priors and reconstruct completely missing or hidden object parts, a capability most clearly reflected in the lower Chamfer distances. Conversely, the unsupervised variant lacks a mechanism to hallucinate hidden structures and acts instead as a tight surface fitter for visible fragments, achieving competitive metrics especially for IoU. This fragment-fitting behavior translates well to real-world data, yielding clean but partial object decompositions on cluttered ScanNet++ scenes (Fig.~\ref{fig:scannet_qualitative}). We also show qualitative results on Replica scenes in Fig.~\ref{fig:replica_qualitative}.

\begin{table}[t]
    \centering
    \setlength{\tabcolsep}{6pt}
    \begin{tabular}{lcc| cccccc}
        \toprule
        Method & Occ. & ASE &\textbf{IoU}$\uparrow$ & \textbf{F}~\cite{TankAndTamples}$\uparrow$ & \textbf{L1}$\downarrow$ & \textbf{L2}$\downarrow$ & \textbf{\#\,Prim}.$\downarrow$ \\
        \midrule
        \method{} & \xmark & \xmark & $0.20$ & $0.16$ & $5.85$ & $2.09$ & $6.3$ \\ 
        \method{} Unsupervised & \cmark & \xmark & $0.47$ & $0.13$ & $4.92$ & $1.80$ & $4.0$ \\ 
        \method{} & \cmark & \xmark & $0.48$ & $0.16$ & $4.49$ & $1.62$ & $4.9$ \\ 
        \method{} Unsupervised & \cmark & \cmark & $0.53$ & $0.16$ & $4.48$ & $1.77$ & $3.5$ \\ 
        \method{} & \cmark & \cmark & $0.54$ & $0.18$ & $3.98$ & $1.39$ & $4.3$ \\ 
        \bottomrule

    \end{tabular}
    \vspace{5px}
    \caption{\textbf{Comparison of Supervised and Unsupervised Robust Models}. We compare methods on partial object point clouds from ASE. Occ. denotes fine-tuning with occlusion augmentations on ShapeNet, ASE fine-tuning on partial object point clouds from ASE.}
    \label{tab:occlusion-unsup}
\end{table}

\begin{figure*}
    \centering
    \setlength{\tabcolsep}{2pt}
    \begin{tabular}{ccc}
        & \textbf{Point Cloud} & \textbf{Superquadrics} \\
         \rotatebox{90}{\hspace{30pt}\textbf{3f1e1610de}} &
        \includegraphics[width=0.45\linewidth]{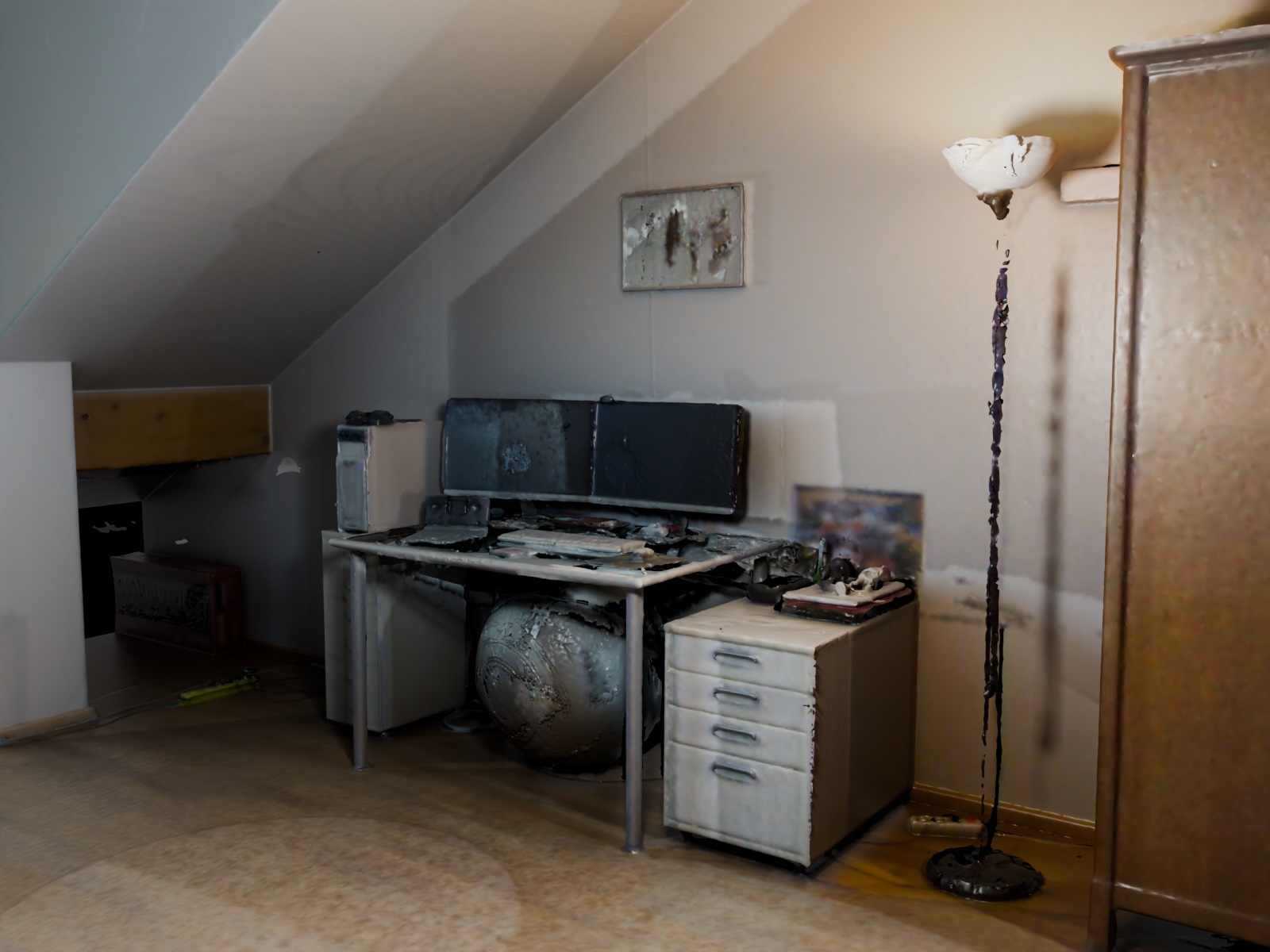} &
        \includegraphics[width=0.45\linewidth]{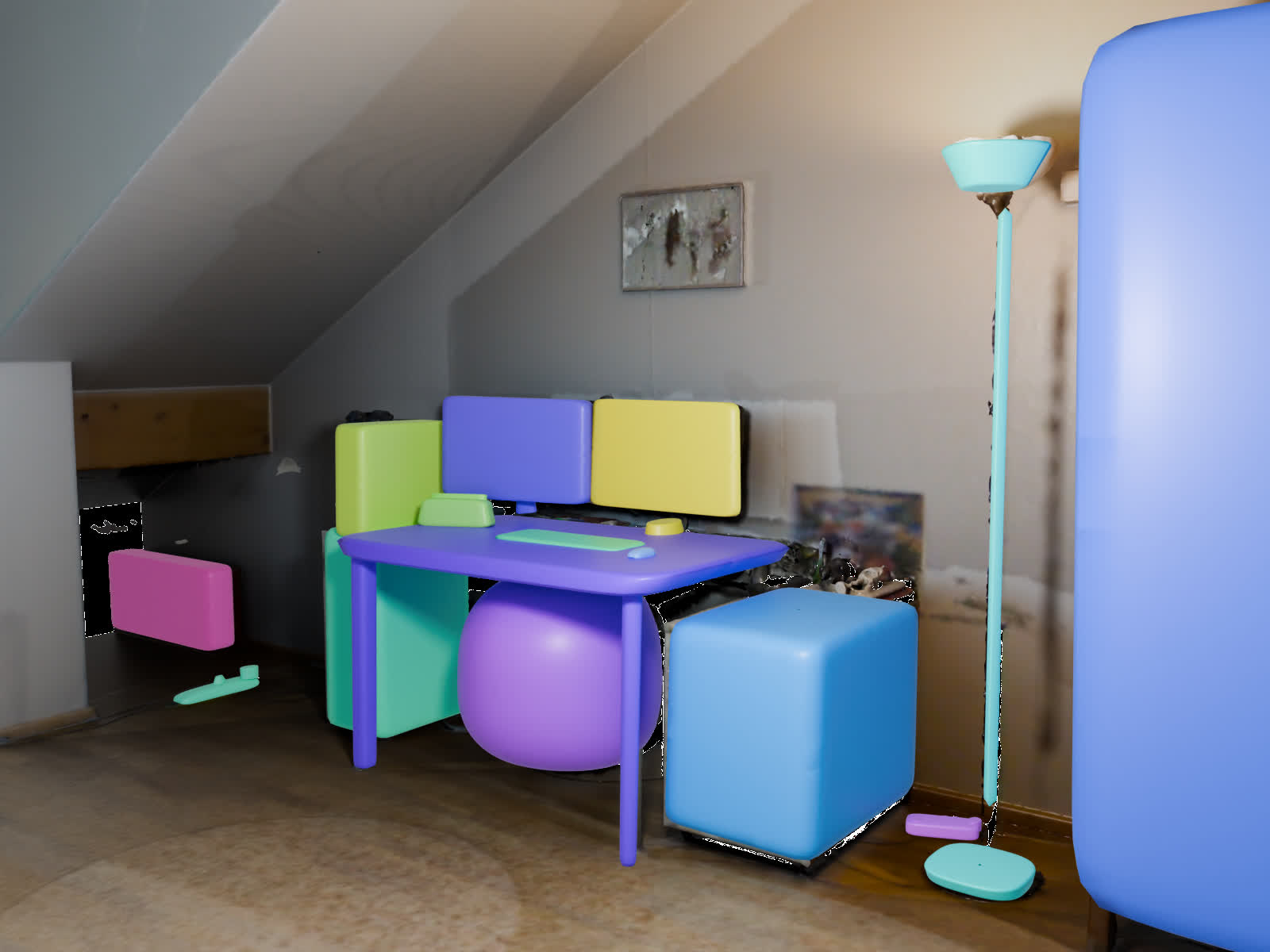} \\
        \rotatebox{90}{\hspace{30pt}\textbf{4d451d9c36}} &
        \includegraphics[width=0.45\linewidth]{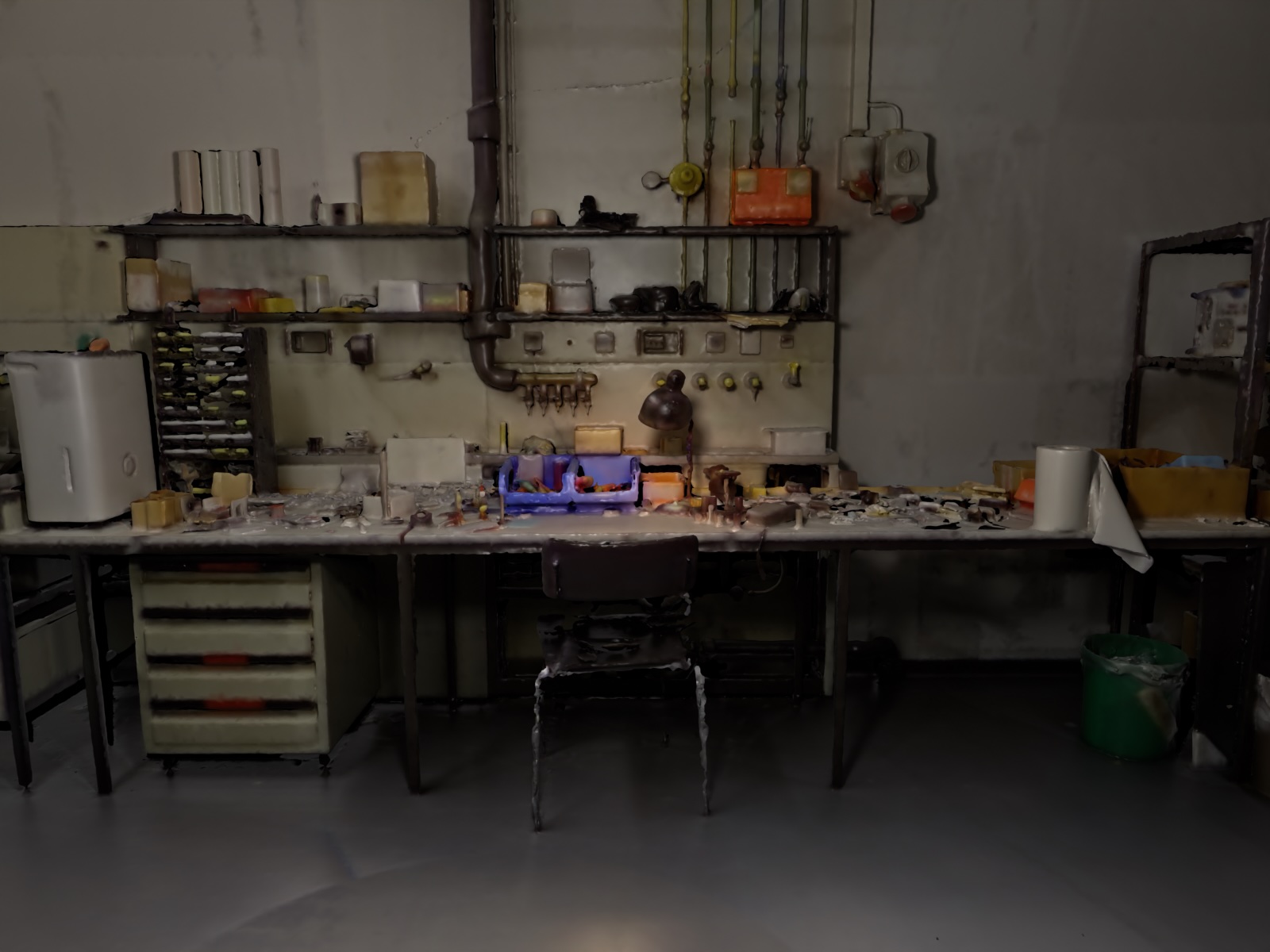} &
        \includegraphics[width=0.45\linewidth]{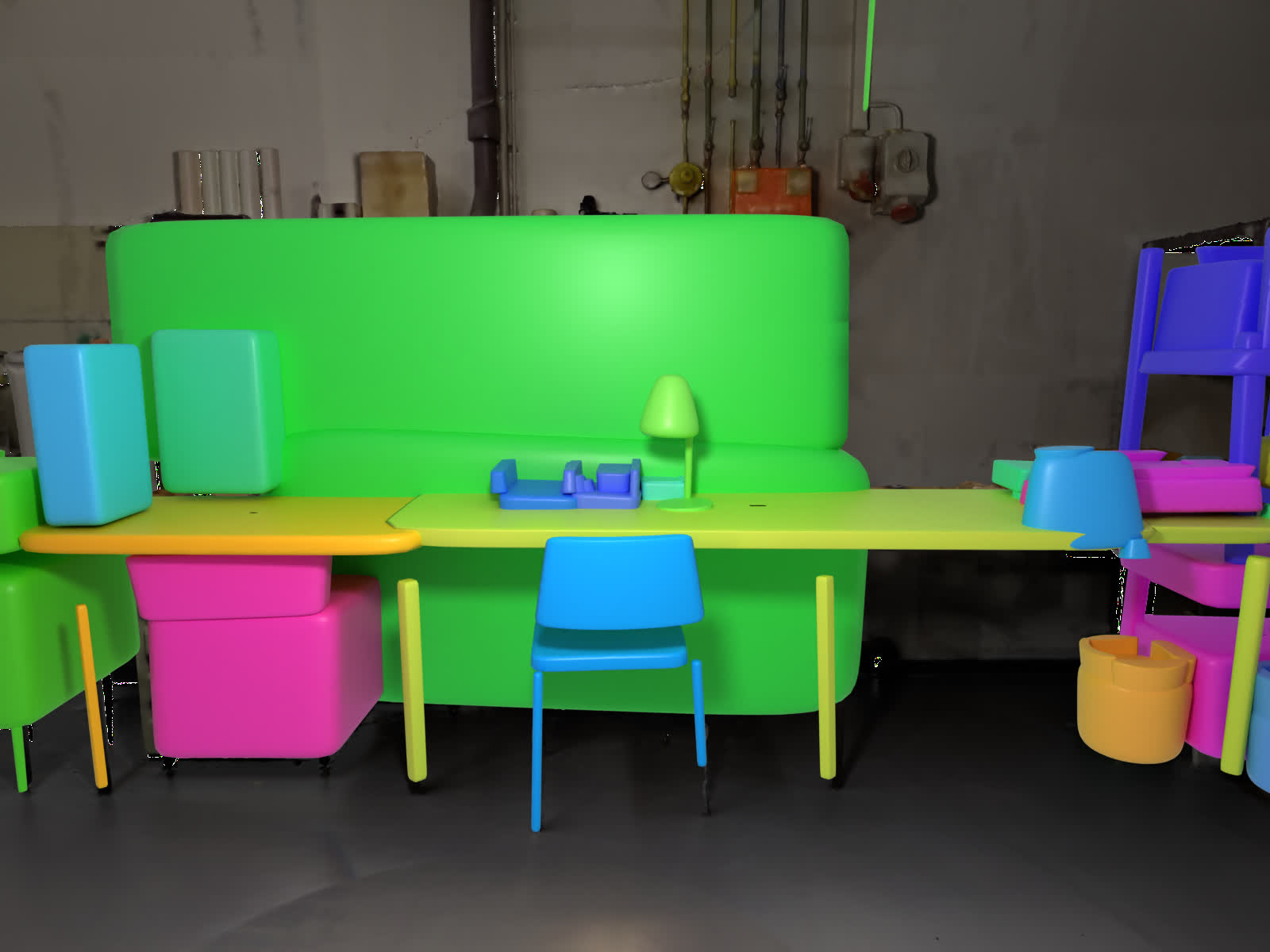} \\
        \rotatebox{90}{\hspace{30pt}\textbf{6b40d1a939}} &
        \includegraphics[width=0.45\linewidth]{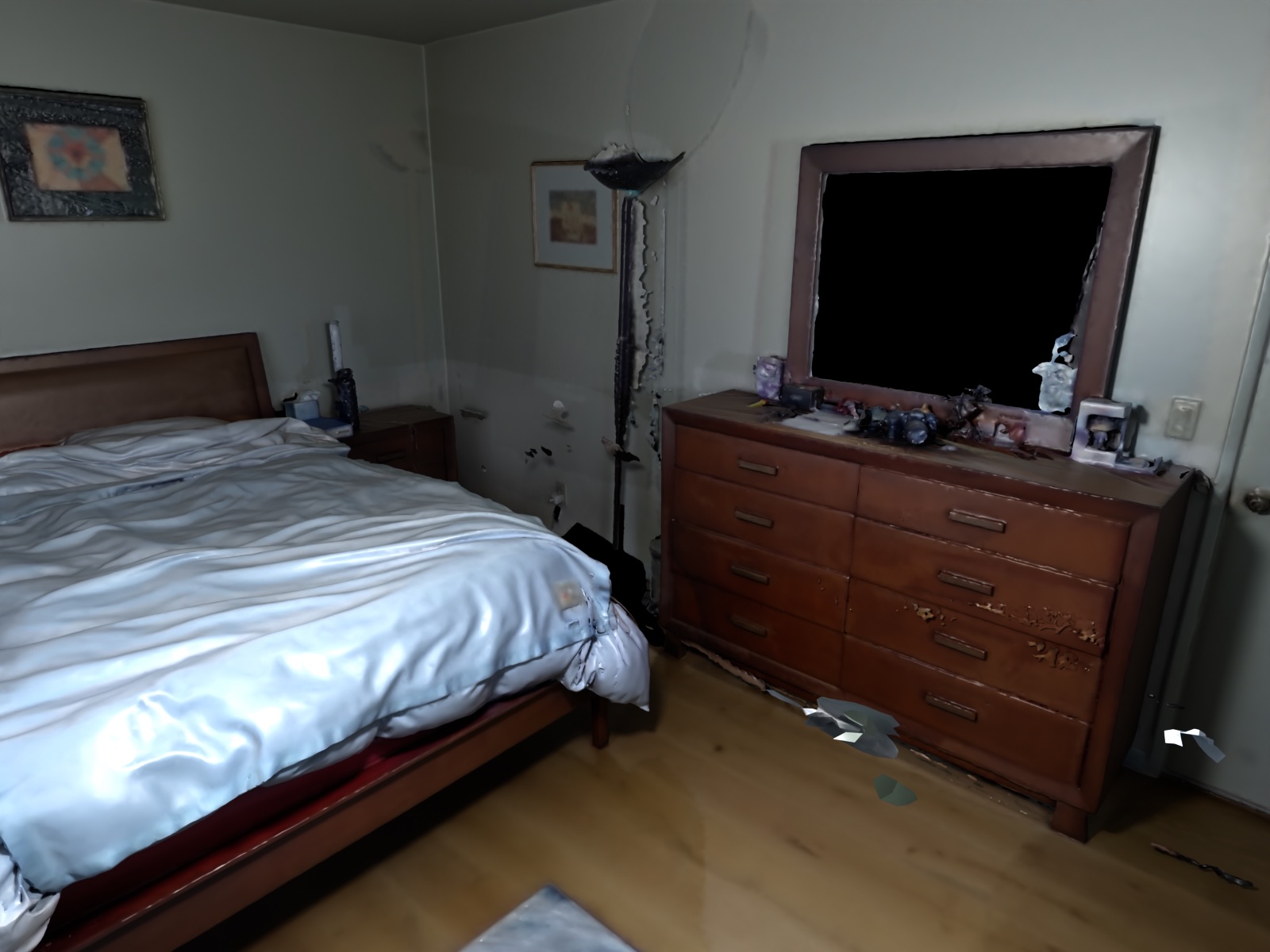} &
        \includegraphics[width=0.45\linewidth]{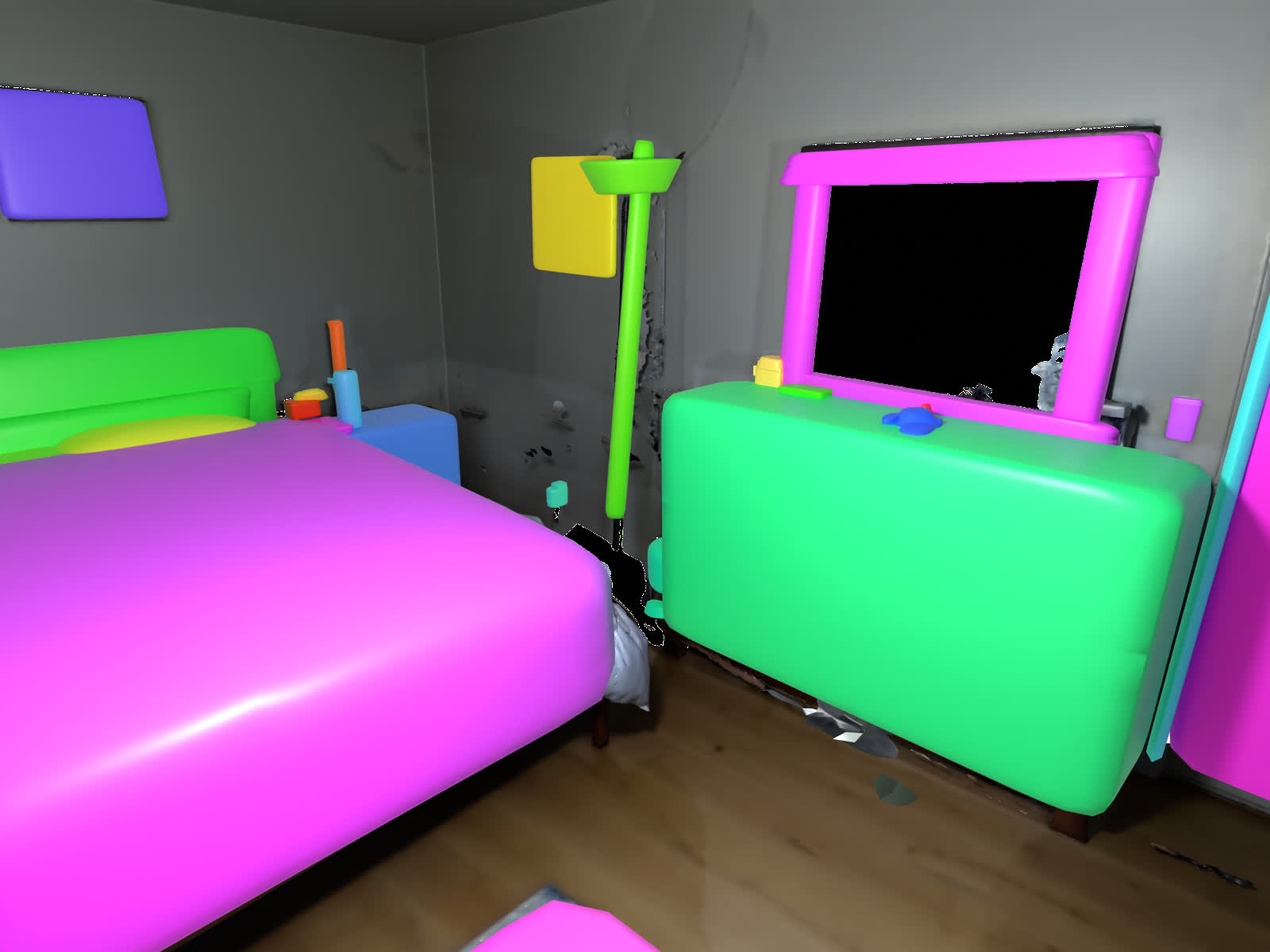} \\
        \rotatebox{90}{\hspace{30pt}\textbf{95748dd597}} &
        \includegraphics[width=0.45\linewidth]{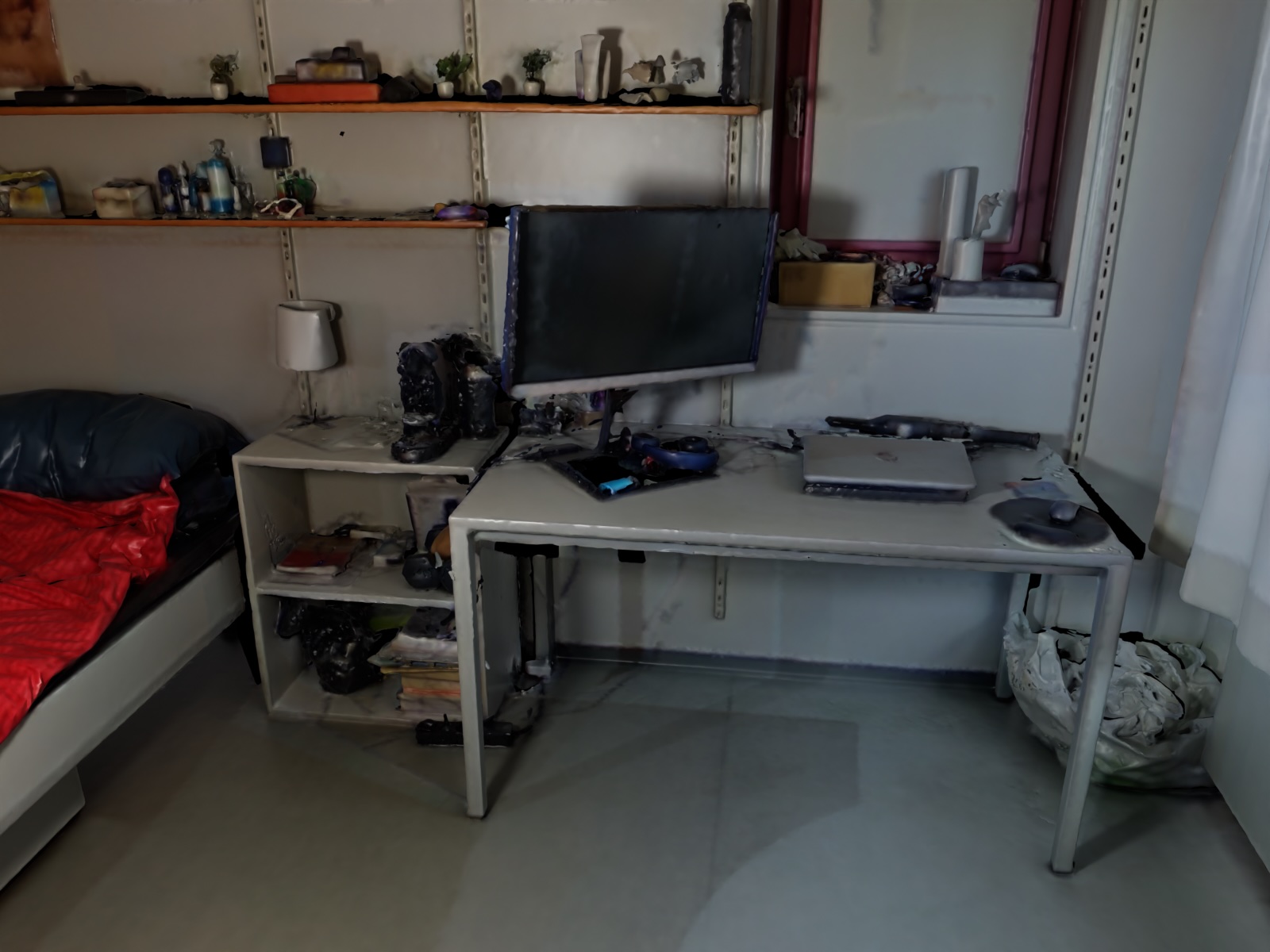} &
        \includegraphics[width=0.45\linewidth]{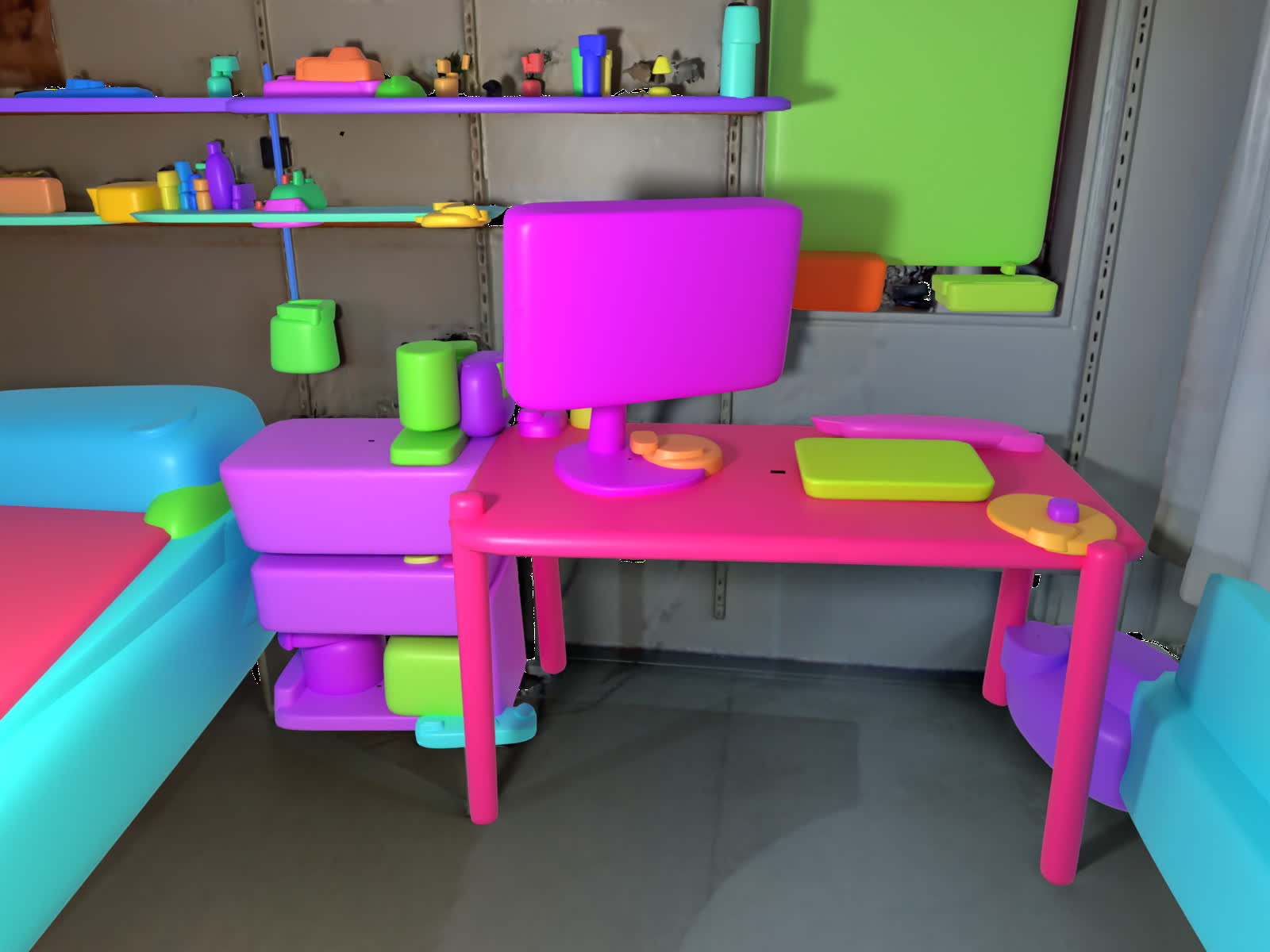}
    \end{tabular}
    \caption{\textbf{Robustness in real-world point clouds.} We show four example scene from ScanNet++ where each object is decomposed into a set of superquadrics.}
    \label{fig:scannet_qualitative}
\end{figure*}

\begin{figure*}
    \centering
    \setlength{\tabcolsep}{2pt}
    \begin{tabular}{ccc}
        & \textbf{Point Cloud} & \textbf{Superquadrics} \\
         \rotatebox{90}{\hspace{40pt}\textbf{Room 0}} &
        \includegraphics[width=0.45\linewidth]{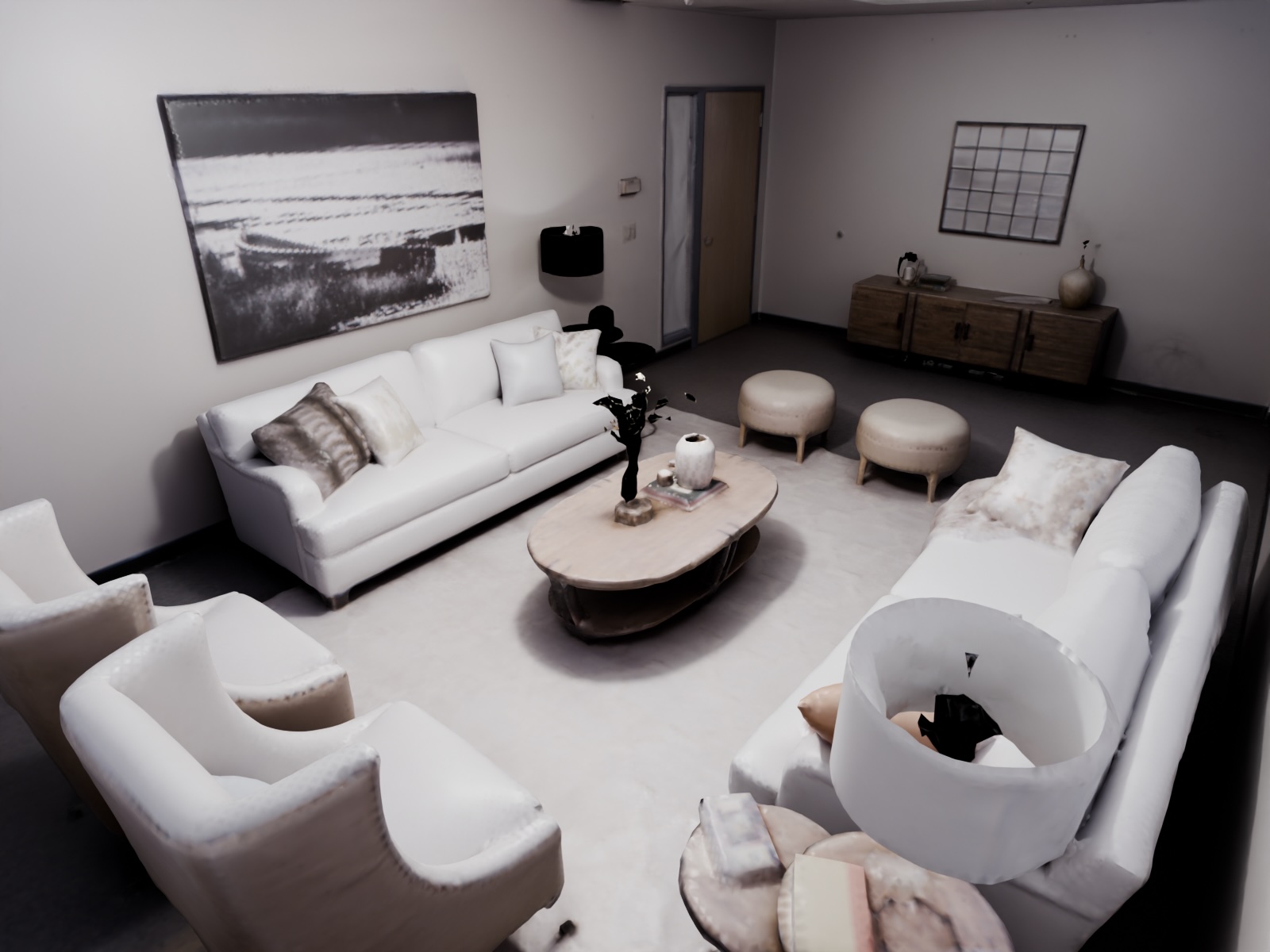} &
        \includegraphics[width=0.45\linewidth]{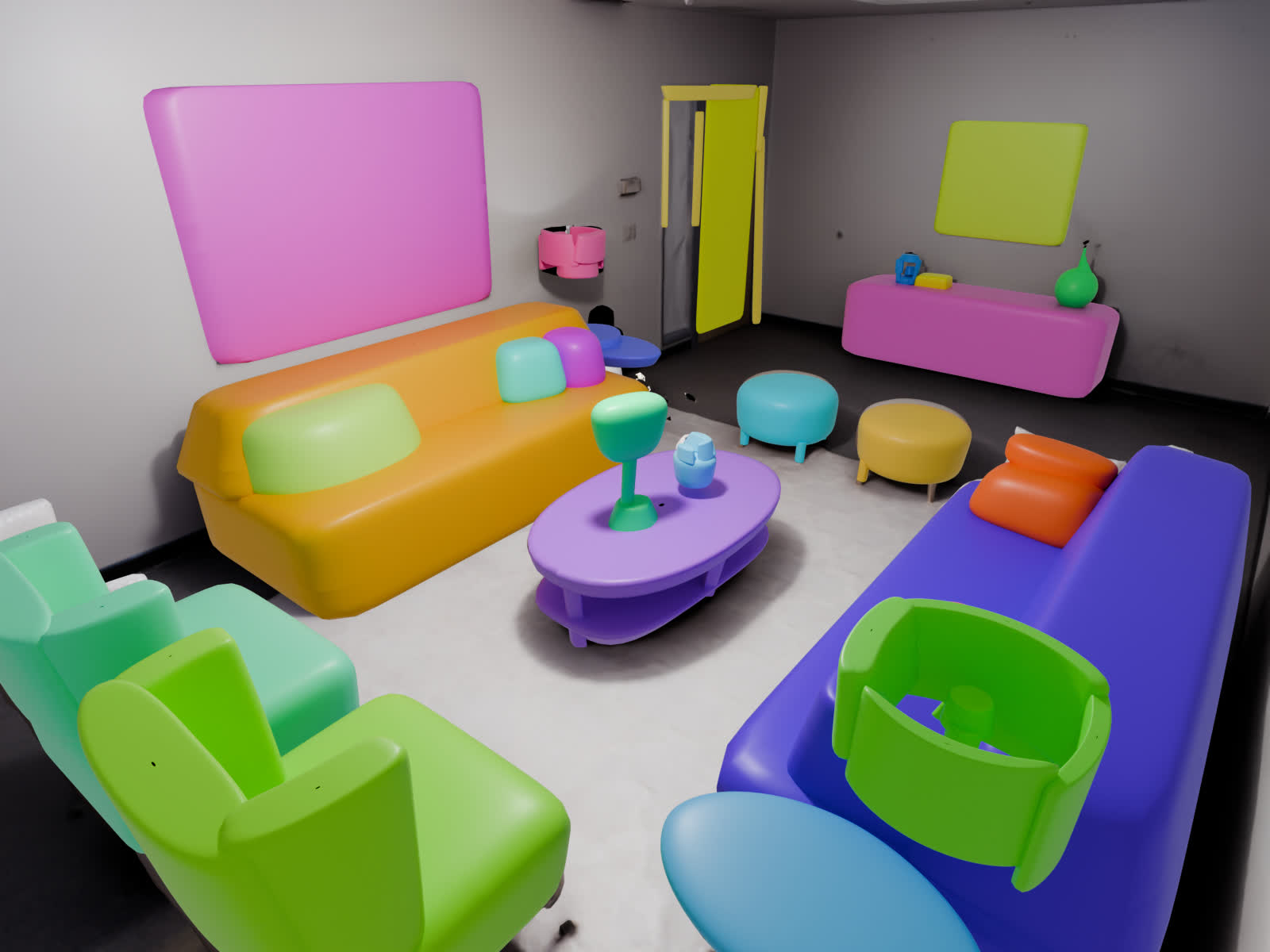} \\
         \rotatebox{90}{\hspace{40pt}\textbf{Office 1}} &
        \includegraphics[width=0.45\linewidth]{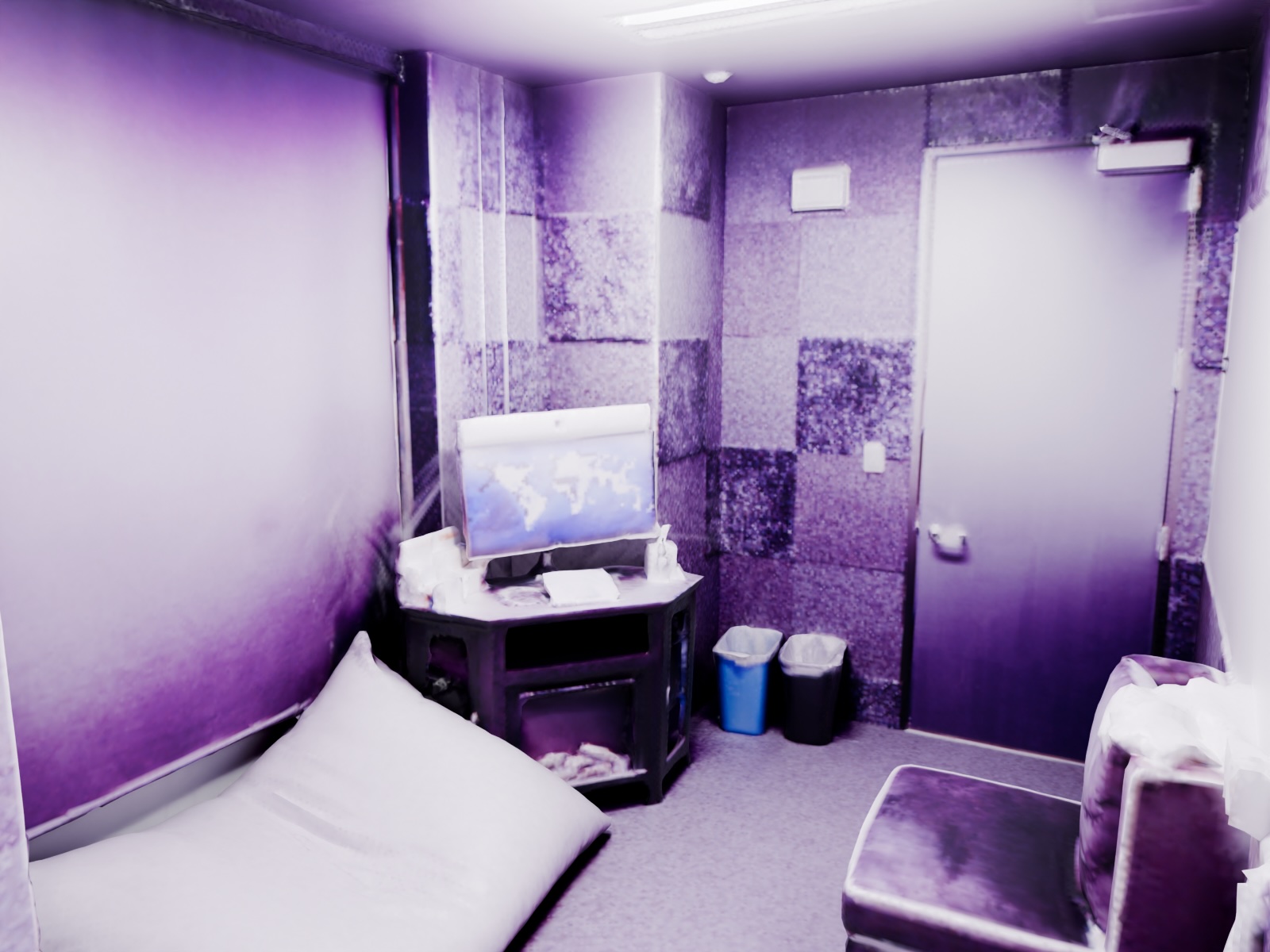} &
        \includegraphics[width=0.45\linewidth]{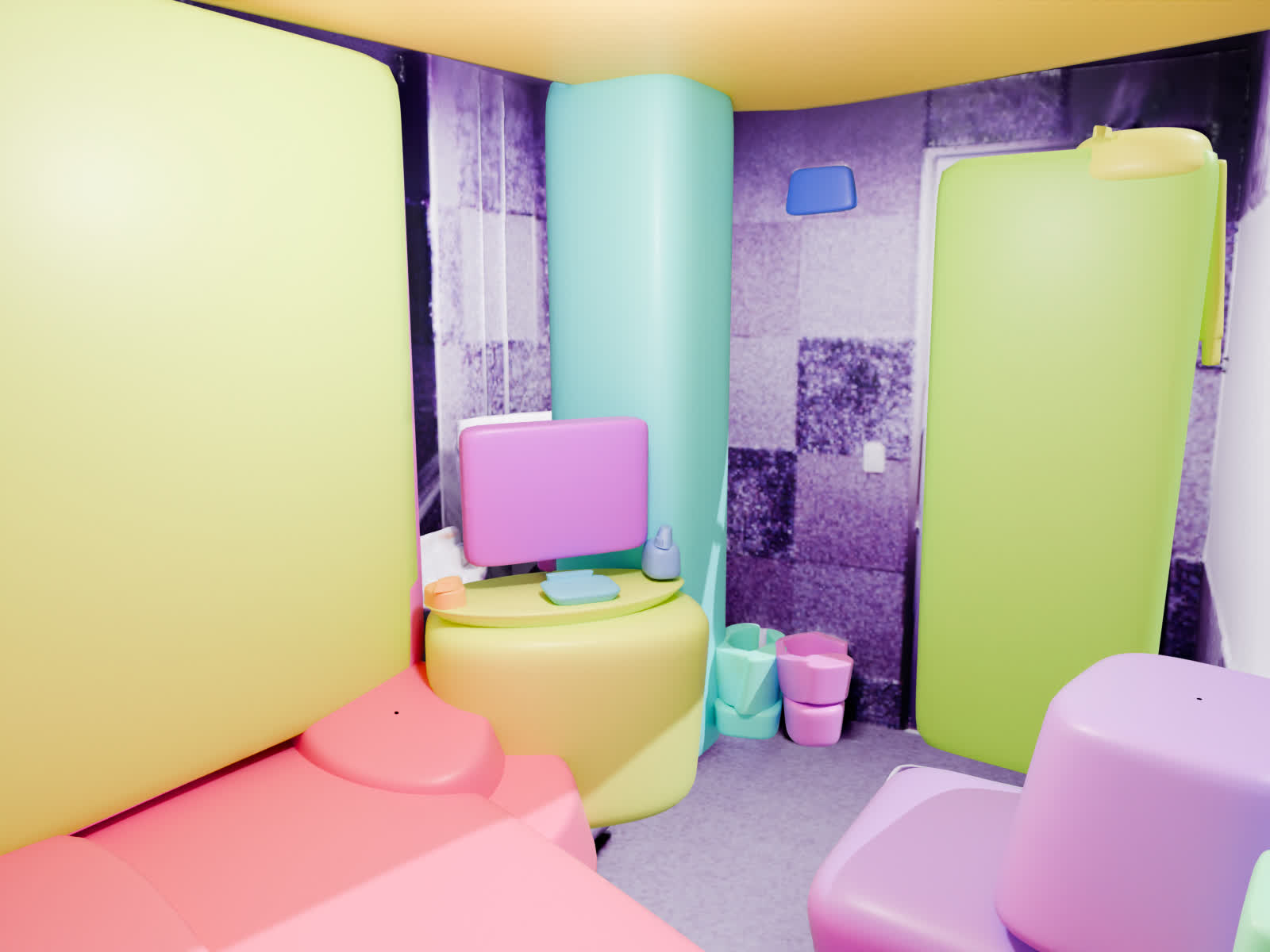}
    \end{tabular}
    \caption{\textbf{Robustness in real-world point clouds.} We show four examples scene from Replica where each object is decomposed into a set of superquadrics.}
    \label{fig:replica_qualitative}
\end{figure*}

\subsection{SDF Activation Function}
\label{sdf-activation}
To improve stability and robustness to outliers, we pass the distances $d(\mathbf{x}_i; \Theta_k)$ through a truncated robust kernel :
\begin{equation*}
\psi(d) =
\begin{cases}
\delta\,\tanh\!\left(d/\delta\right), & d \ge 0,\\[1.2ex]
-\omega\,\delta\,\tanh\!\left(d/\delta\right), & d < 0.
\end{cases}
\end{equation*}
where $\delta$ is the truncation hyperparameter, always set to $\delta=0.05$ in our experiments. We then asymmetrically penalize positive distances (outside the primitive), encouraging a tight boundary fit.

\begin{table}[b]
    \centering
    \setlength{\tabcolsep}{4pt}
    \begin{tabular}{c | ccccccc}
        \toprule
        Weight $\omega$ & \textbf{IoU}$\uparrow$ & \textbf{F}~\cite{TankAndTamples}$\uparrow$ & \textbf{L1}$\downarrow$ & \textbf{L2}$\downarrow$ & \textbf{Normals}$\uparrow$ & \textbf{\#\,Prim}.$\downarrow$ & \textbf{Overlap}$\downarrow$ \\
        \midrule
        1 & 0.74 & 0.41 & 1.50 & 0.046 & 0.83 & 5.61 & 11\% \\ 
        0.5 & 0.73 & 0.39 & 1.50 & 0.043 & 0.84 & 5.61 & 9\% \\ 
        0.1 & 0.72 & 0.37 & 1.55 & 0.044 & 0.85 & 5.64 & 8\% \\ 
        0.01 & 0.71 & 0.36 & 1.56 & 0.044 & 0.85 & 5.61 & 8\% \\ 
        \bottomrule
    \end{tabular}
    \vspace{5px}
    \caption{\textbf{SDF Asymmetry Ablation Study}. $\lsparsity$ and $\lexist$ are kept for all experiments as they play a crucial role in the model's architecture. Results are reported on the ShapeNet test set. L1 and L2 are multiplied by $10^2$.
    }
    \label{sdf-relu-ablation}
\end{table}

In Tab.~\ref{sdf-relu-ablation}, we report an ablation study on the negative weight ($\omega$) value. Larger values improve IoU and F-score, but slightly degrade normal consistency and increase overlap. Smaller values yield better normal alignment and reduced overlap at the cost of reconstruction accuracy. We select a weight of 0.1 as a good trade-off between reconstruction quality and surface fidelity, as reflected by normal completeness.

By normal completeness, we mean the following: for each ground-truth point, we find its nearest neighbor on the reconstructed superquadric surfaces and compute the absolute dot product between their normals. This metric evaluates how well the predicted normals align with the ground truth at locations where surfaces are expected to be reconstructed.

\section{Additional Qualitative Results}\label{sec:quali-supp}
\subsection{Baselines in comparison on ShapeNet}
\begin{figure}[t!]
\centering
\begin{small}
\rotatebox{90}{\footnotesize \shortstack{Input \\ Pointcloud}}
\includegraphics[height=1.7cm,trim={200px 100px 190px 130px},clip]{figures/shapenet_jpg/pc_3782.jpg}\hspace{-9px}
\includegraphics[height=1.7cm,trim={120px 70px 220px 80px},clip]{figures/shapenet_jpg/pc_1469.jpg}
\includegraphics[height=1.7cm,trim={220px 120px 220px 120px},clip]{figures/shapenet_jpg/pc_5749.jpg}
\includegraphics[height=1.7cm,trim={220px 120px 220px 90px},clip]{figures/shapenet_jpg/pc_5983.jpg}
\includegraphics[height=1.7cm,trim={260px 120px 230px 210px},clip]{figures/shapenet_jpg/pc_6367.jpg}
\includegraphics[height=1.7cm,trim={240px 150px 260px 220px},clip]{figures/shapenet_jpg/pc_38.jpg}\\
\rotatebox{90}{\footnotesize \hspace{8px}\shortstack{\\ CSA~\cite{Yang2021UnsupervisedLF}}}
\includegraphics[height=1.7cm,trim={200px 100px 190px 130px},clip]{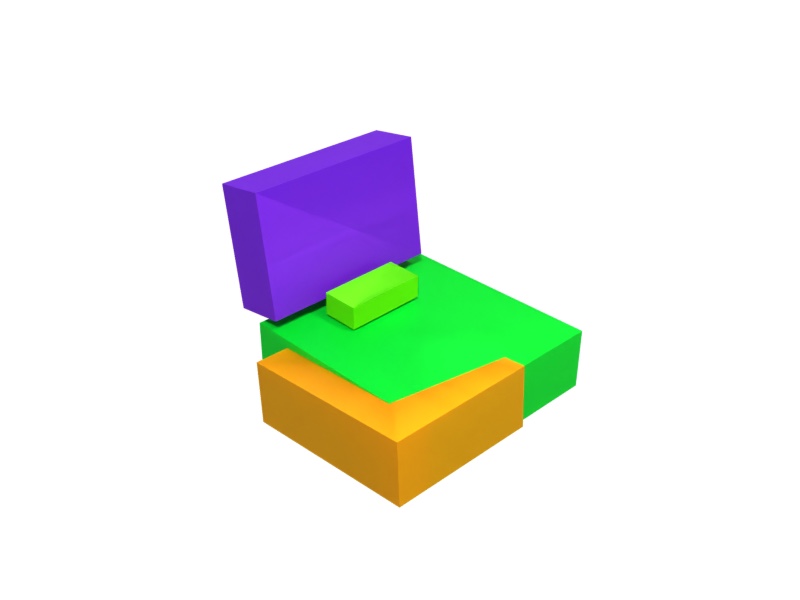}\hspace{-9px}
\includegraphics[height=1.7cm,trim={120px 70px 220px 80px},clip]{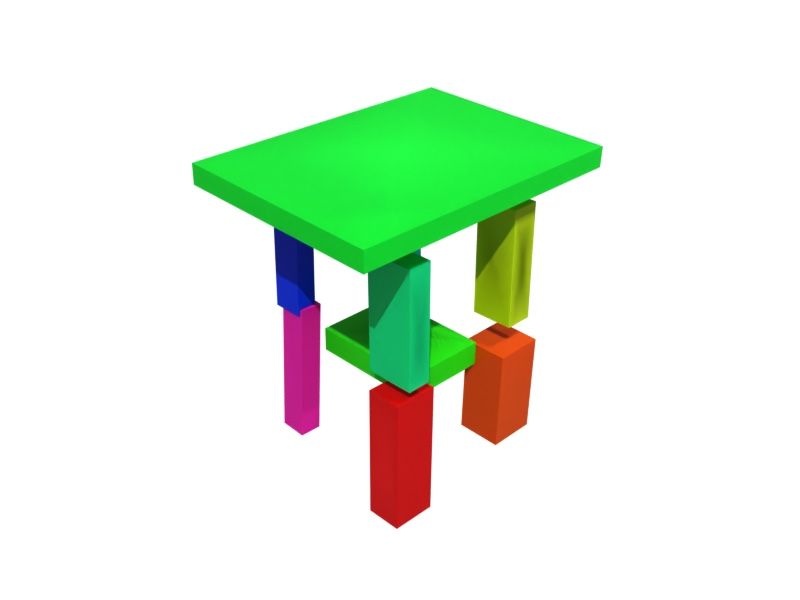}
\includegraphics[height=1.7cm,trim={220px 120px 220px 120px},clip]{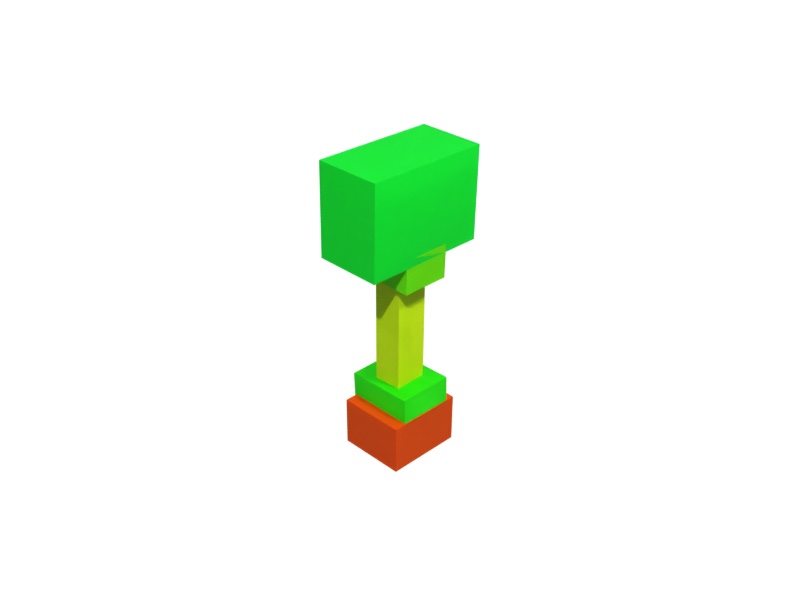}
\includegraphics[height=1.7cm,trim={220px 120px 220px 90px},clip]{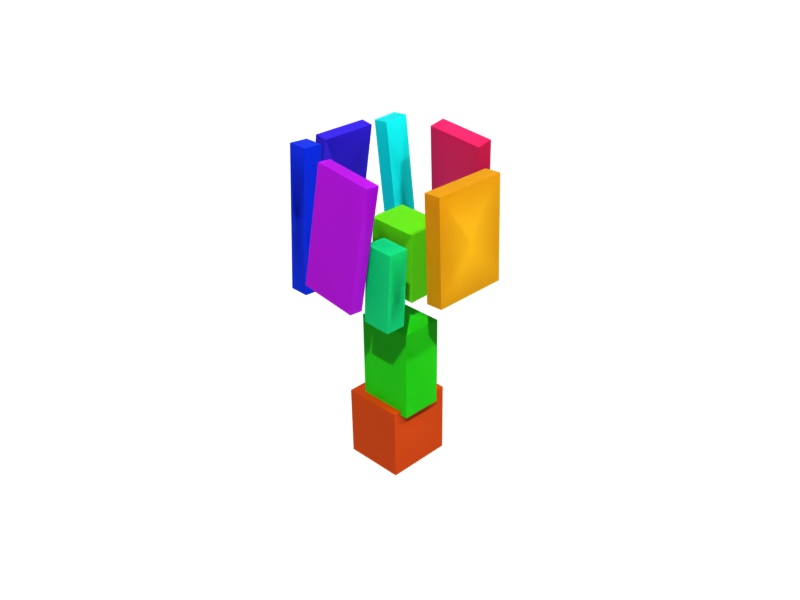}
\includegraphics[height=1.7cm,trim={220px 110px 180px 200px},clip]{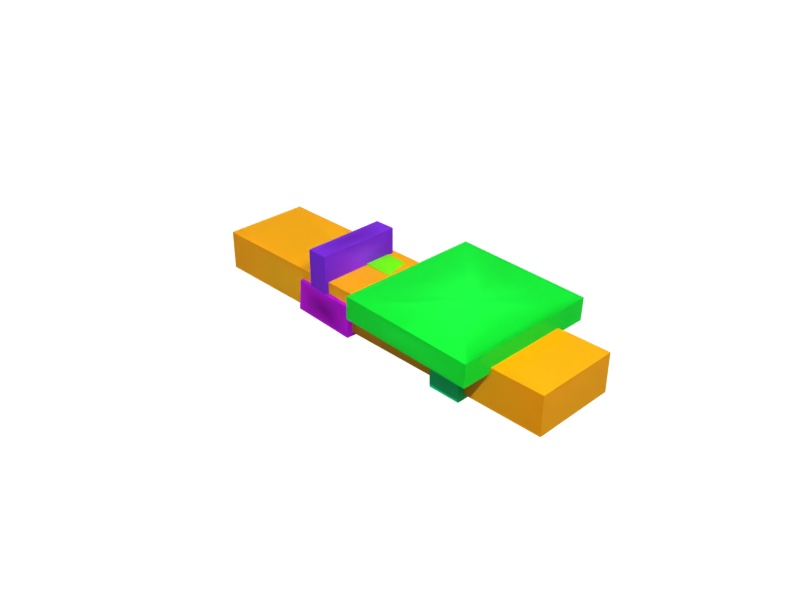}
\includegraphics[height=1.7cm,trim={240px 150px 260px 200px},clip]{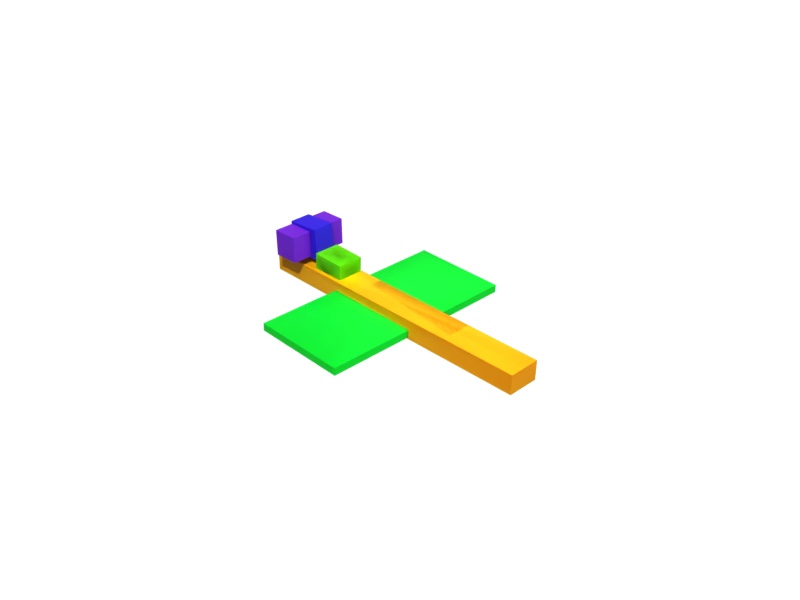}\\
\rotatebox{90}{\footnotesize \hspace{8px}\shortstack{\\ SQ~\cite{paschalidou2019superquadrics}}}
\includegraphics[height=1.7cm,trim={200px 100px 190px 130px},clip]{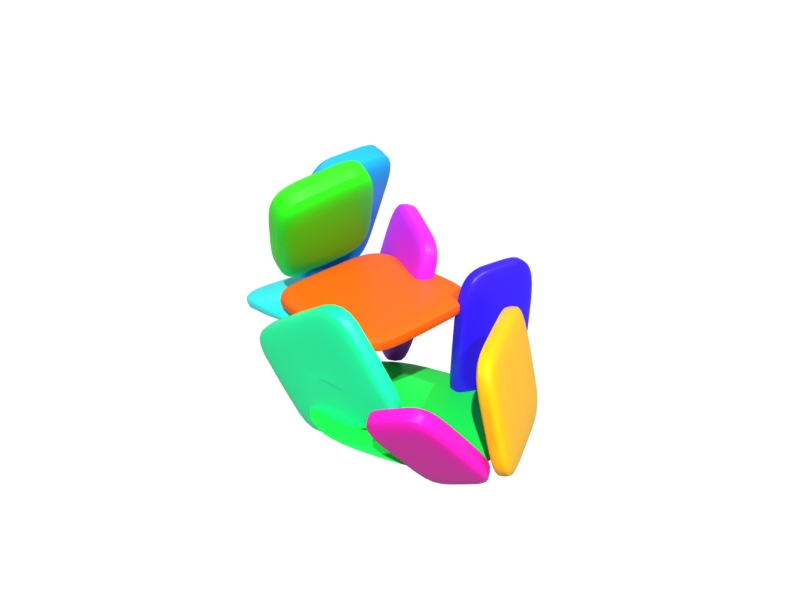}\hspace{-9px}
\includegraphics[height=1.7cm,trim={120px 70px 220px 80px},clip]{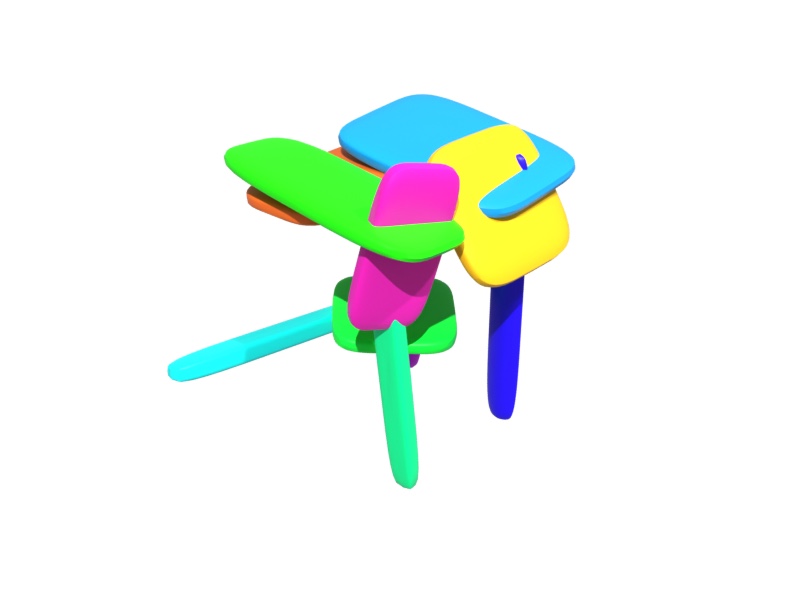}
\includegraphics[height=1.7cm,trim={220px 120px 220px 120px},clip]{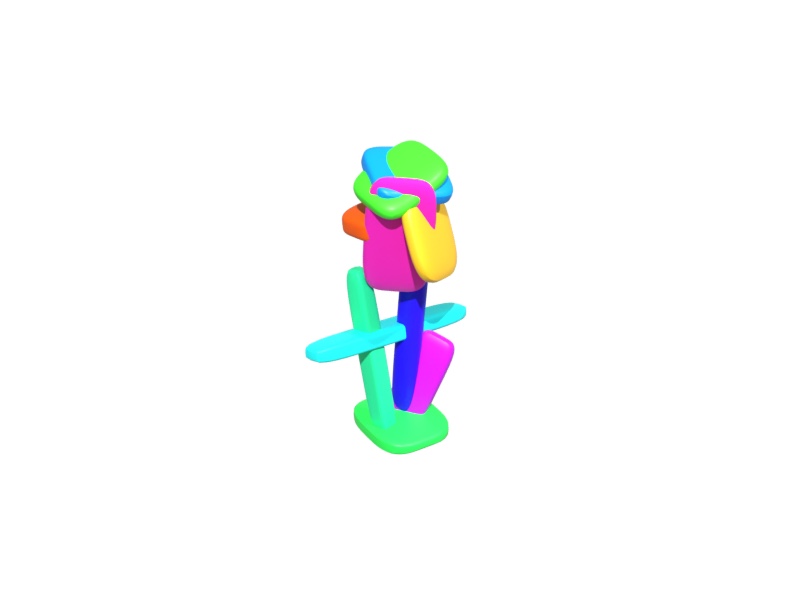}
\includegraphics[height=1.7cm,trim={220px 120px 220px 90px},clip]{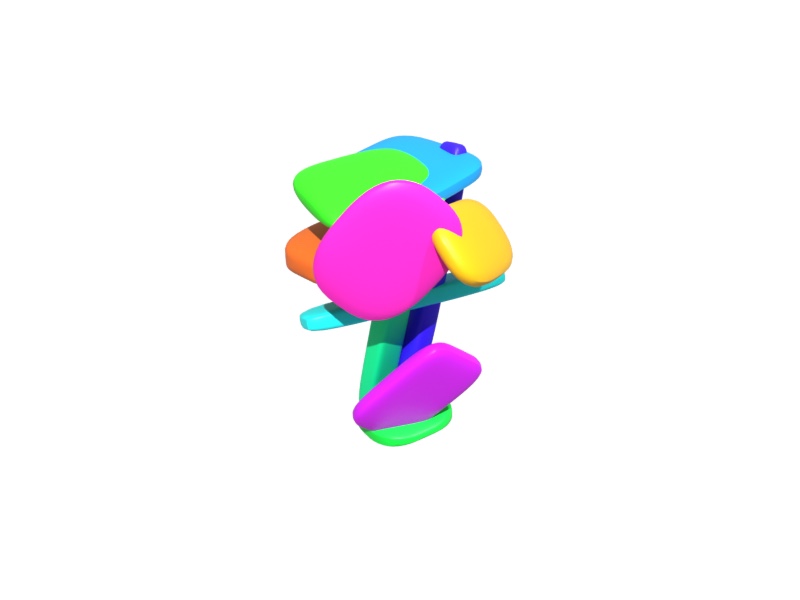}
\includegraphics[height=1.7cm,trim={260px 120px 230px 210px},clip]{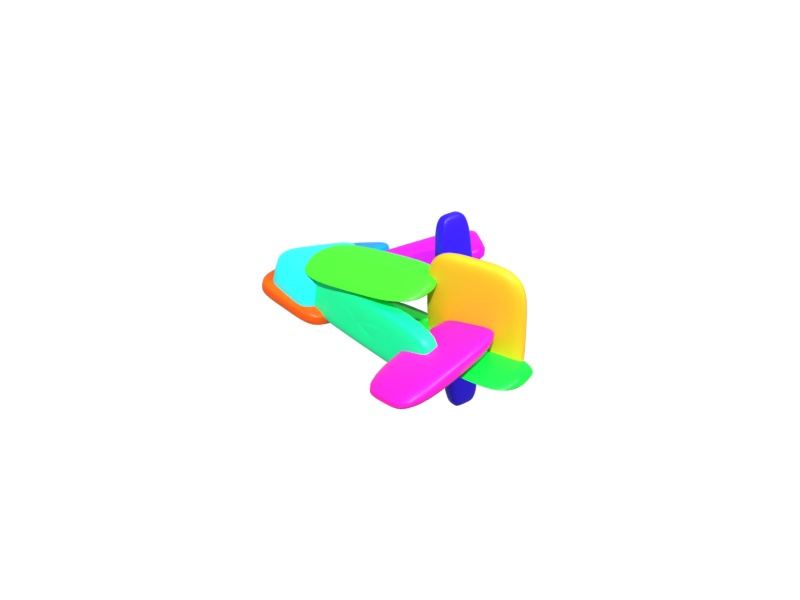}
\includegraphics[height=1.7cm,trim={240px 150px 260px 200px},clip]{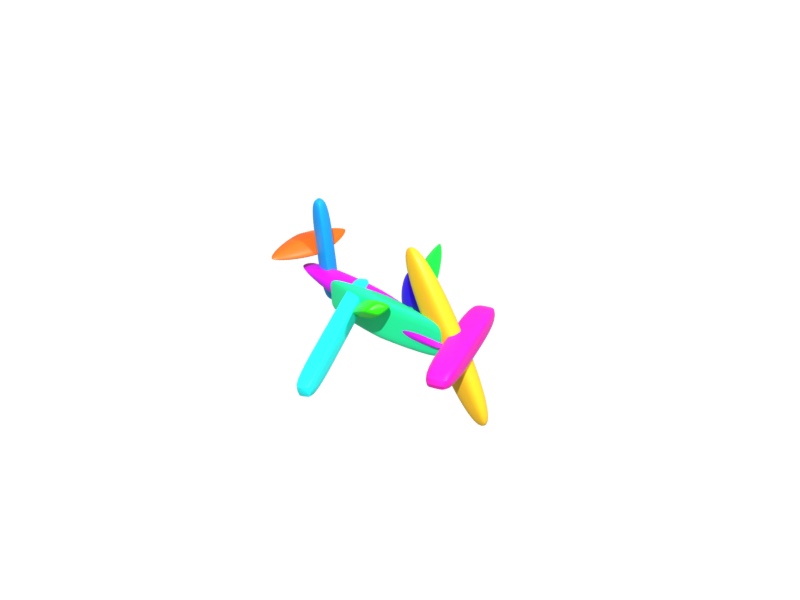}\\
\rotatebox{90}{\footnotesize \hspace{8px}\shortstack{\\ EMS~\cite{liu2022robust}}}
\includegraphics[height=1.7cm,trim={200px 100px 190px 130px},clip]{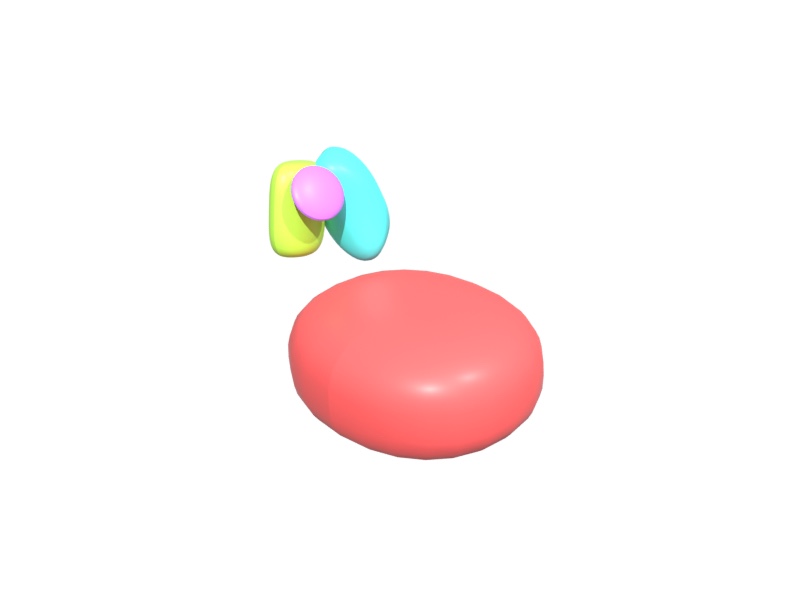}\hspace{-9px}
\includegraphics[height=1.7cm,trim={120px 70px 220px 80px},clip]{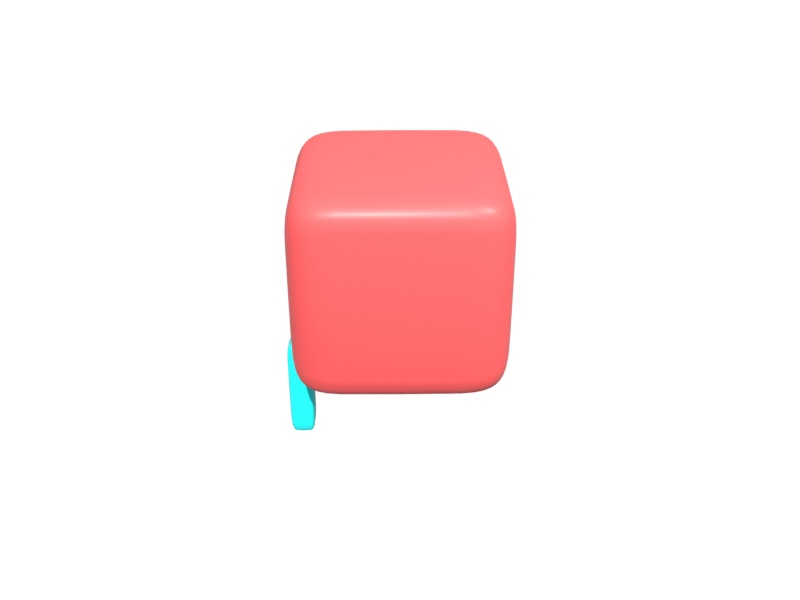}
\includegraphics[height=1.7cm,trim={220px 120px 220px 120px},clip]{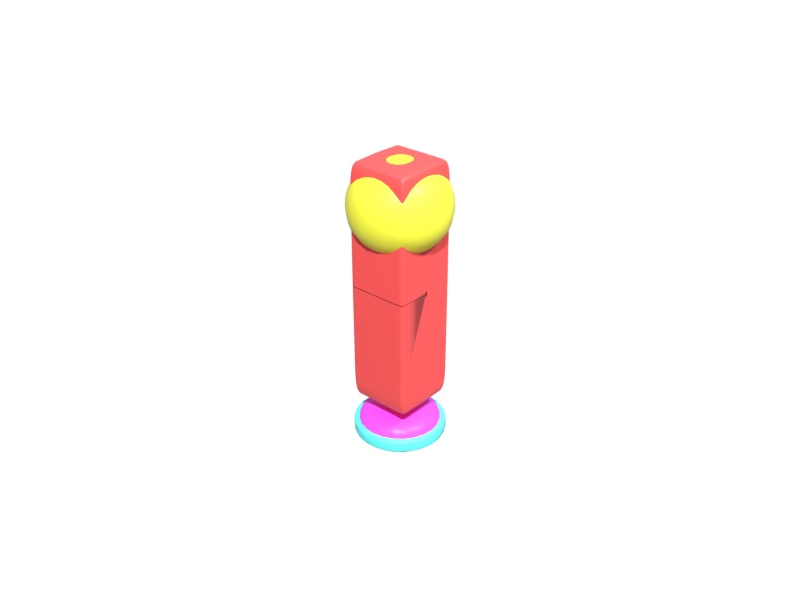}
\includegraphics[height=1.7cm,trim={220px 120px 220px 90px},clip]{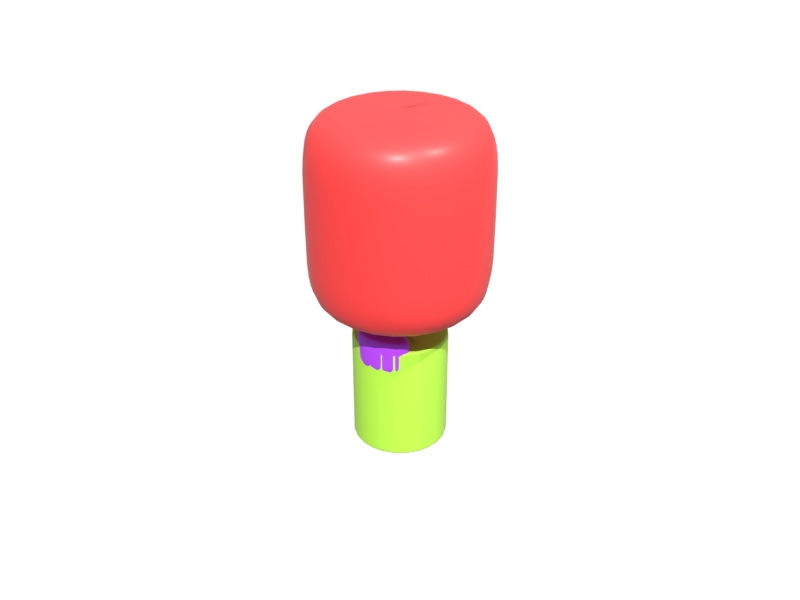}
\includegraphics[height=1.7cm,trim={260px 120px 230px 210px},clip]{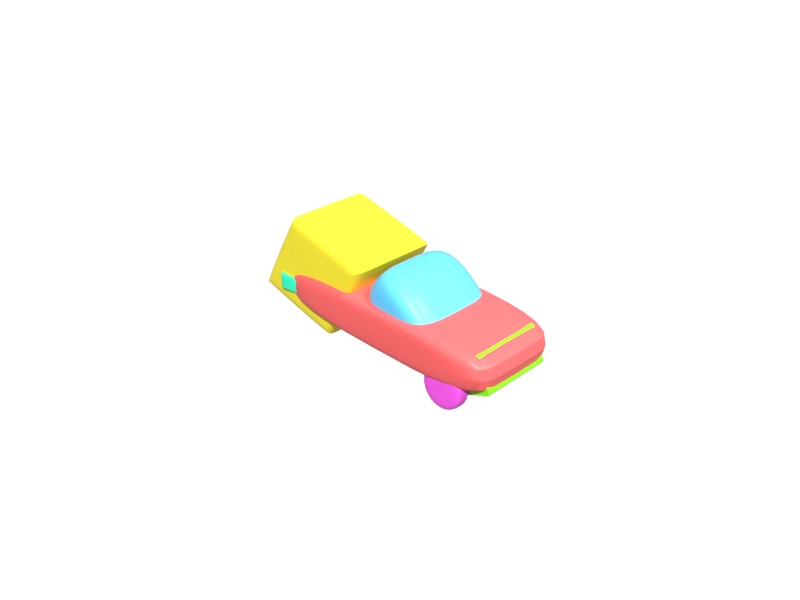}
\includegraphics[height=1.7cm,trim={240px 150px 260px 220px},clip]{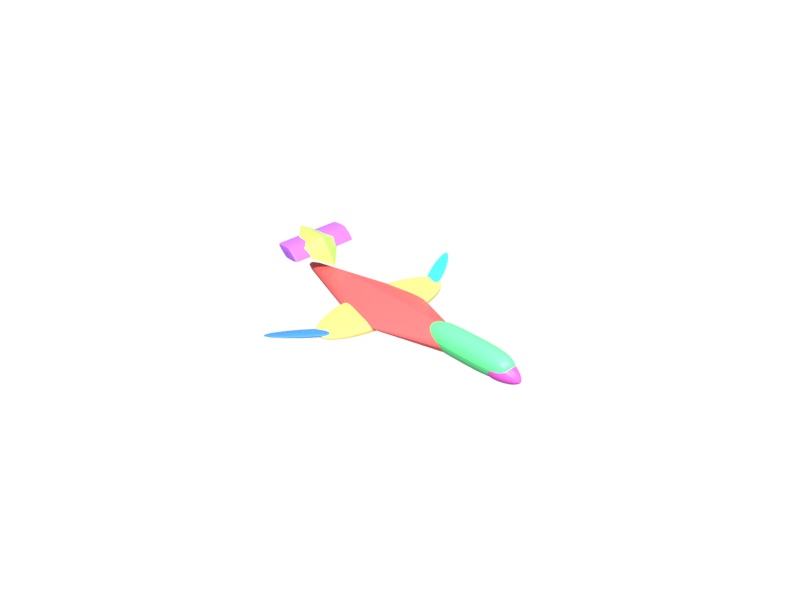}\\

\rotatebox{90}{\footnotesize \shortstack{Marching \\ Prim.\cite{liu2023marching}}}
\includegraphics[height=1.7cm,trim={200px 100px 190px 130px},clip]{figures/shapenet_jpg/mp_3782.jpg}\hspace{-9px}
\includegraphics[height=1.7cm,trim={120px 70px 220px 80px},clip]{figures/shapenet_jpg/mp_1469.jpg}
\includegraphics[height=1.7cm,trim={220px 120px 220px 120px},clip]{figures/shapenet_jpg/mp_5749.jpg}
\includegraphics[height=1.7cm,trim={220px 120px 220px 90px},clip]{figures/shapenet_jpg/mp_5983.jpg}
\includegraphics[height=1.7cm,trim={260px 120px 230px 210px},clip]{figures/shapenet_jpg/mp_6367.jpg}
\includegraphics[height=1.7cm,trim={240px 150px 260px 220px},clip]{figures/shapenet_jpg/mp_38.jpg}\\
\rotatebox{90}{\footnotesize \shortstack{ \hspace{2px}SuperDec\\ \cite{fedele2025superdec}}}\hspace{-4px}
\includegraphics[height=1.7cm,trim={200px 100px 190px 130px},clip]{figures/shapenet_jpg/superdec_3782.jpg}\hspace{-9px}
\includegraphics[height=1.7cm,trim={120px 70px 220px 80px},clip]{figures/shapenet_jpg/superdec_1469.jpg}
\includegraphics[height=1.7cm,trim={220px 120px 220px 120px},clip]{figures/shapenet_jpg/superdec_5749.jpg}
\includegraphics[height=1.7cm,trim={220px 120px 220px 90px},clip]{figures/shapenet_jpg/superdec_5983.jpg}
\includegraphics[height=1.7cm,trim={260px 120px 230px 210px},clip]{figures/shapenet_jpg/superdec_6367.jpg}
\includegraphics[height=1.7cm,trim={240px 150px 260px 220px},clip]{figures/shapenet_jpg/superdec_38.jpg}\\
\rotatebox{90}{\footnotesize \shortstack{ \hspace{2px}\method{}\\\scriptsize{(No T\&B)}}}\hspace{-4px}
\includegraphics[height=1.7cm,trim={200px 100px 190px 130px},clip]{figures/shapenet_jpg/superflex_no_def_3782.jpg}\hspace{-9px}
\includegraphics[height=1.7cm,trim={120px 70px 220px 80px},clip]{figures/shapenet_jpg/superflex_no_def_1469.jpg}
\includegraphics[height=1.7cm,trim={220px 120px 220px 120px},clip]{figures/shapenet_jpg/superflex_no_def_5749.jpg}
\includegraphics[height=1.7cm,trim={220px 120px 220px 90px},clip]{figures/shapenet_jpg/superflex_no_def_5983.jpg}
\includegraphics[height=1.7cm,trim={260px 120px 230px 210px},clip]{figures/shapenet_jpg/superflex_no_def_6367.jpg}
\includegraphics[height=1.7cm,trim={240px 150px 260px 220px},clip]{figures/shapenet_jpg/superflex_no_def_38.jpg}\\

\rotatebox{90}{\footnotesize \shortstack{\\ \hspace{5px}\method{}}}\hspace{-4px}
\includegraphics[height=1.7cm,trim={200px 100px 190px 130px},clip]{figures/shapenet_jpg/superflex_3782.jpg}\hspace{-9px}
\includegraphics[height=1.7cm,trim={120px 70px 220px 80px},clip]{figures/shapenet_jpg/superflex_1469.jpg}
\includegraphics[height=1.7cm,trim={220px 120px 220px 120px},clip]{figures/shapenet_jpg/superflex_5749.jpg}
\includegraphics[height=1.7cm,trim={220px 120px 220px 90px},clip]{figures/shapenet_jpg/superflex_5983.jpg}
\includegraphics[height=1.7cm,trim={260px 120px 230px 210px},clip]{figures/shapenet_jpg/superflex_6367.jpg}
\includegraphics[height=1.7cm,trim={240px 150px 260px 220px},clip]{figures/shapenet_jpg/superflex_38.jpg}\\
\end{small}
\caption{\textbf{Qualitative Results on ShapeNet.} \emph{Top row:} input point clouds. \emph{Below:} the outputs of baselines and our methods. Different colors indicate different primitives. For \name{}, we compare variants with and without tapering and bending (T\&B) deformation parameters.}
\label{fig:quali-shapenet-sup}
\end{figure}

We provide an expanded qualitative comparison of \method{} against state-of-the-art baselines across a wide range of ShapeNet categories. As shown in Figure~\ref{fig:quali-shapenet-sup}, our method consistently yields superior decompositions. While rigid primitive methods often struggle to capture organic curves, leading to misalignments or the use of redundant primitives, \method{} leverages bending and tapering to achieve high-fidelity reconstructions while using fewer primitives.

\subsection{Qualitative Results of \method{} on ShapeNet and ABO}
\begin{figure}[t!]
\centering
\begin{small}
\rotatebox{90}{\footnotesize \shortstack{Input \\ Pointcloud}}
\includegraphics[width=0.14\linewidth,trim={600px 500px 600px 200px},clip]{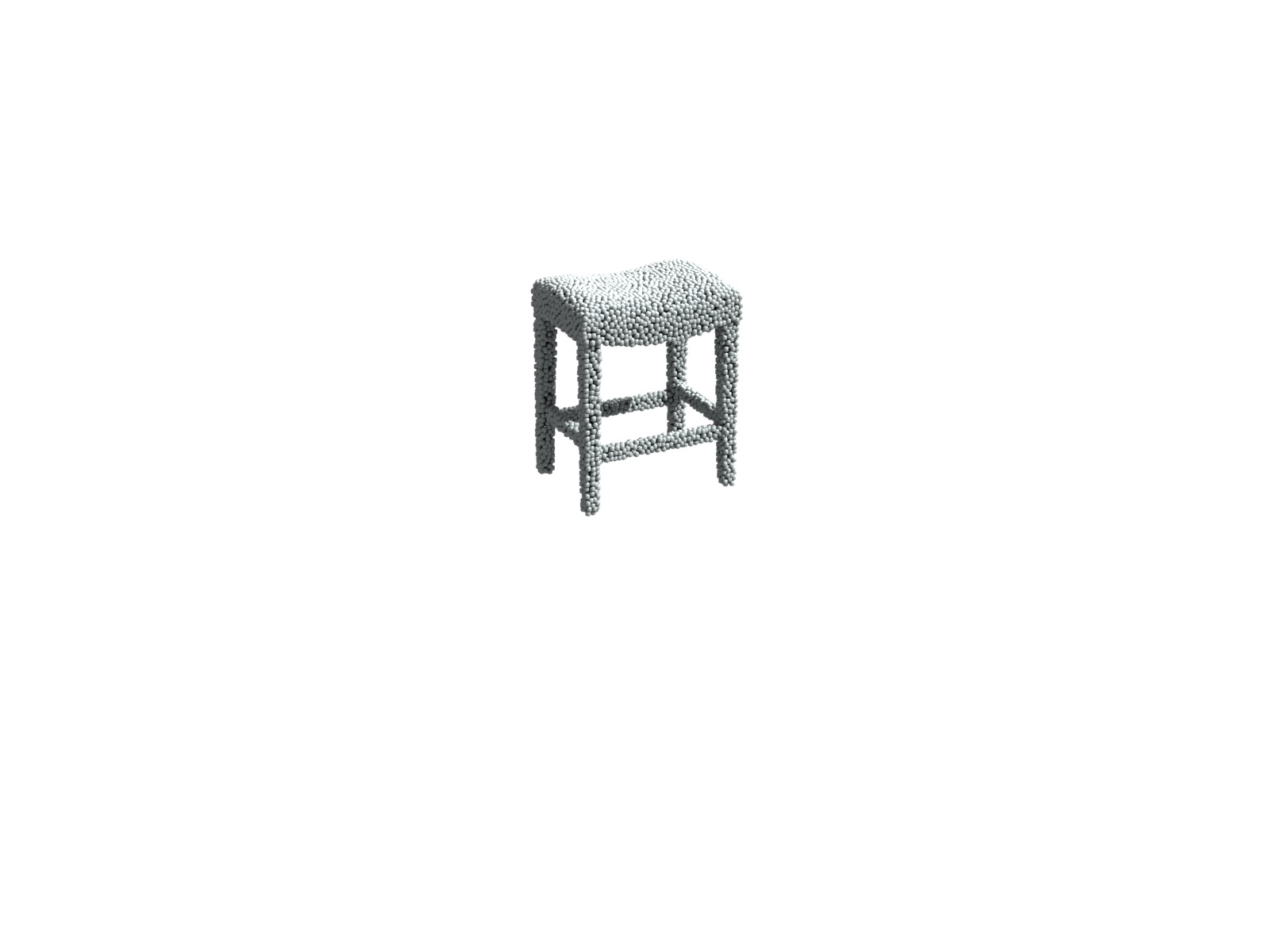}
\includegraphics[width=0.125\linewidth,trim={600px 500px 600px 200px},clip]{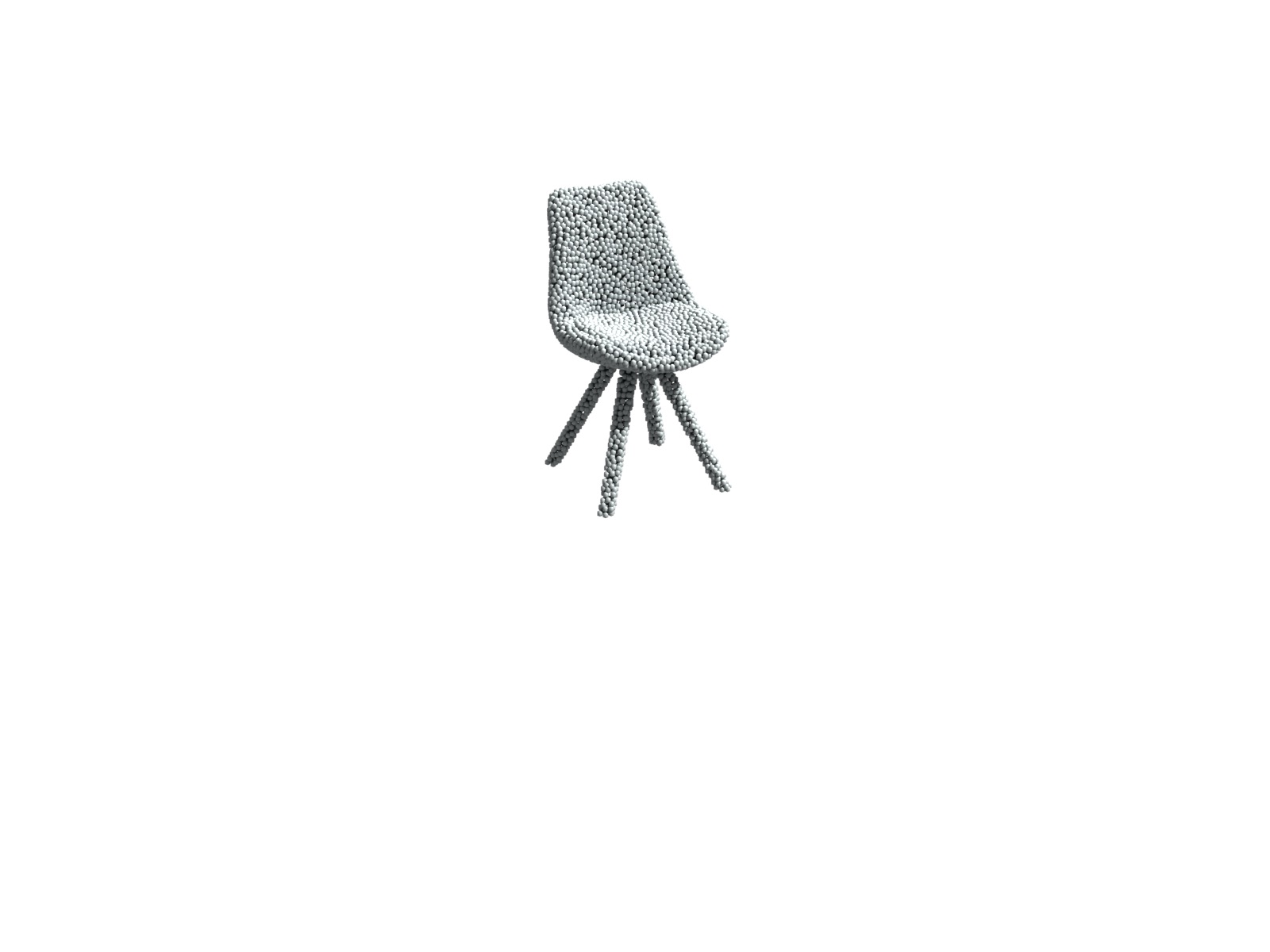}
\includegraphics[width=0.11\linewidth,trim={600px 450px 600px 200px},clip]{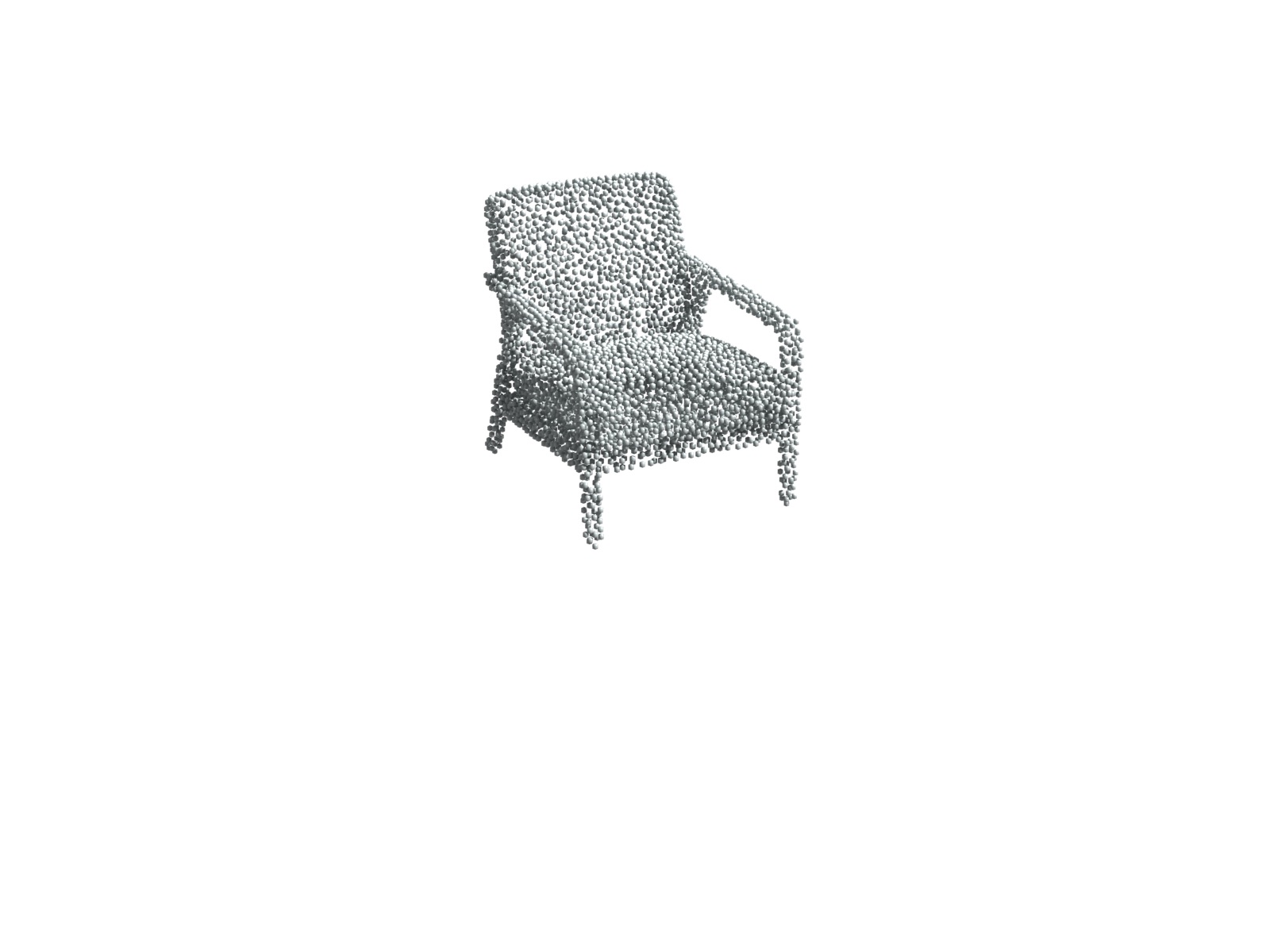}
\includegraphics[width=0.115\linewidth,trim={600px 500px 600px 100px},clip]{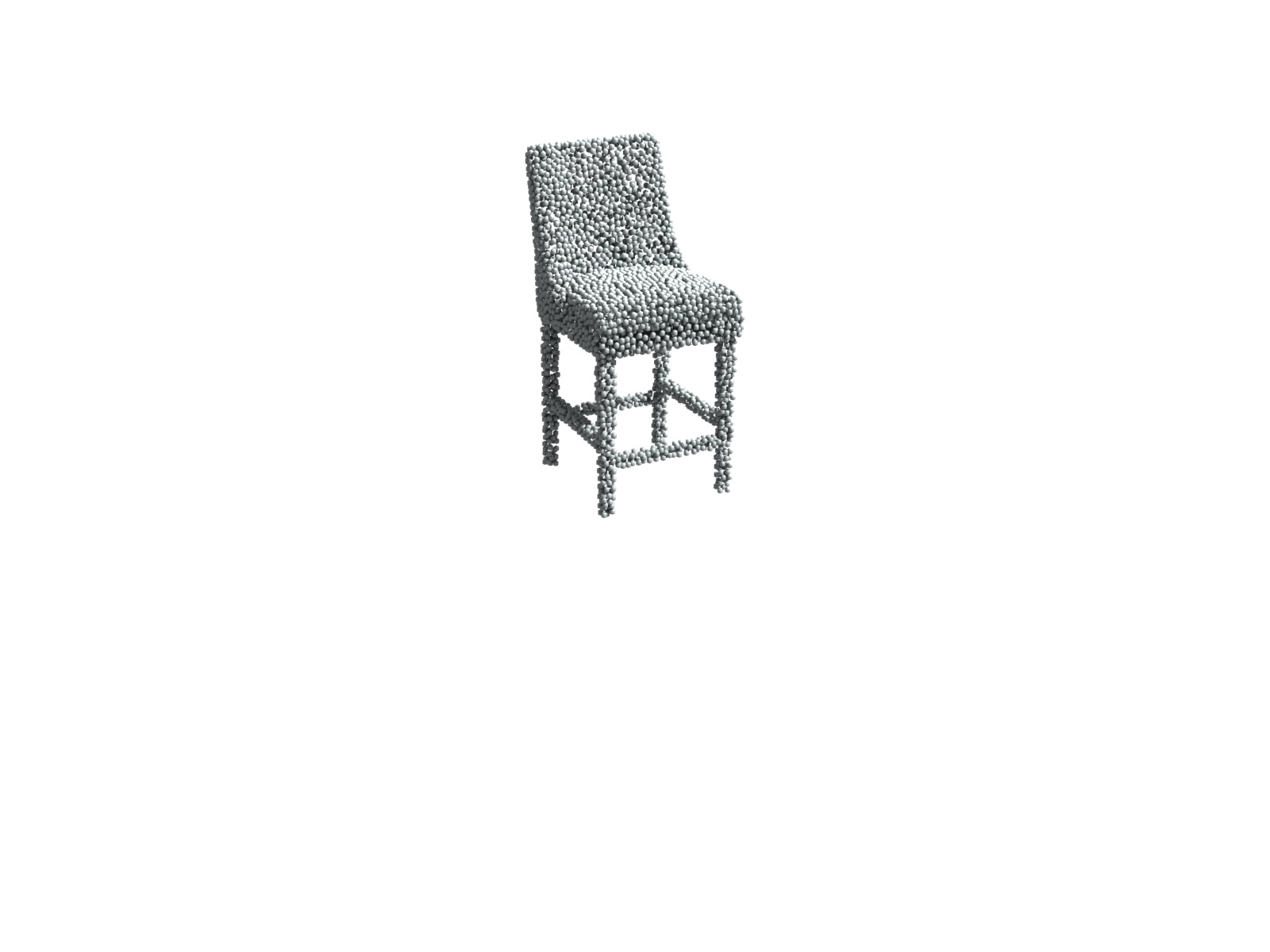}
\includegraphics[width=0.115\linewidth,trim={500px 400px 500px 1px},clip]{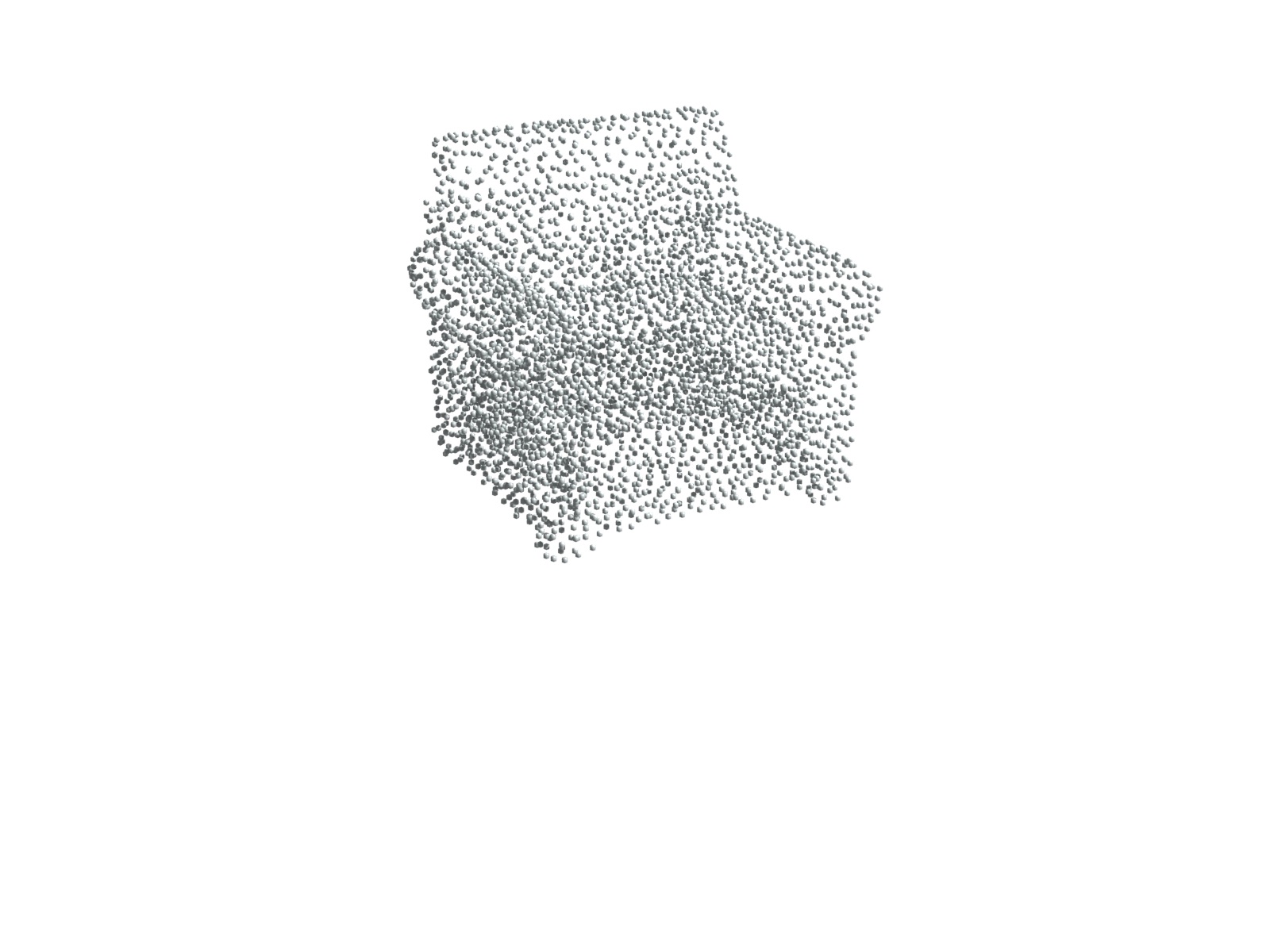}
\includegraphics[width=0.125\linewidth,trim={550px 450px 550px 200px},clip]{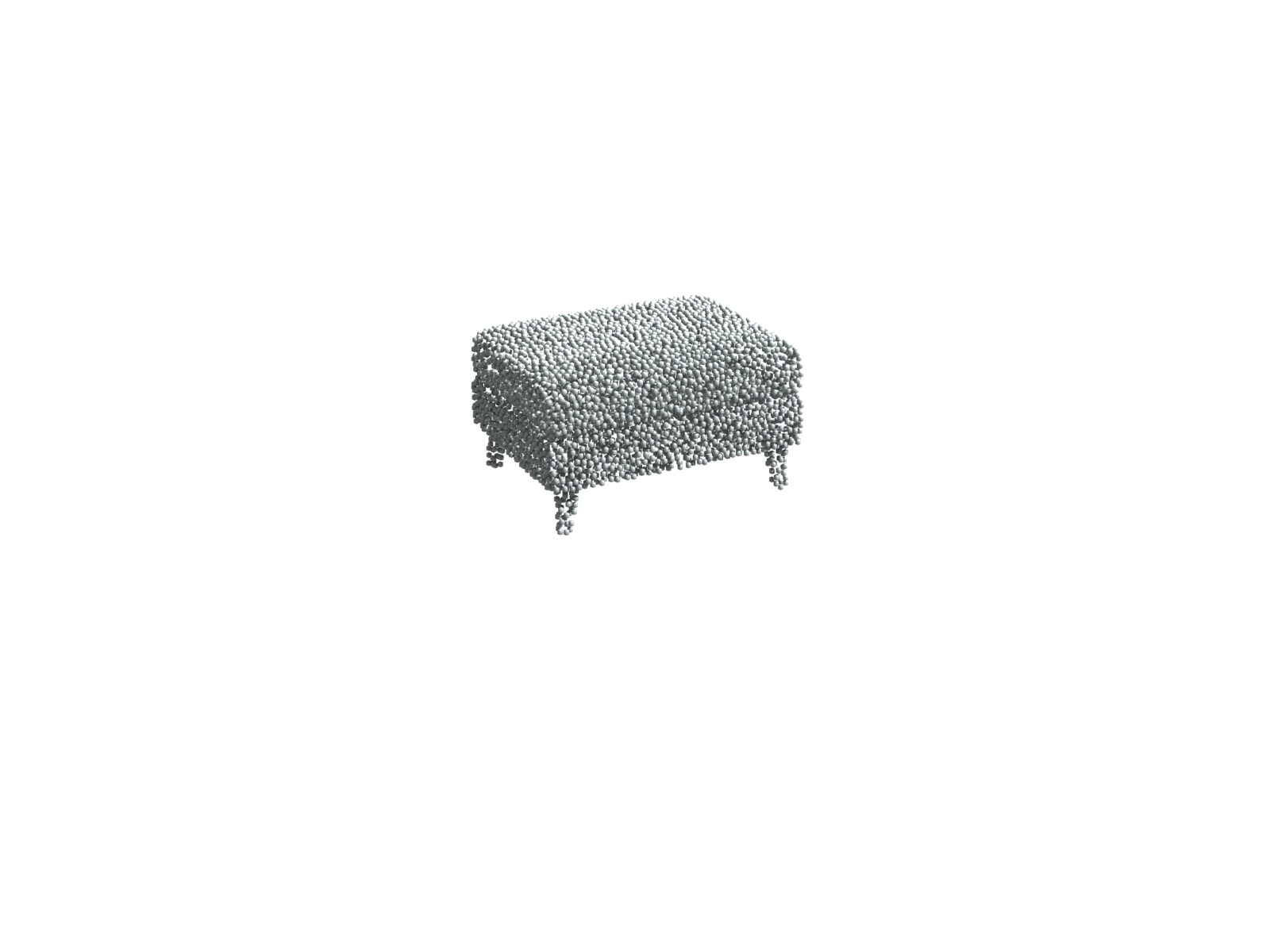}
\includegraphics[width=0.16\linewidth,trim={300px 300px 350px 1px},clip]{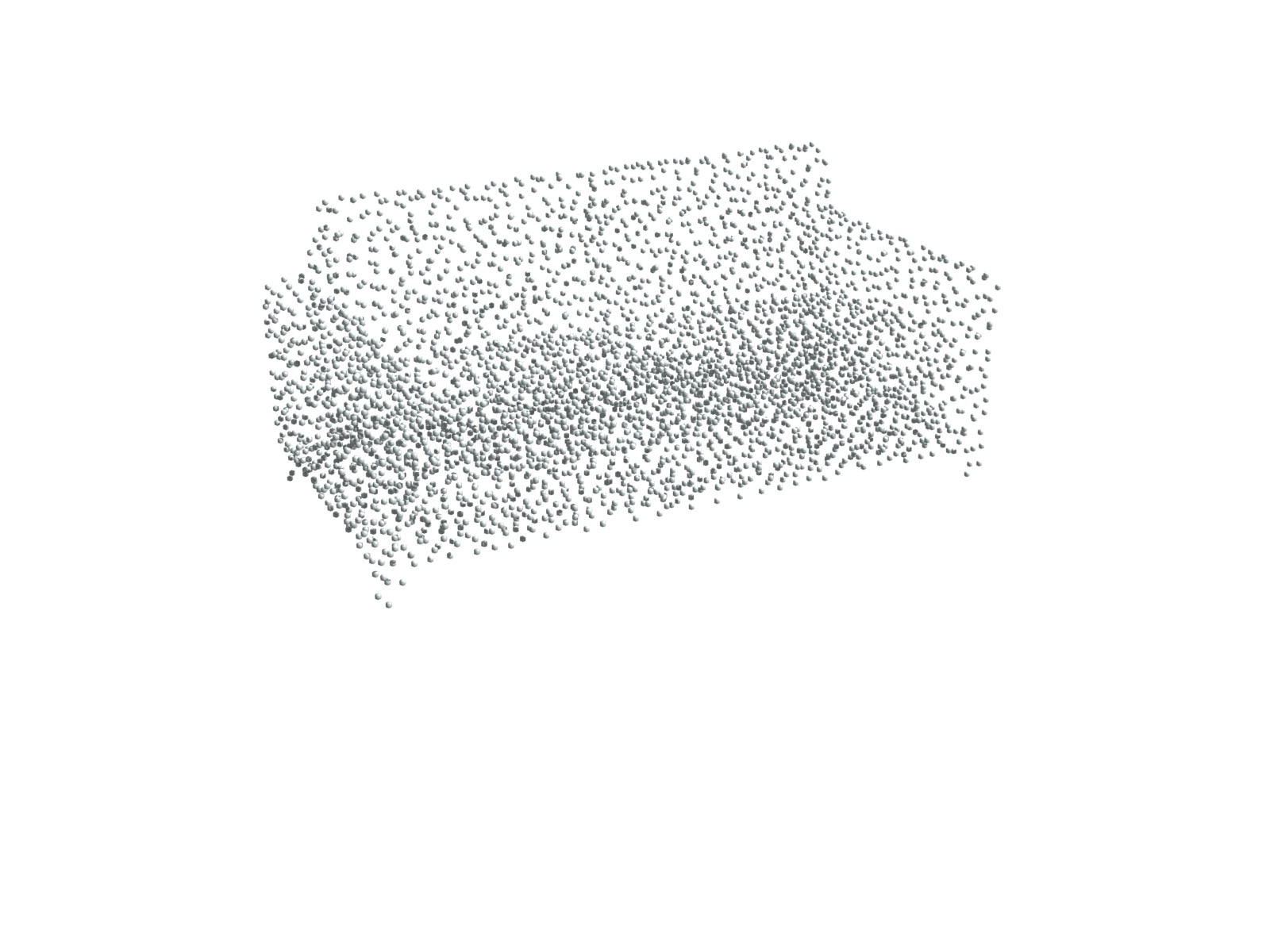}\\
\vspace{-10px}
\rotatebox{90}{\footnotesize \shortstack{\\ \hspace{5px}\method{}}}\hspace{4px} 
\includegraphics[width=0.14\linewidth,trim={600px 500px 600px 200px},clip]{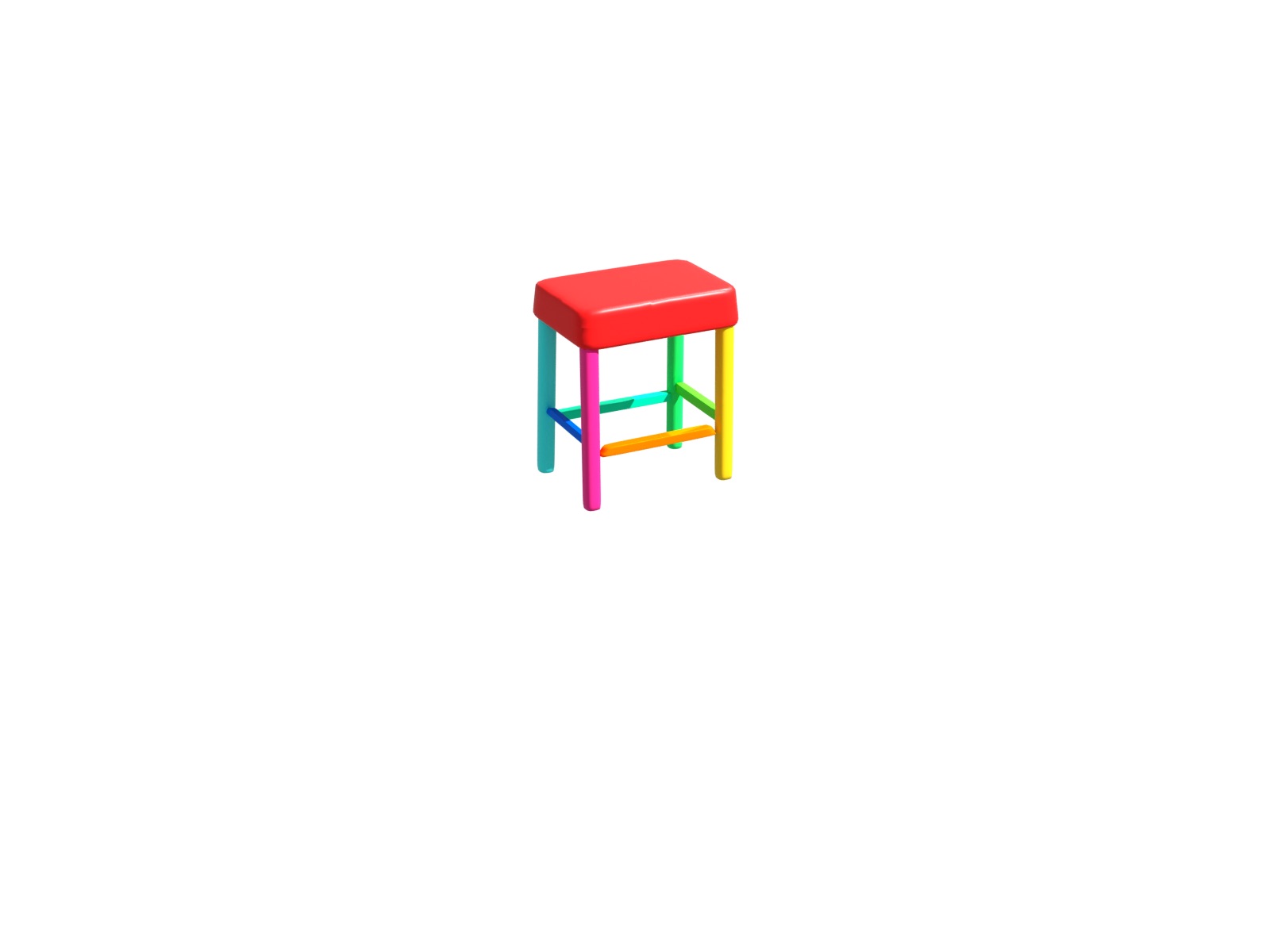}
\includegraphics[width=0.125\linewidth,trim={600px 500px 600px 200px},clip]{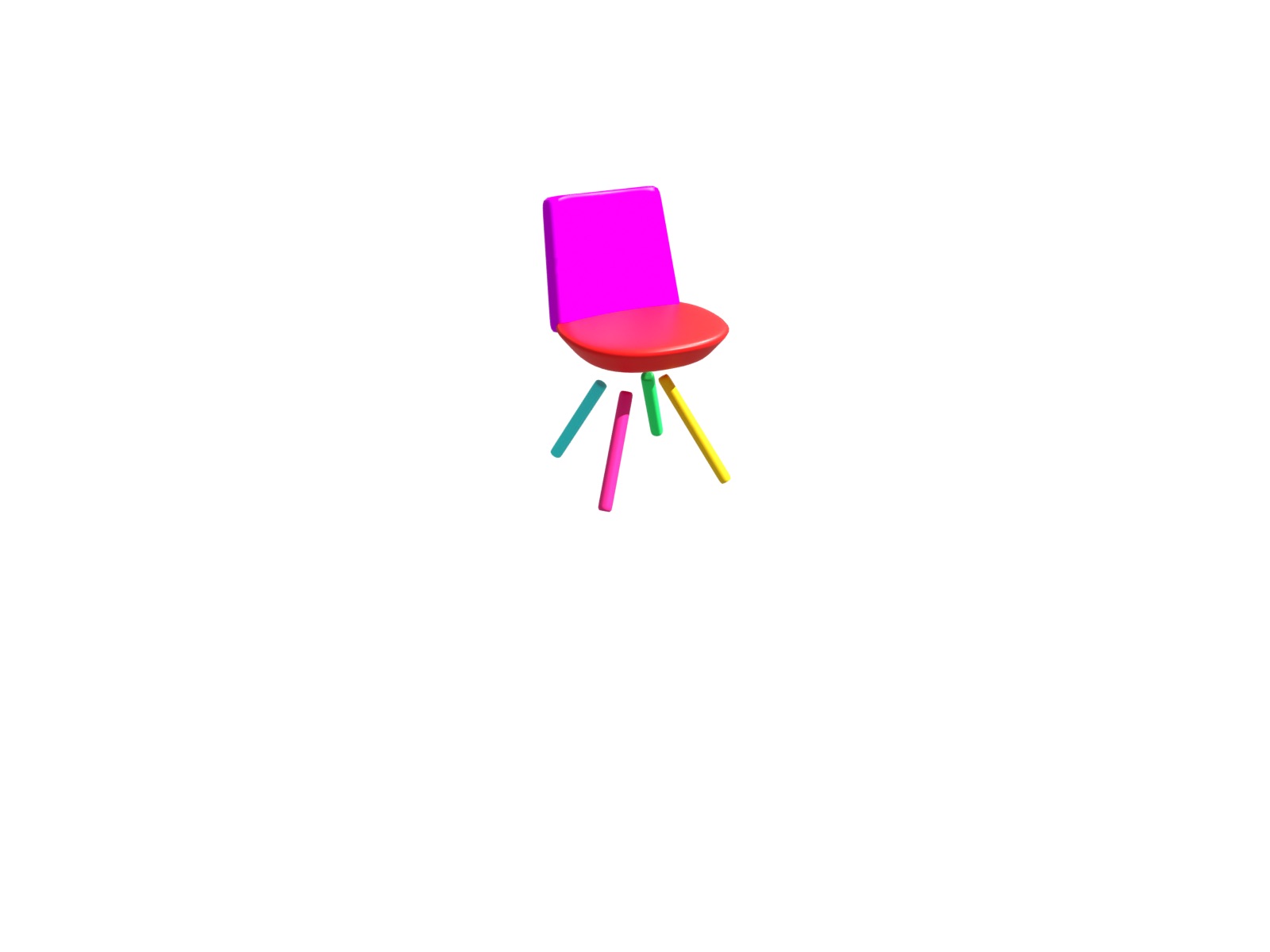}
\includegraphics[width=0.11\linewidth,trim={600px 450px 600px 200px},clip]{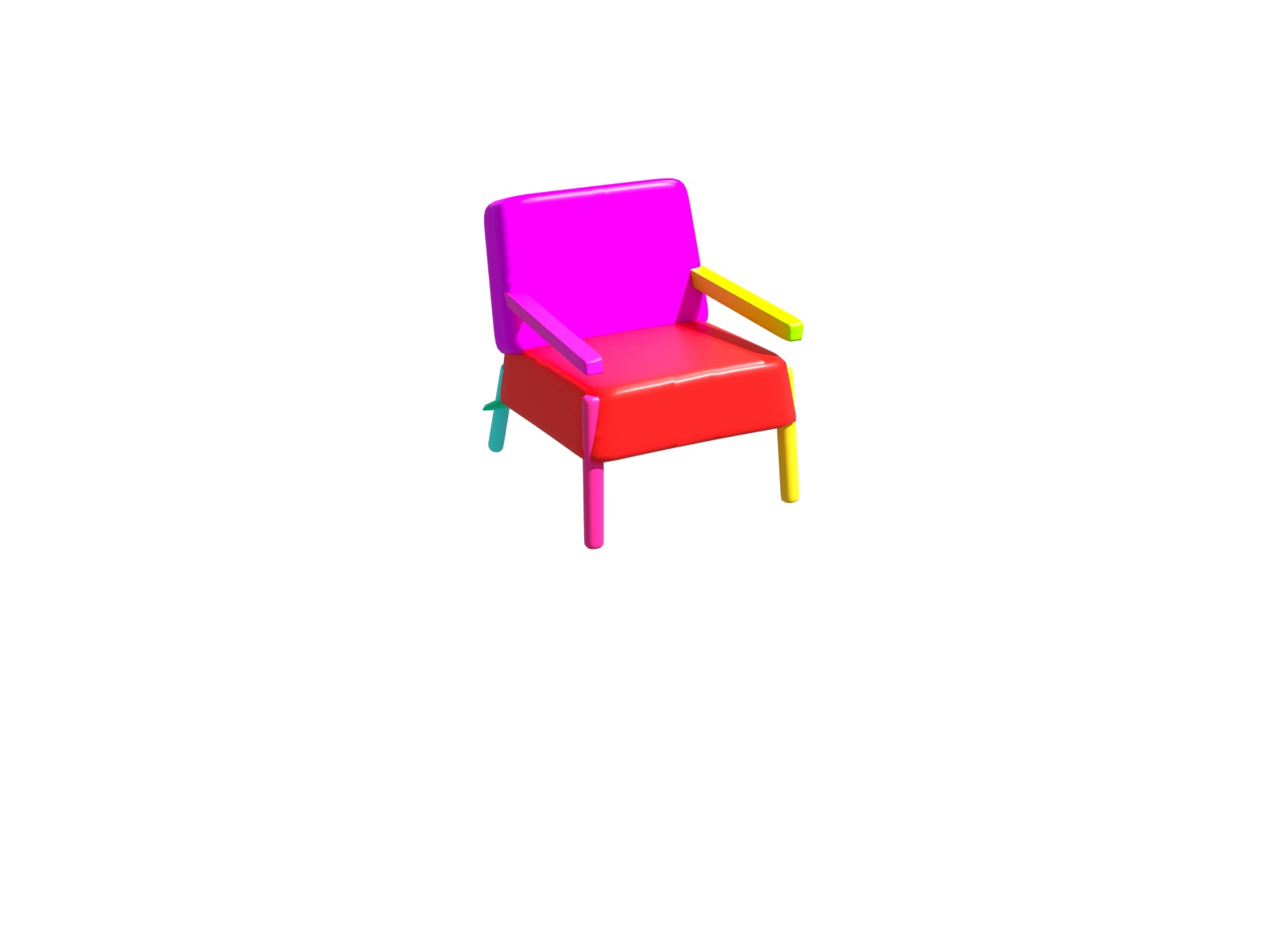}
\includegraphics[width=0.115\linewidth,trim={600px 500px 600px 100px},clip]{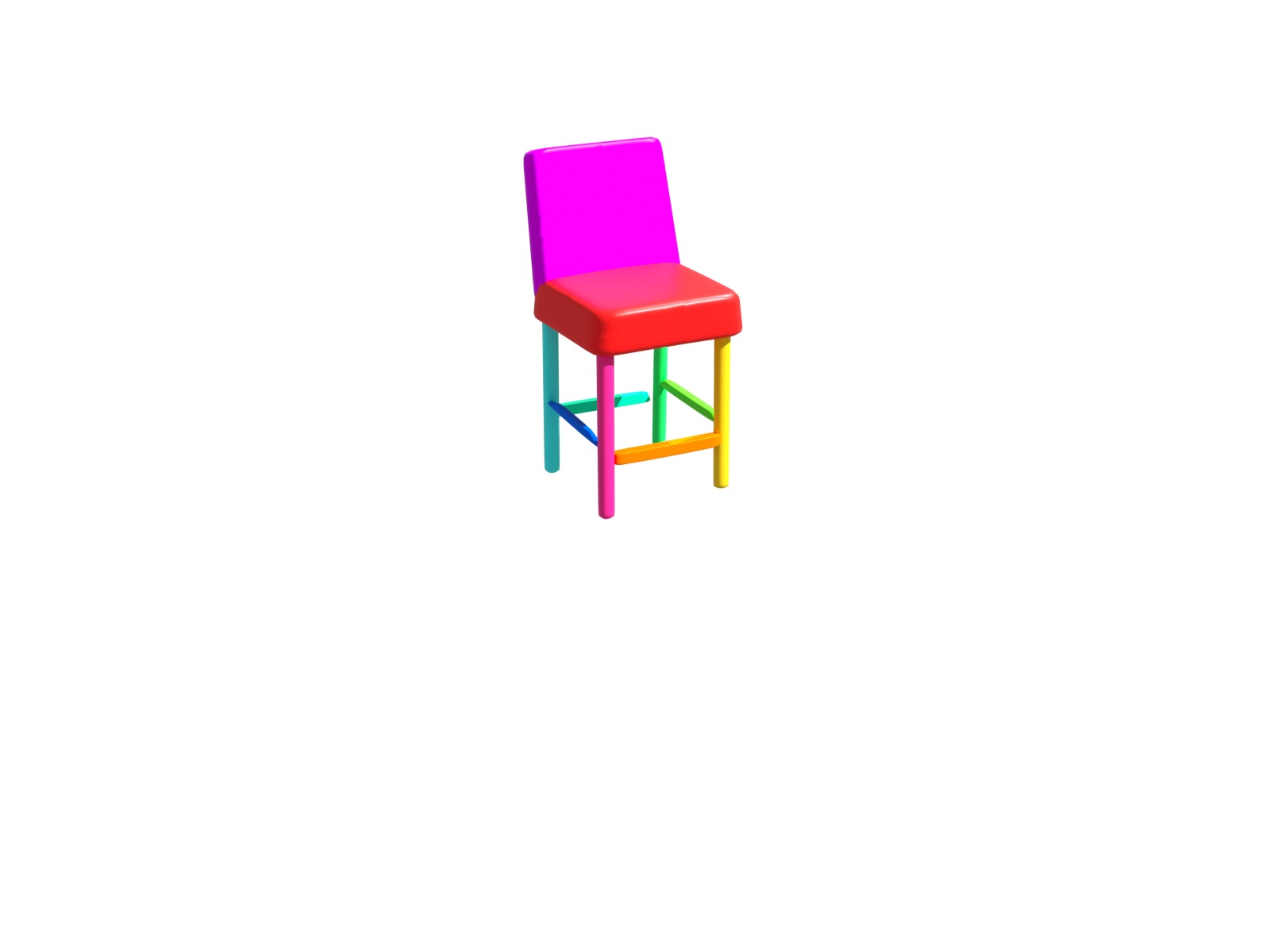}
\includegraphics[width=0.115\linewidth,trim={500px 400px 500px 1px},clip]{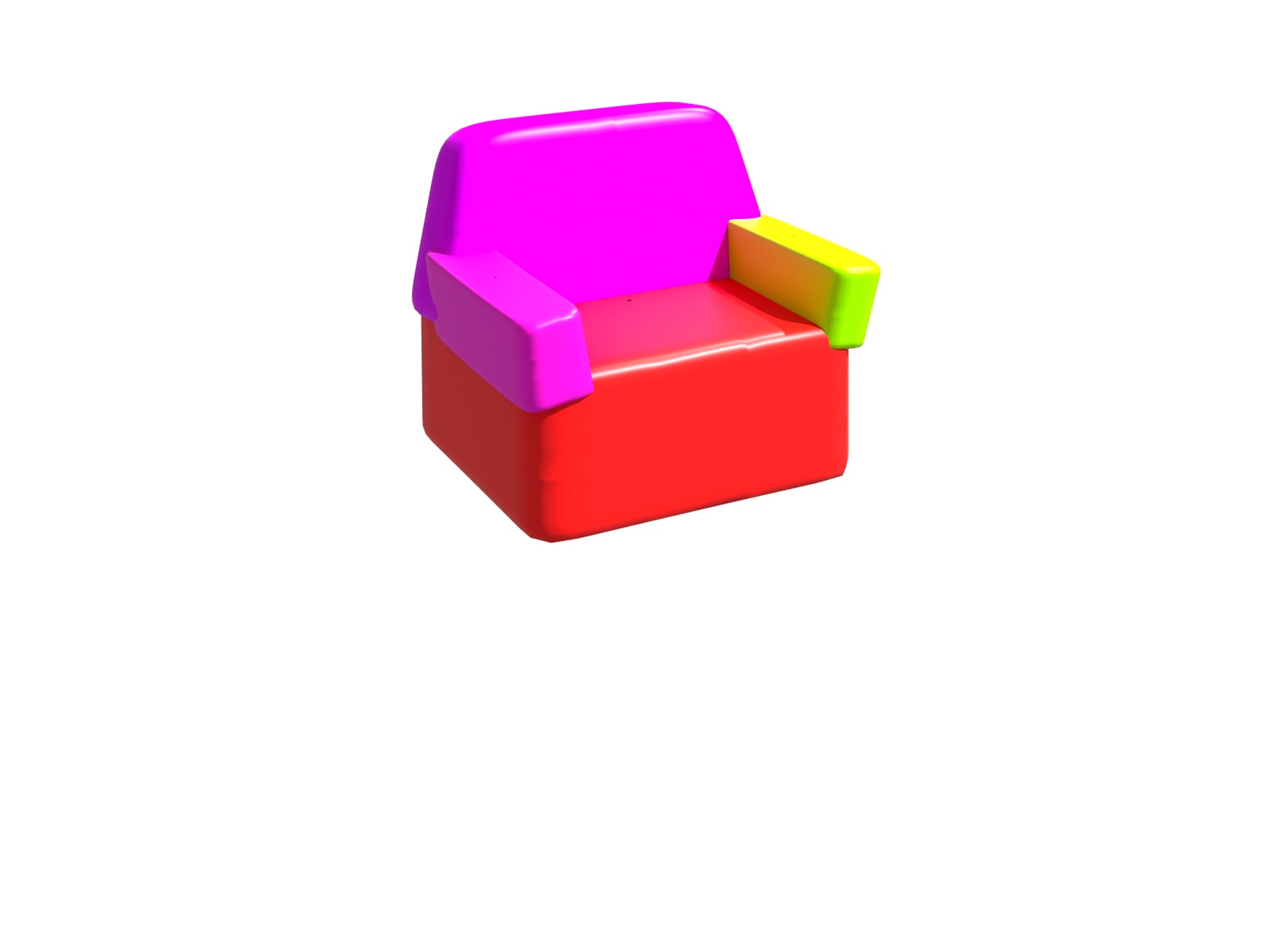}
\includegraphics[width=0.125\linewidth,trim={550px 450px 550px 200px},clip]{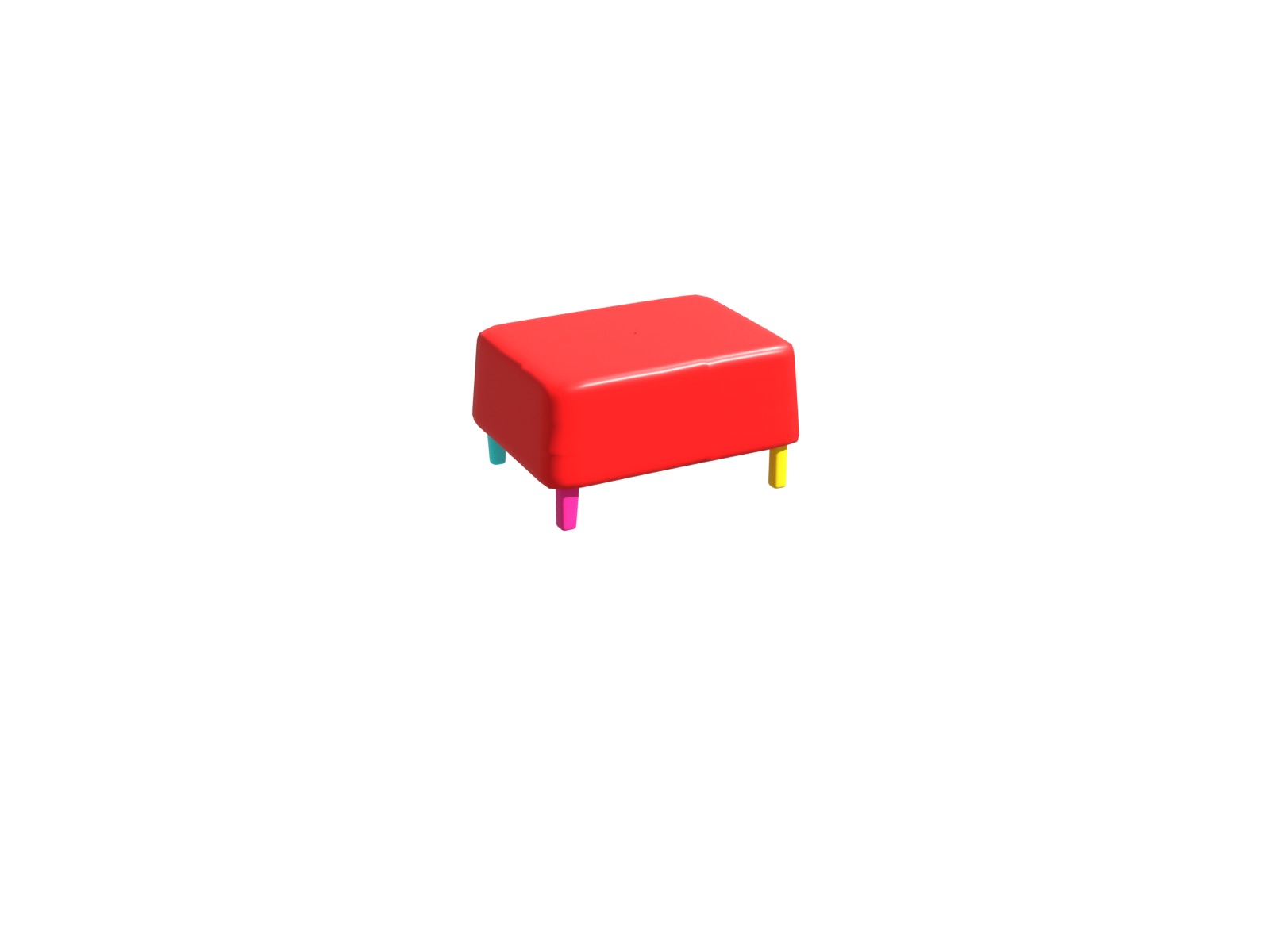}
\includegraphics[width=0.16\linewidth,trim={300px 300px 350px 1px},clip]{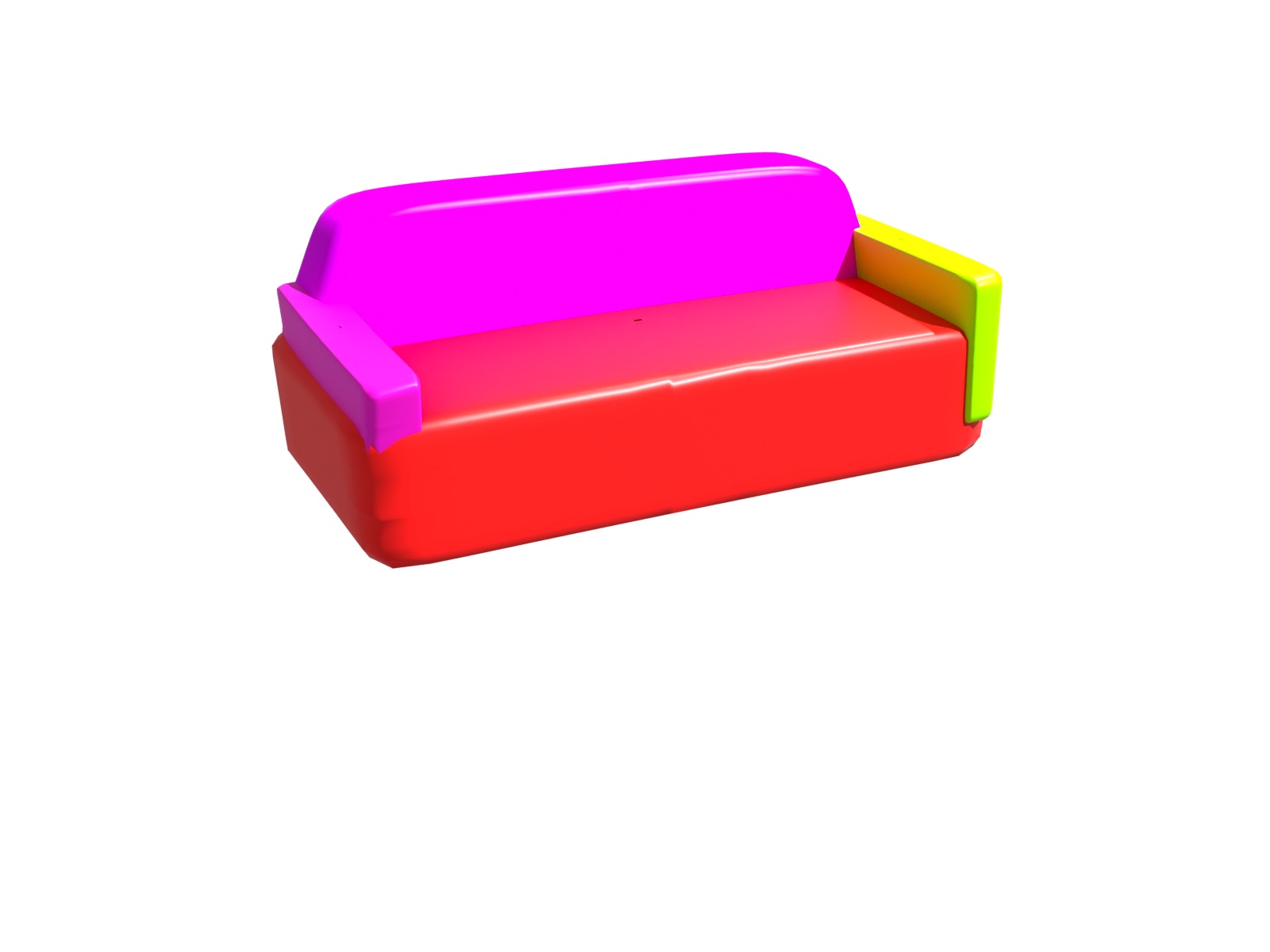}\\
\vspace{-10px}
\rotatebox{90}{\footnotesize \shortstack{\hspace{5px}\method{}\\ Optimized}}\hspace{-4px}
\includegraphics[width=0.14\linewidth,trim={600px 500px 600px 200px},clip]{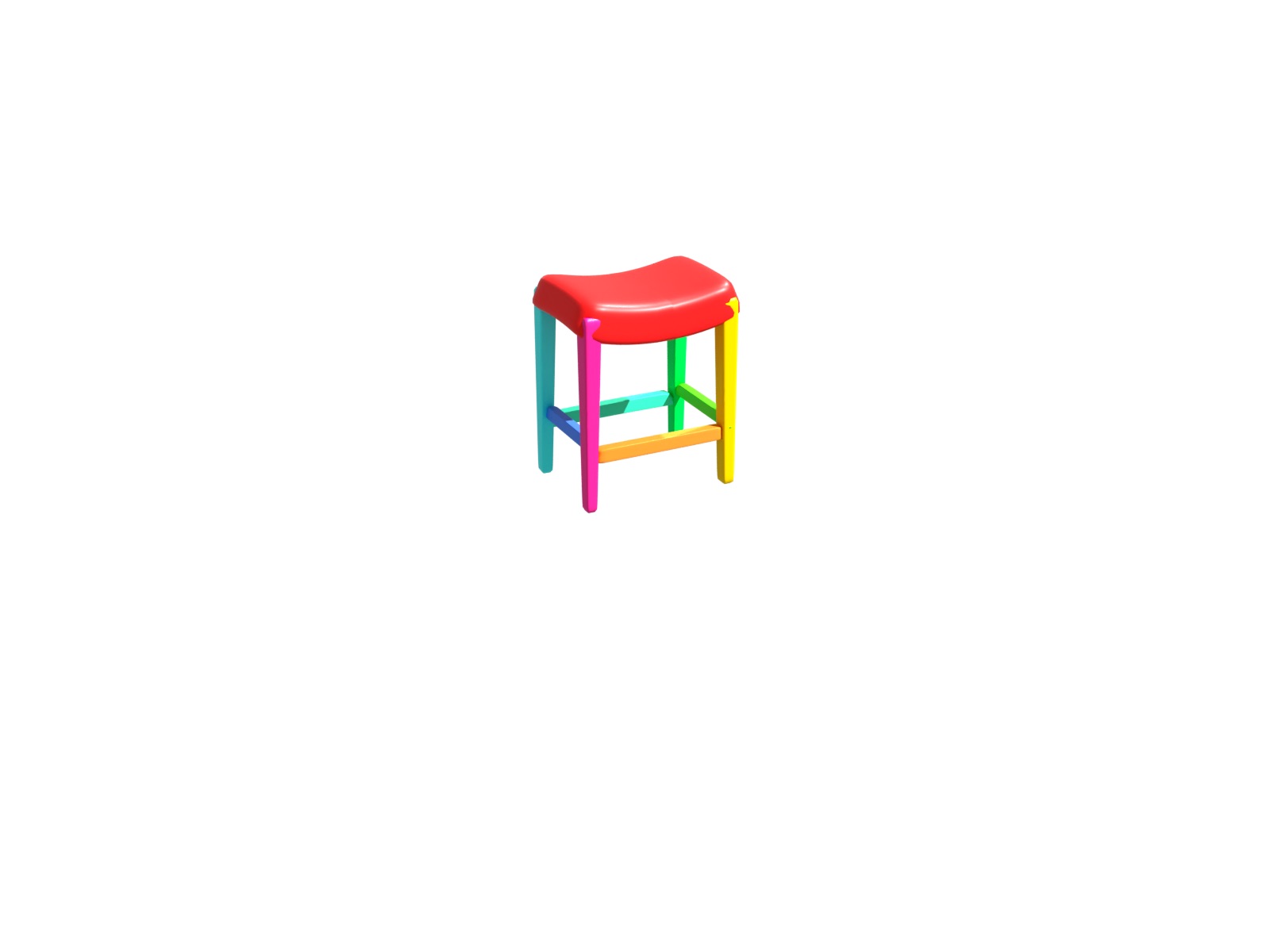}
\includegraphics[width=0.125\linewidth,trim={600px 500px 600px 200px},clip]{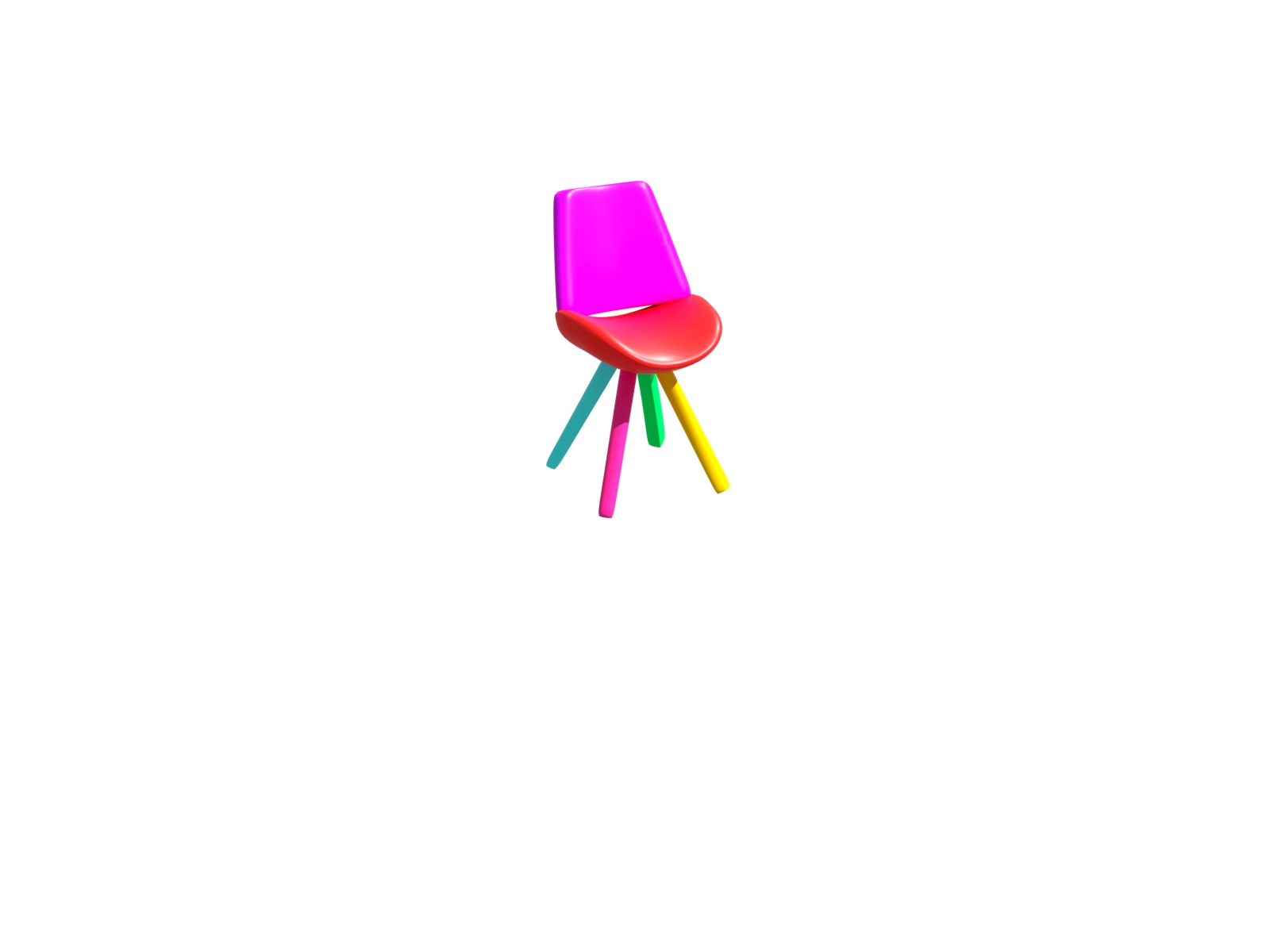}
\includegraphics[width=0.11\linewidth,trim={600px 450px 600px 200px},clip]{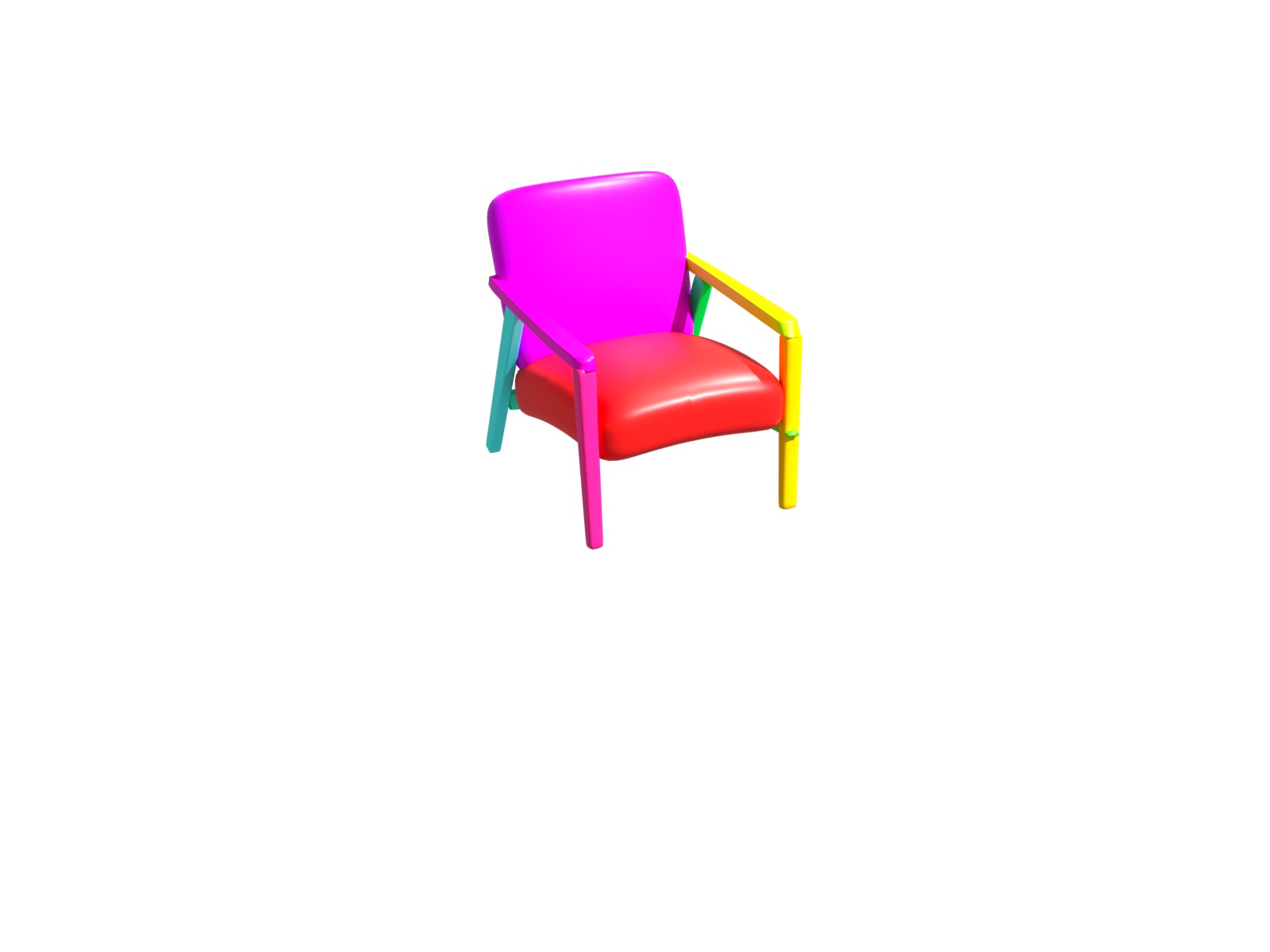}
\includegraphics[width=0.115\linewidth,trim={600px 500px 600px 100px},clip]{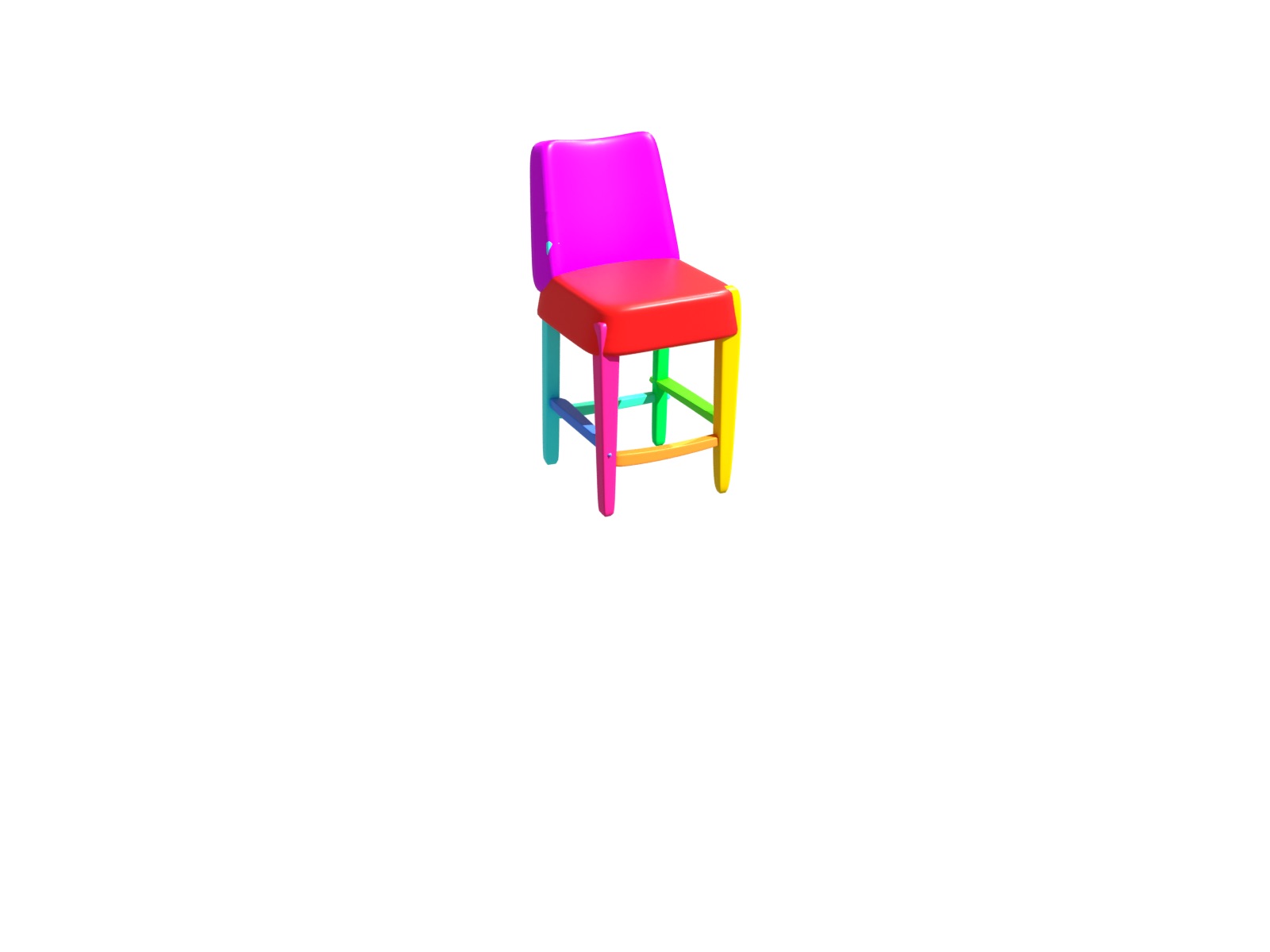}
\includegraphics[width=0.115\linewidth,trim={500px 400px 500px 1px},clip]{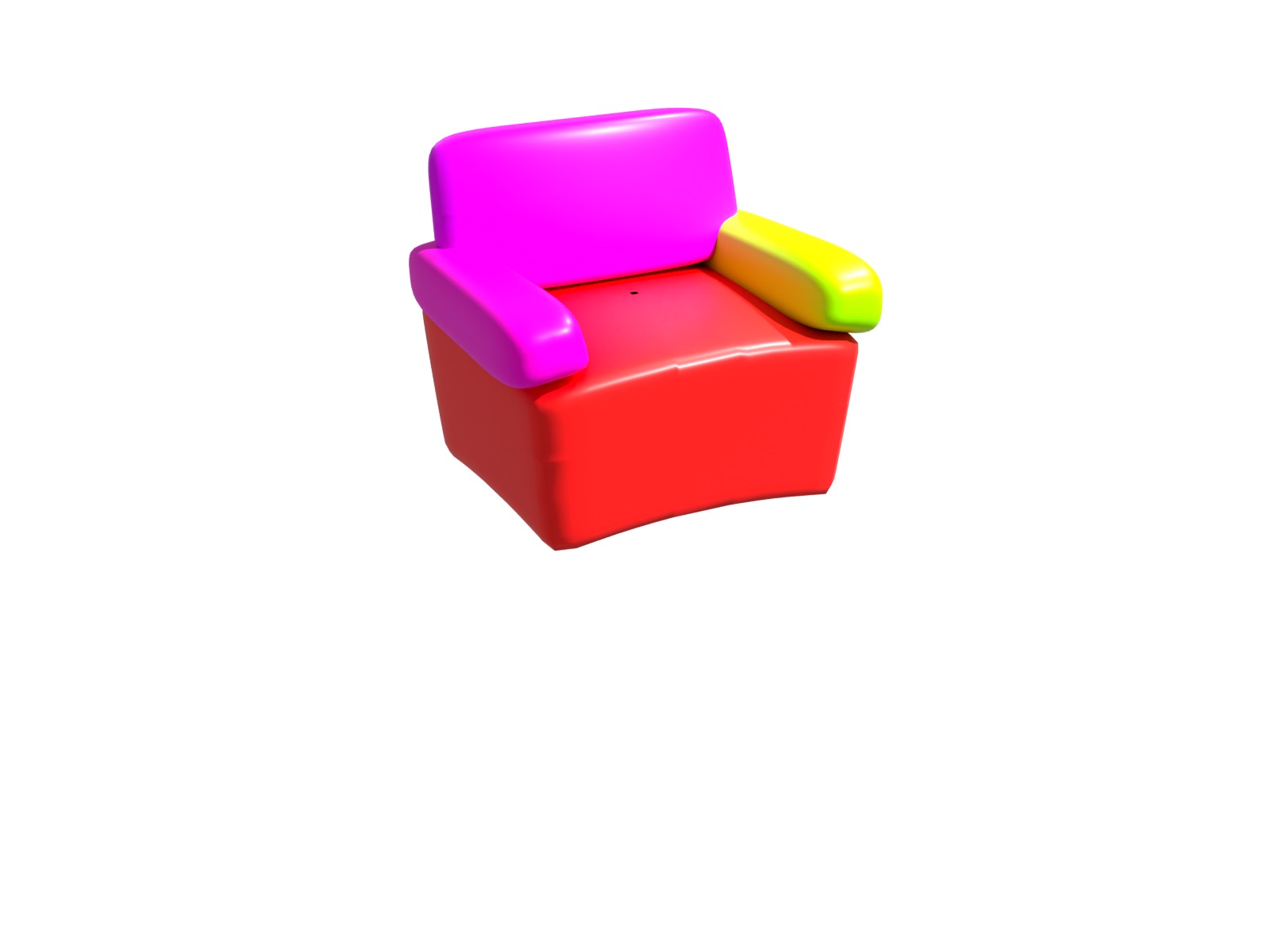}
\includegraphics[width=0.125\linewidth,trim={550px 450px 550px 200px},clip]{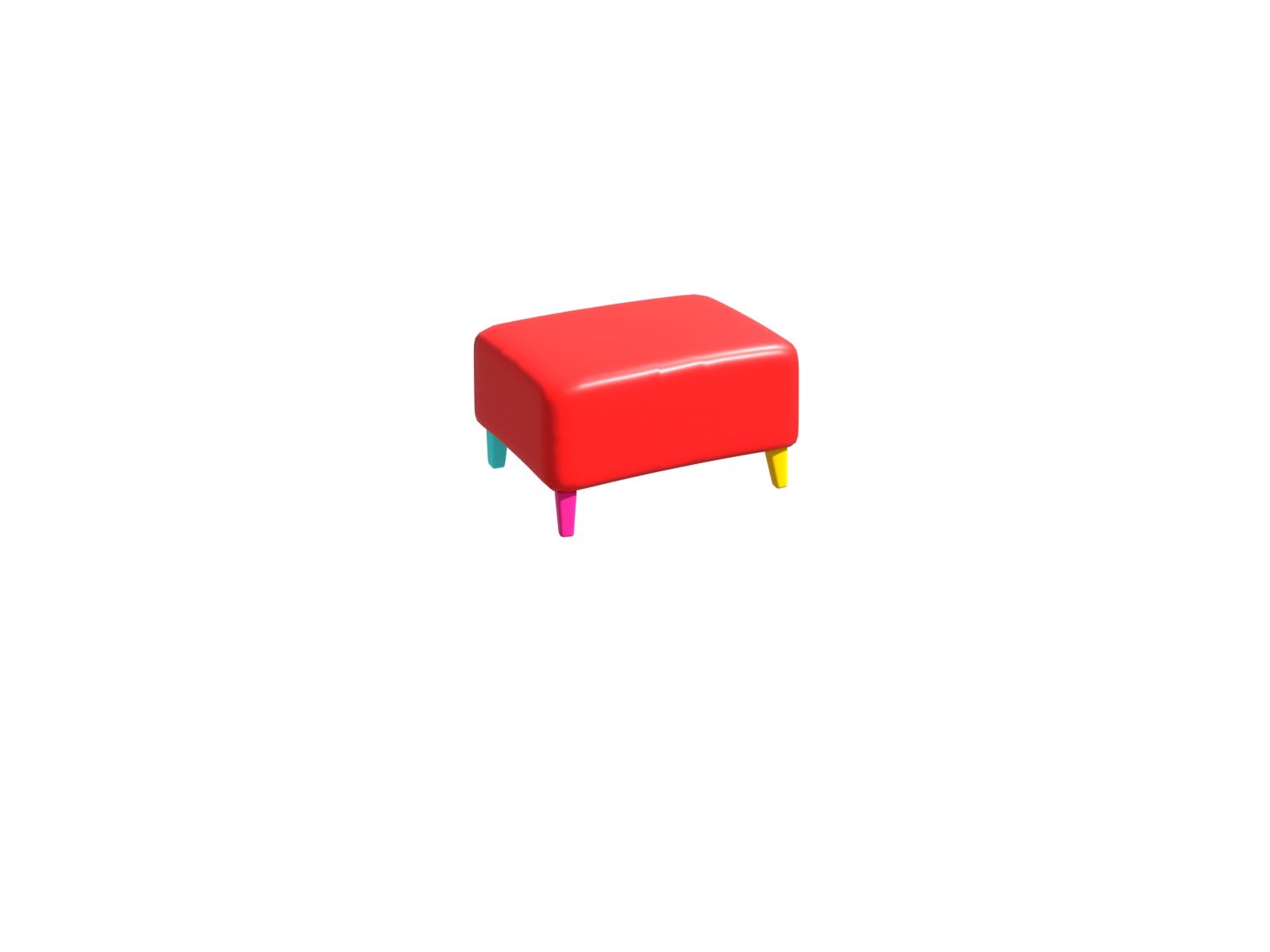}
\includegraphics[width=0.16\linewidth,trim={300px 300px 350px 1px},clip]{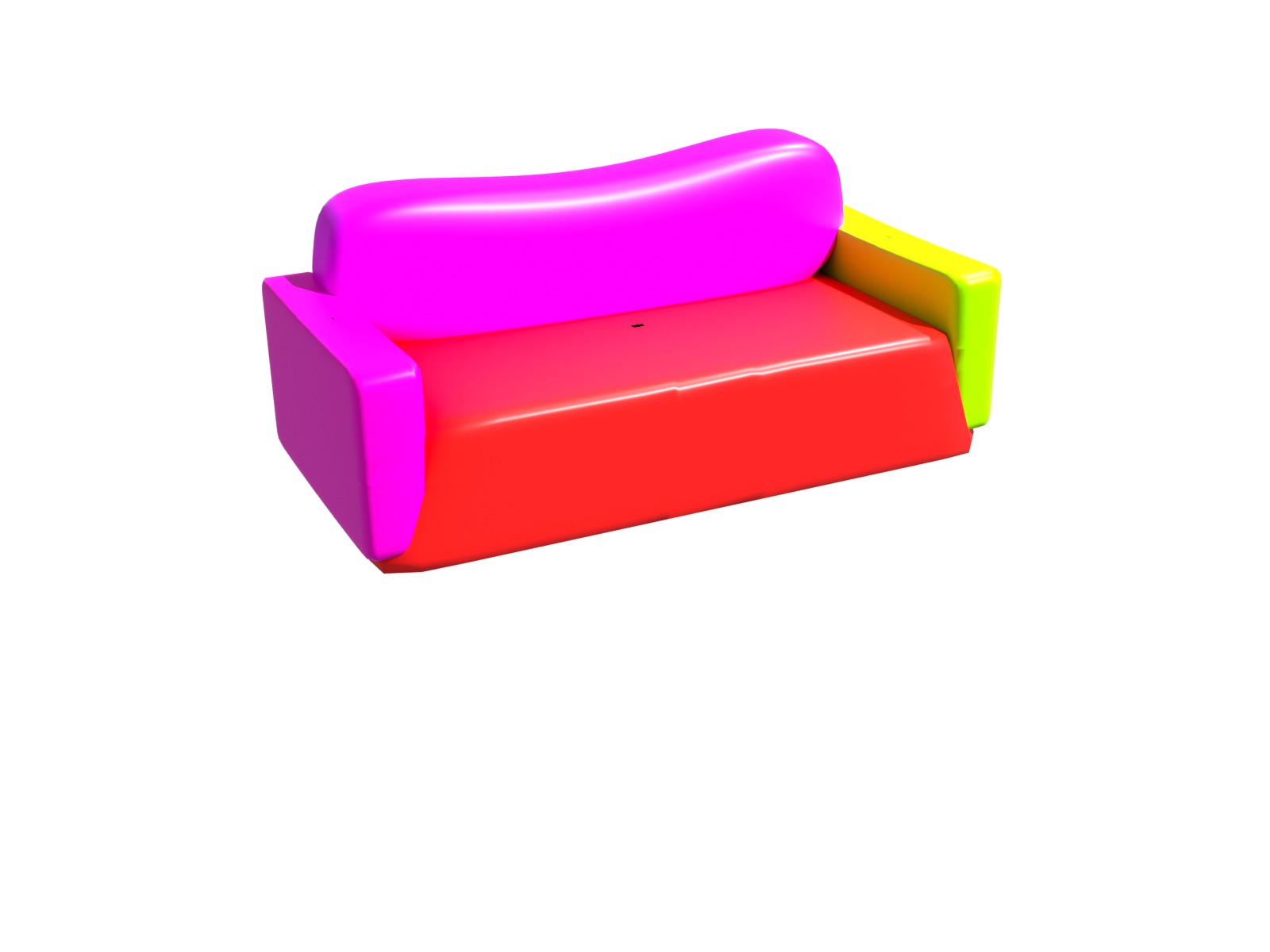}\\

\vspace{0.4cm}
\rotatebox{90}{\footnotesize \shortstack{Input \\ Pointcloud}}
\includegraphics[width=0.125\linewidth, height=1.7cm,trim={520px 480px 520px 200px},clip]{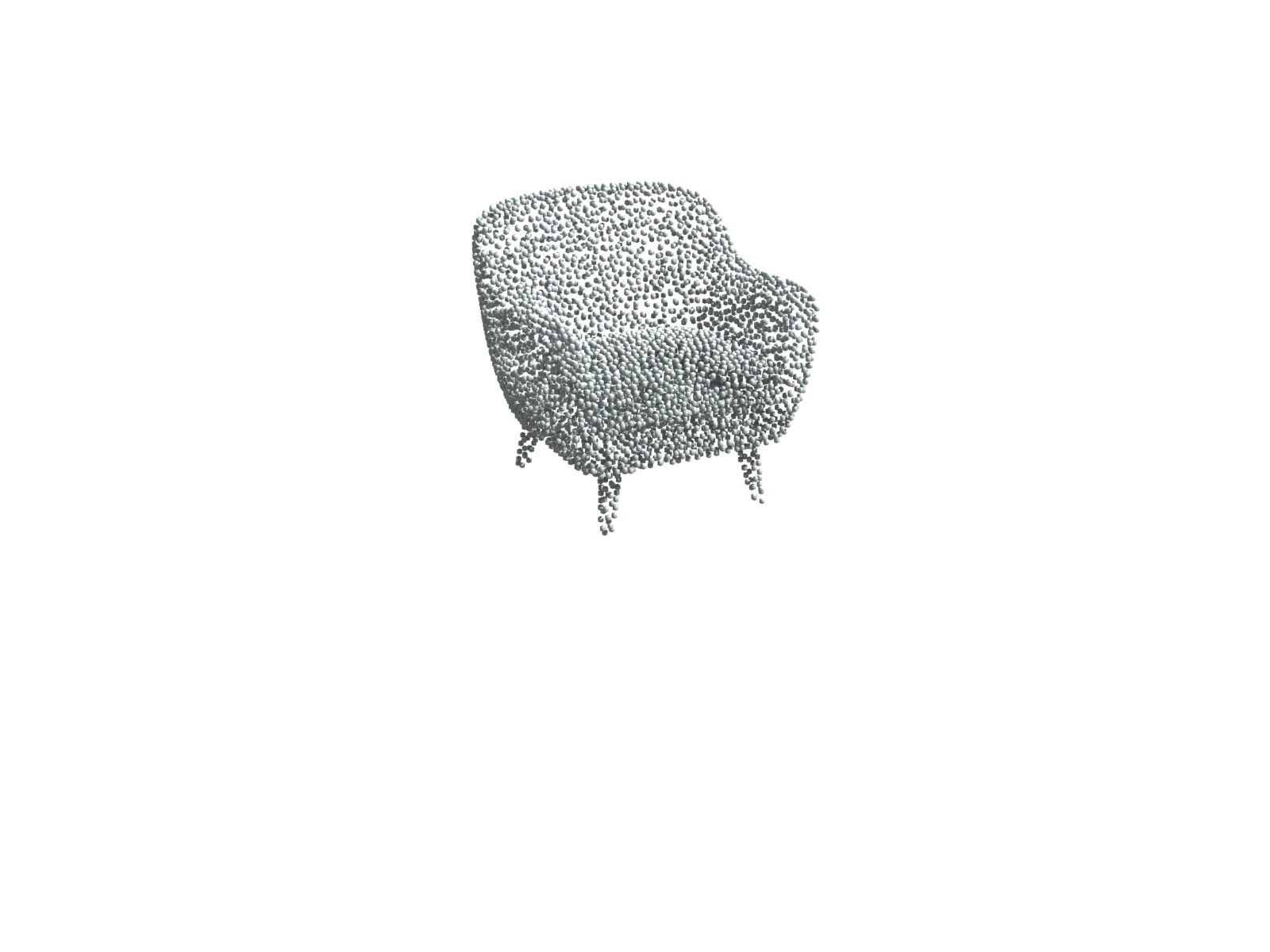}
\includegraphics[width=0.125\linewidth, trim={500px 400px 500px 100px},clip]{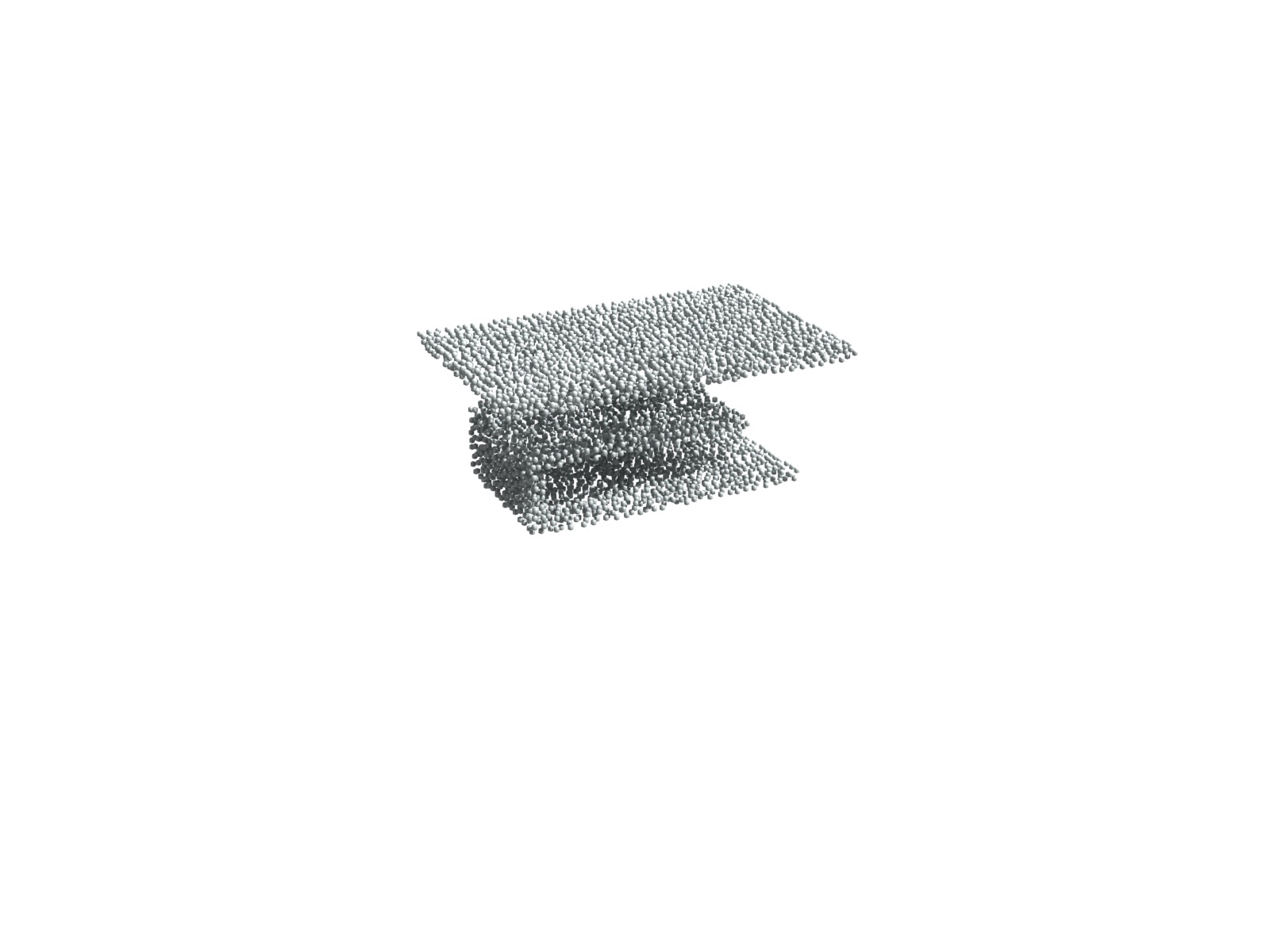}
\includegraphics[width=0.125\linewidth, height=1.7cm,trim={520px 450px 520px 150px},clip]{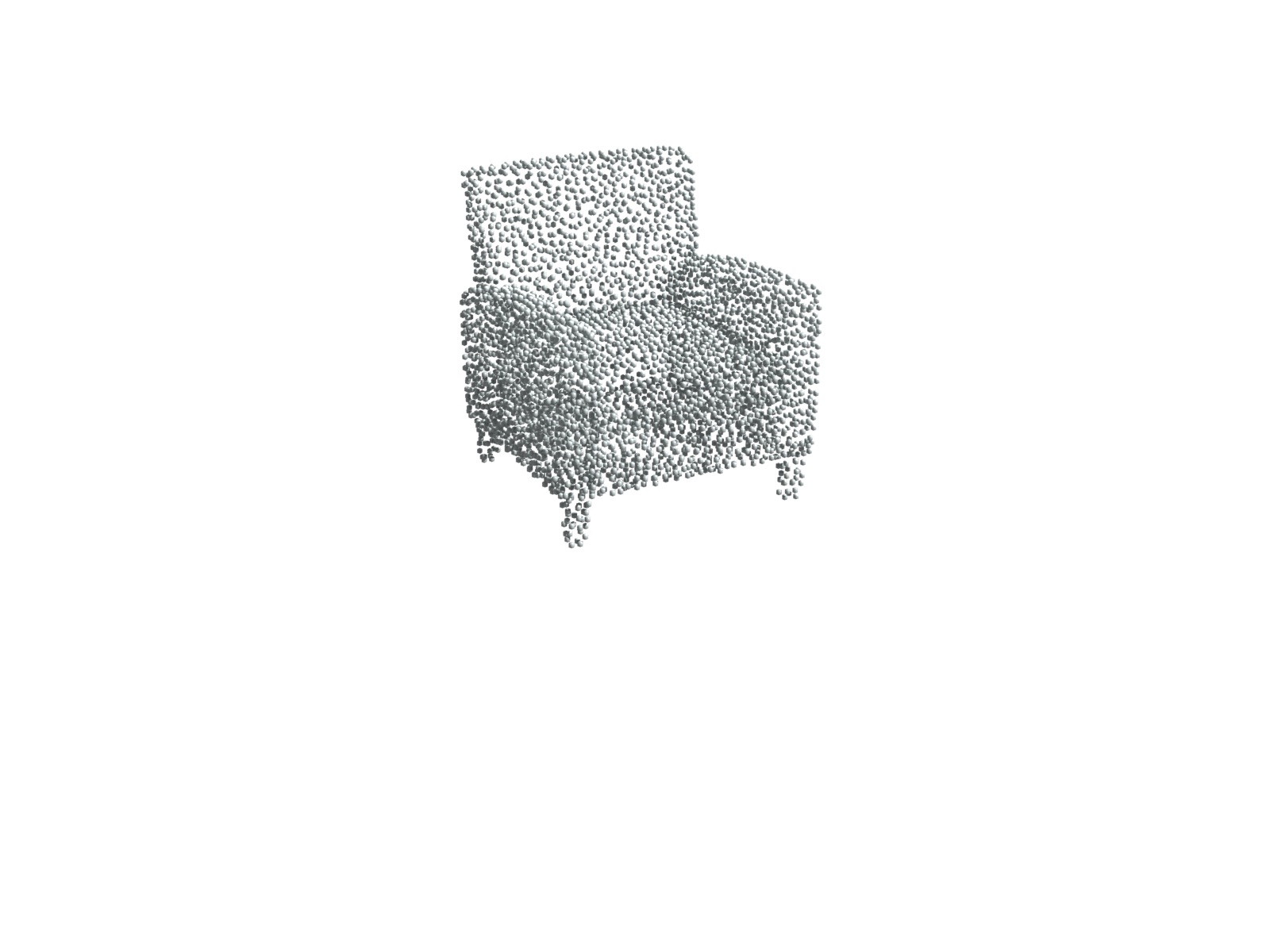}
\includegraphics[width=0.125\linewidth, trim={250px 200px 200px 1px},clip]{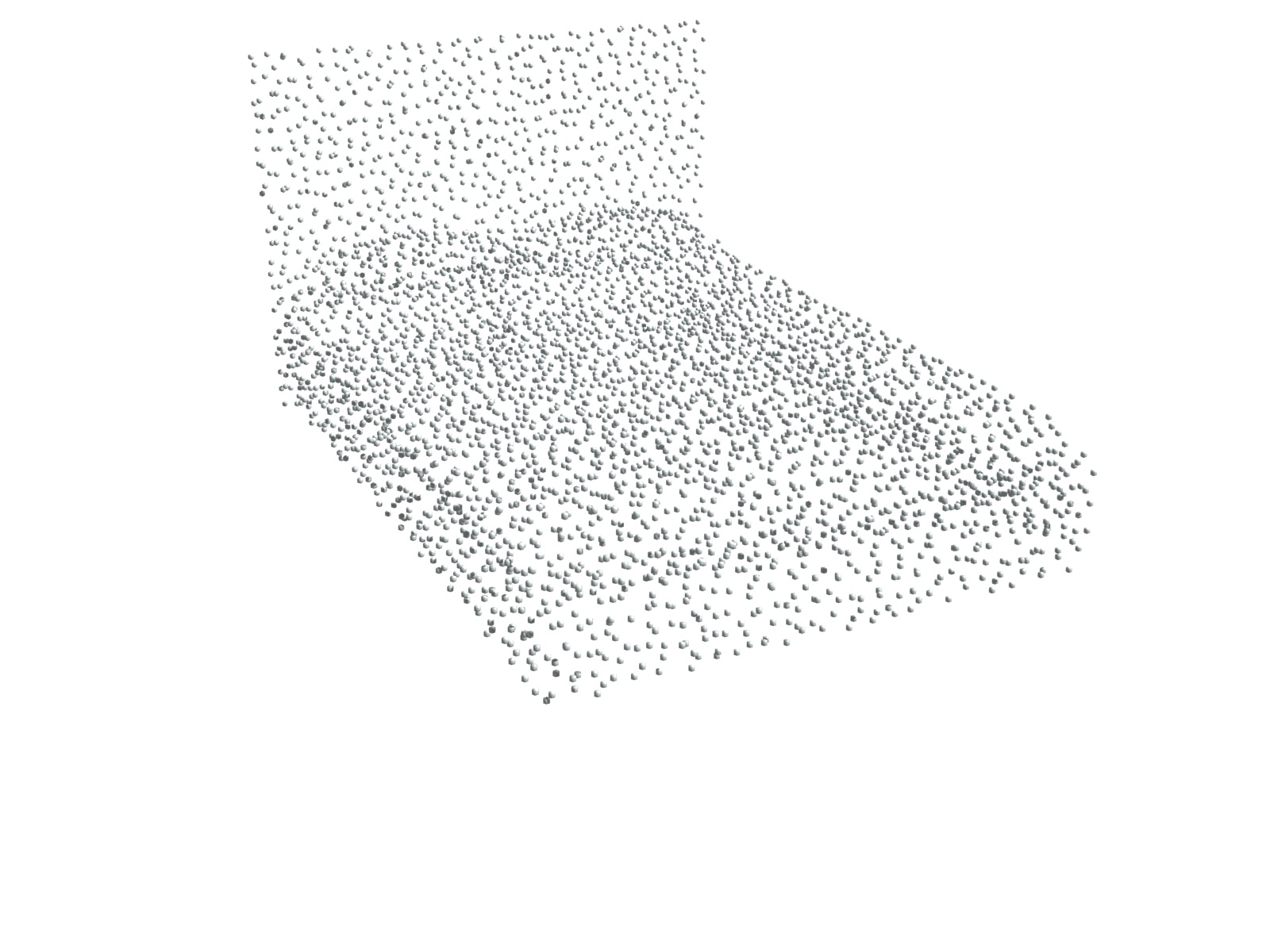}
\includegraphics[width=0.125\linewidth, trim={550px 450px 500px 300px},clip]{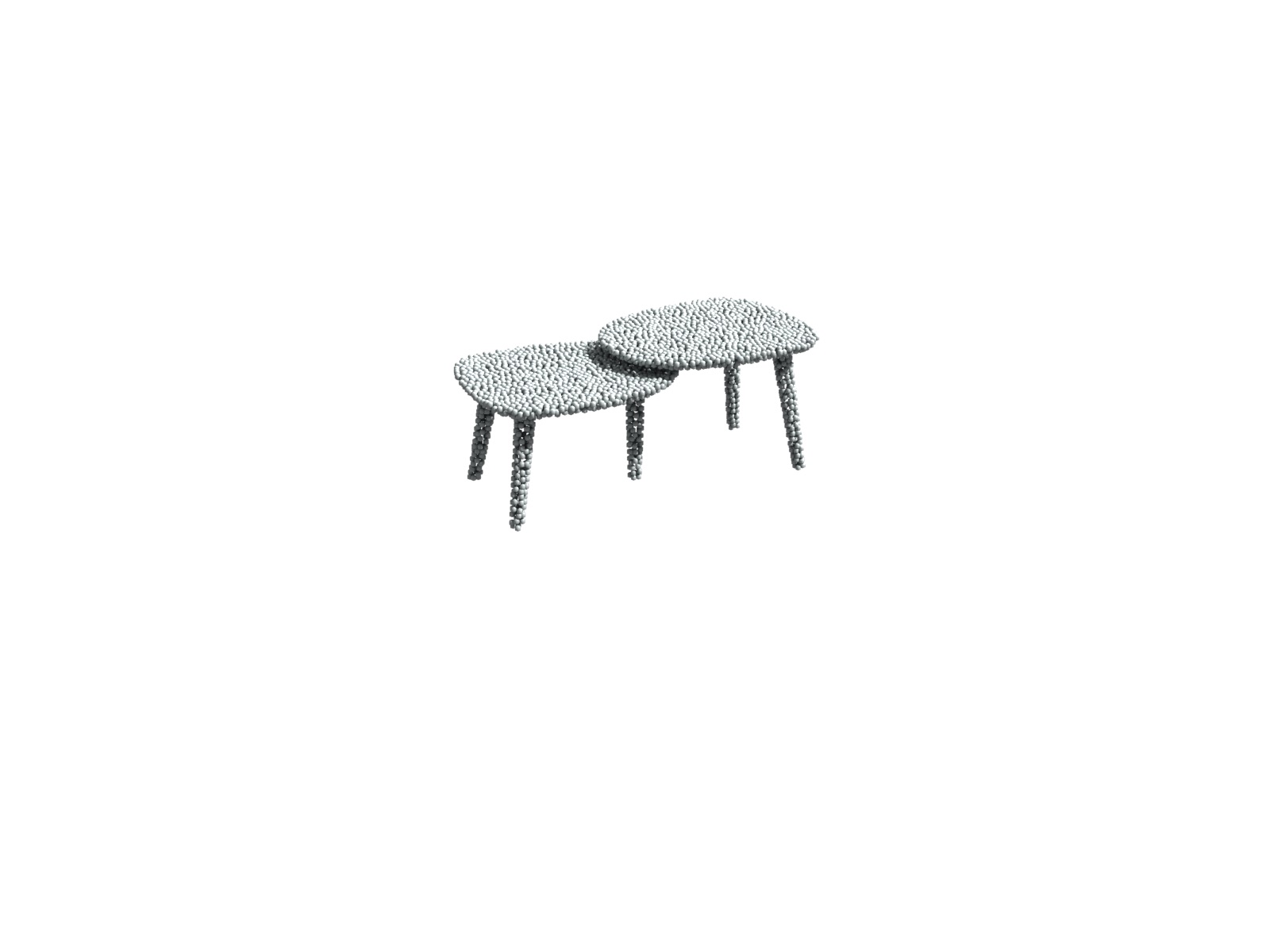}
\includegraphics[width=0.125\linewidth,trim={520px 480px 520px 200px},clip]{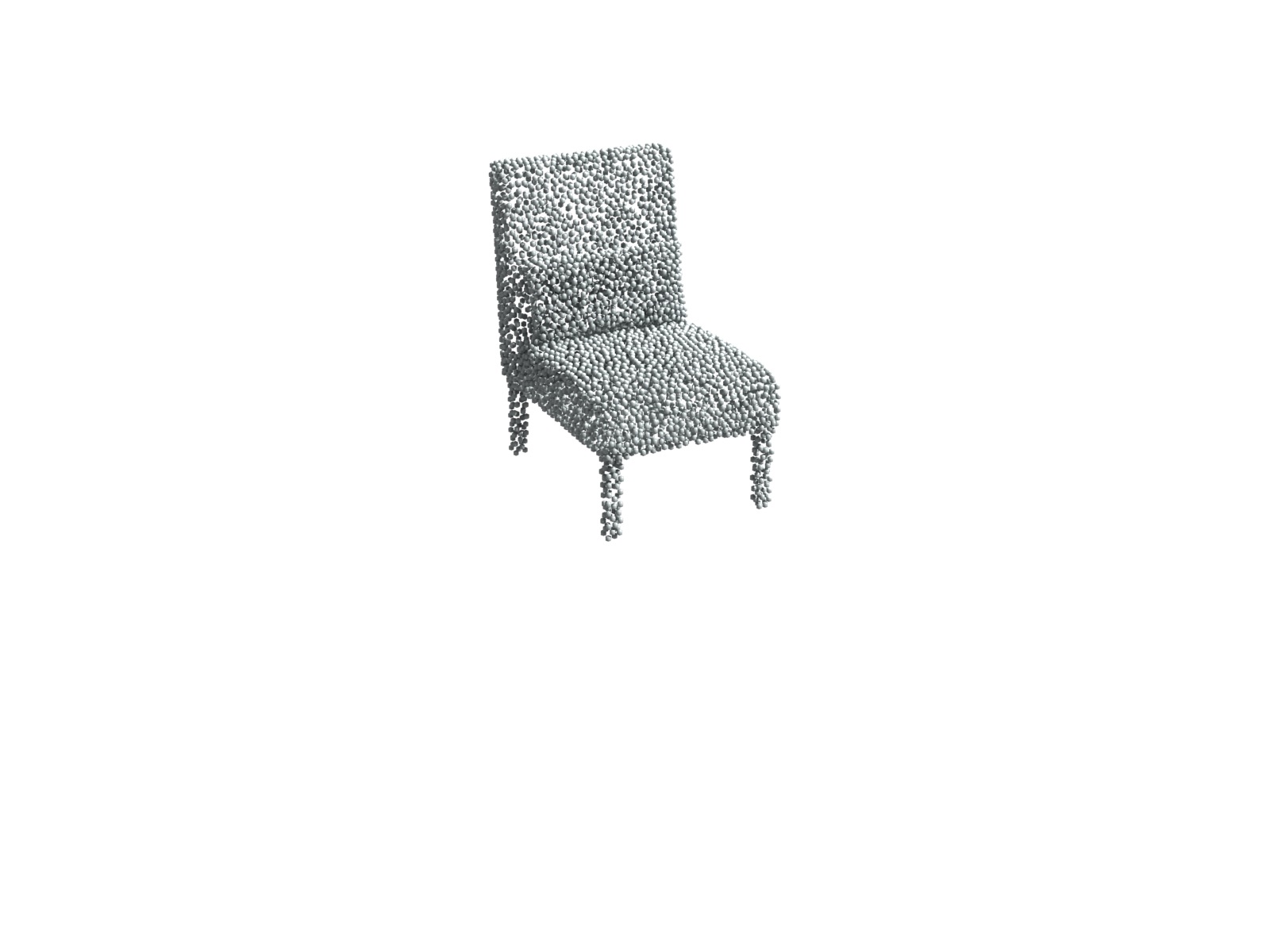}
\includegraphics[width=0.125\linewidth, trim={520px 480px 520px 200px},clip]{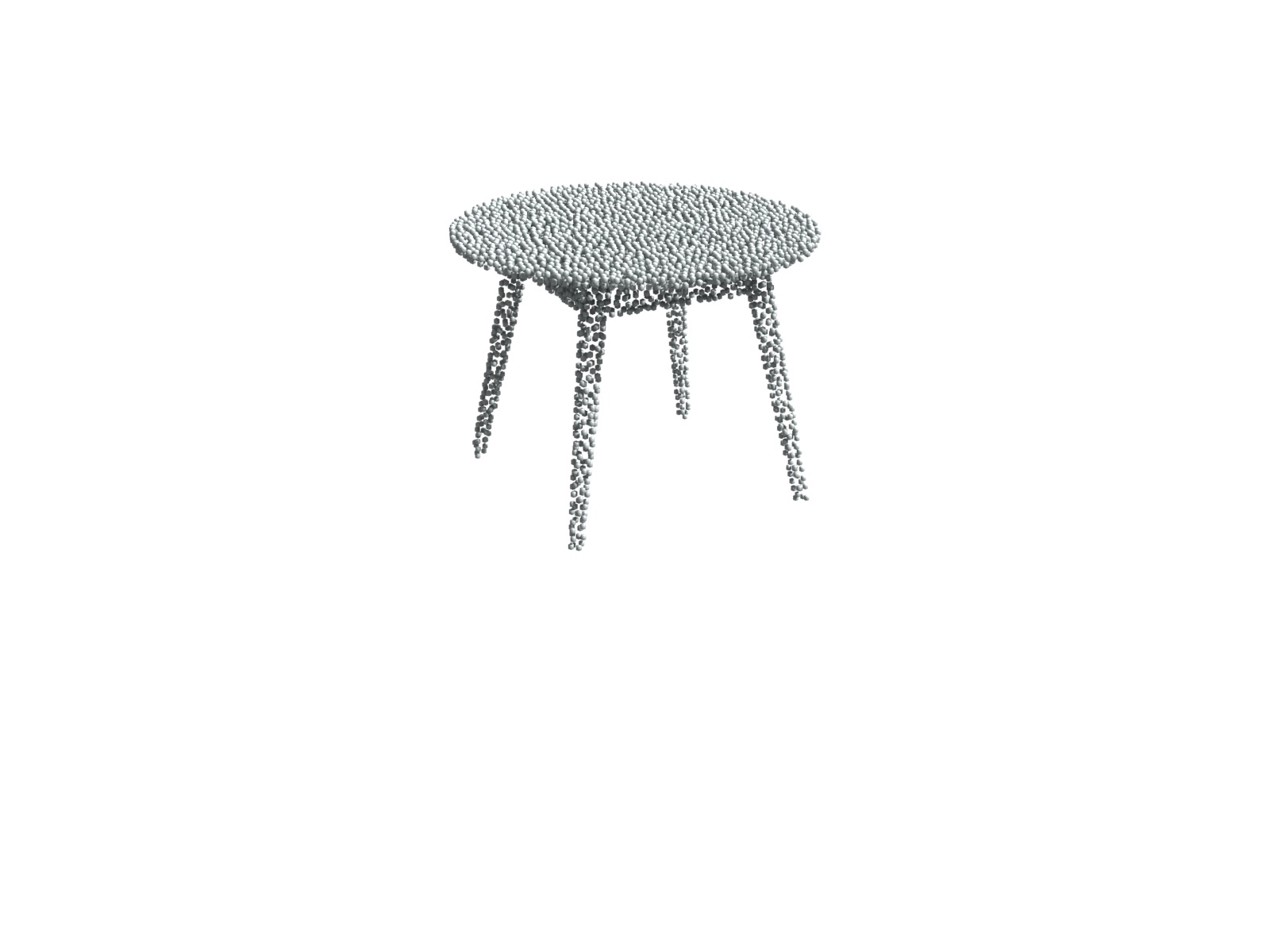}\\
\vspace{-2px}
\rotatebox{90}{\footnotesize \shortstack{\\ \hspace{5px}\method{}}}\hspace{4px}
\includegraphics[width=0.125\linewidth, height=1.7cm,trim={520px 480px 520px 200px},clip]{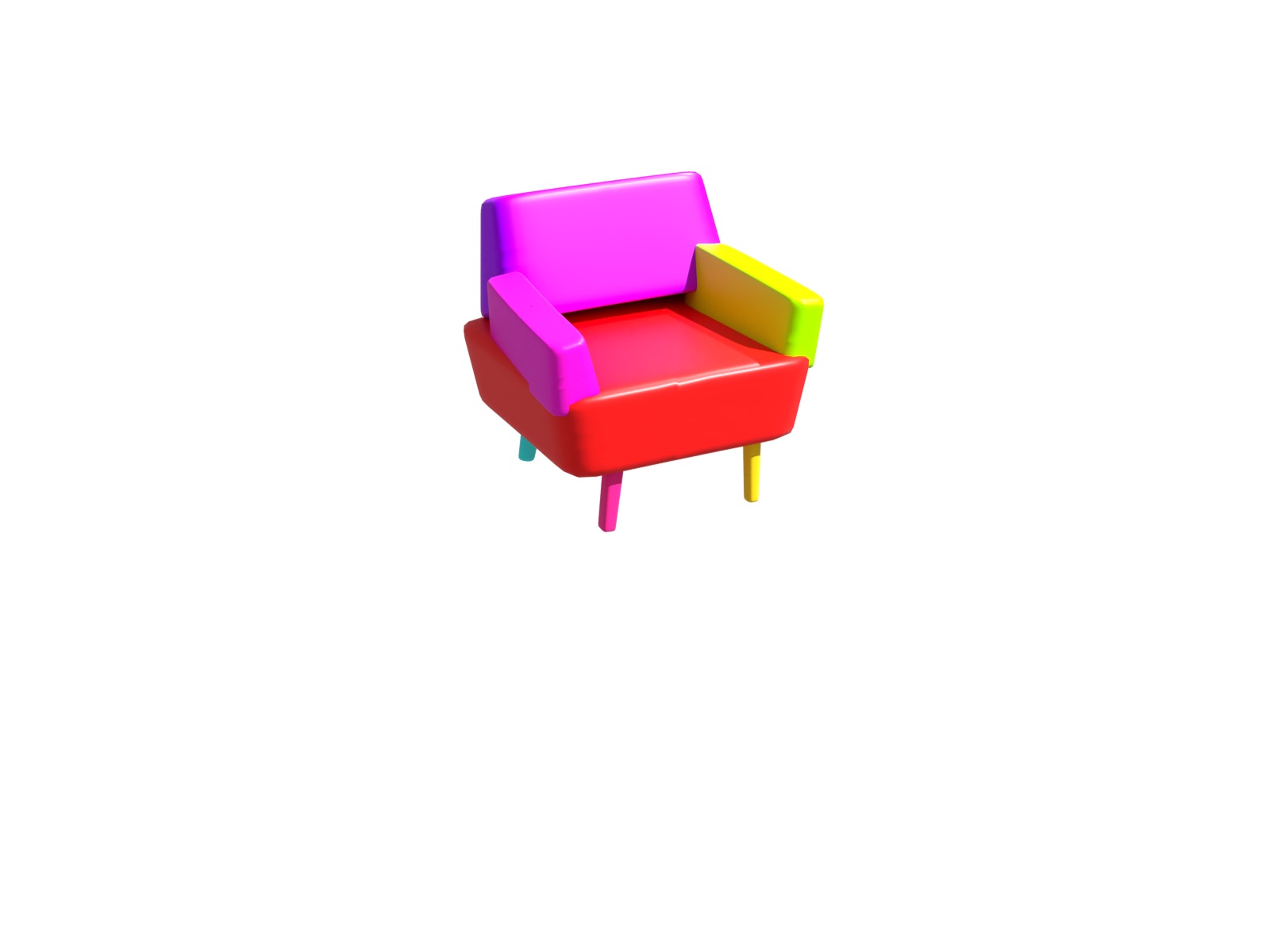}
\includegraphics[width=0.125\linewidth, trim={500px 400px 500px 100px},clip]{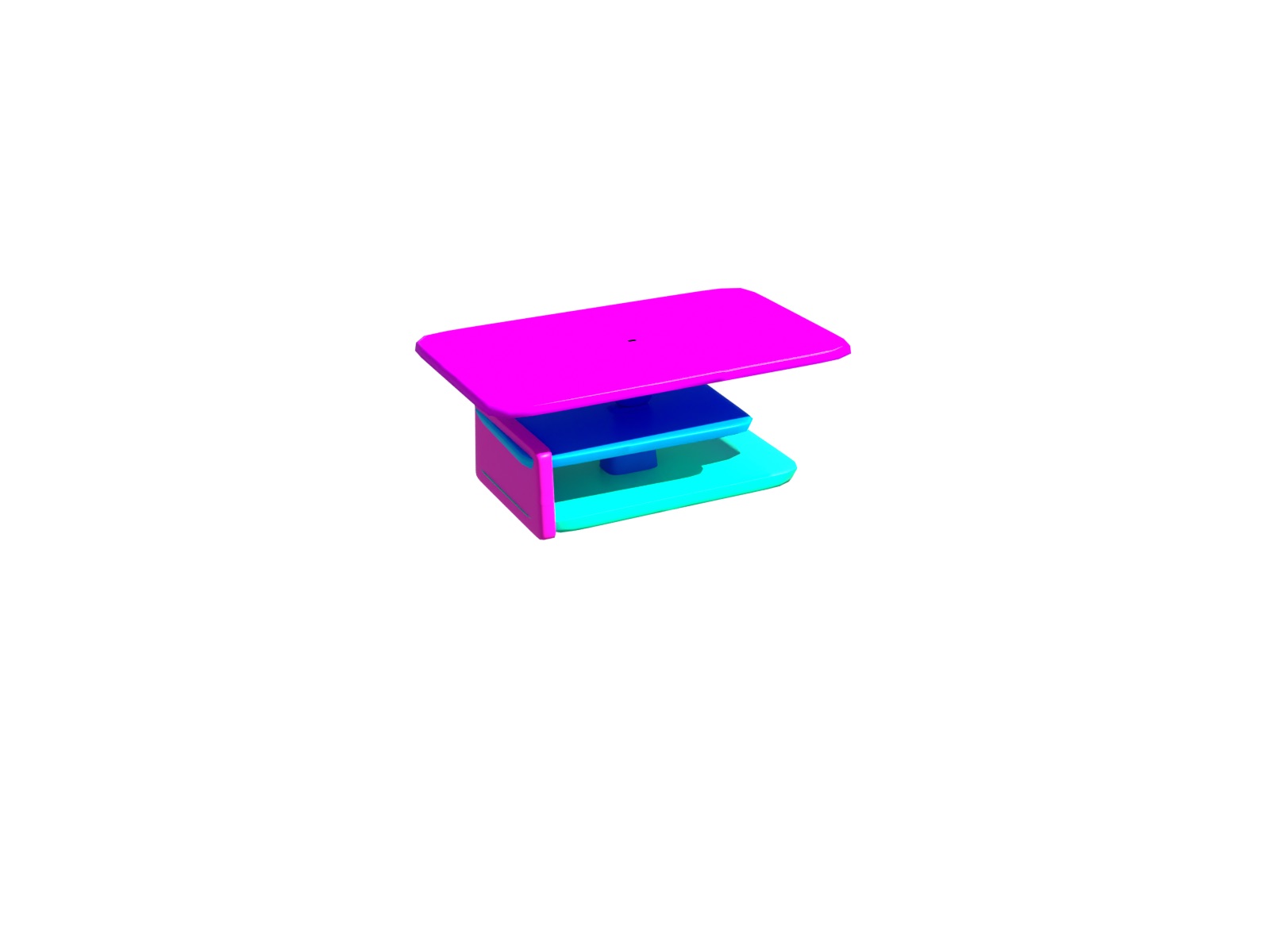}
\includegraphics[width=0.125\linewidth, height=1.7cm,trim={520px 450px 520px 150px},clip]{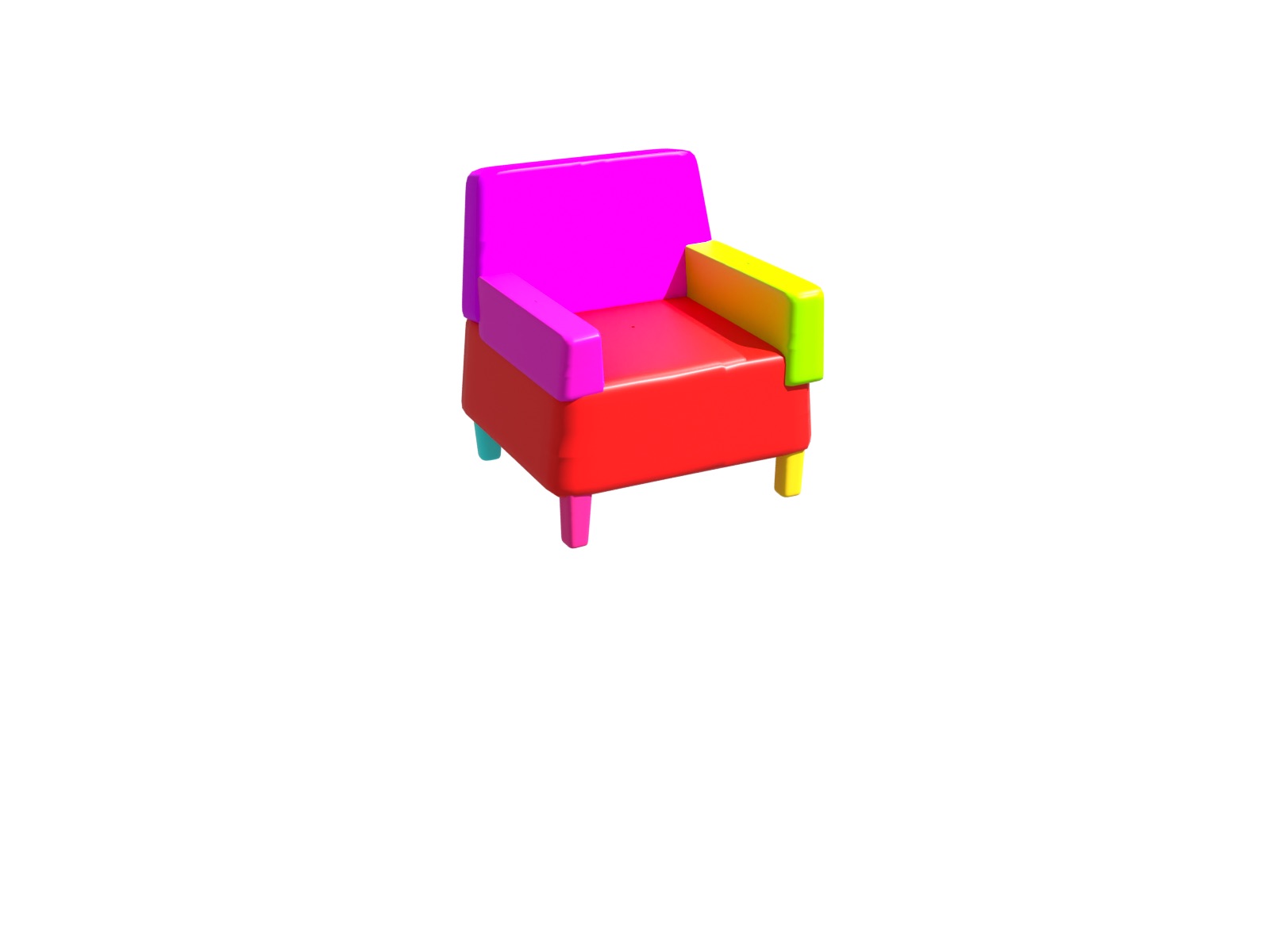}
\includegraphics[width=0.125\linewidth, trim={250px 200px 200px 1px},clip]{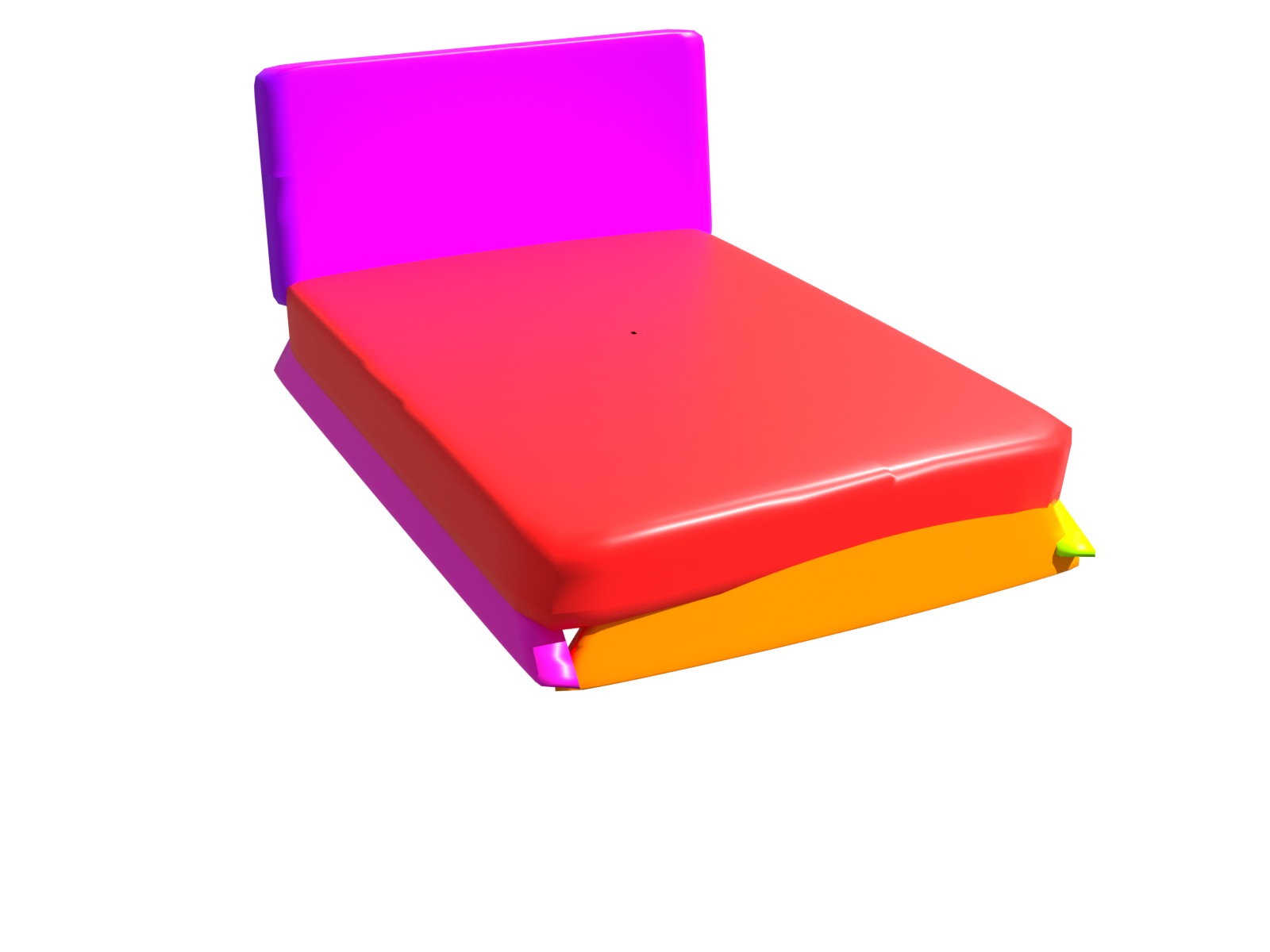}
\includegraphics[width=0.125\linewidth, trim={550px 450px 500px 300px},clip]{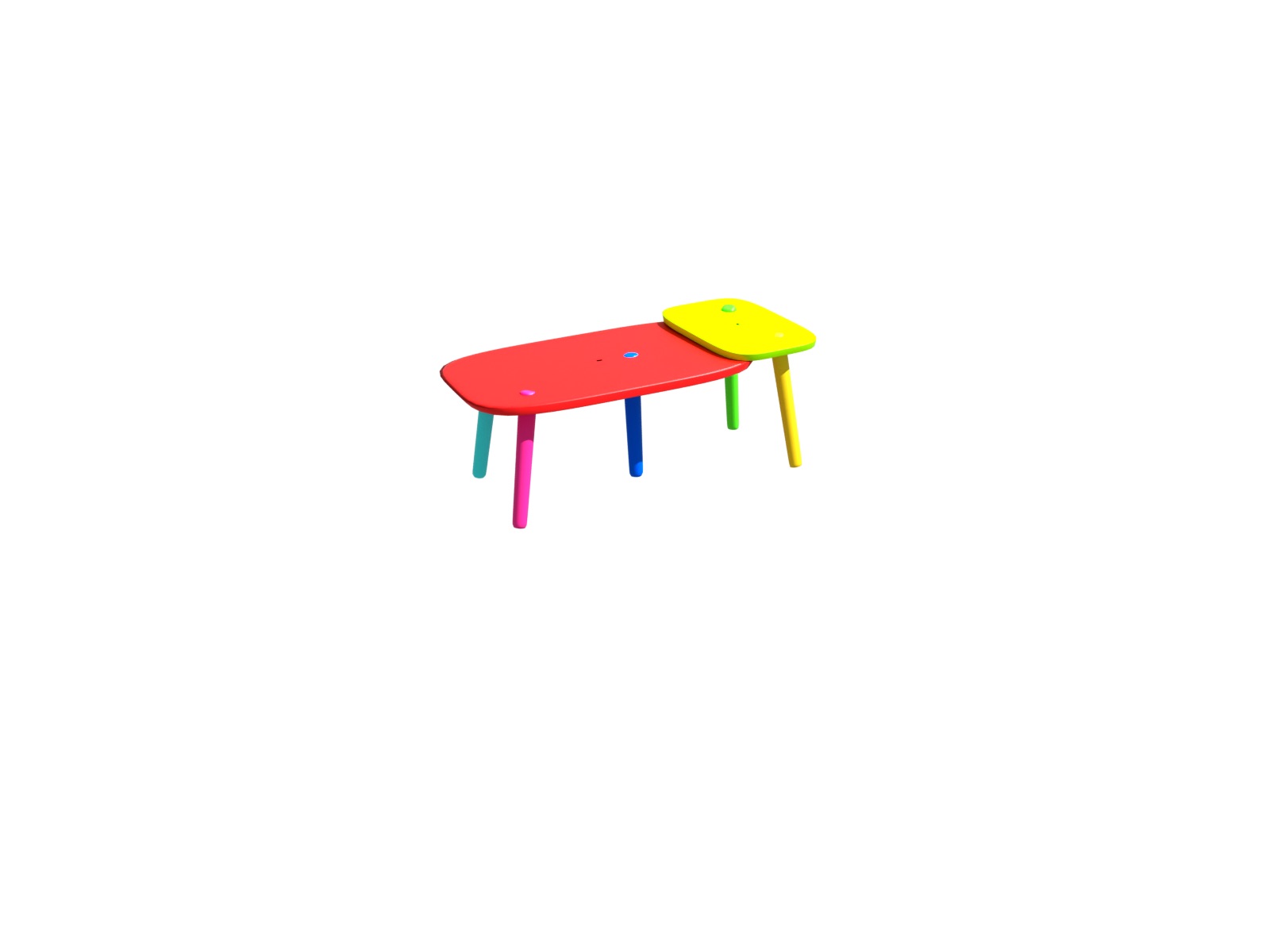}
\includegraphics[width=0.125\linewidth,trim={520px 480px 520px 200px},clip]{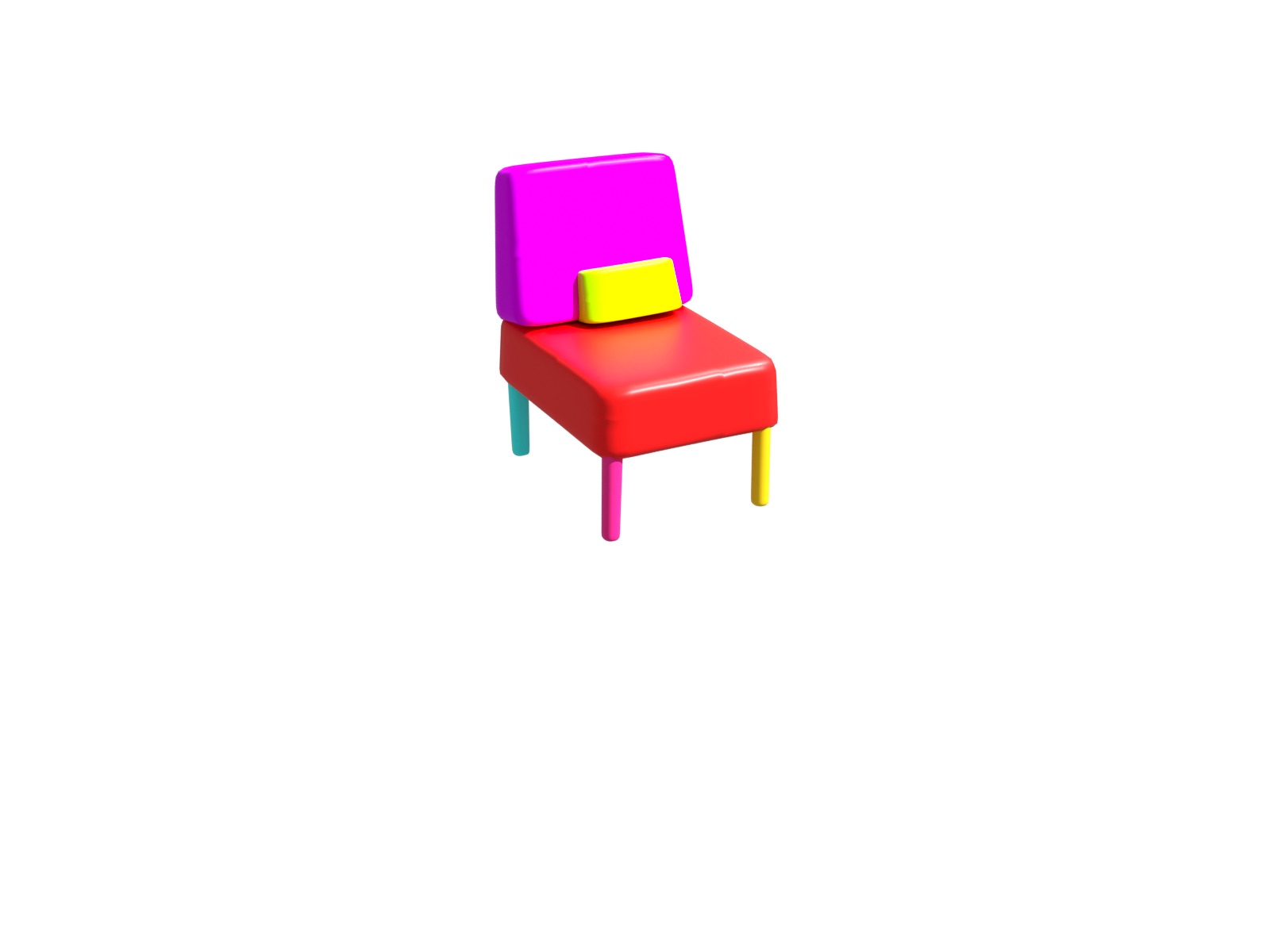}
\includegraphics[width=0.125\linewidth, trim={520px 480px 520px 200px},clip]{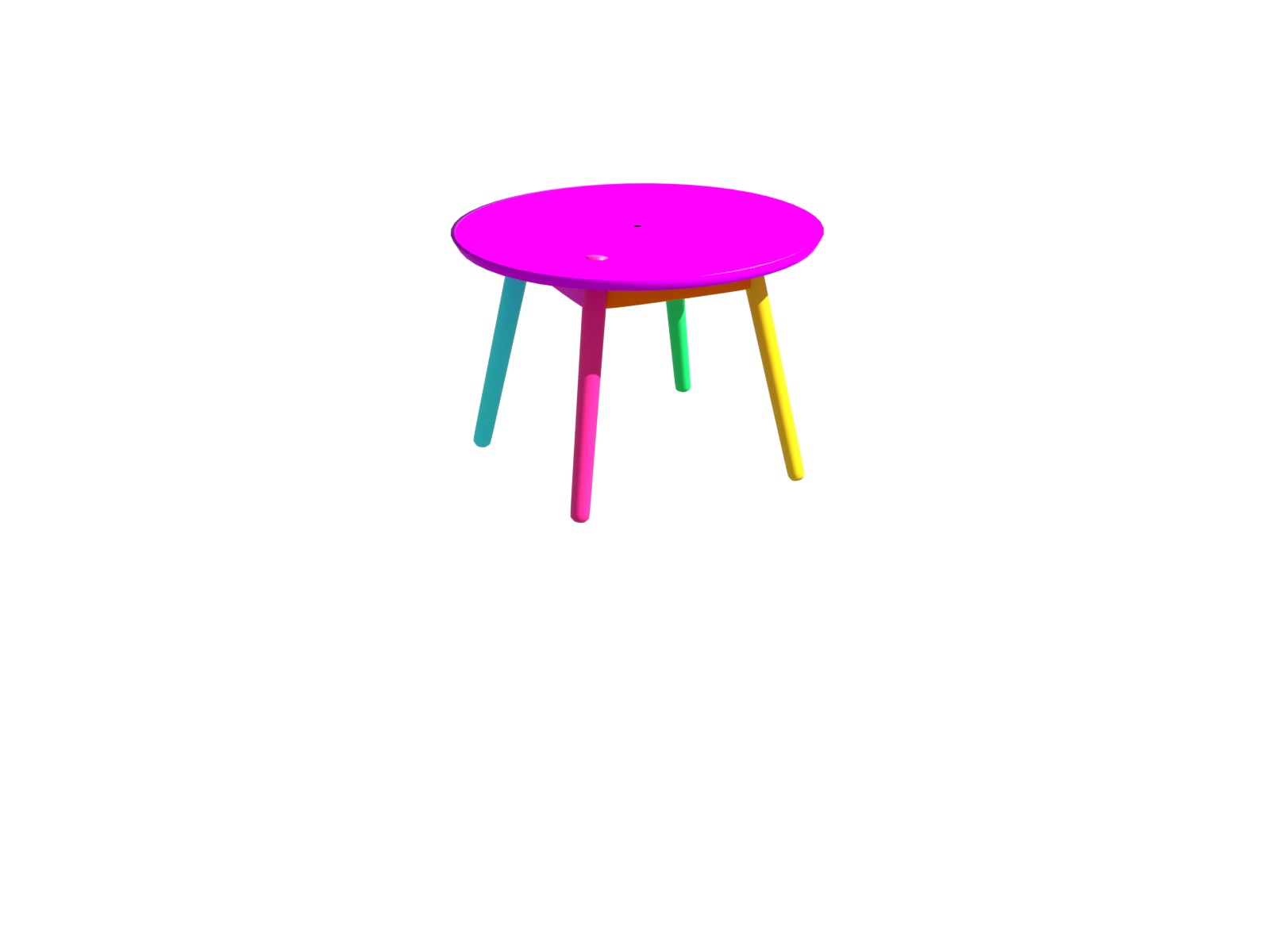}\\
\vspace{-2px}
\rotatebox{90}{\footnotesize \shortstack{\hspace{5px}\method{}\\ Optimized}}\hspace{-4px}
\includegraphics[width=0.125\linewidth, height=1.7cm,trim={520px 480px 520px 200px},clip]{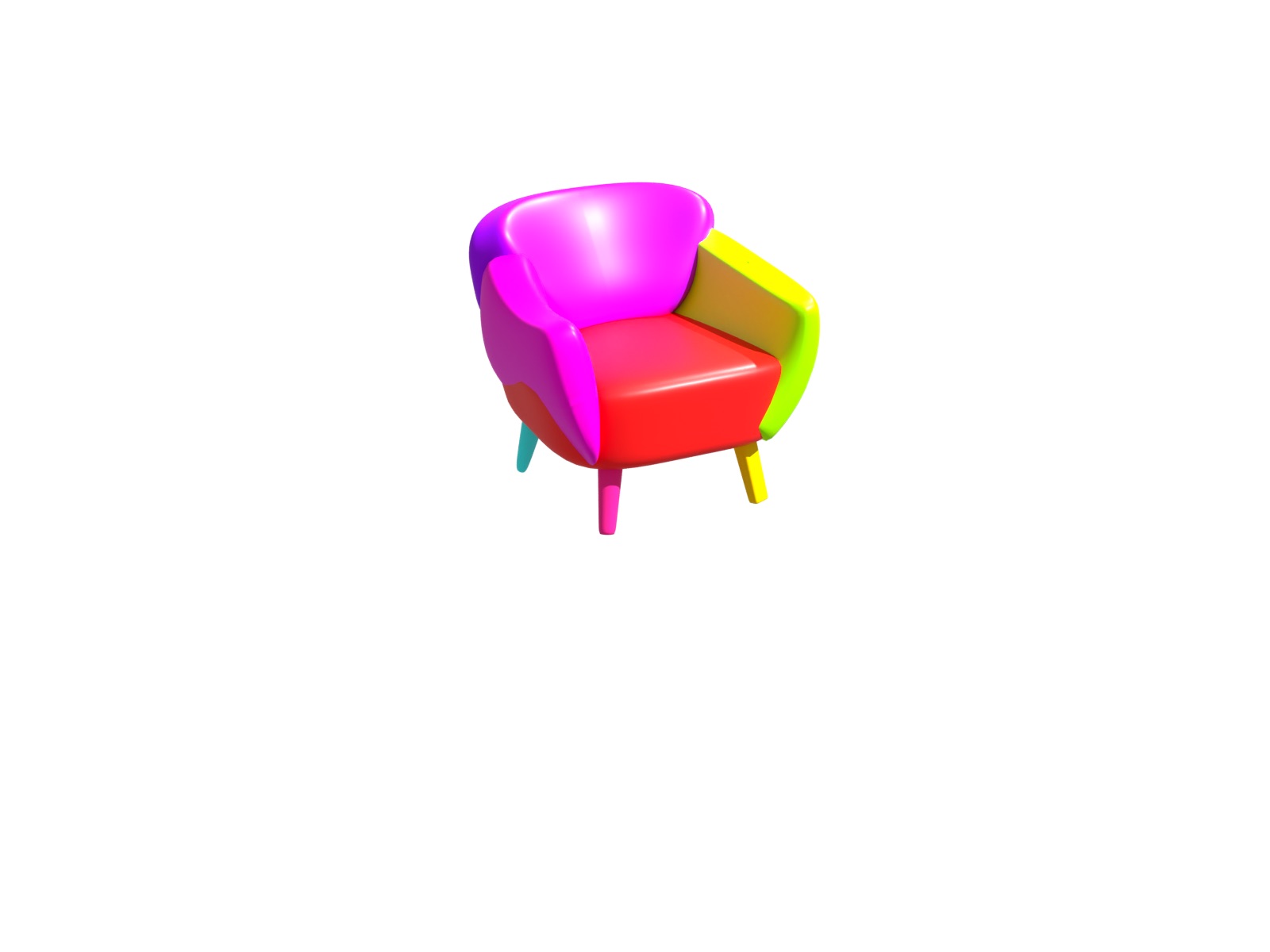}
\includegraphics[width=0.125\linewidth, trim={500px 400px 500px 100px},clip]{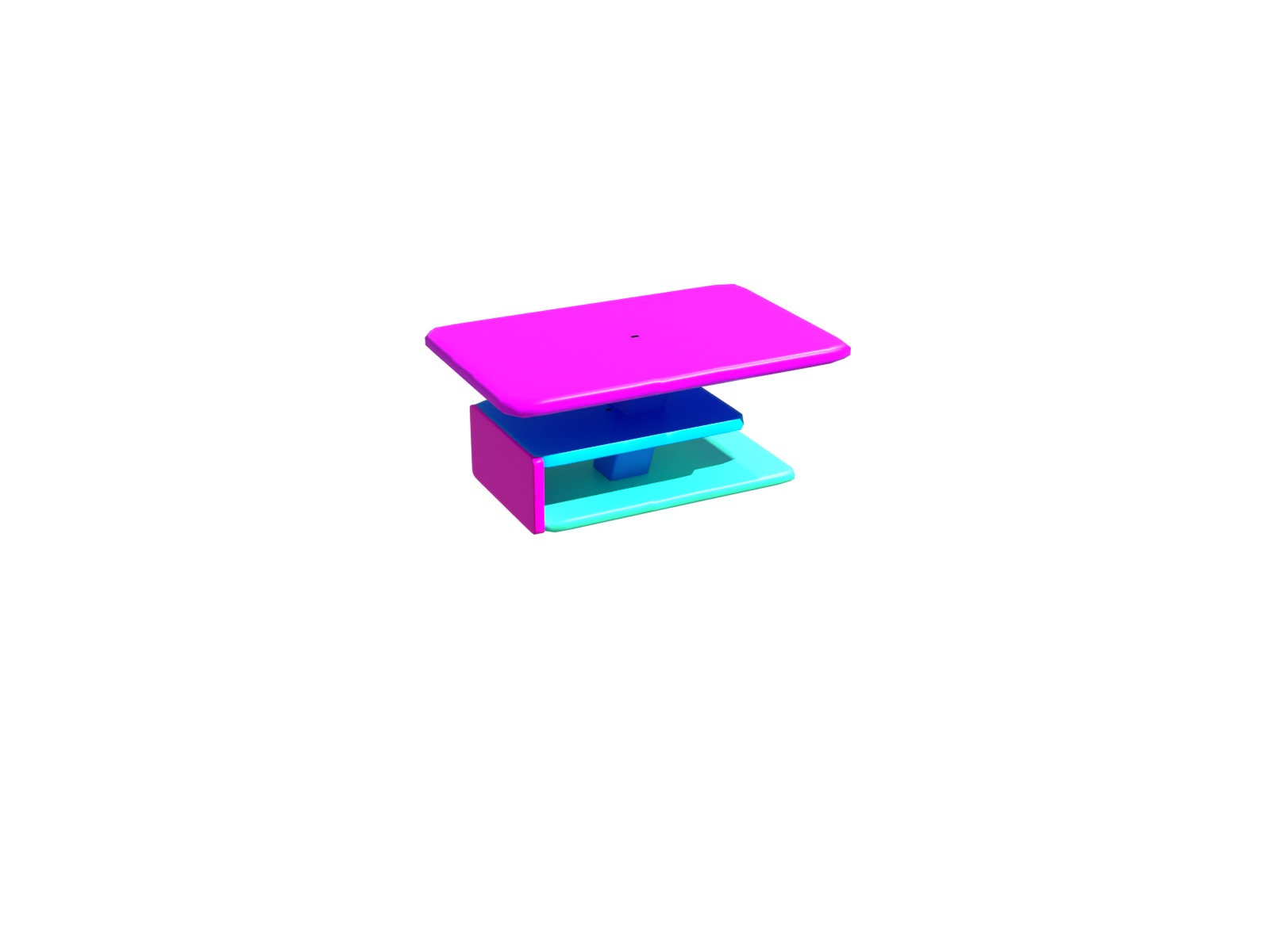}
\includegraphics[width=0.125\linewidth, height=1.7cm,trim={520px 450px 520px 150px},clip]{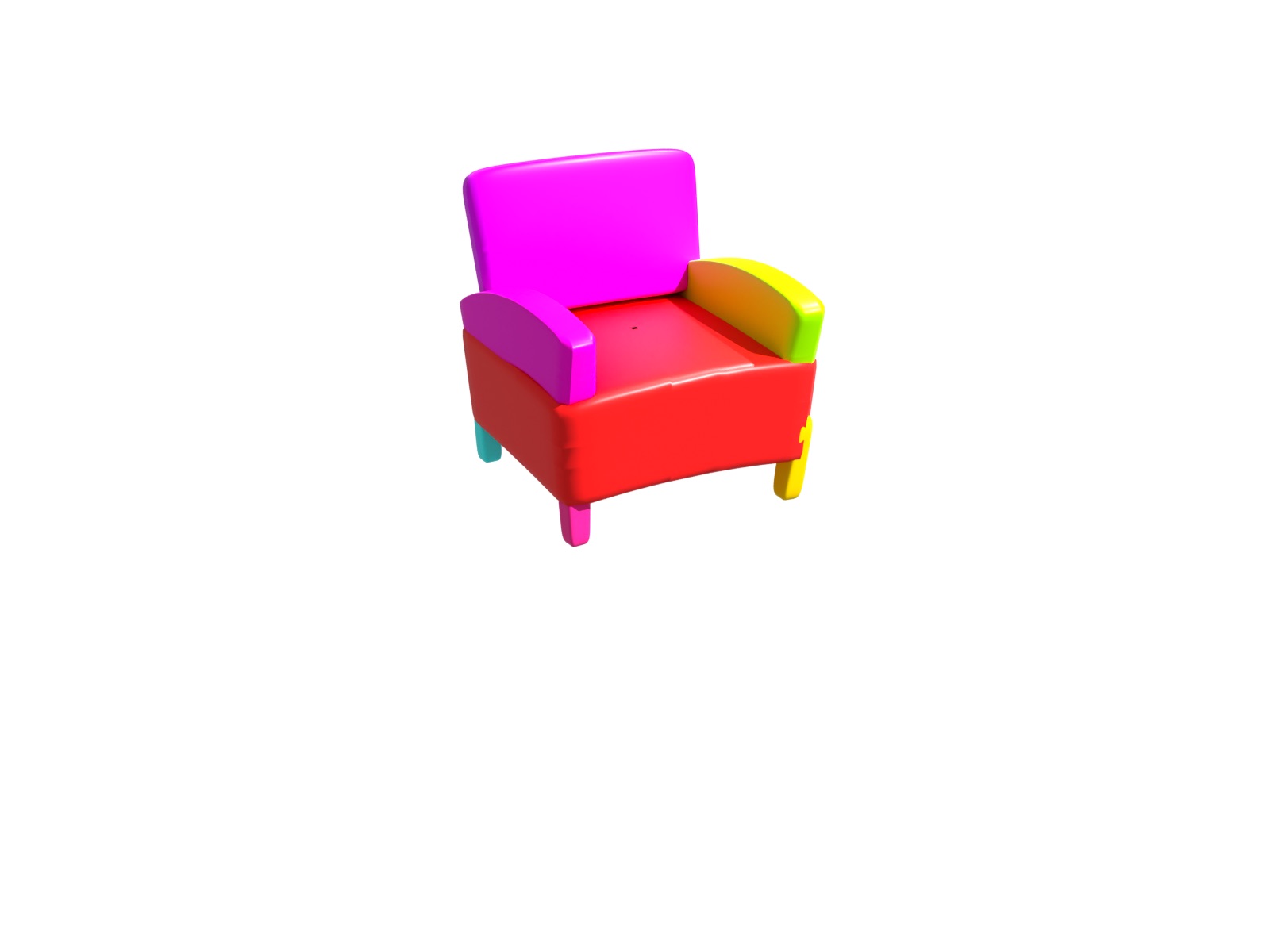}
\includegraphics[width=0.125\linewidth, trim={250px 200px 200px 1px},clip]{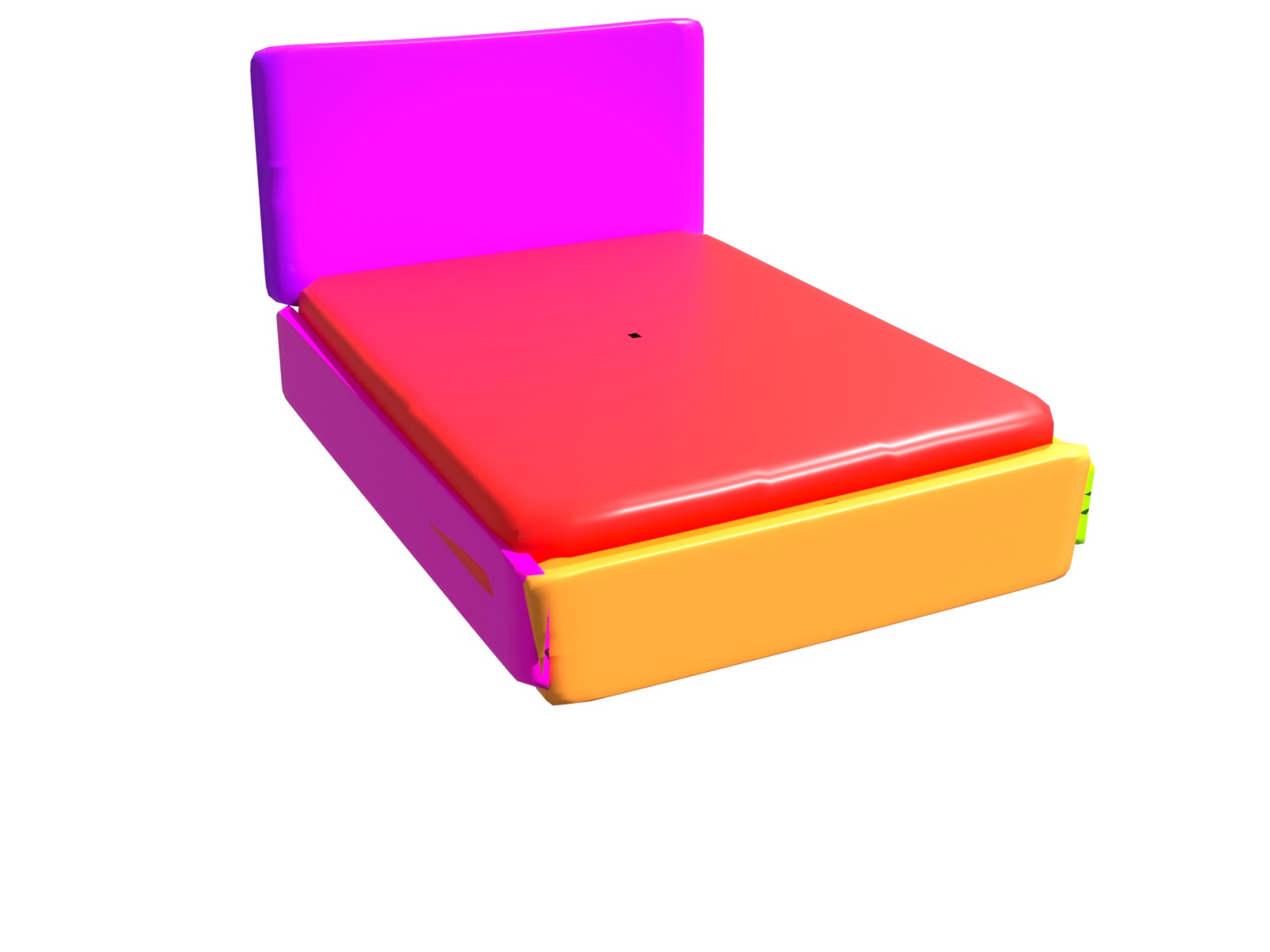}
\includegraphics[width=0.125\linewidth, trim={550px 450px 500px 300px},clip]{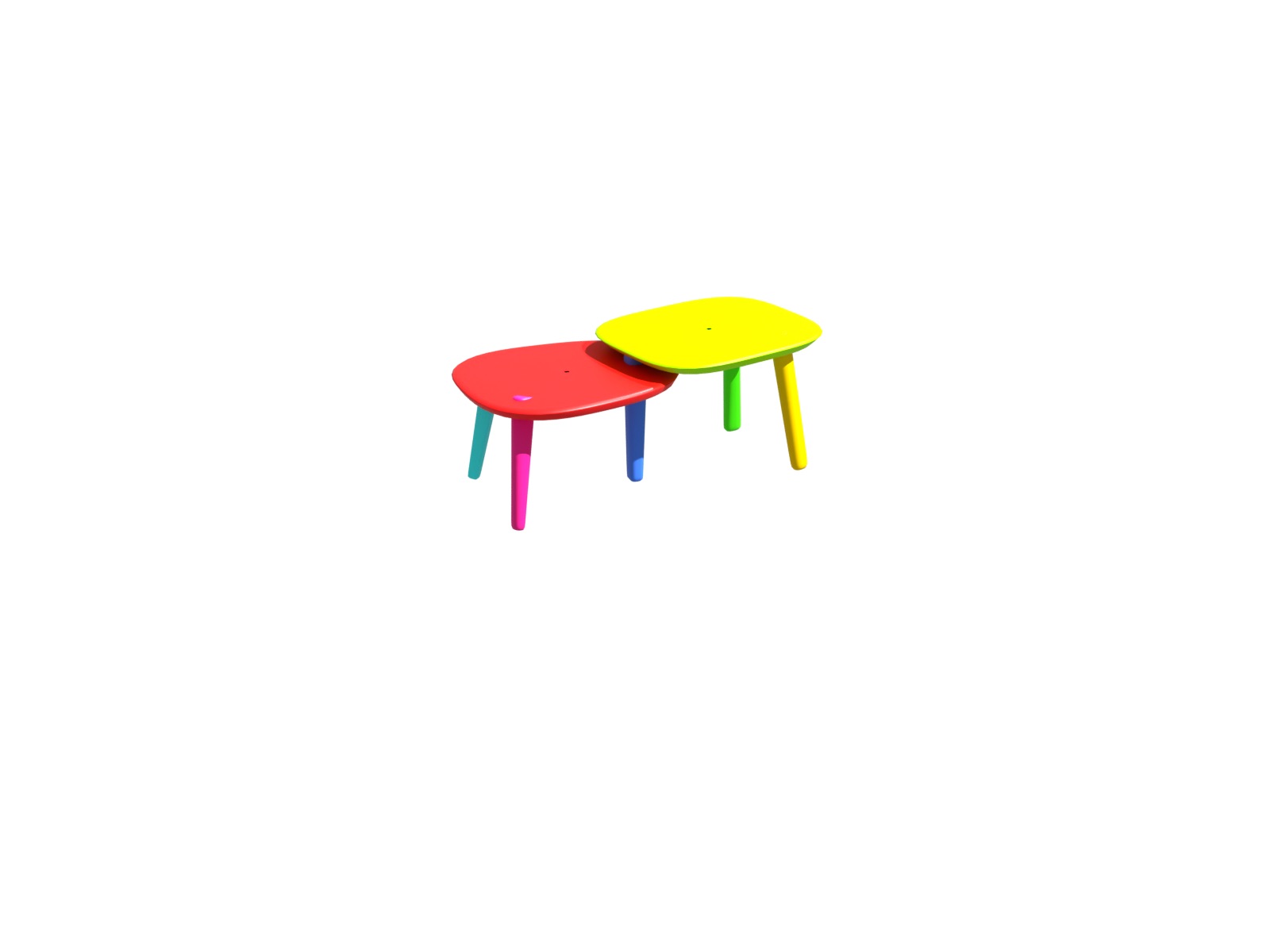}
\includegraphics[width=0.125\linewidth,trim={520px 480px 520px 200px},clip]{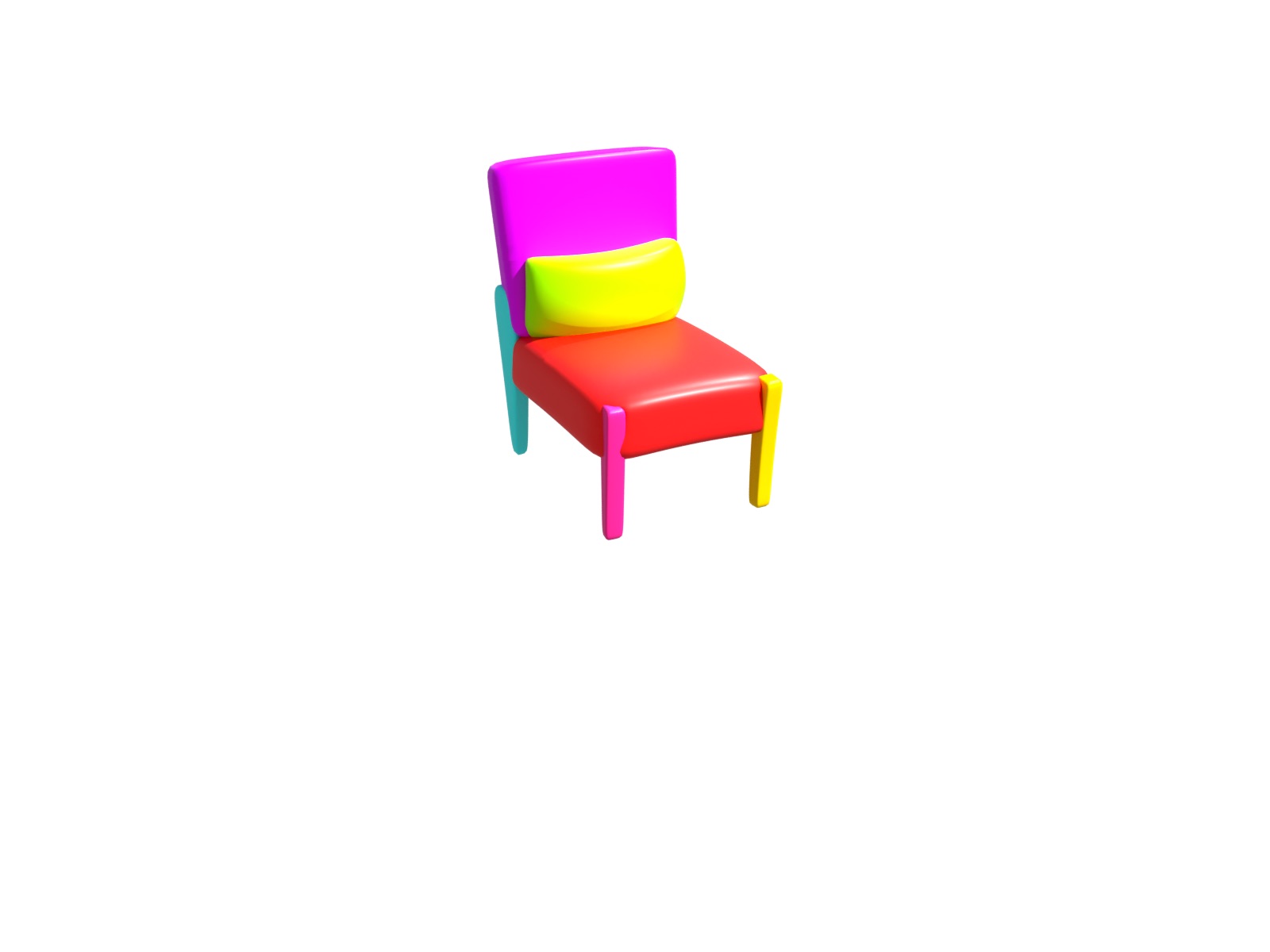}
\includegraphics[width=0.125\linewidth, trim={520px 480px 520px 200px},clip]{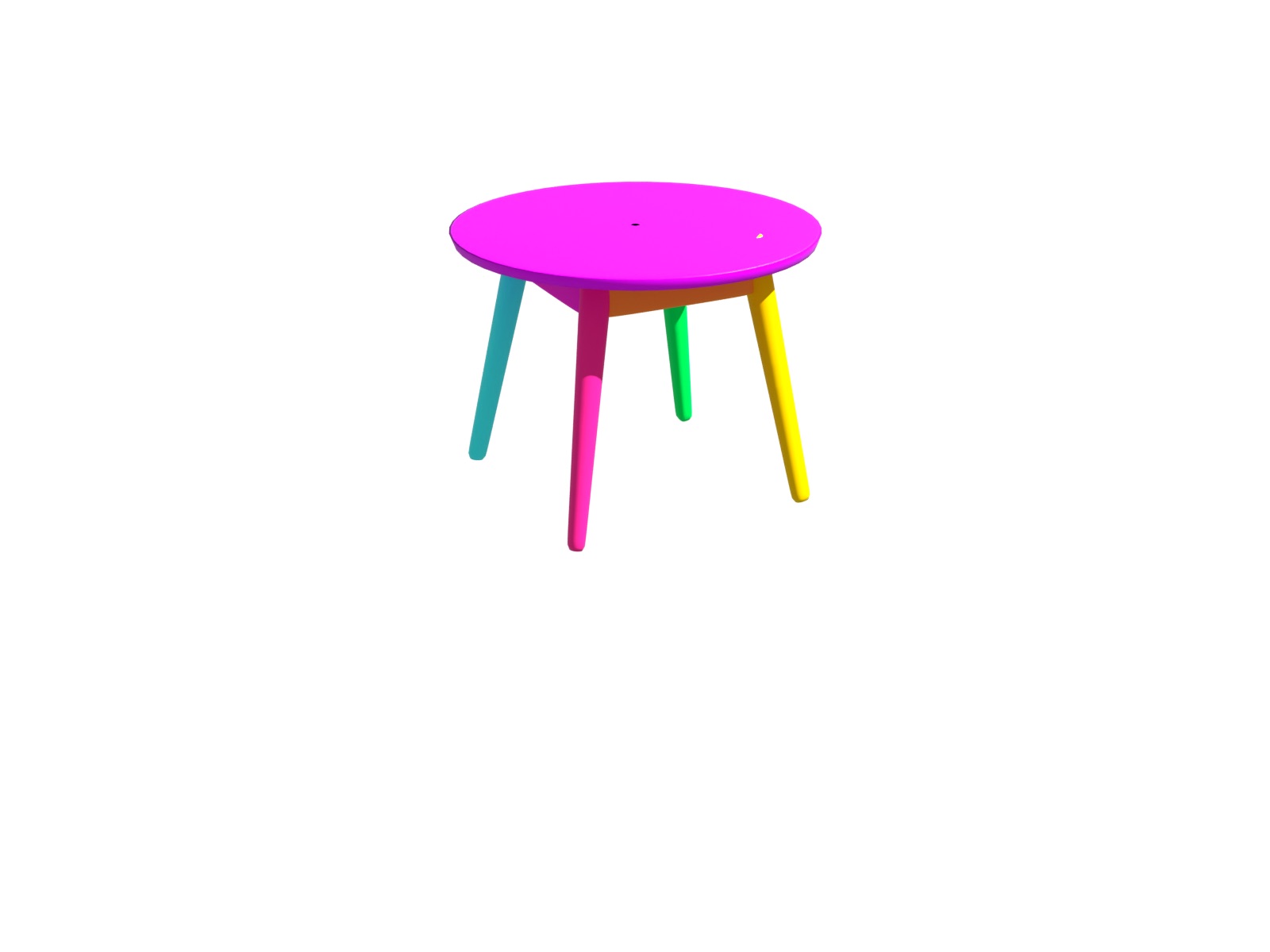}\\
\end{small}
\caption{\textbf{Qualitative results on ABO.} In each block, the first row shows the input point clouds, the second row shows the direct predictions of \method{}, and the third row shows the corresponding optimized results. Different colors denote different primitives.}
\label{fig:quali-superflex-abo}
\end{figure}

\begin{figure}[t]
\centering
\begin{small}
\rotatebox{90}{\footnotesize \shortstack{Input \\ Pointcloud}}
\includegraphics[height=1.7cm,trim={420px 200px 480px 200px},clip]{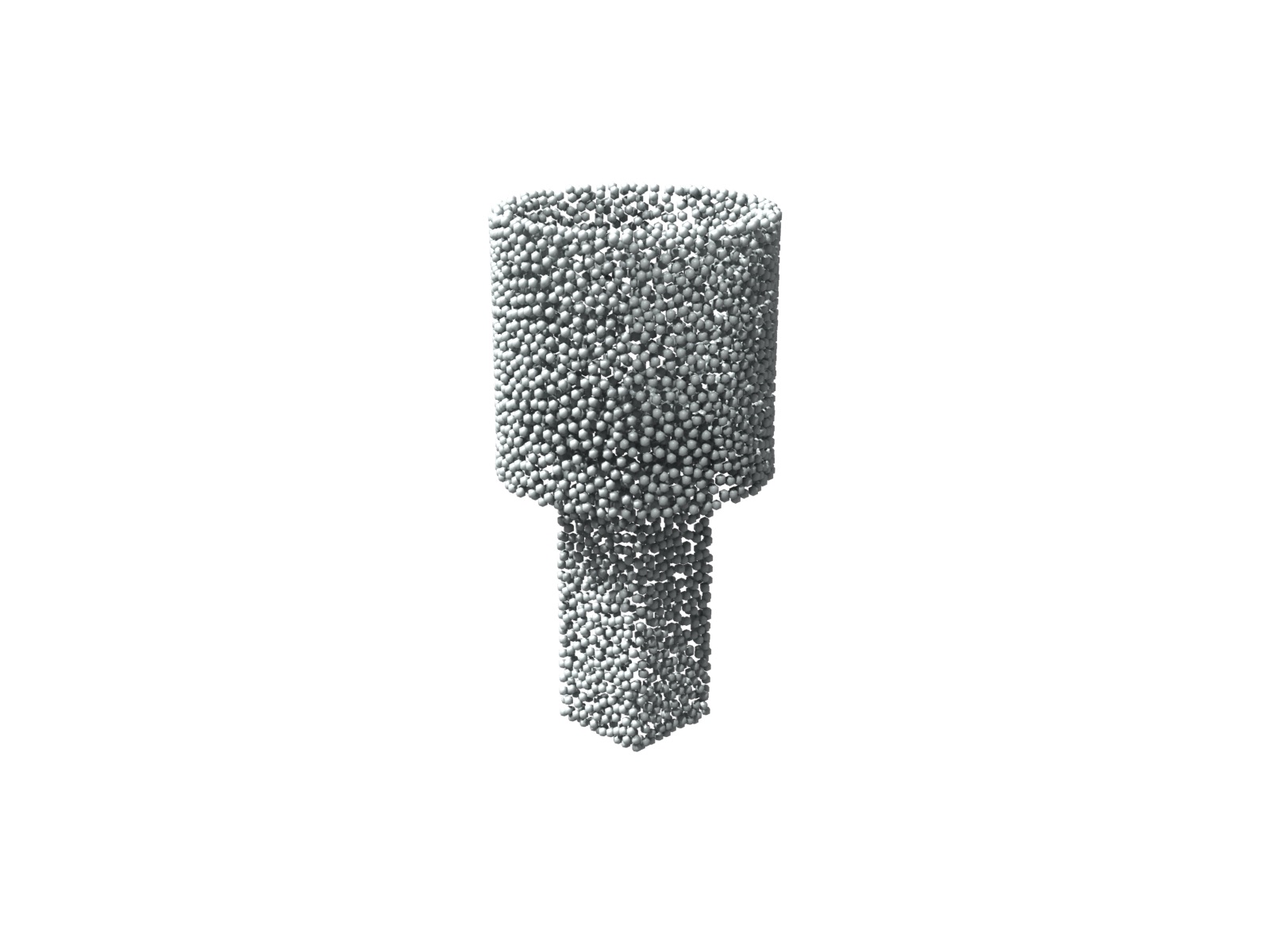}
\includegraphics[height=1.7cm,trim={440px 160px 440px 120px},clip]{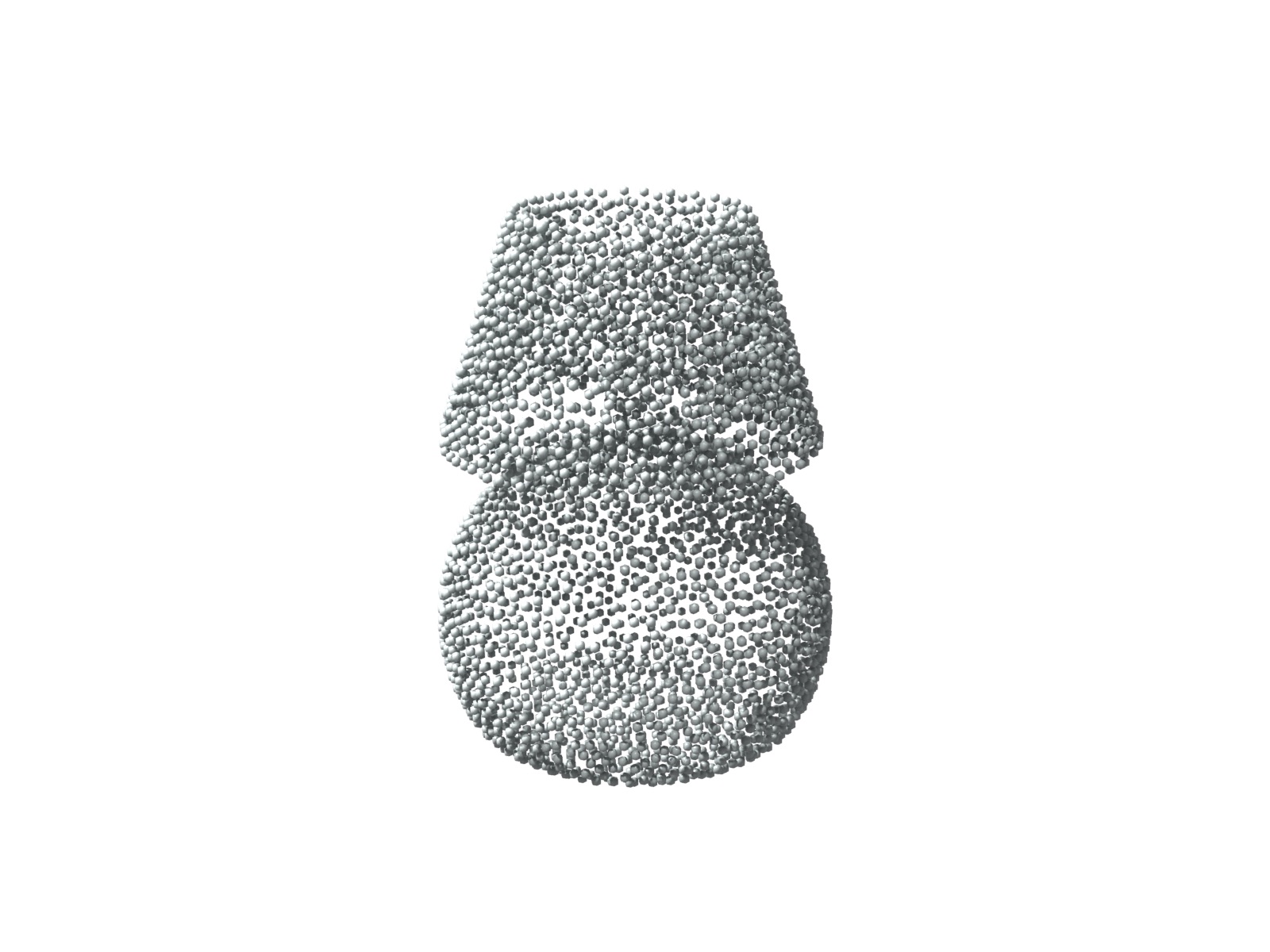}
\includegraphics[height=1.7cm,trim={420px 140px 420px 200px},clip]{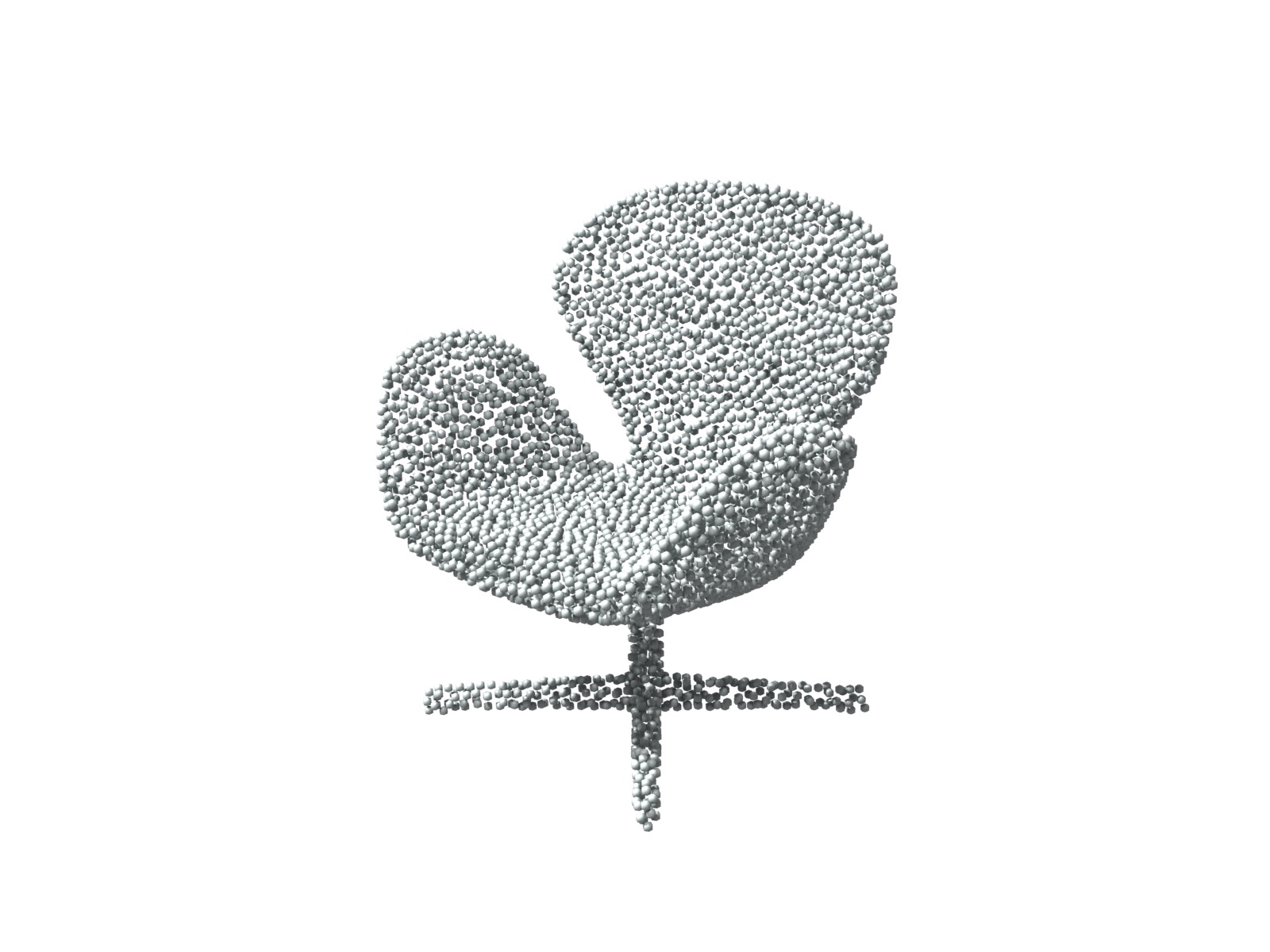}
\includegraphics[height=1.7cm,trim={440px 160px 440px 120px},clip]{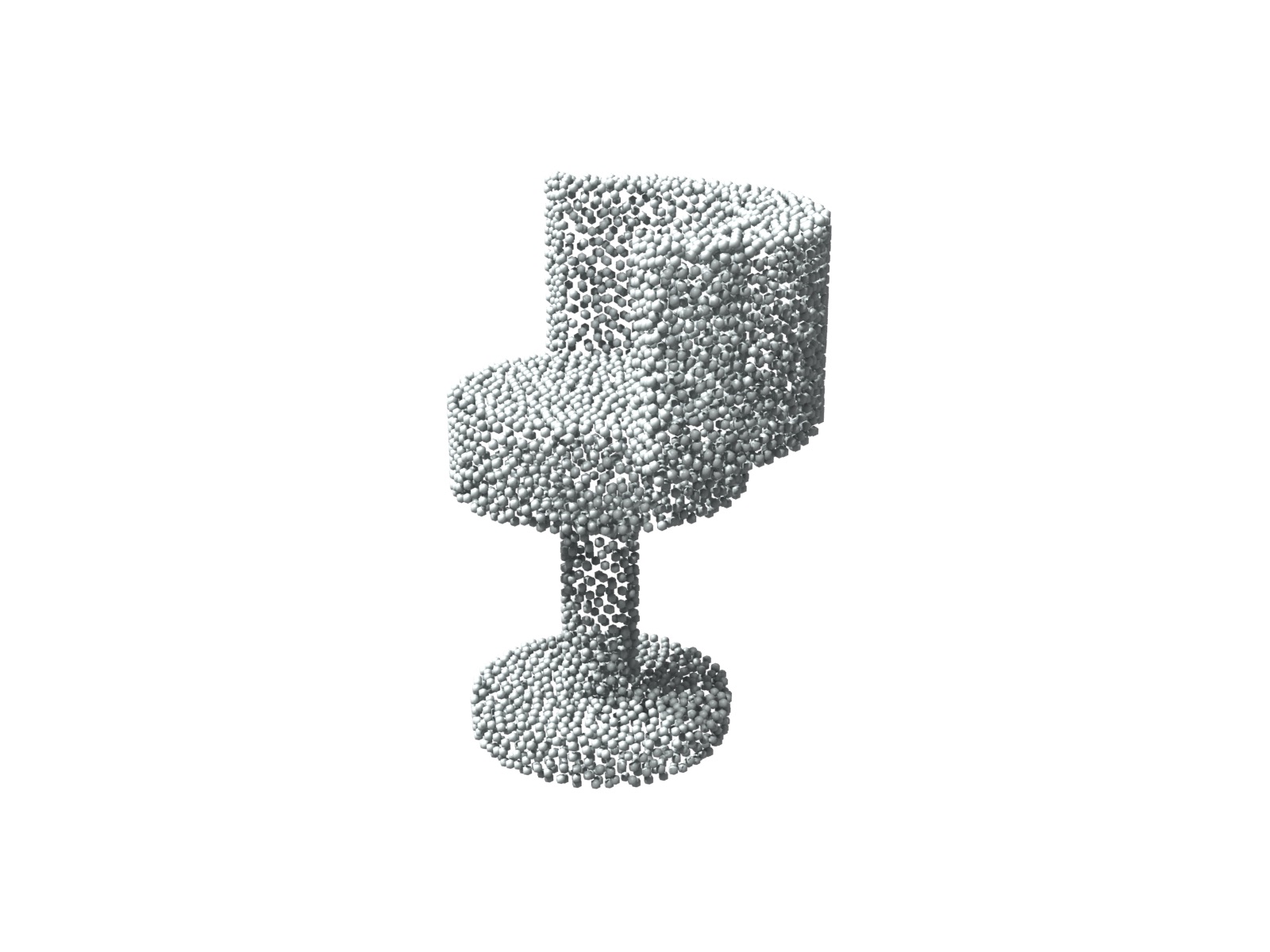}
\includegraphics[height=1.7cm,trim={420px 180px 420px 220px},clip]{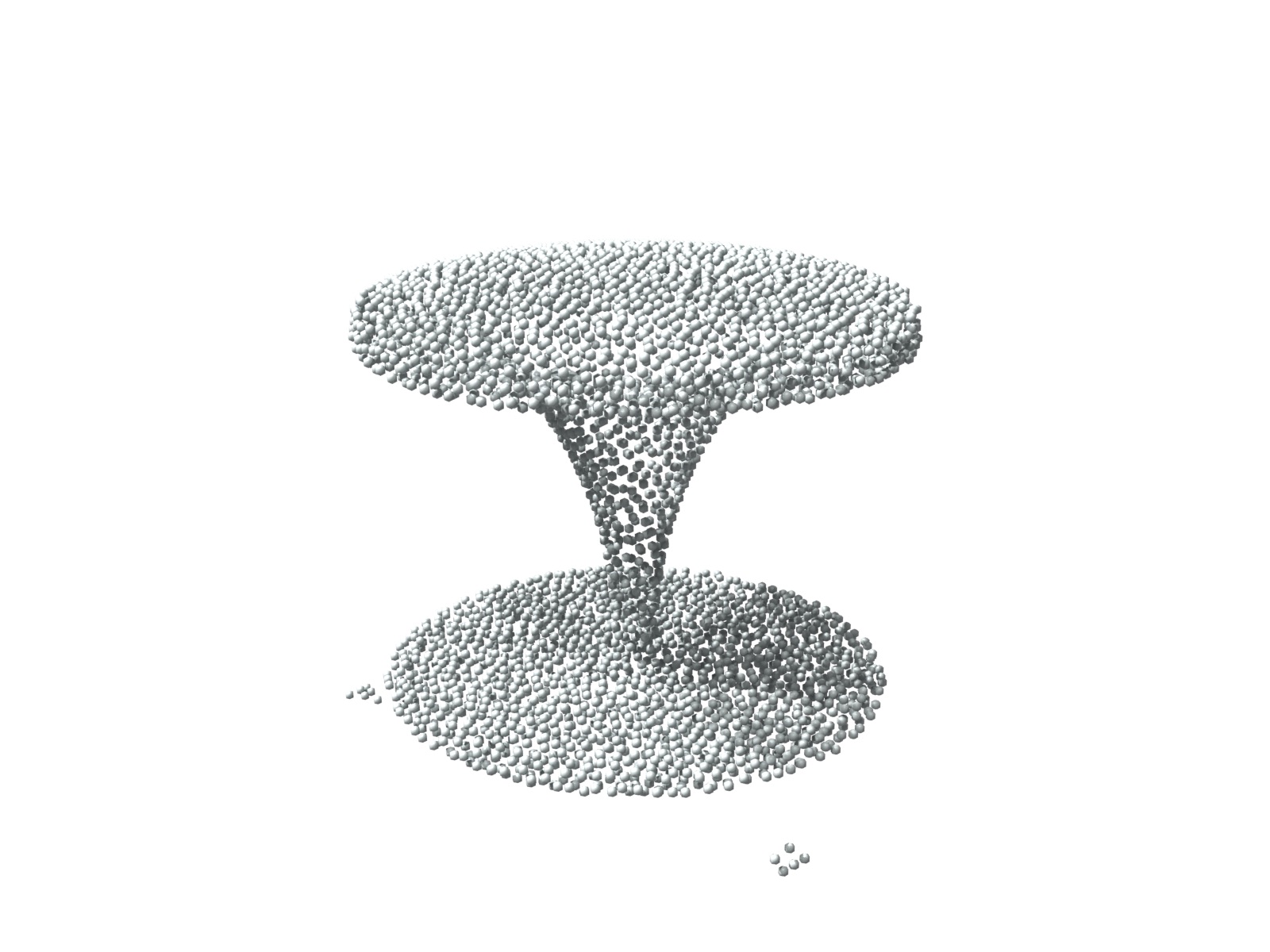}
\includegraphics[height=1.7cm,trim={400px 280px 400px 220px},clip]{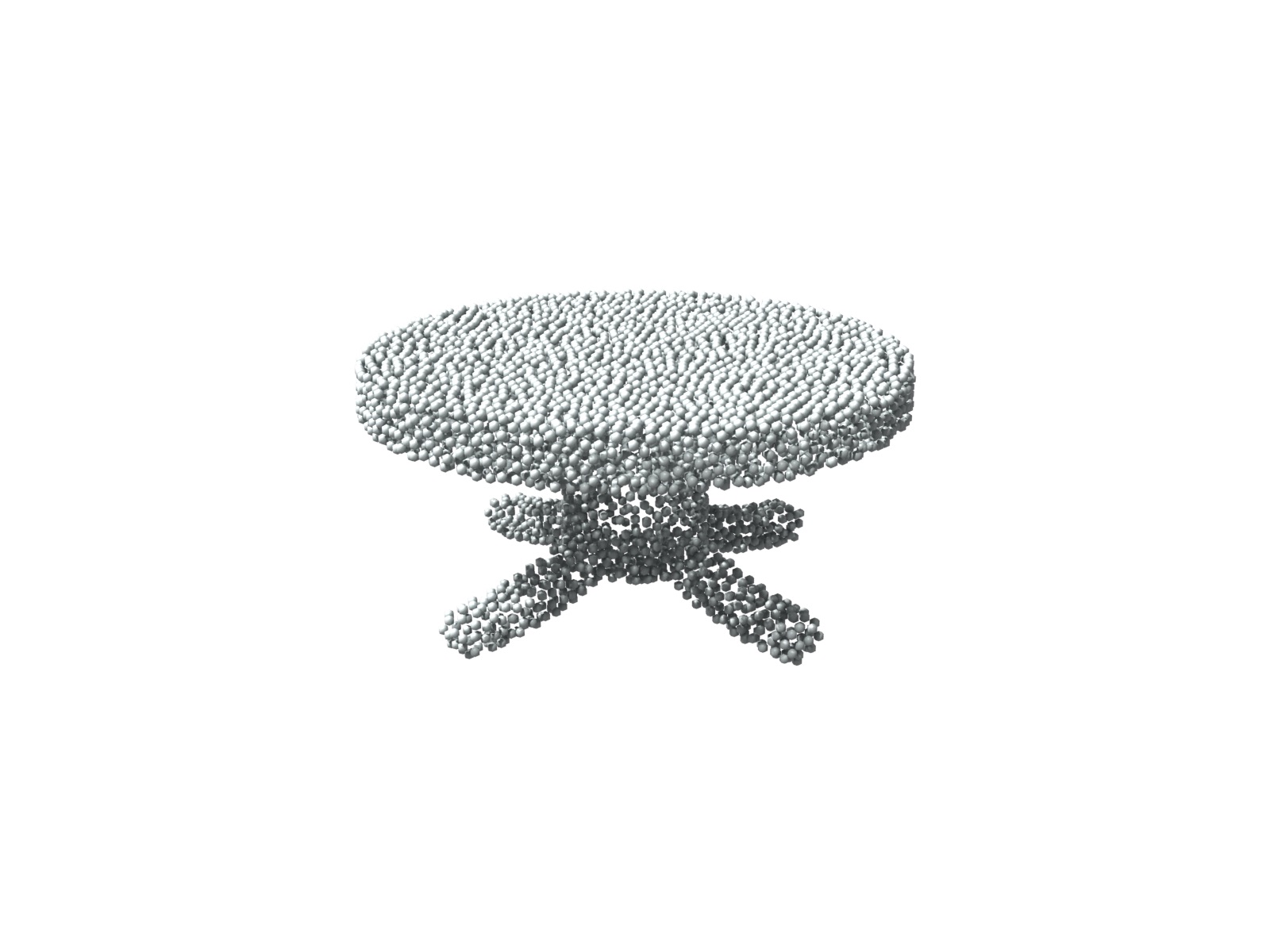}
\includegraphics[height=1.7cm,trim={440px 160px 440px 120px},clip]{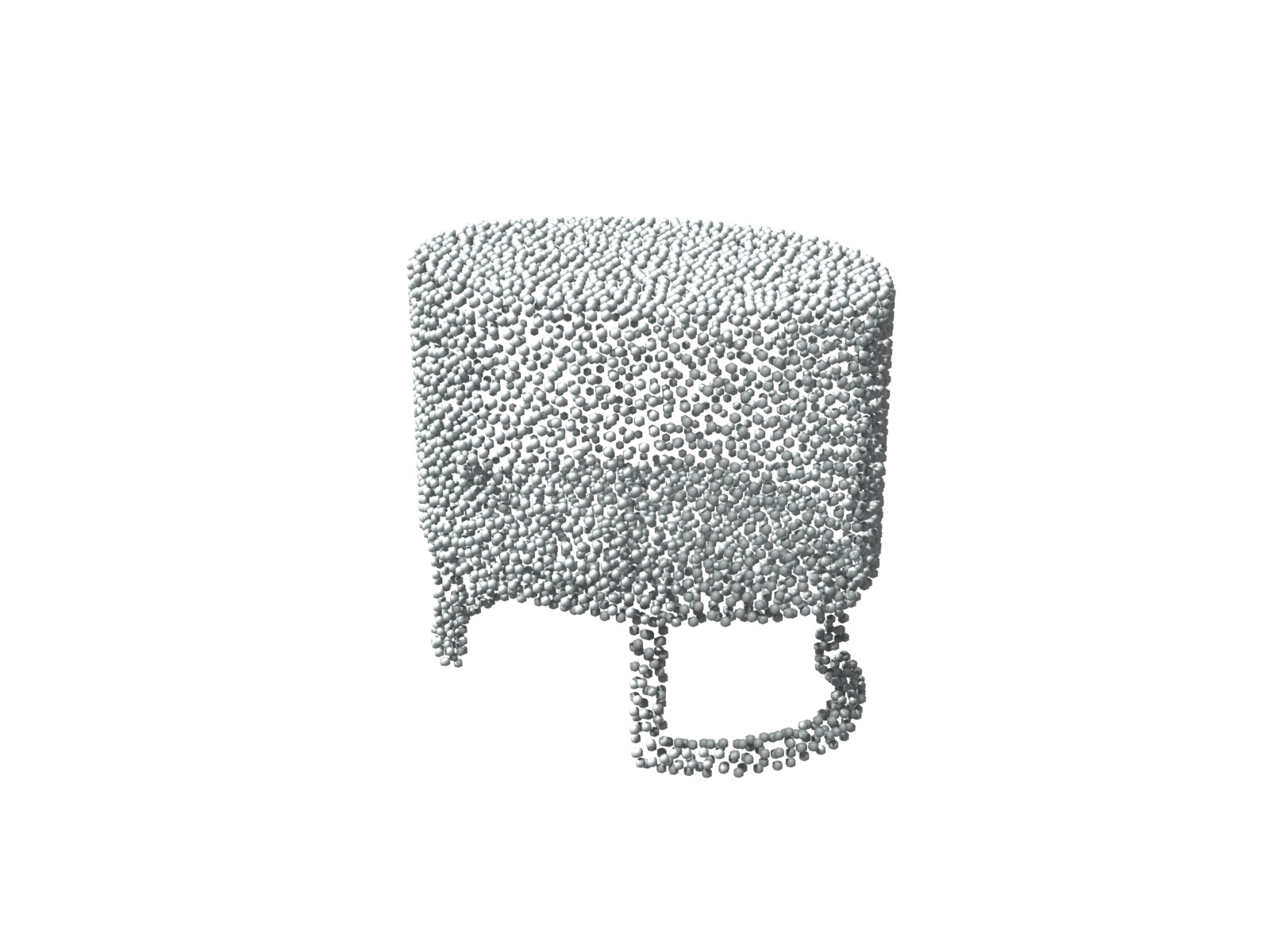}\\

\rotatebox{90}{\footnotesize \shortstack{ \hspace{2px}\method{}\\\scriptsize{(No T\&B)}}}\hspace{-4px}
\includegraphics[height=1.7cm,trim={420px 200px 480px 200px},clip]{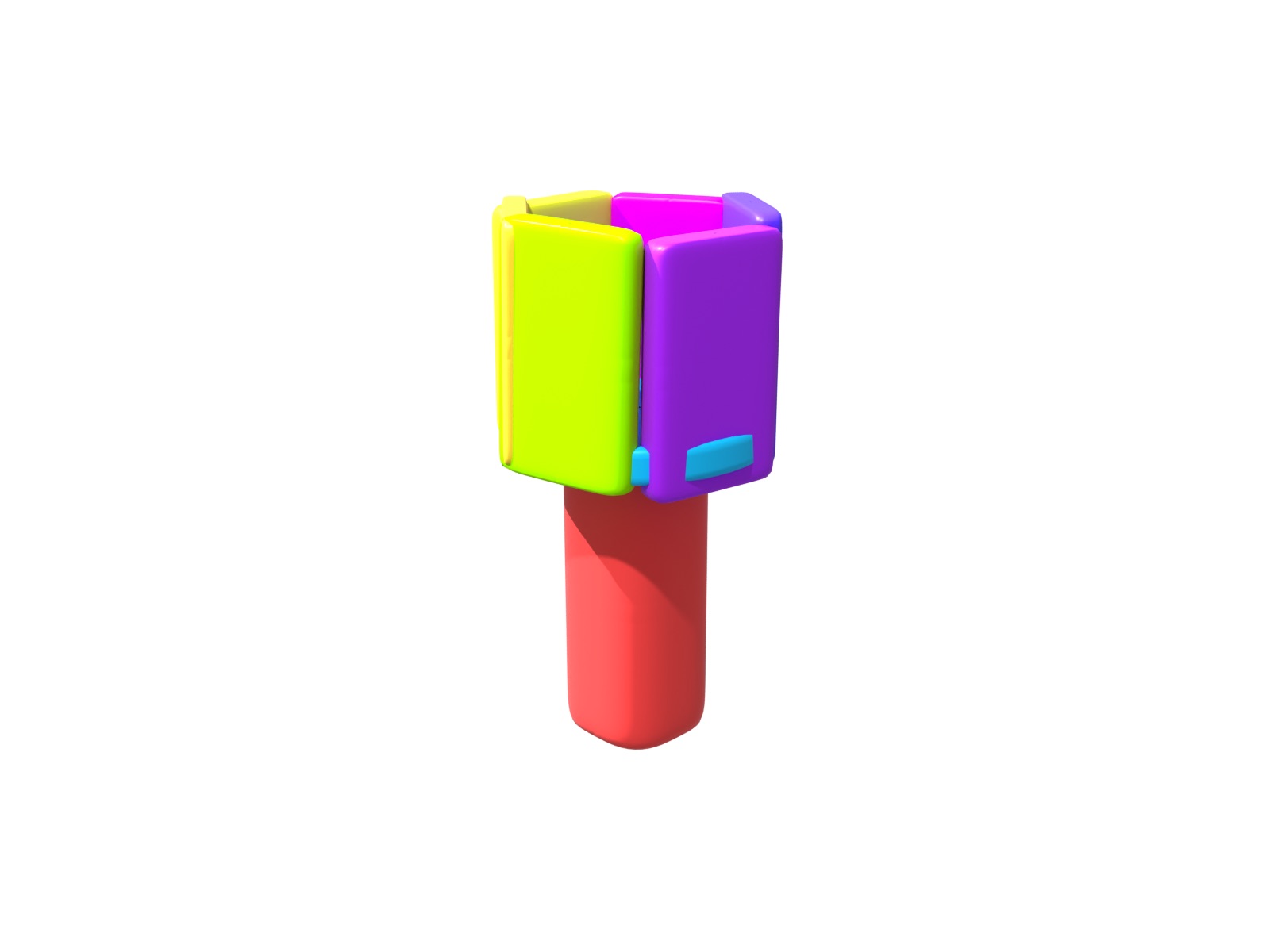}
\includegraphics[height=1.7cm,trim={440px 160px 440px 120px},clip]{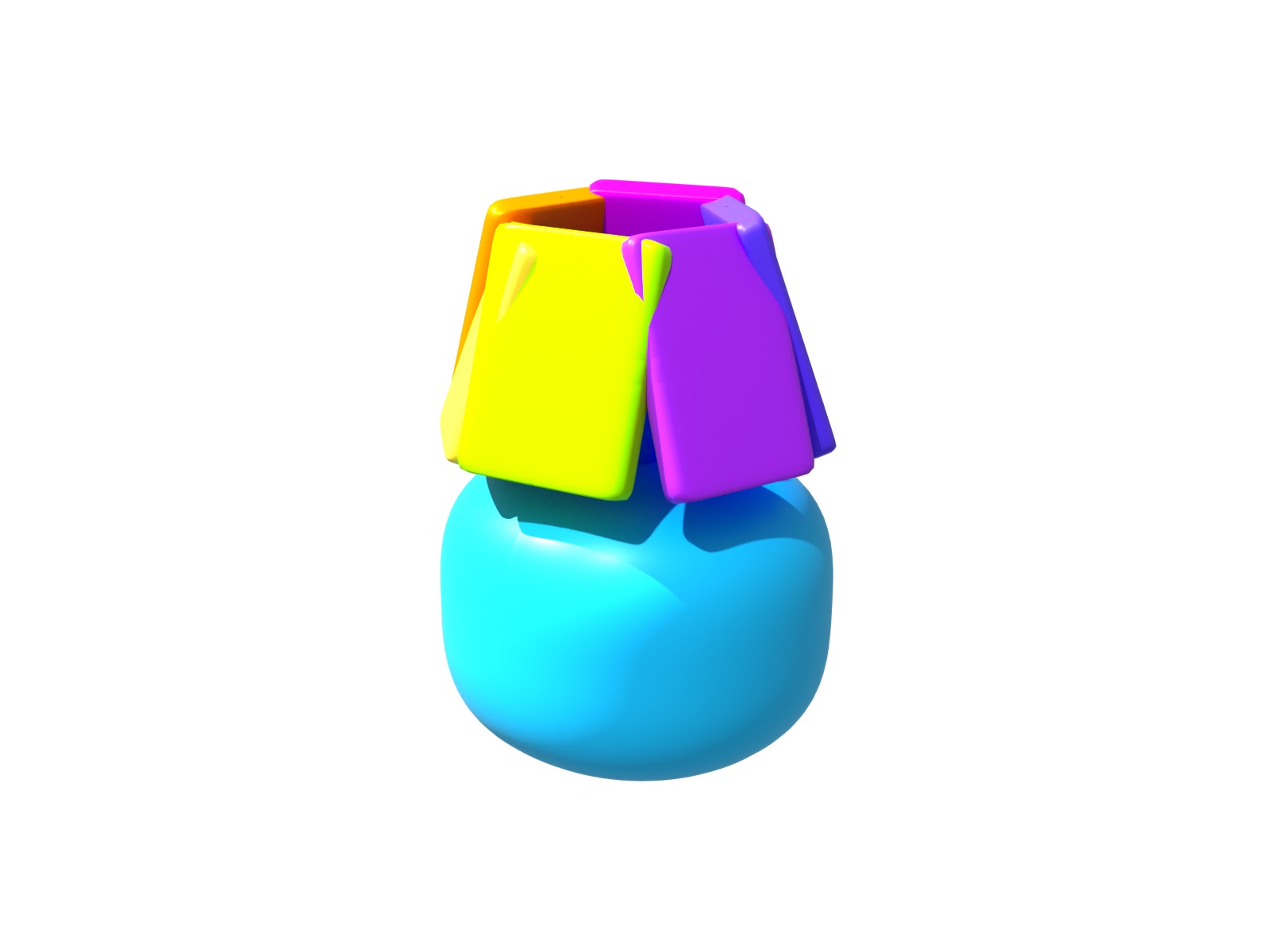}
\includegraphics[height=1.7cm,trim={420px 140px 420px 200px},clip]{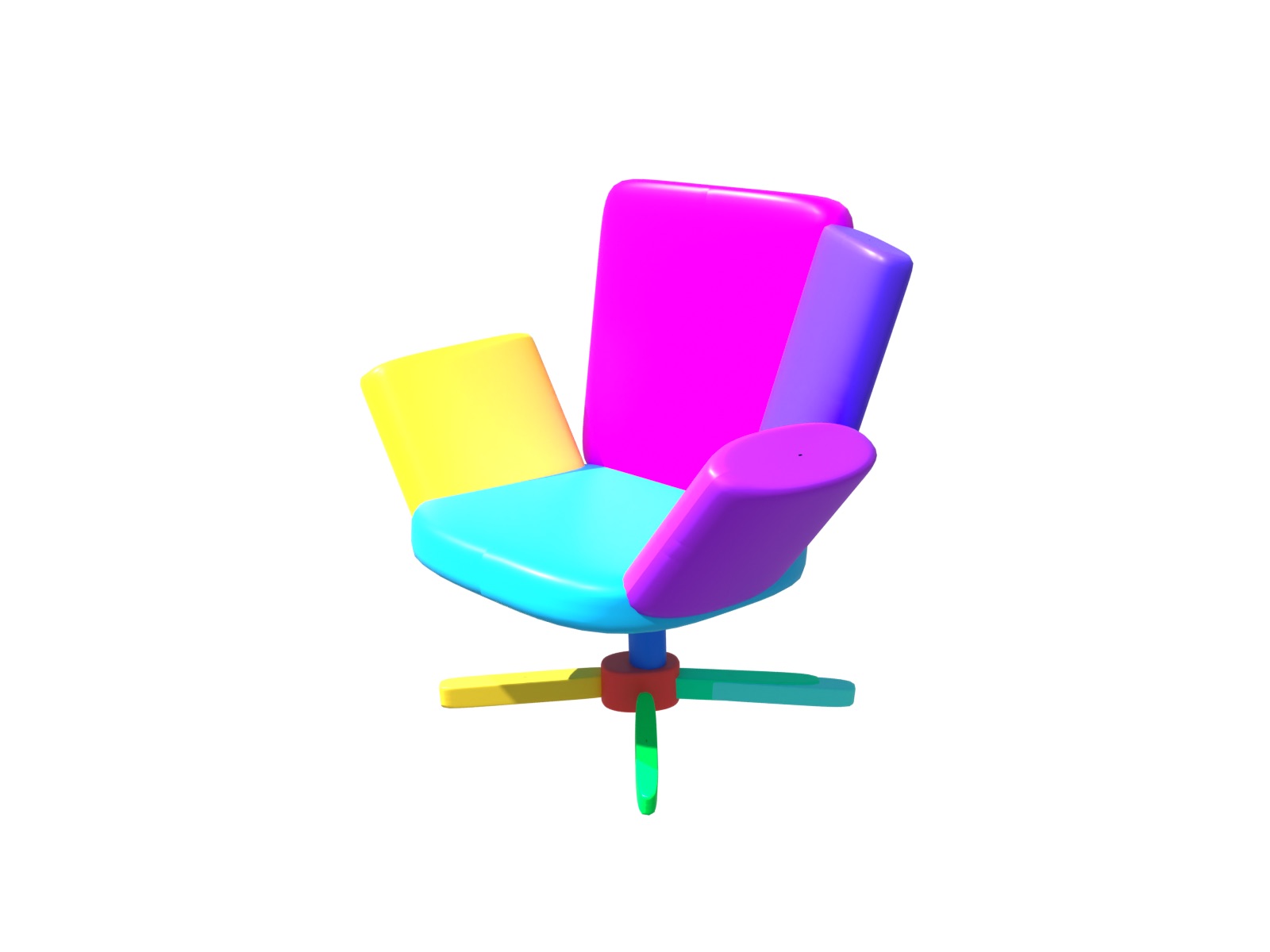}
\includegraphics[height=1.7cm,trim={440px 160px 440px 120px},clip]{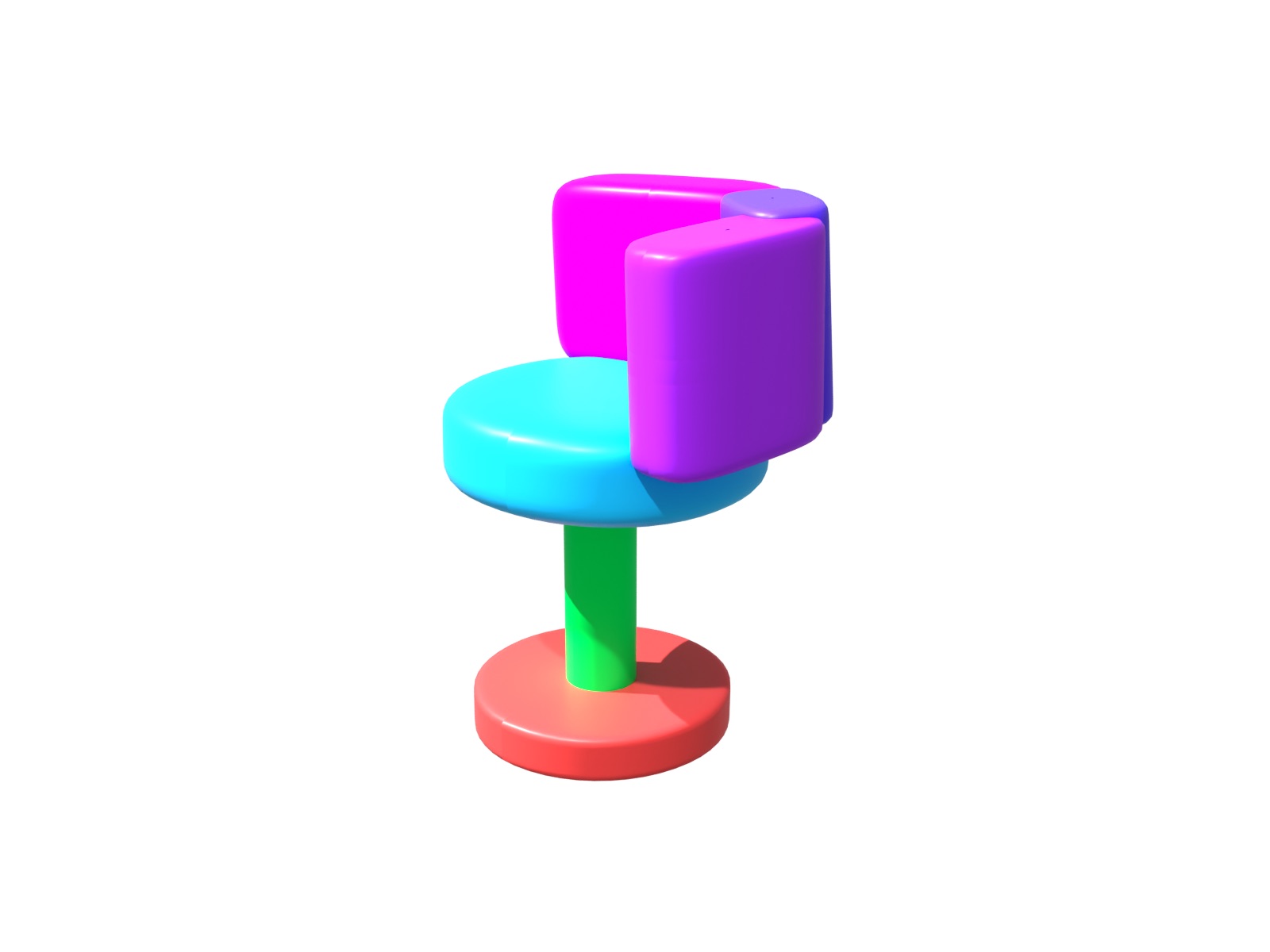}
\includegraphics[height=1.7cm,trim={420px 180px 420px 200px},clip]{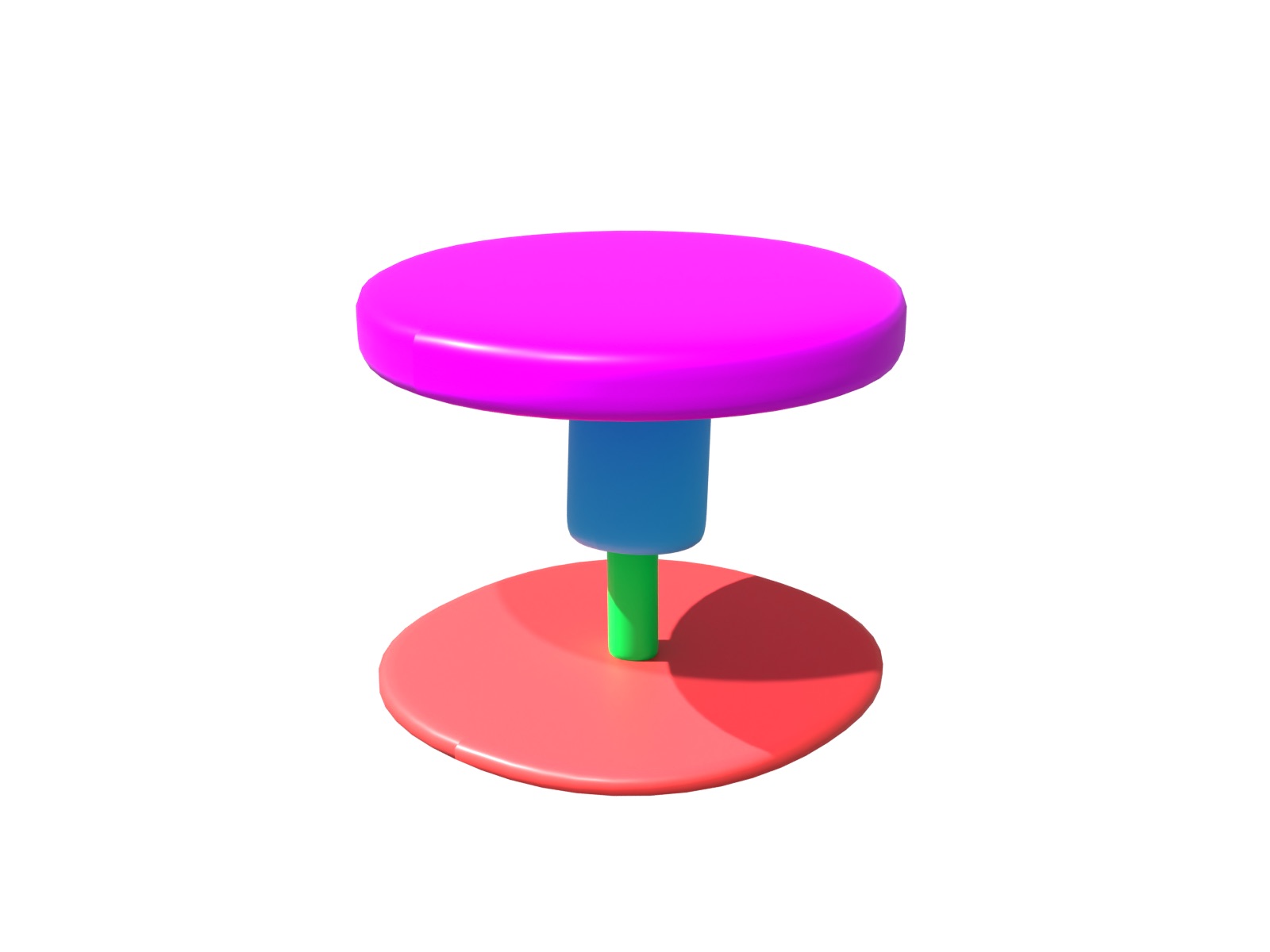}
\includegraphics[height=1.7cm,trim={400px 280px 400px 220px},clip]{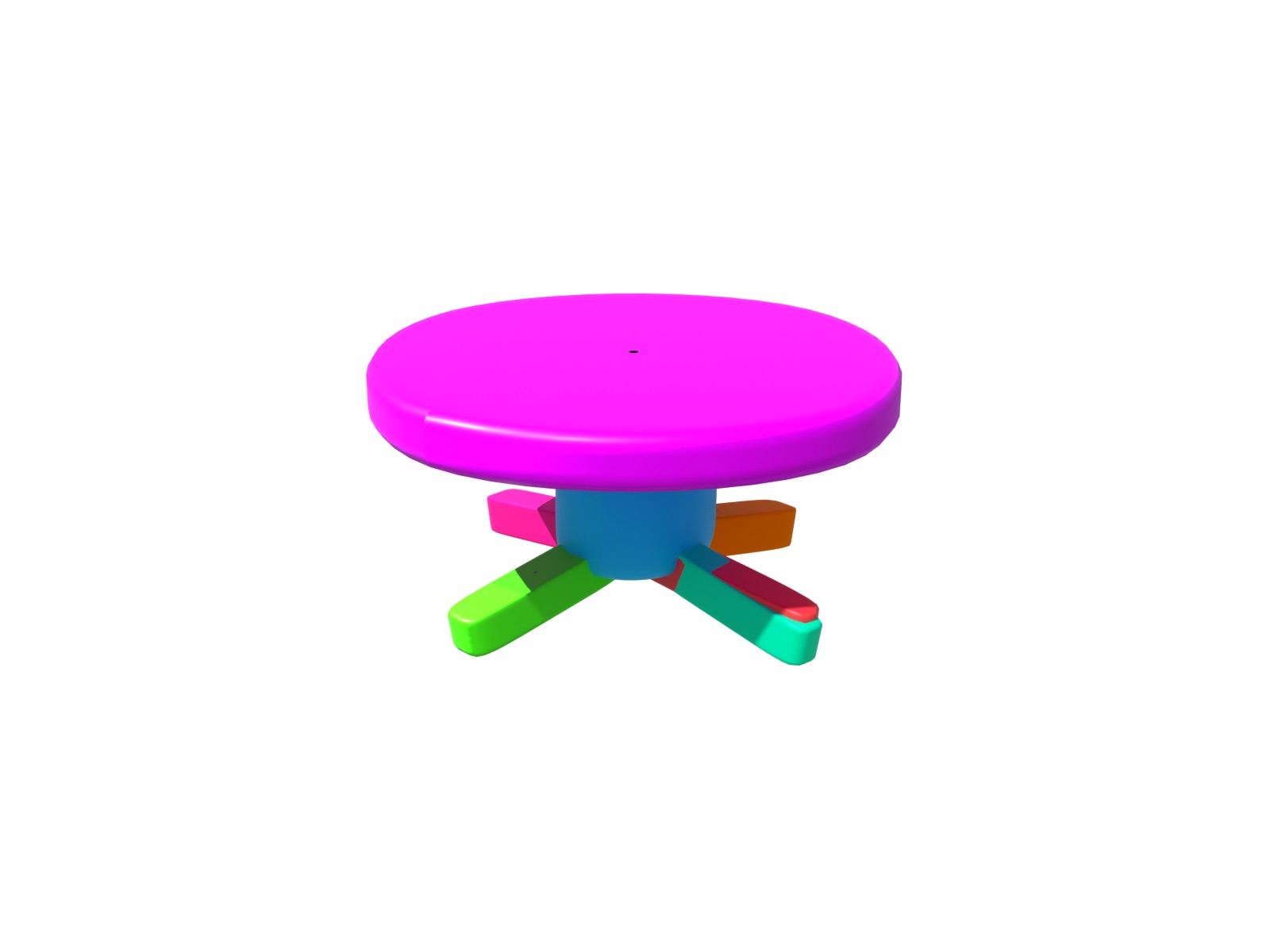}
\includegraphics[height=1.7cm,trim={440px 160px 440px 120px},clip]{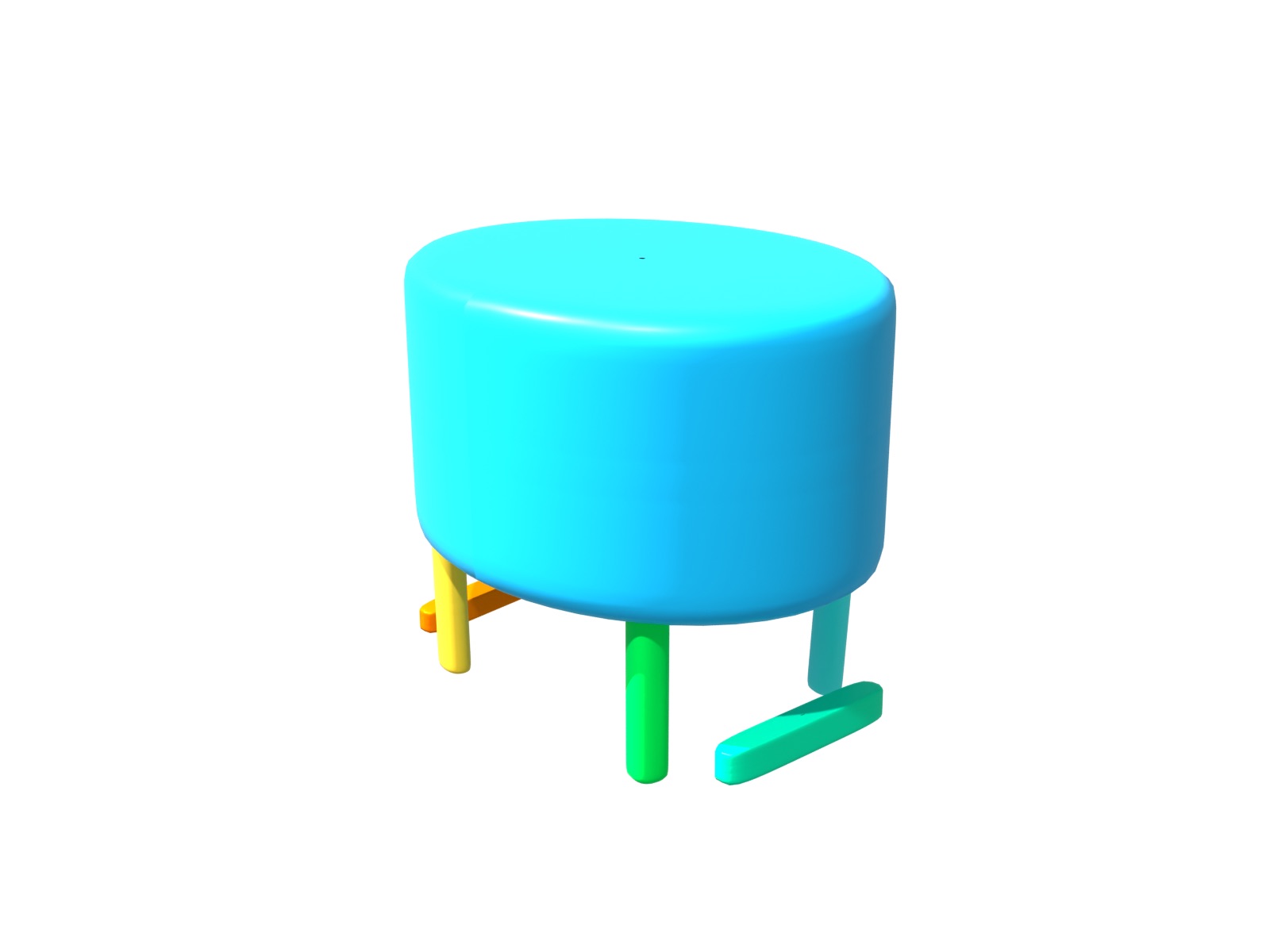}\\

\rotatebox{90}{\footnotesize \shortstack{\\ \hspace{5px}\method{}}}\hspace{4px}
\includegraphics[height=1.7cm,trim={420px 200px 480px 200px},clip]{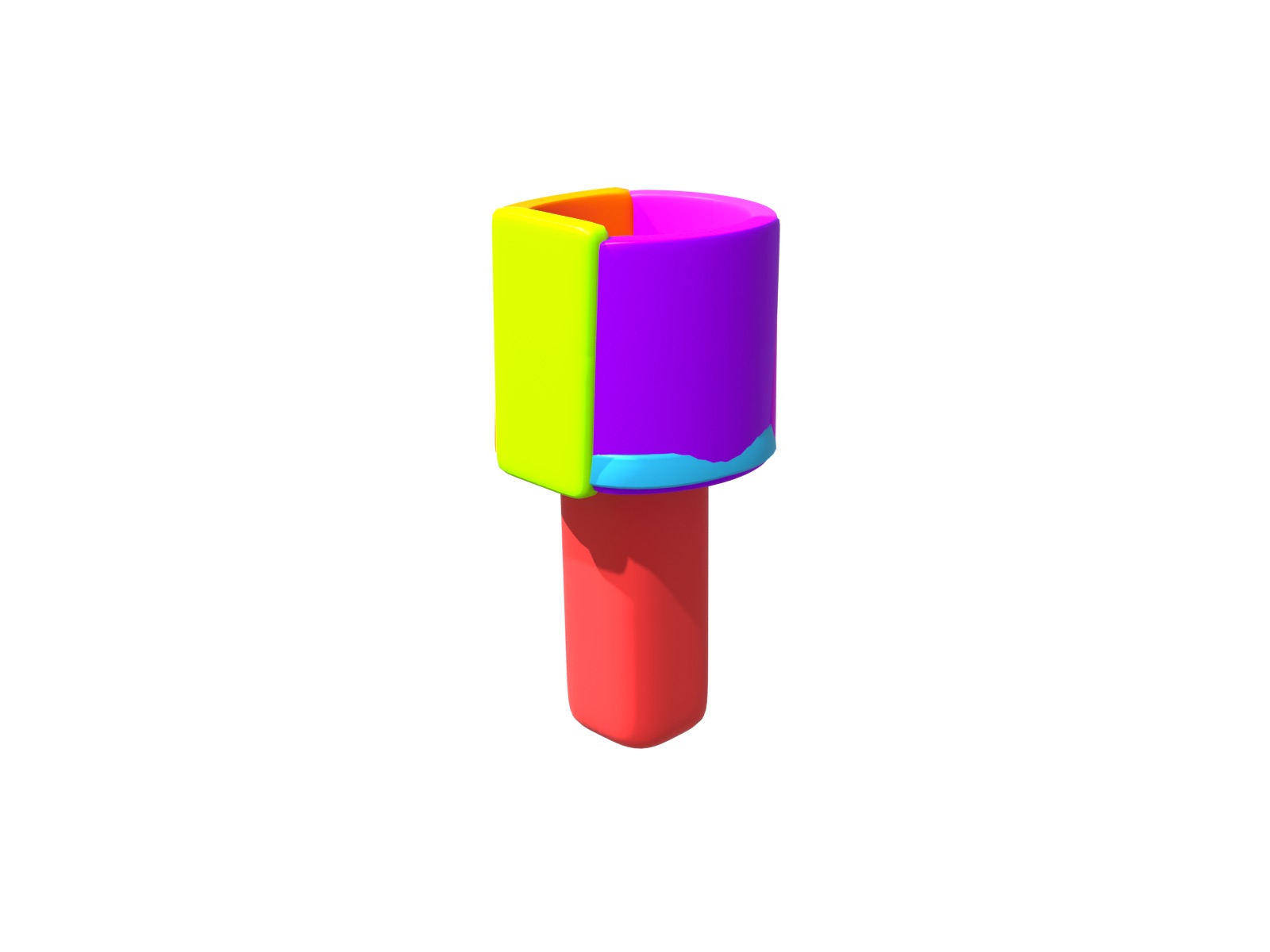}
\includegraphics[height=1.7cm,trim={440px 160px 440px 120px},clip]{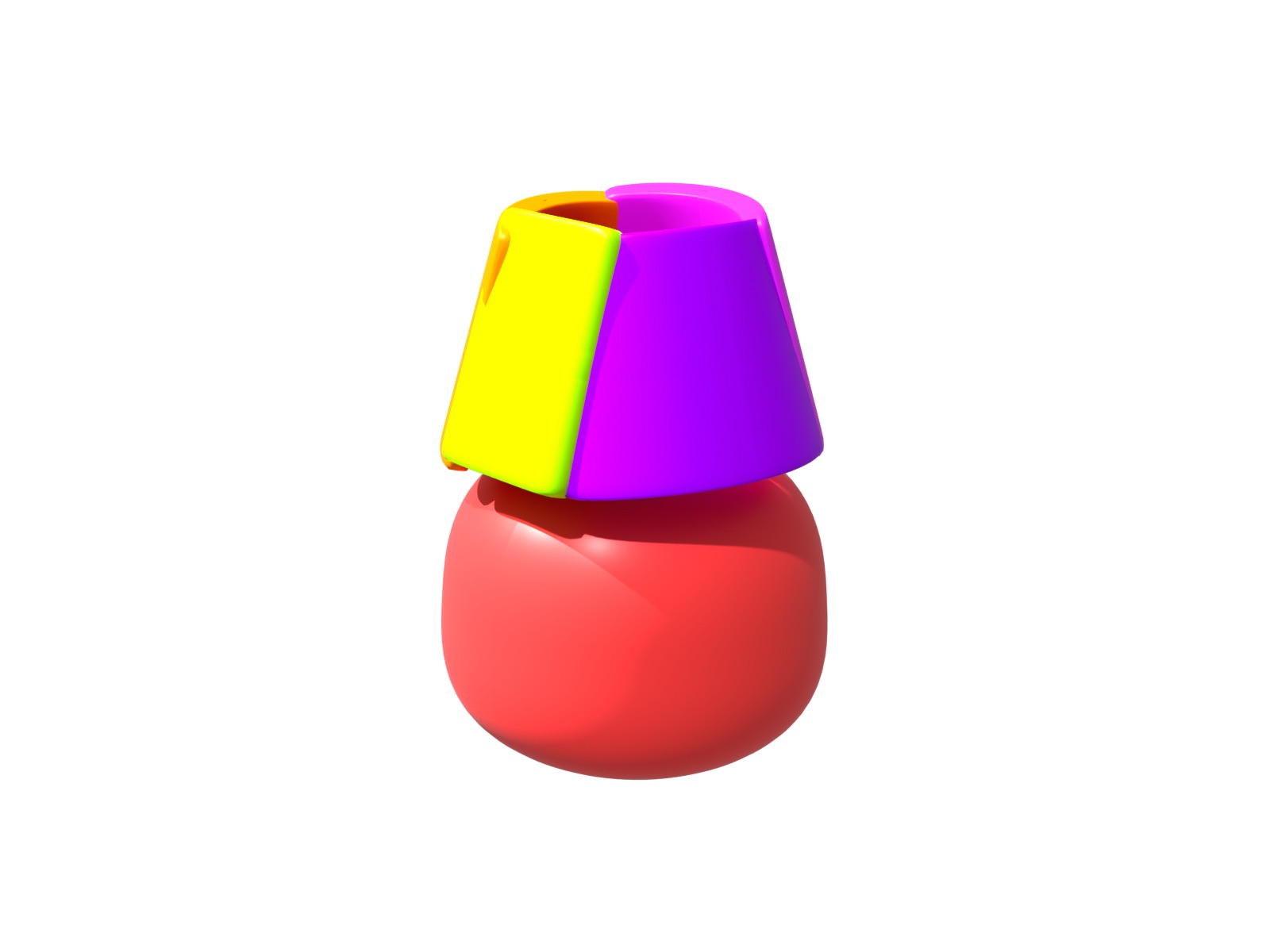}
\includegraphics[height=1.7cm,trim={420px 140px 420px 200px},clip]{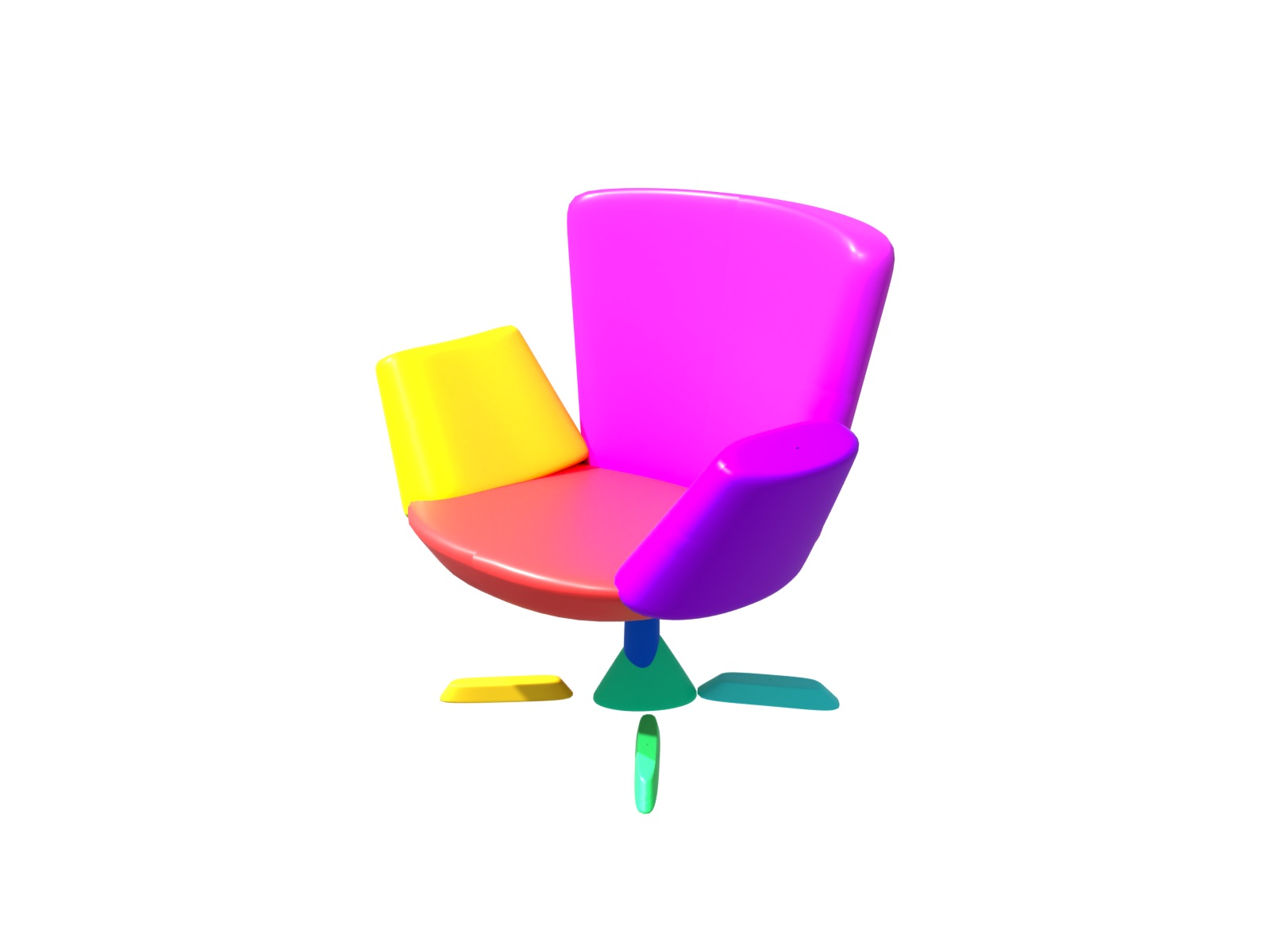}
\includegraphics[height=1.7cm,trim={440px 160px 440px 120px},clip]{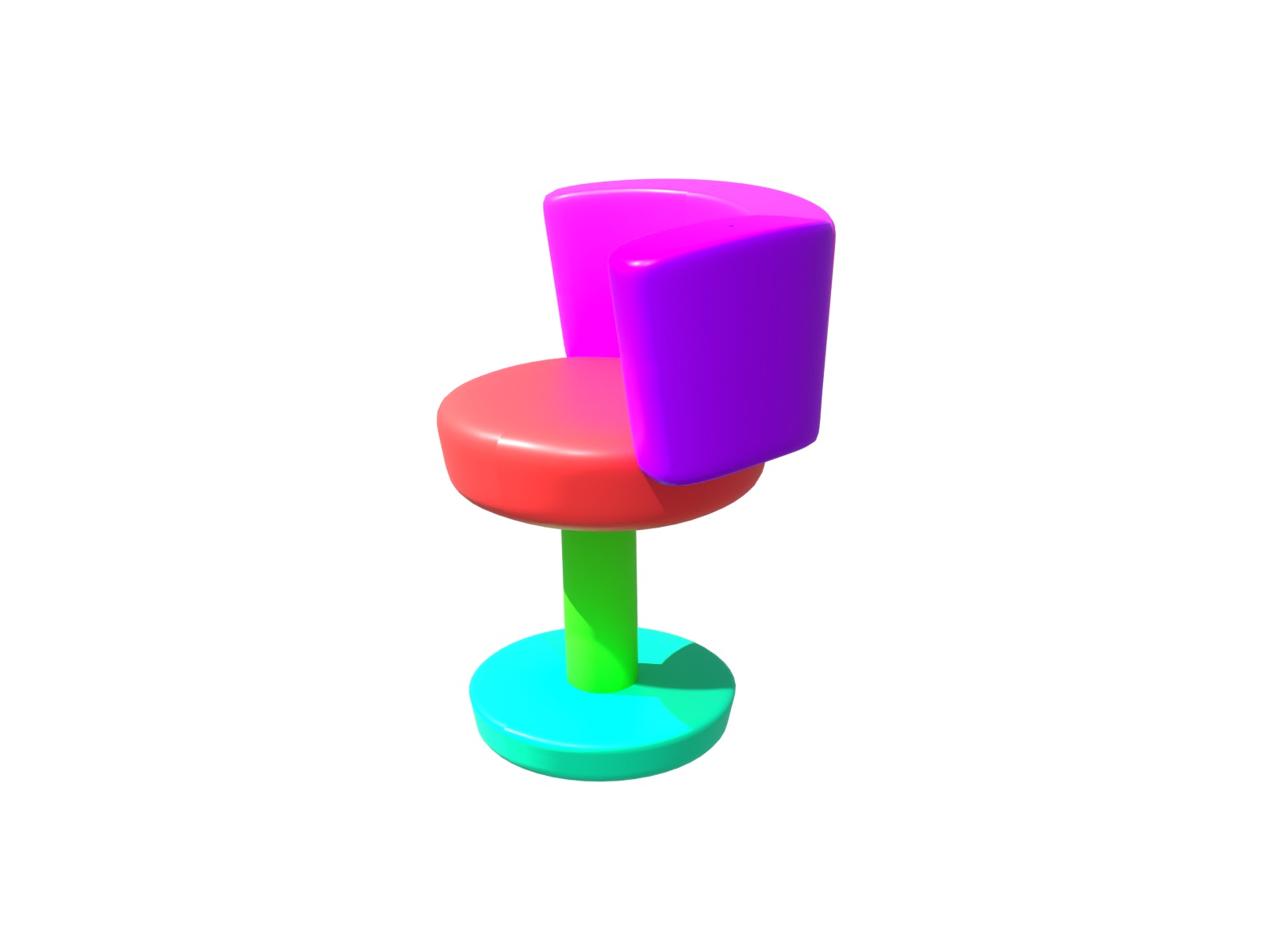}
\includegraphics[height=1.7cm,trim={420px 180px 420px 200px},clip]{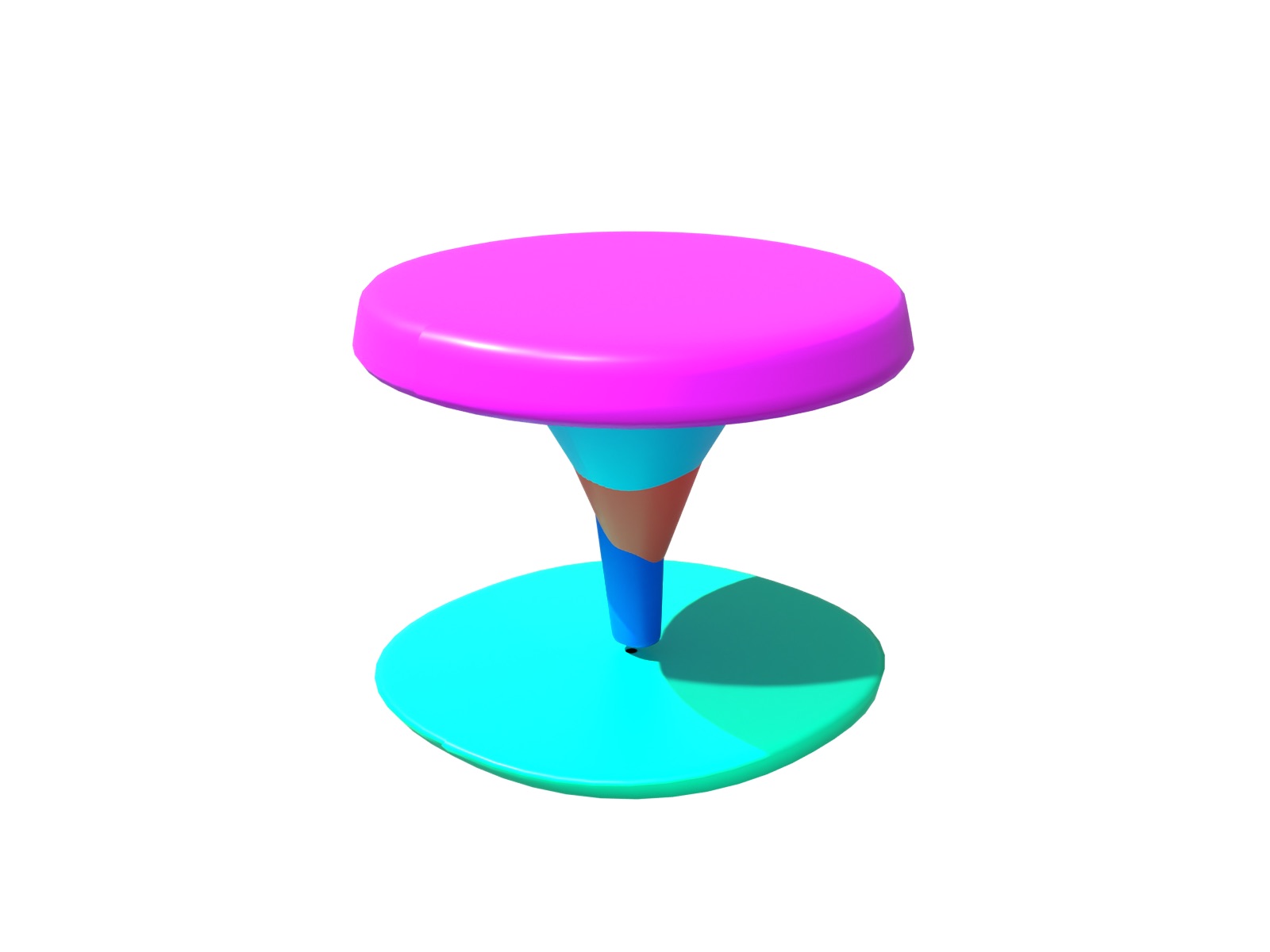}
\includegraphics[height=1.7cm,trim={400px 280px 400px 220px},clip]{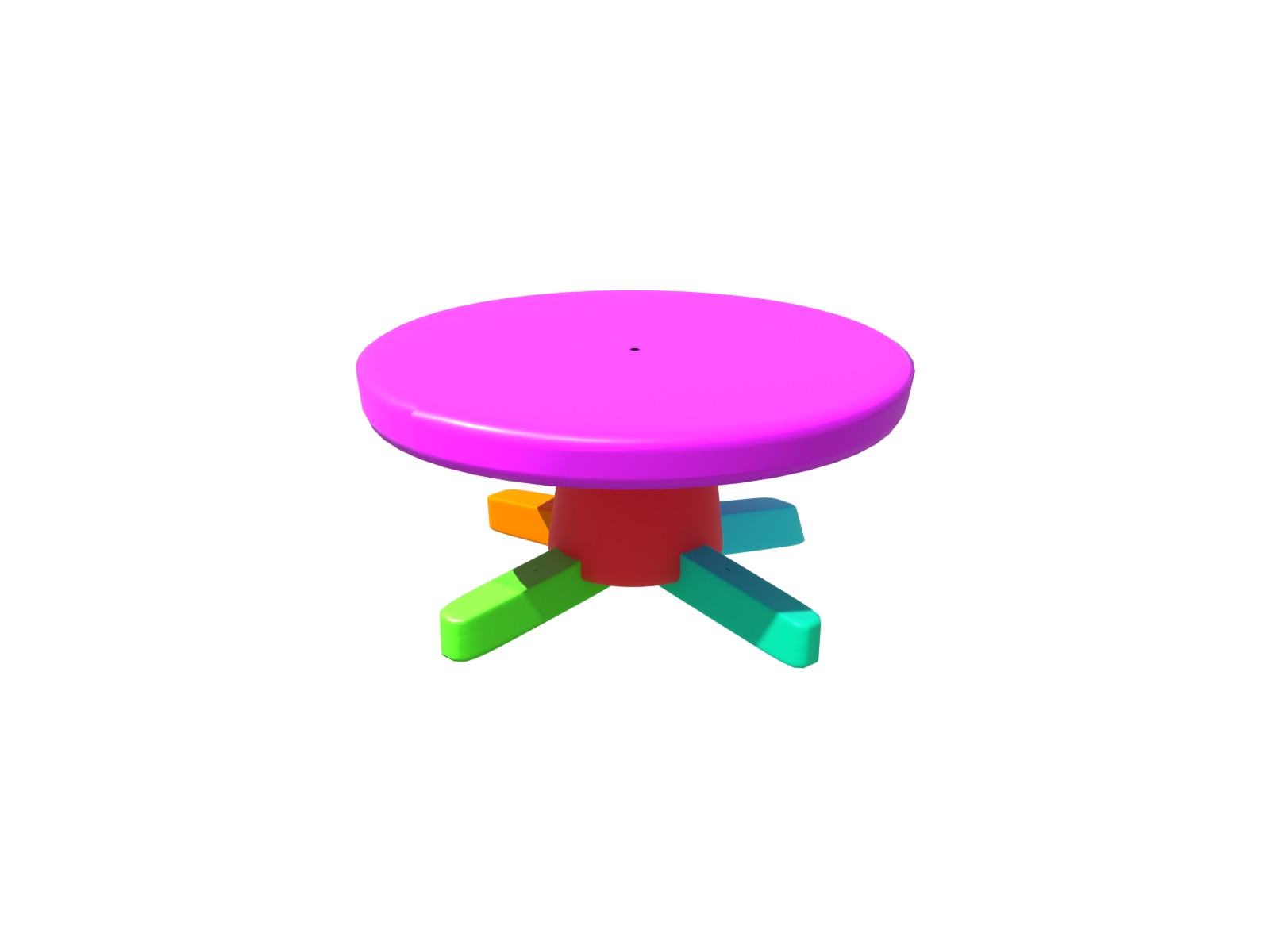}
\includegraphics[height=1.7cm,trim={440px 160px 440px 120px},clip]{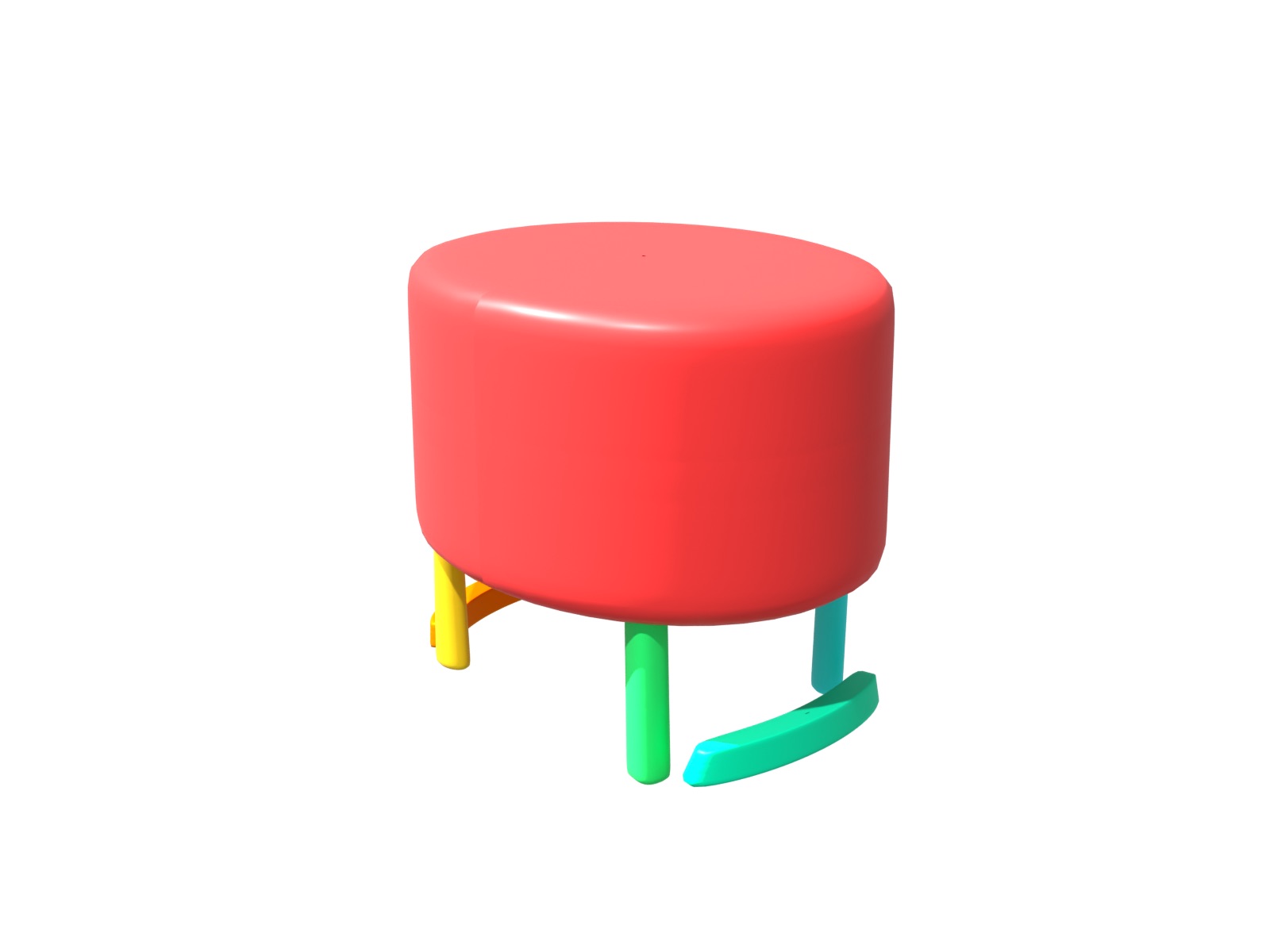}\\

\rotatebox{90}{\footnotesize \shortstack{\hspace{5px}\method{}\\ Optimized}}\hspace{-4px}
\includegraphics[height=1.7cm,trim={420px 200px 480px 200px},clip]{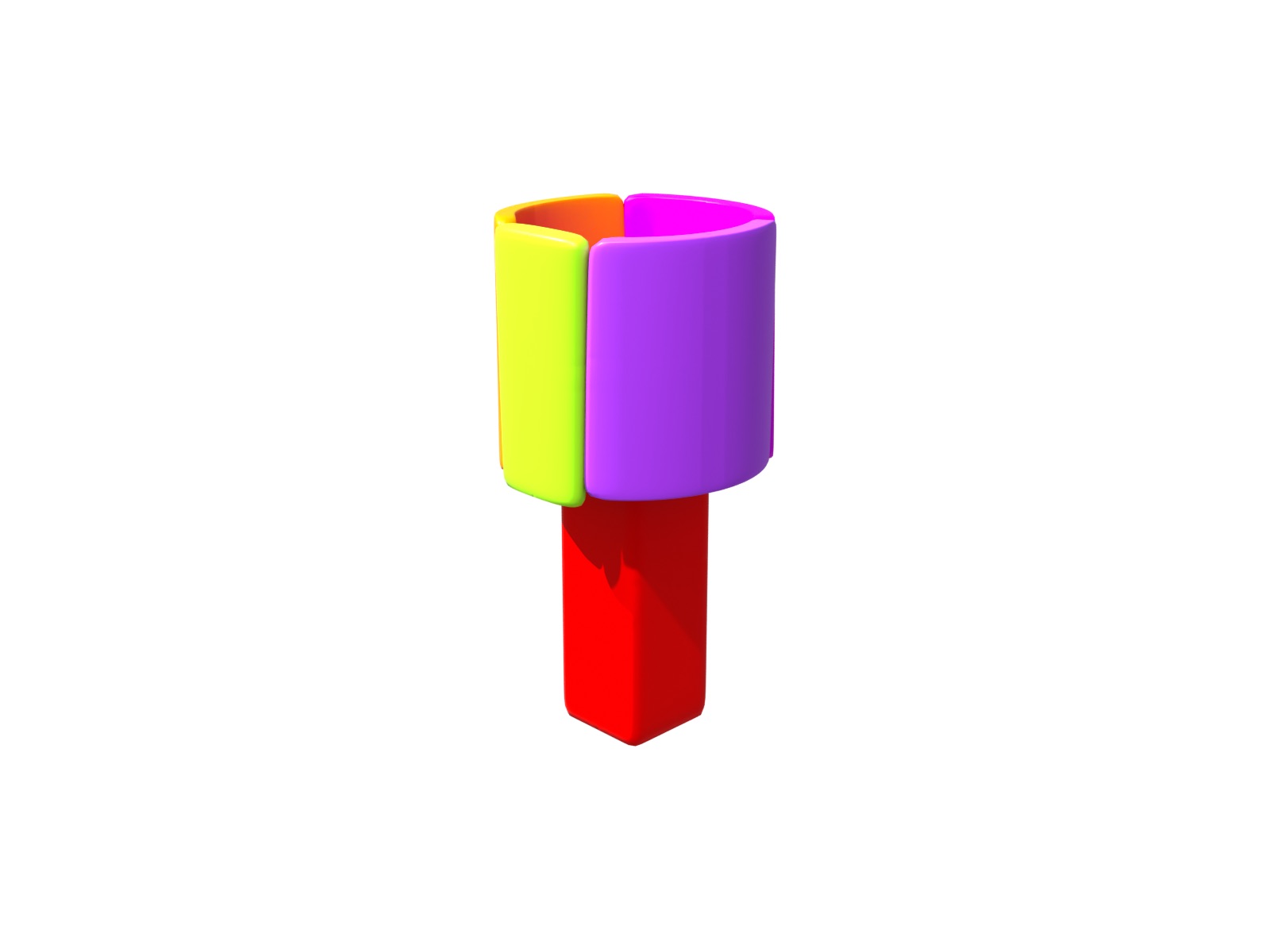}
\includegraphics[height=1.7cm,trim={440px 160px 440px 120px},clip]{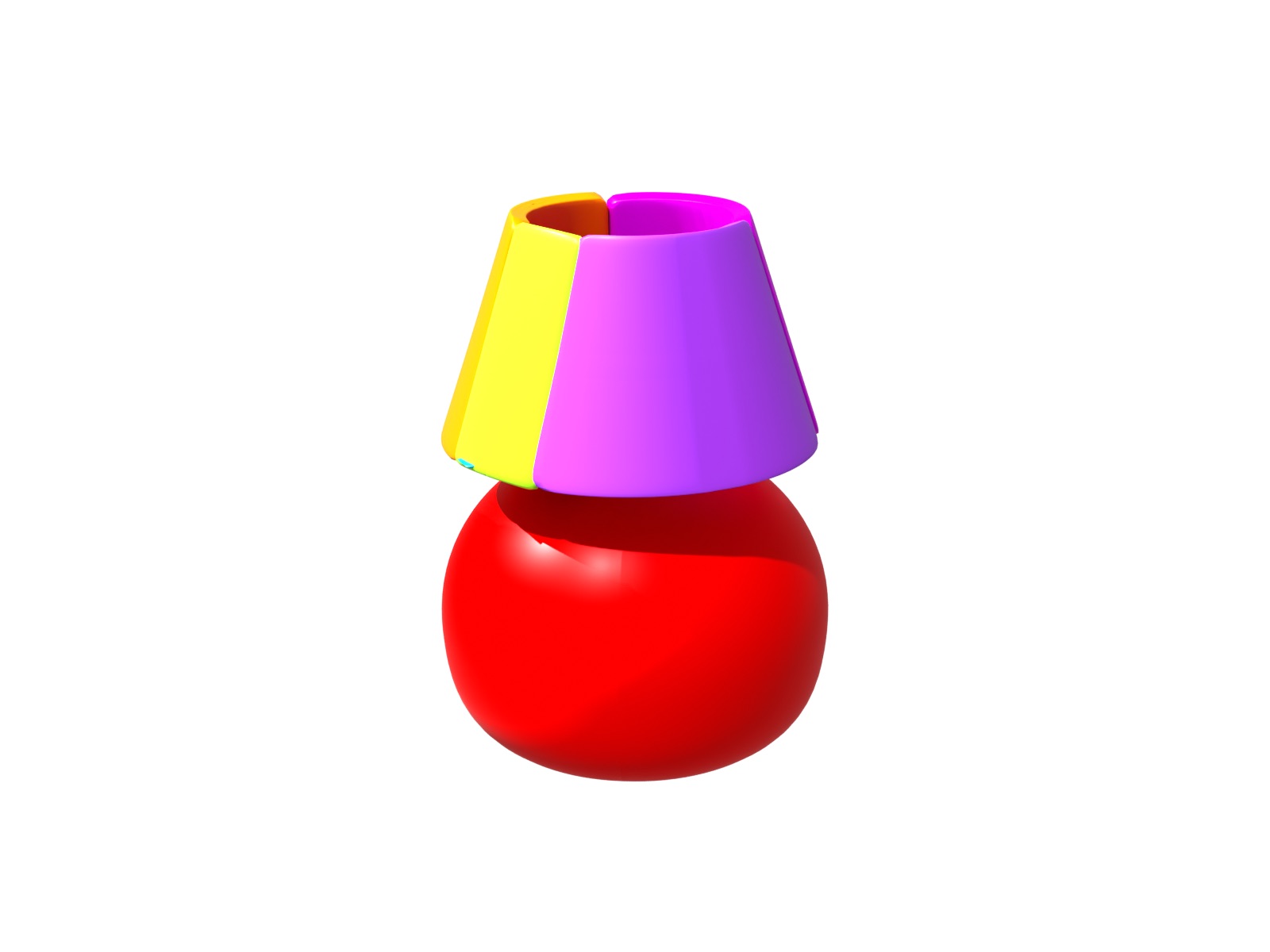}
\includegraphics[height=1.7cm,trim={420px 140px 420px 200px},clip]{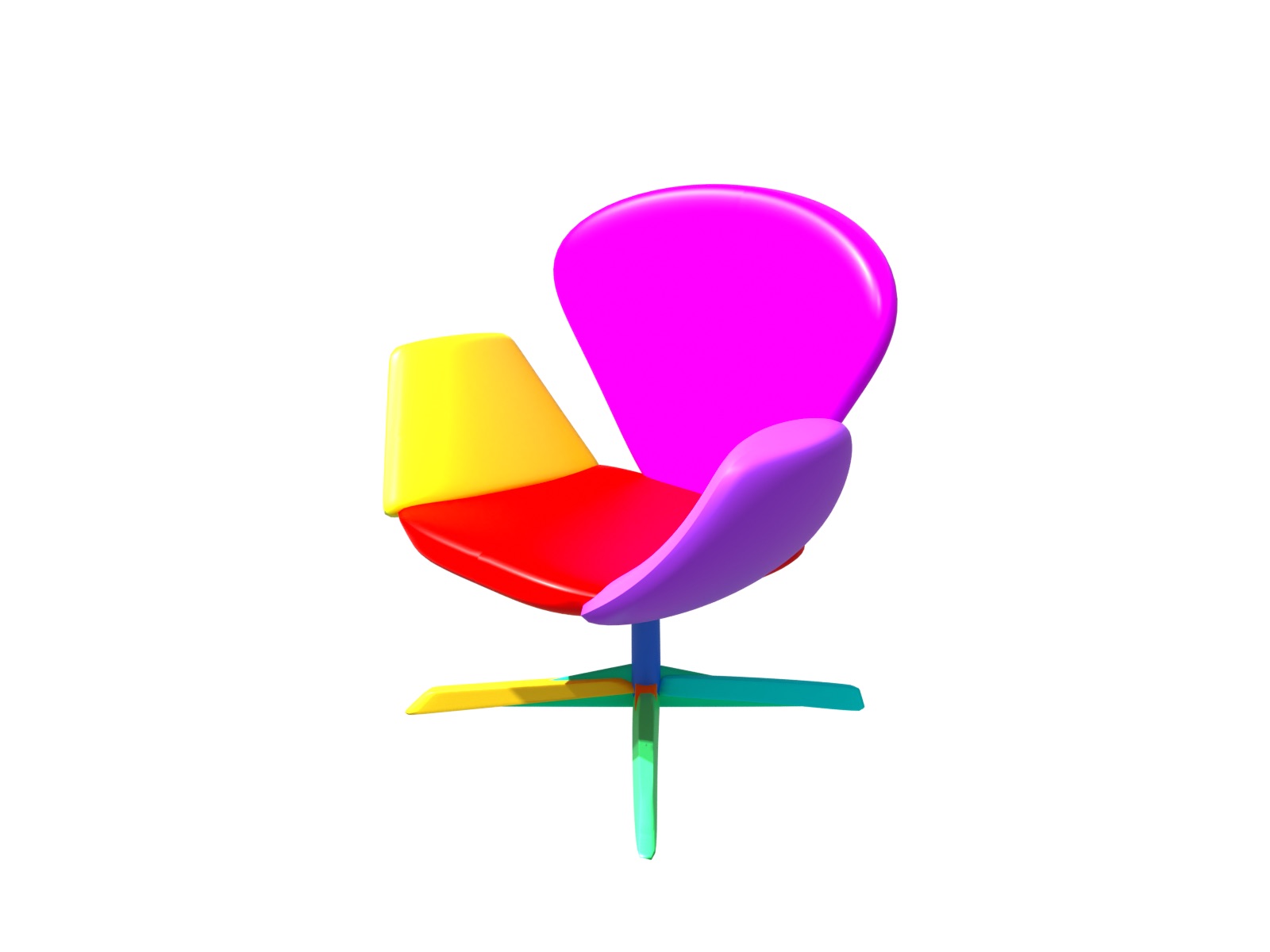}
\includegraphics[height=1.7cm,trim={440px 160px 440px 120px},clip]{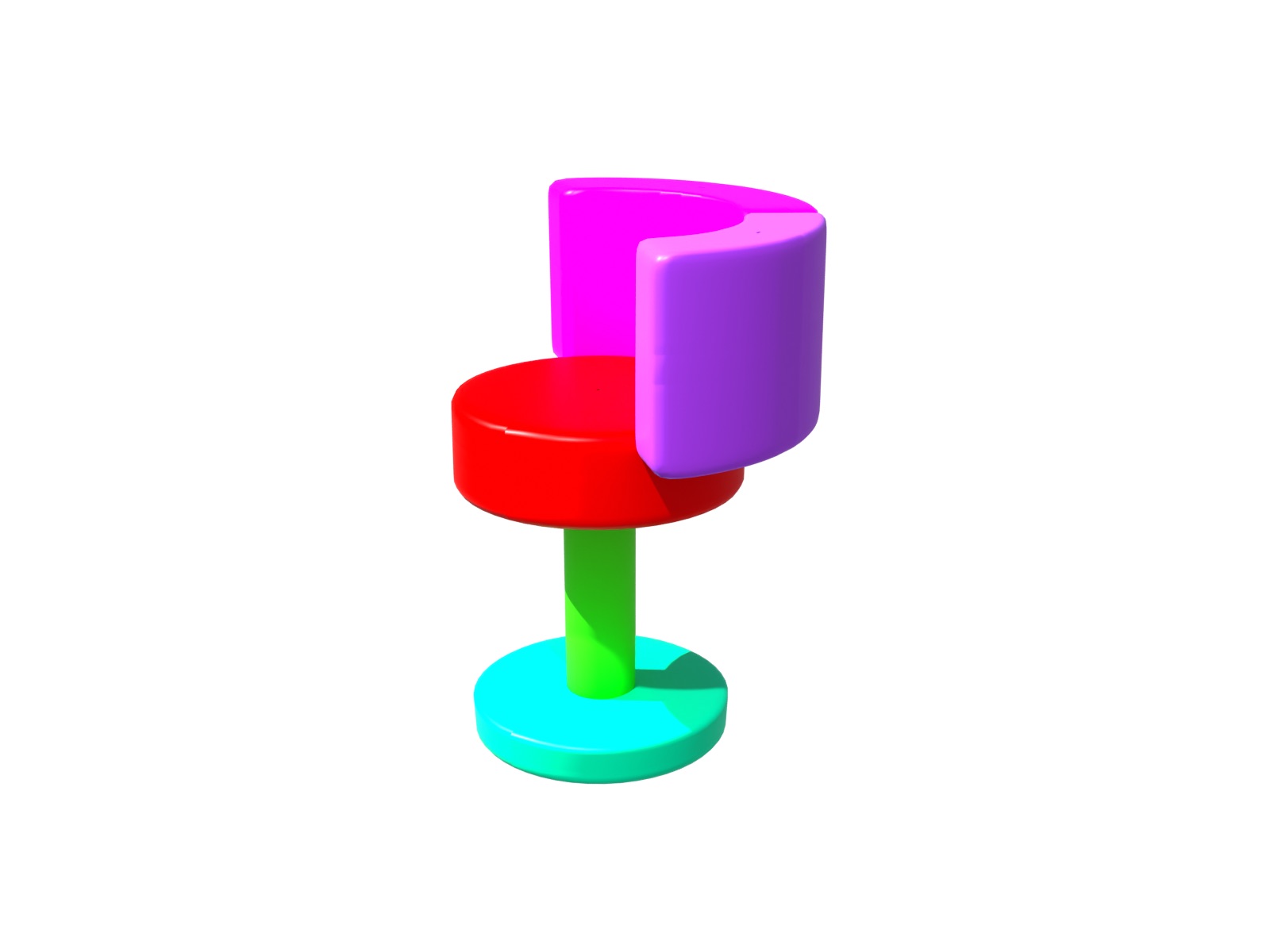}
\includegraphics[height=1.7cm,trim={420px 180px 420px 200px},clip]{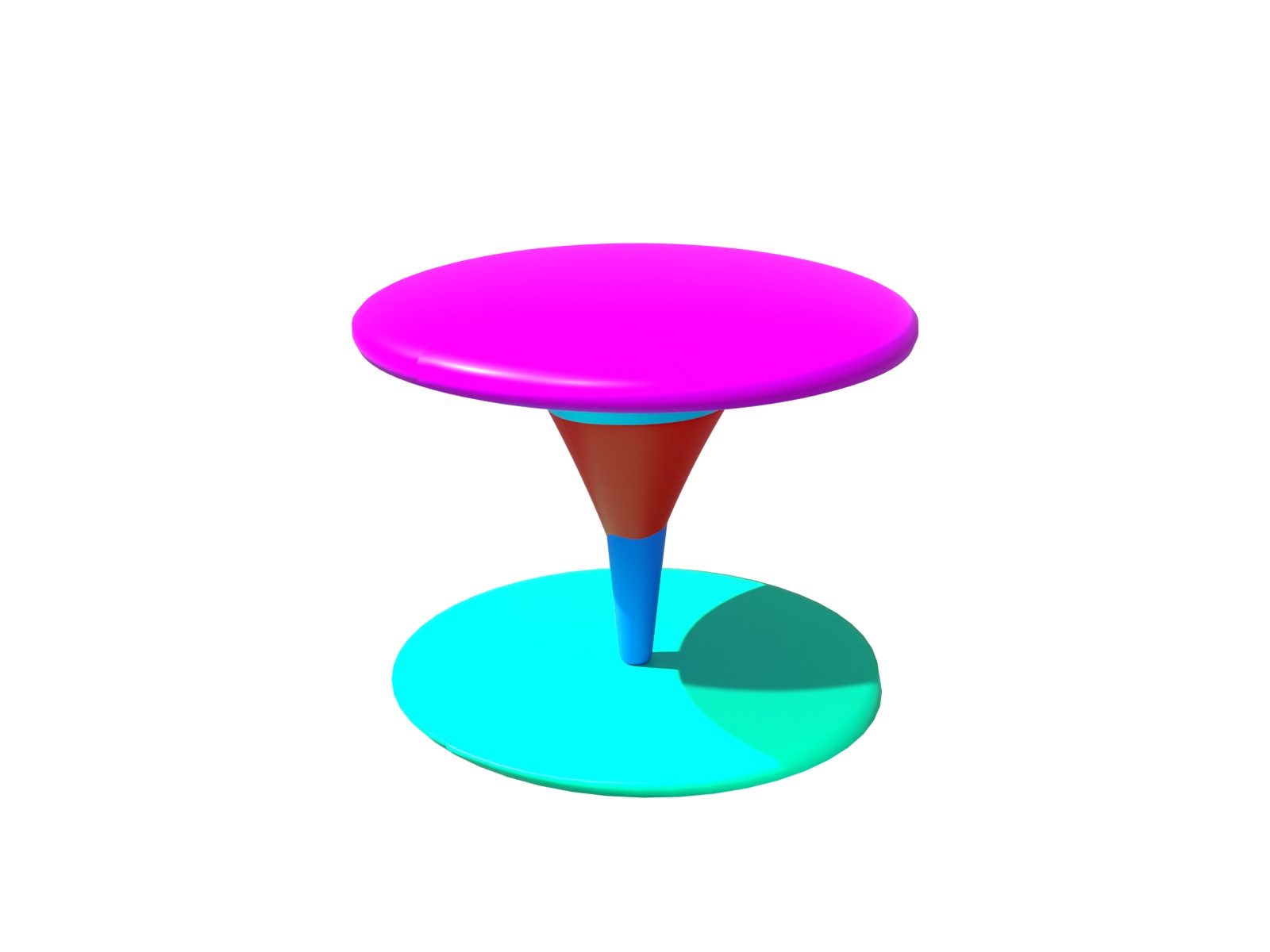}
\includegraphics[height=1.7cm,trim={400px 280px 400px 220px},clip]{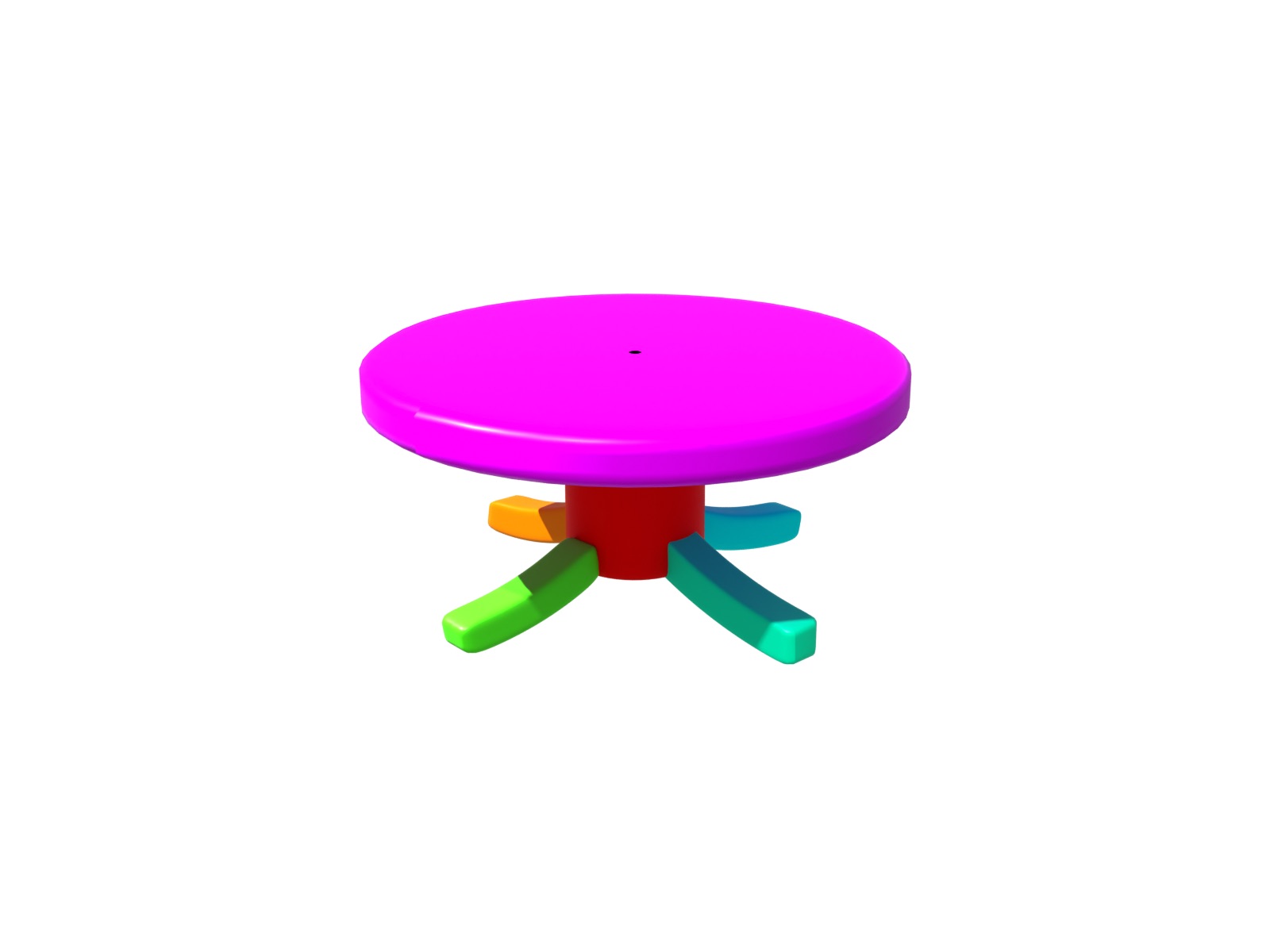}
\includegraphics[height=1.7cm,trim={440px 160px 440px 120px},clip]{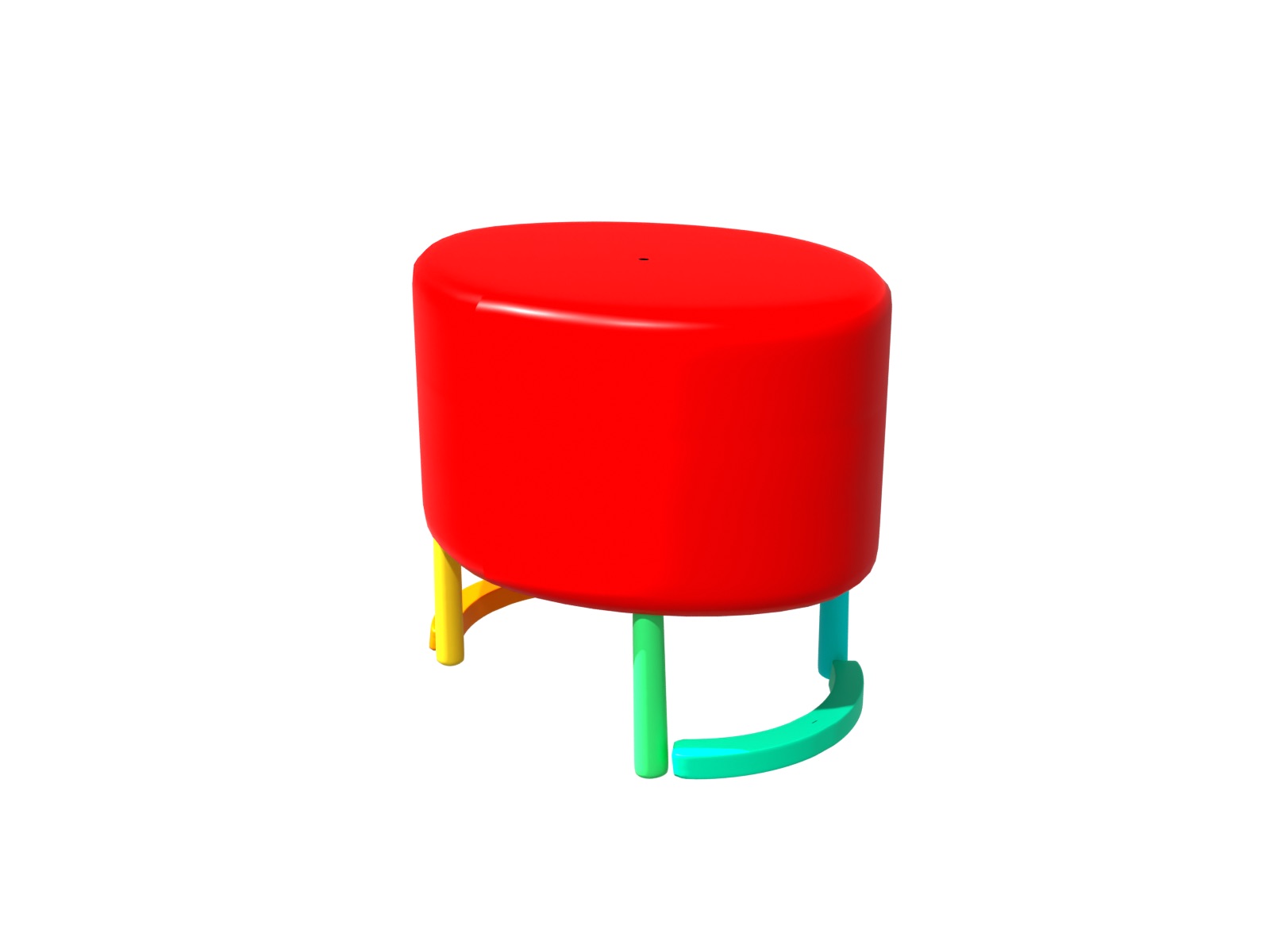}\\
\end{small}
\caption{\textbf{Additional Qualitative Results on ShapeNet.} \emph{Top row:} input point clouds. \emph{Below:} outputs of our method. Different colors indicate different primitives.}
\label{fig:quali-superflex}
\end{figure}

This section illustrates the effectiveness of our two-stage approach: a robust feed-forward prediction followed by optimization-based refinement. In Figure~\ref{fig:quali-superflex-abo} and Figure~\ref{fig:quali-superflex}, we compare the raw outputs of $\method{}^{TB}$ with the results after our optimization procedure on objects from the Amazon Berkeley Objects (ABO) dataset~\cite{collins2022abo} and ShapeNet~\cite{chang2015shapenet}. The feed-forward model successfully identifies the global topology and provides a near-optimal initial placement of primitives. Subsequently, the optimization stage "tightens" the fit by fine-tuning the deformation parameters, significantly reducing the reconstruction error while maintaining the structural integrity of the initial decomposition.

\section{3D Scene Decomposition from RGB Frames}
\label{sec:sam3}
As already shown in the main paper, our method can be used in combination with 3D instance segmentation methods to decompose point clouds of entire scenes into superquadric primitives. While this generally works well, it still requires running a 3D reconstruction of the full scene, which may not always be feasible.

\begin{figure}[t!]
    \centering
    \begin{overpic}[width=\linewidth]
    {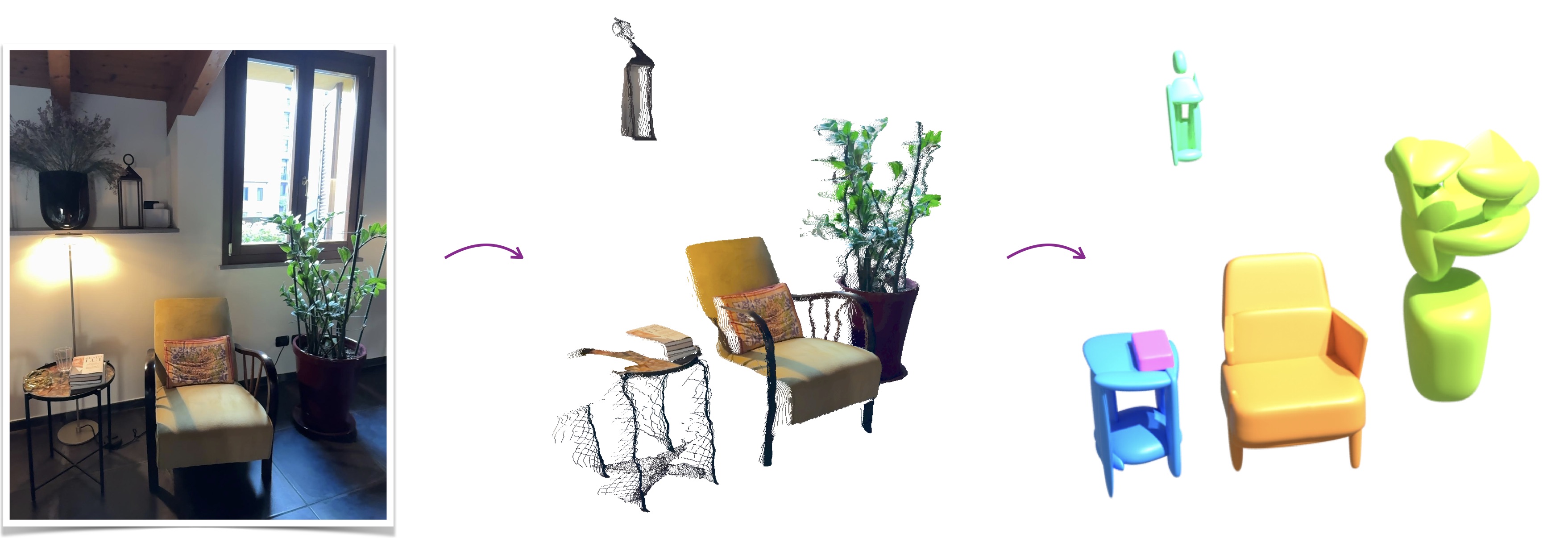}
    \put(1.2,-2.0){{\footnotesize \colorbox{white}{Input RGB Image}}}
    \put(35,-2.0){{\footnotesize \colorbox{white}{Object Point Maps}}}
    \put(74,-2.0){{\footnotesize \colorbox{white}{Superquadrics}}}
    \put(27,20){{\footnotesize \colorbox{white}{SAM3}}}
    \put(27,15){{\footnotesize \colorbox{white}{MoGE}}}
    \put(61,20.3){{\footnotesize \colorbox{white}{\method{}}}}
    \end{overpic}
    \vspace{3px}
    \caption{\textbf{\method{} from a single RGB Image.} From an input RGB frame, we first compute instance point maps using SAM3 and MoGE. Then, we apply \method{} on top of each object's point map.}
    \label{fig:moge-sam3}
\end{figure}

\noindent In this section, inspired by how SAM3D~\cite{sam3dteam2025sam3d3dfyimages} proposed to reconstruct multi-object scenes from a single RGB frame, we experiment how \method{} can be used in the same monocular setting. We propose a modular pipeline (Fig.~\ref{fig:moge-sam3}) that integrates SAM3~\cite{carion2025sam3segmentconcepts} for object segmentation and MoGE~\cite{wang2025moge2} for lifting 2D masks into partial 3D point clouds. While single-view observations suffer from inherent self-occlusion, our robust model, fine-tuned with occlusion augmentations, can still infer complete object geometries as sets of coherent deformable primitives. These results suggest that \method{} has the potential to serve as a robust, compact, and interpretable alternative to traditional full-scene reconstructions even from monocular observations.

\end{document}